\documentclass[acmsmall, screen]{acmart}

\AtBeginDocument{%
  \providecommand\BibTeX{{%
    \normalfont B\kern-0.5em{\scshape i\kern-0.25em b}\kern-0.8em\TeX}}}

\usepackage{color,xcolor}
\usepackage{graphicx}
\usepackage{amsmath}
\usepackage{amssymb}
\usepackage{colortbl}

\def\ie{\textit{i.e.}}

\def\et{\textit{et al.}}

\definecolor{mygray}{gray}{.9}
\definecolor{mypink}{rgb}{.99,.91,.95}
\definecolor{mycyan}{cmyk}{.3,0,0,0}
\definecolor{tabcolor}{rgb}{.5,.5,.5}

\begin{document}

\title{Hierarchical and Progressive Image Matting}

\author{Yu Qiao}
\authornote{Both authors contributed equally to this research.}
\email{qiaoyu2020@mail.dlut.edu.cn}
\author{Yuhao Liu}
\authornotemark[1]
\email{yuhaoliu7456@gmail.com}
\affiliation{%
	\institution{Dalian University of Technologyy}
	\streetaddress{2 linggong road}
	\city{Dalian}
	\state{Liaoning}
	\country{China}
	\postcode{116024}
}

\author{Ziqi Wei}
\authornote{Corresponding author.}
\email{ziqi.wei@ia.ac.cn}
\affiliation{%
	\institution{CAS Key Laboratory of Molecular Imaging, Institute of Automation}
	\city{Beijing}
	\country{China}
	\postcode{100190}
}

\author{Yuxin Wang}
\email{wyx@dlut.edu.cn}
\affiliation{%
	\institution{Dalian University of Technology}
	\streetaddress{2 linggong road}
	\city{Dalian}
	\state{Liaoning}
	\country{China}
	\postcode{116024}
}

\author{Qiang Cai}
\email{caiq@th.btbu.edu.cn}
\affiliation{%
	\institution{Beijing Technology And Business University}
	\city{Beijing}
	\country{China}
	\postcode{100048}
}

\author{Guofeng Zhang}
\email{zhangguofeng@wonxing.com}
\affiliation{%
	\institution{Wonxing Technology}
	\streetaddress{14 Haitian 2 Road}
	\city{Shenzhen}
	\state{Guangdong}
	\country{China}
	\postcode{518110}
}

\author{Xin Yang}
\authornotemark[2]
\email{xinyang@dlut.edu.cn}
\affiliation{%
	\institution{Dalian University of Technologyy}
	\streetaddress{2 linggong road}
	\city{Dalian}
	\state{Liaoning}
	\country{China}
	\postcode{116024}
}

\renewcommand{\shortauthors}{Xin, et al.}

\begin{abstract}
Most matting researches resort to advanced semantics to achieve high-quality alpha mattes, and direct low-level features combination is usually explored to complement alpha details. However, we argue that appearance-agnostic integration can only provide biased foreground details and alpha mattes require different-level feature aggregation for better pixel-wise opacity perception. In this paper, we propose an end-to-end Hierarchical and Progressive Attention Matting Network (\emph{HAttMatting++}), which can better predict the opacity of the foreground from single RGB images without additional input. Specifically, we utilize channel-wise attention to distill pyramidal features and employ spatial attention at different levels to filter appearance cues. This progressive attention mechanism can estimate alpha mattes from adaptive semantics and semantics-indicated boundaries. We also introduce a hybrid loss function fusing Structural SIMilarity (SSIM), Mean Square Error (MSE), Adversarial loss, and sentry supervision to guide the network to further improve the overall foreground structure. Besides, we construct a large-scale and challenging image matting dataset comprised of $59,600$ training images and $1000$ test images (a total of $646$ distinct foreground alpha mattes), which can further improve the robustness of our hierarchical and progressive aggregation model. Extensive experiments demonstrate that the proposed \emph{HAttMatting++} can capture sophisticated foreground structures and achieve state-of-the-art performance with single RGB images as input.
\end{abstract}



\maketitle

\section{Introduction}
\label{sec:introduction}
Image matting refers to precisely estimating the foreground opacity from an input image. This problem as well as its inverse process (known as image composition) have been well studied by both academia and industry. Image matting serves as a prerequisite technology for a broad set of applications, including online image editing, mixed reality, and film production. Formally, it is modeled by solving the following image synthesis equation:
\begin{equation}
\label{image_synthesis}
\textit{I}_{z} = \alpha_{z}F_{z}+(1 - \alpha_{z})B_{z}, \quad \alpha_{z}\in[0,1]	
\end{equation}
where $z$ denotes the pixel position in the input image $\textit{I}$. $\alpha_{z}$, $\textit{F}_{z}$ and $\textit{B}_{z}$ refer to the alpha estimation, foreground (\emph{FG}) and background (\emph{BG}) at pixel $z$ separately. The problem is highly ill-posed. For each pixel in a given \emph{RGB} image, the only observed value is the input $I_{z}$, which has R, G, B channels. $F_{z}$ and $B_{z}$ also have $3$ channels, and the alpha value $\alpha_{z}$ has one channel. Thus there are $7$ values need to be solved, but only 3 values are known.
\begin{figure}[t]
\setlength{\tabcolsep}{2pt}\small{
	\begin{tabular}{cccc}
		\includegraphics[scale=0.048]{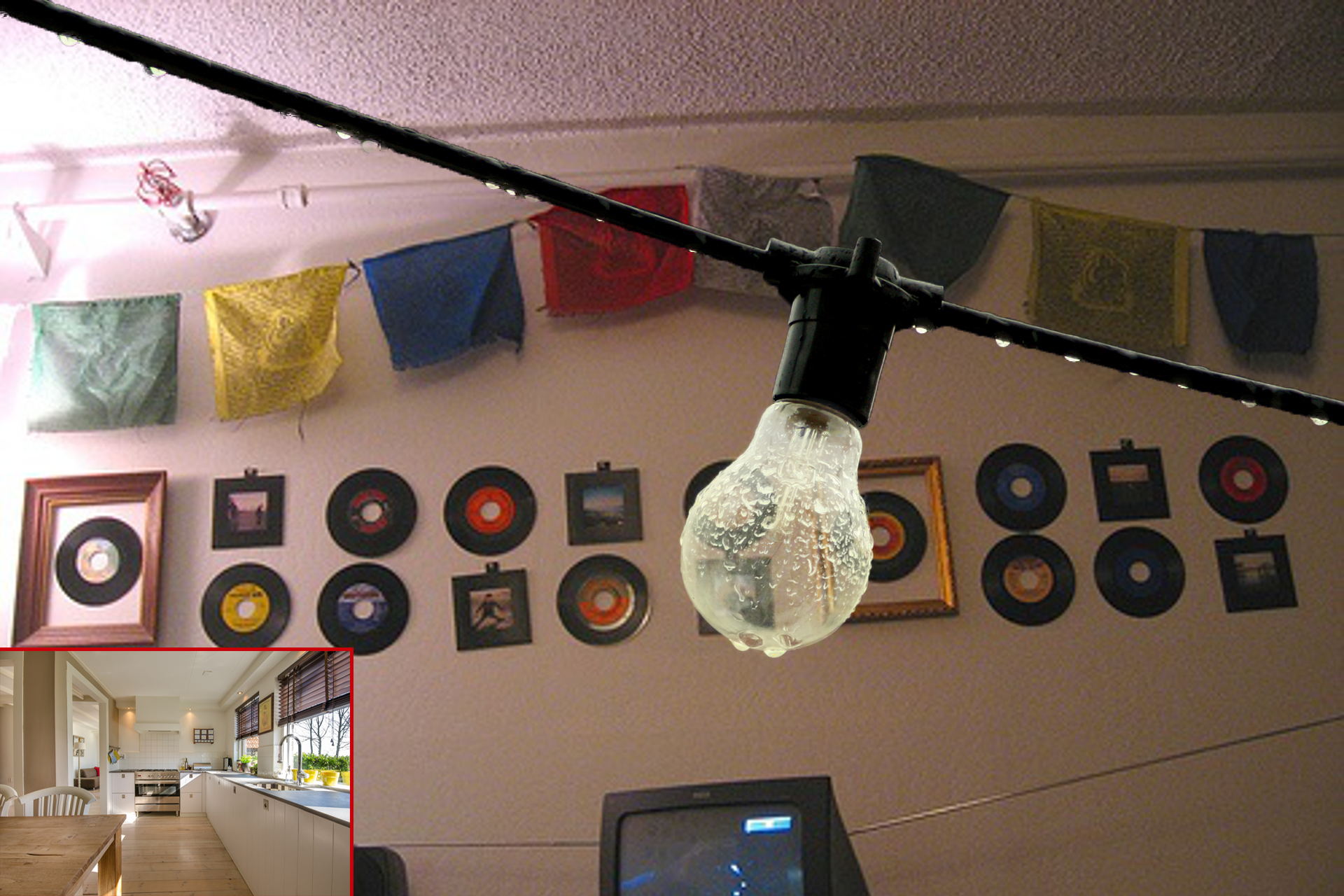} &
		\includegraphics[scale=0.048]{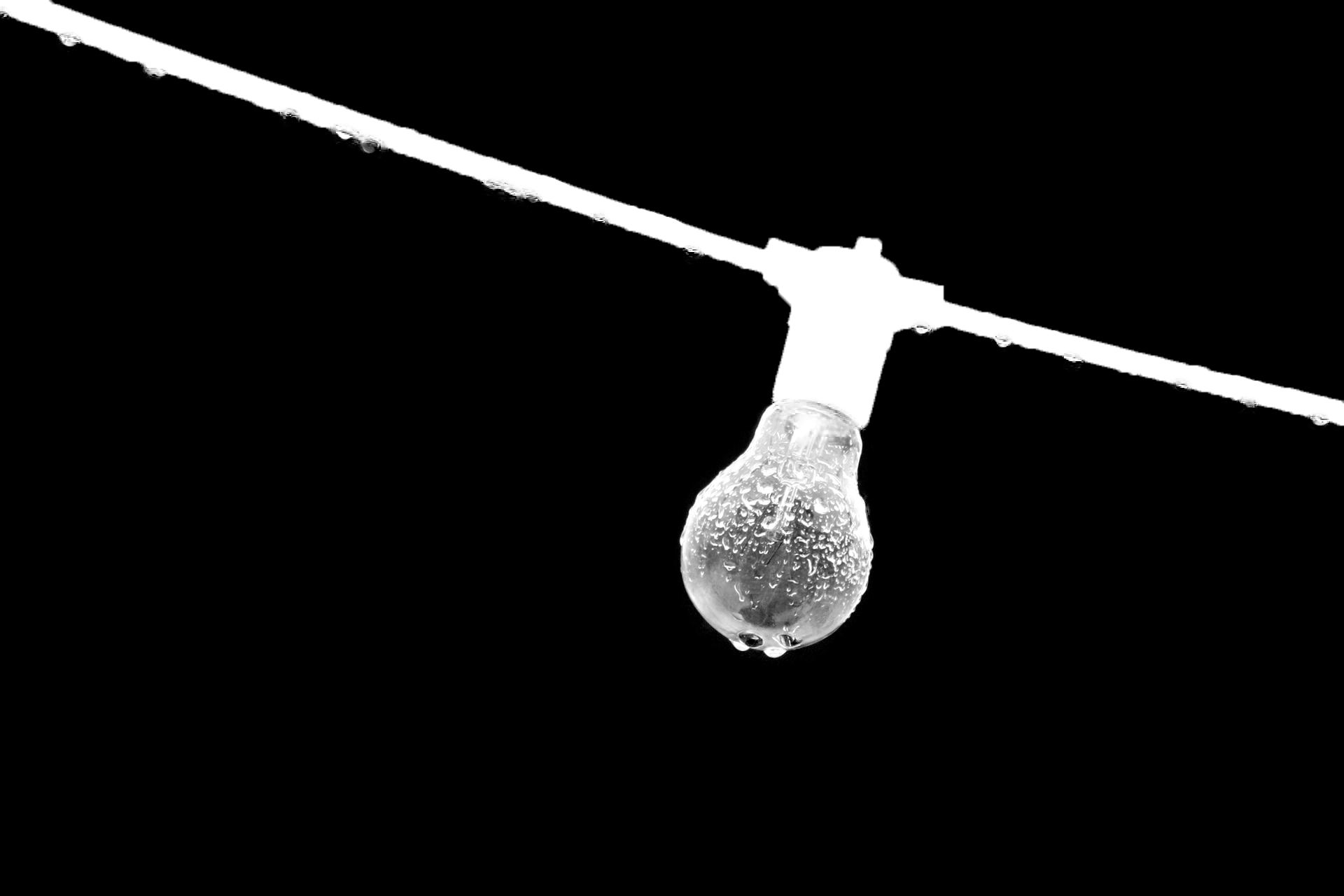} &
		\includegraphics[scale=0.048]{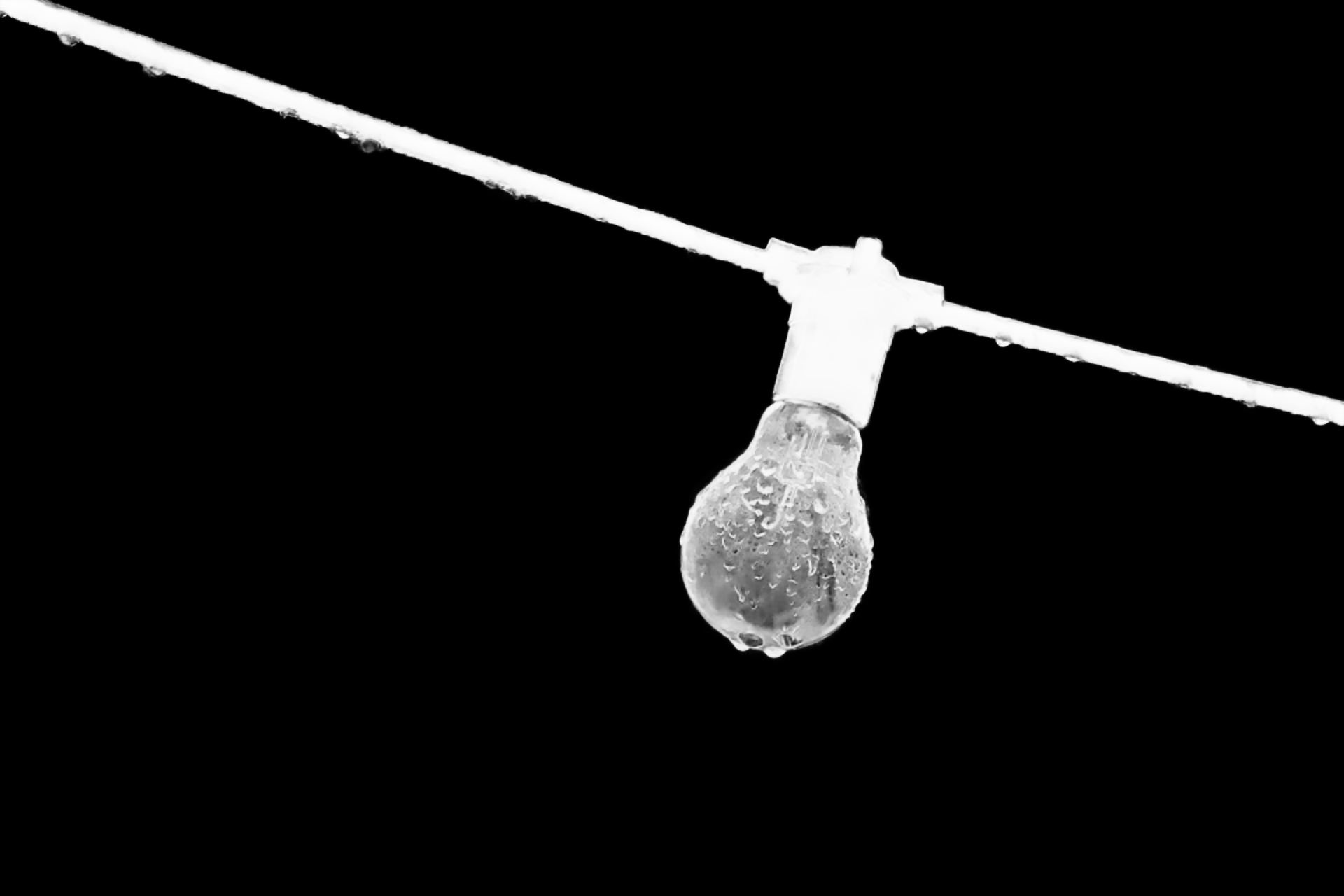} &
		\includegraphics[scale=0.048]{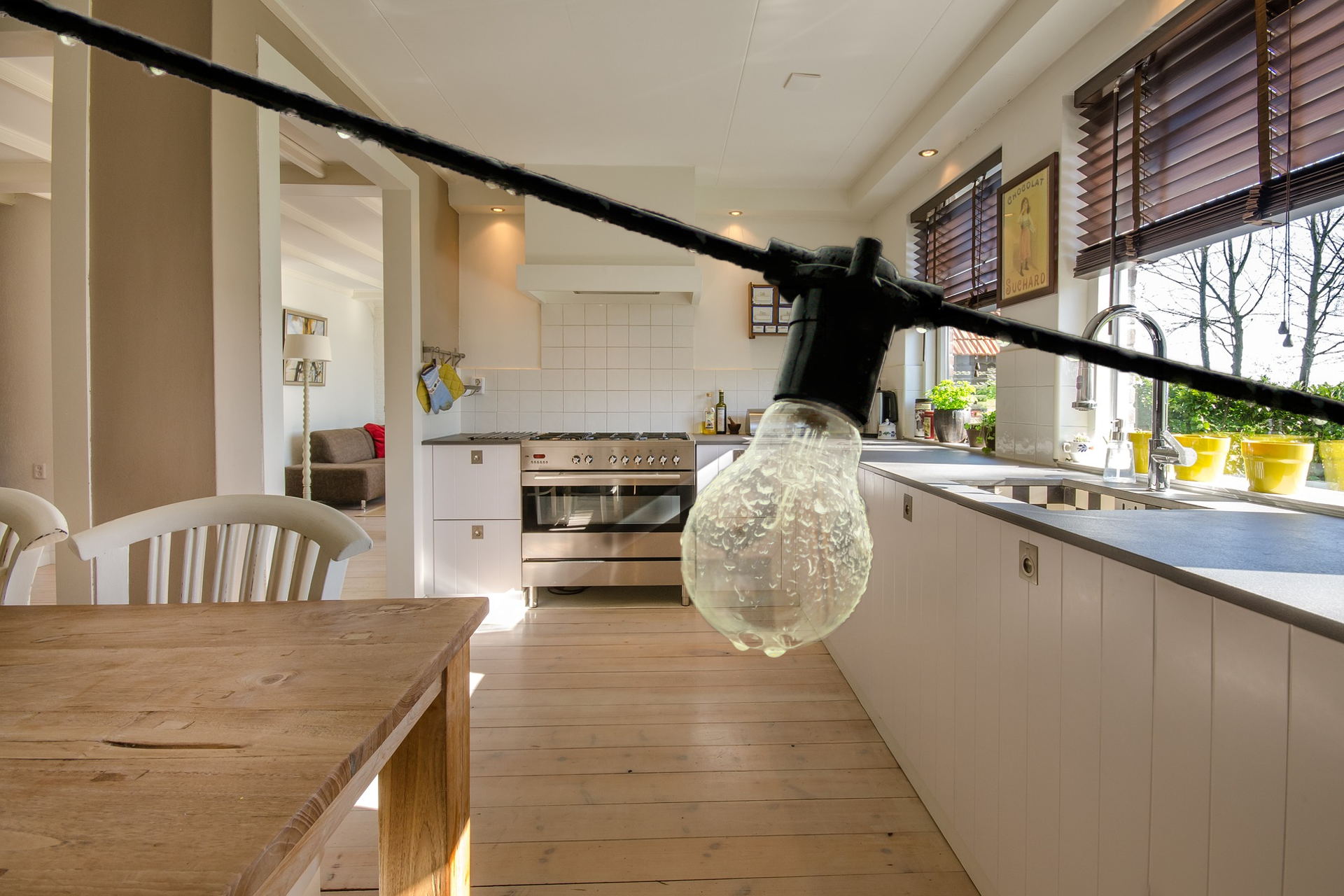} \\
		
		\includegraphics[scale=0.0614]{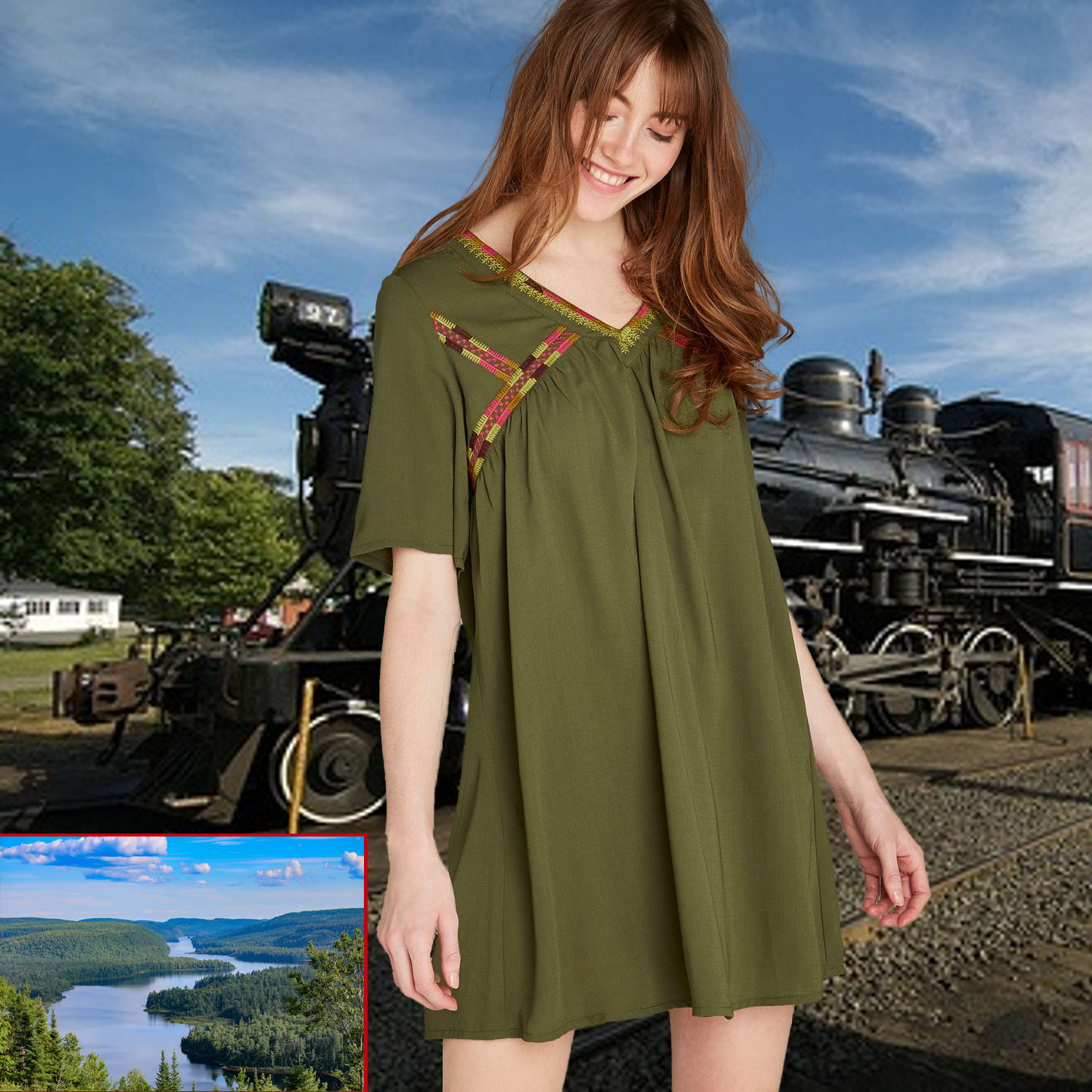} &
		\includegraphics[scale=0.0614]{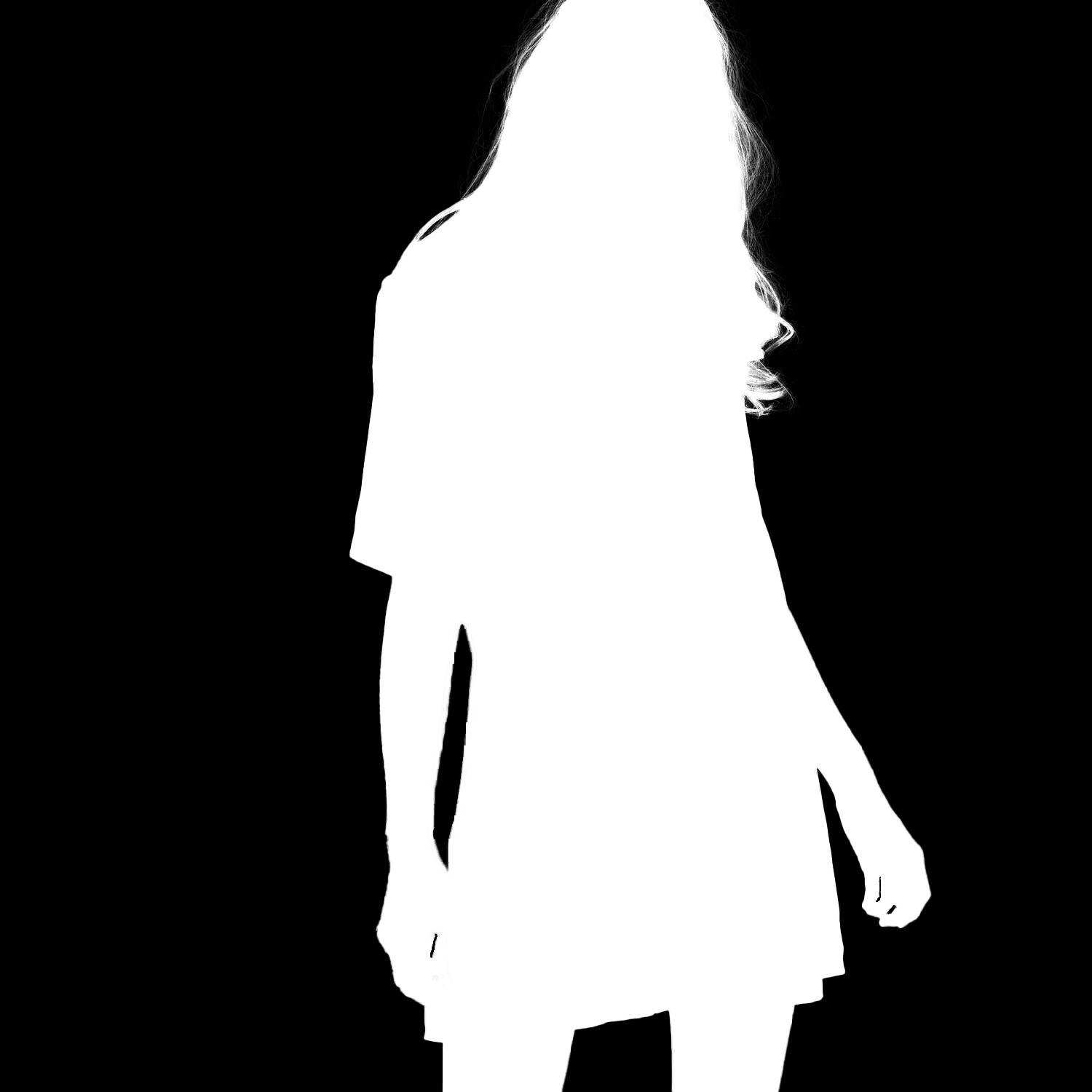} &
		\includegraphics[scale=0.0614]{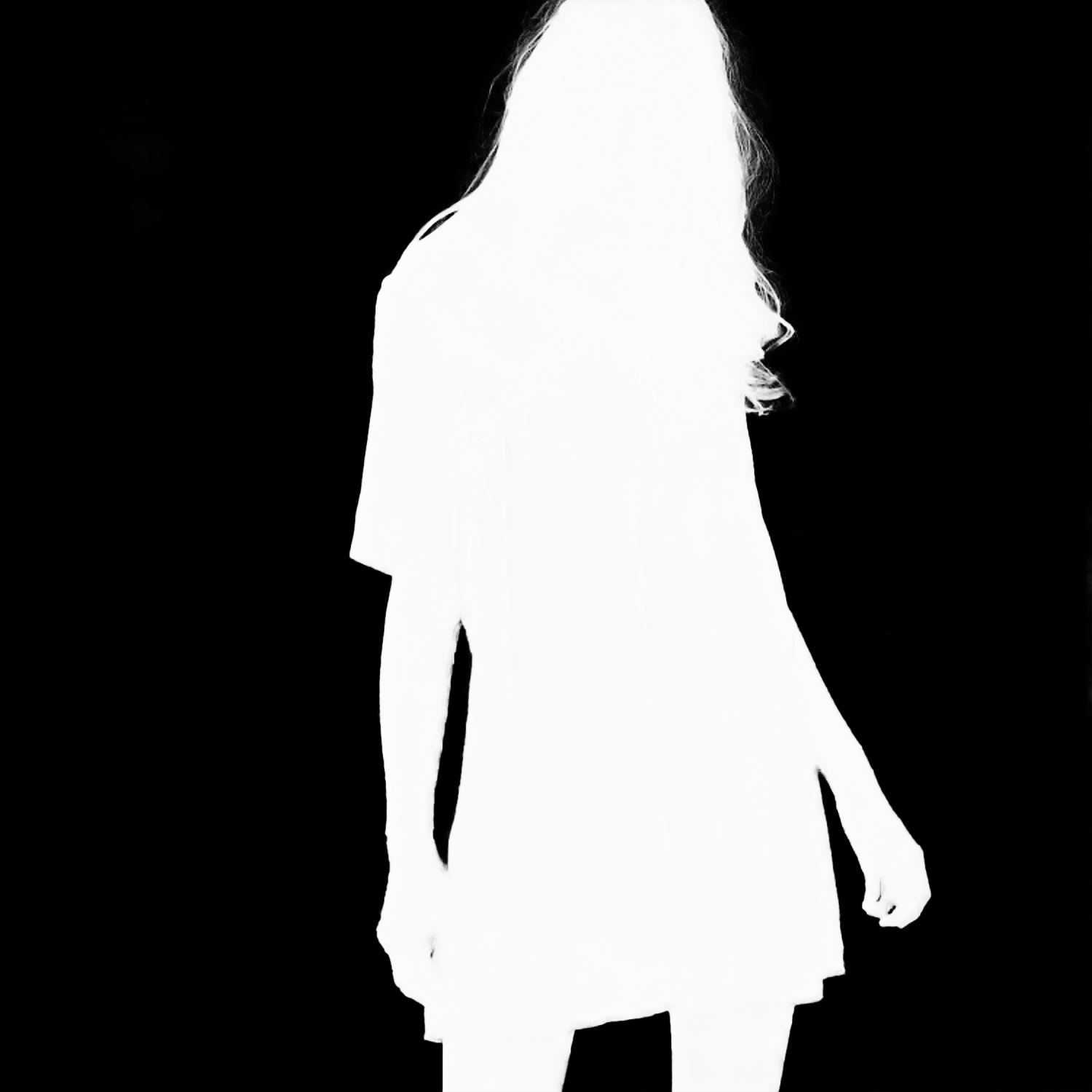} &
		\includegraphics[scale=0.0614]{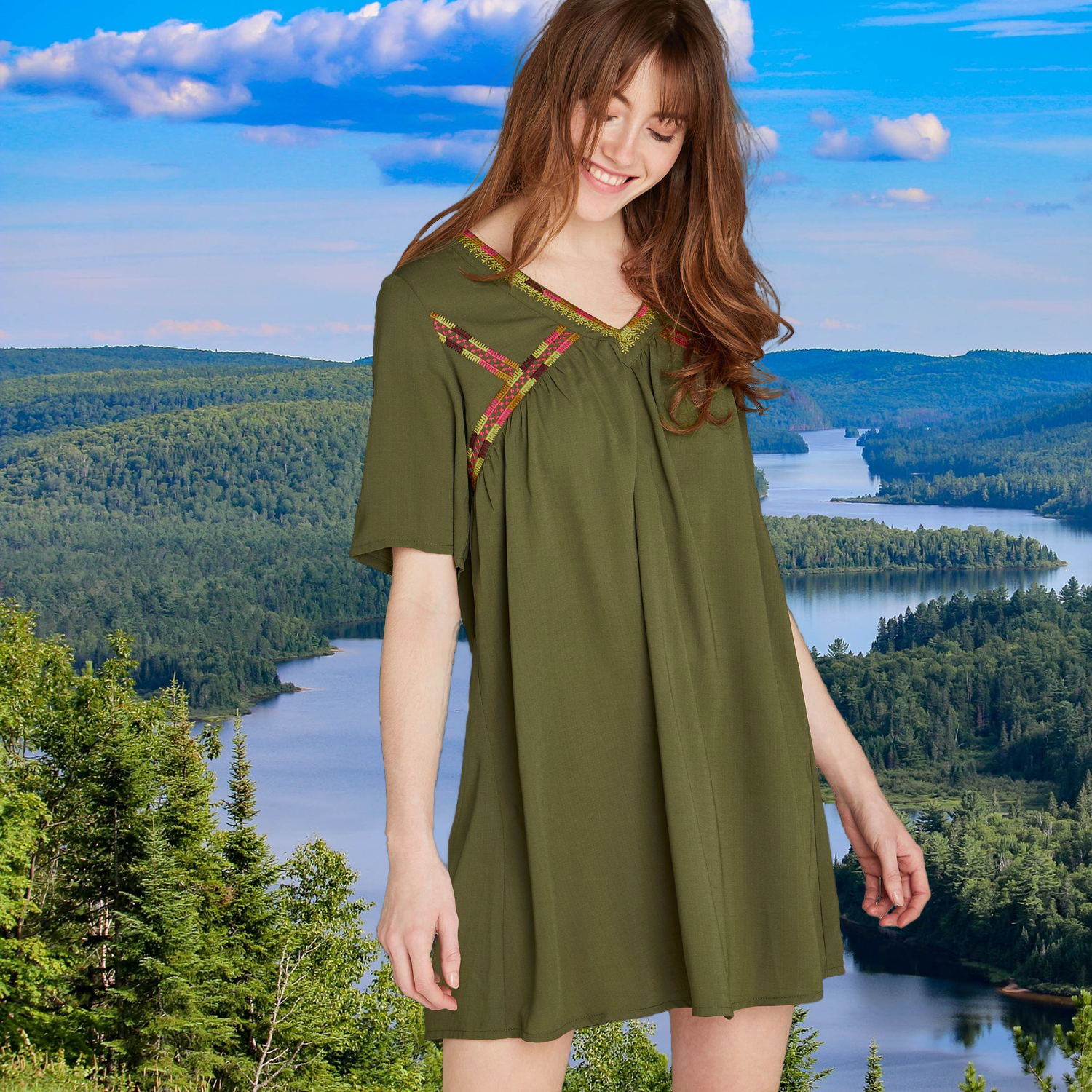} \\
		
		Input Image & Ground Truth & Our Results & Composition \\
\end{tabular}}
\caption{The alpha mattes produced by our \emph{HAttMatting++} and the corresponding composition results with some internet images as new background (as shown in the left bottom of Input Image). }
\label{fig:teaser}
\end{figure}

In this paper, we argue that the structure of \emph{FG} resides in two aspects: adaptive semantics and refined boundaries, corresponding to $\alpha_z=1$ and $\alpha_z\in(0,1)$ in Eq.~\ref{image_synthesis}, respectively. To solve such a pixel-wise and ill-posed regression problem, most matting methods usually introduce user-provided trimaps as assistant input. The trimap is composed of black, gray, and white, representing \emph{BG}, transition region, and absolute \emph{FG} separately. The transition region indicates \emph{FG} boundaries, combined with absolute \emph{FG} to jointly guide matting algorithms. Given an \emph{RGB} image and the corresponding trimap, traditional matting methods~\cite{Levin2007A, Chen2013KNN} explore hand-crafted representations or color distribution to predict an alpha matte. However, the color features are inapplicable for structure representation, possibly resulting in artifacts and loss of details when \emph{FG} and \emph{BG} have indistinguishable colors.

Deep Image Matting (\emph{DIM})~\cite{Xu2017Deep} formally imports deep learning into matting, and they argue that matting objects share a common structure which can be represented by high-level features. It is noted that \emph{DIM} involves \emph{RGB} images in the refinement stage to combine advanced semantics with appearance cues. Advanced semantics indicate \emph{FG} category and profiles, while appearance cues reveal texture and boundary details. Subsequent matting networks~\cite{cai2019disentangled,hou2019context,hao2019indexnet,dai2021learning} mostly design complicated architectures for advanced semantics extraction and involve appearance cues from the input images or low-level \emph{CNN} features. There are also some other forms of additional input, like prepared \emph{BG} images~\cite{sengupta2020background,lin2021real} and segmentation masks~\cite{yu2021mask}. However, these researches have an obvious dependency on such auxiliary and expensive input for better appearance cues and advanced semantics extraction. A well-defined trimap involves fussy manual labeling efforts and time consumption, which is difficult for novice users in practical applications. Correspondingly, prepared \emph{BG} images or masks share certain constraints to maintain consistency.

Some matting works~\cite{Chen2018SHM,Cho2016Automatic} implement single-image predictions with intermediate segmentation as pseudo trimaps, which partly depress the precision of alpha mattes. The Late Fusion~\cite{Zhang2019CVPR} blends \emph{FG} and \emph{BG} weight maps from the segmentation network with initial \emph{CNN} features to predict alpha mattes. However, when semantic segmentation encounters difficulties, the Late Fusion will compromise. The above methods directly feed advanced semantics and appearance cues to the optimization or fusion stage. The semantic information is extracted from deep layers to describe \emph{FG} profiles, while the appearance cues are mostly from input images or initial layers to focus on texture details and refined boundaries.

In this paper, we hold that the advanced semantics and initial appearance cues require essential adaption before combination, and some secondary appearance cues (such as middle-level features or parts, hands, leaves, ears, etc.) can also be explored to complement potential foreground details. On the one hand, natural image matting is a regression problem substantially and not entirely dependent on image semantics. On the other hand, while appearance cues retain sophisticated image texture, they contain the details outside \emph{FG}. Besides, some potential foreground details can be further enhanced by secondary appearance cues, though the input images or initial CNNs features can provide sufficient low-level information. Consequently, directly fusing advanced semantics and initial appearance cues is slightly flawed, and existing matting networks neglect the profound excavation and distillation of such hierarchical features.

This paper explores advanced semantics and appearance cues synthetically and contributes an end-to-end Hierarchical and Progressive Attention Matting Network (\emph{HAttMatting++}) enabling a progressive manner to aggregate different-level features. Advanced semantics can provide \emph{FG} category and profiles, while appearance cues furnish texture and boundary details. To deeply integrate this hierarchical structure, we perform channel-wise attention on advanced semantics to select matting-adapted features and employ spatial attention on multi-level appearance cues in a top-down manner to filtrate image texture details and finally aggregate them progressively to predict alpha mattes. Moreover, a hybrid loss composed of Mean Square Error~(\emph{MSE}), Structural SIMilarity~(\emph{SSIM})~\cite{Zhou2004Image}, Adversarial Loss~\cite{Goodfellow2014Generative}, and sentry supervision is exploited to optimize the whole network training. Extensive experiments show that our attention-guided hierarchical structure aggregation can attain high-quality alpha mattes with only \emph{RGB} images as input.

The main contributions of this paper are:
\begin{itemize}
\item We present an end-to-end Hierarchical and Progressive Attention Matting Network (\emph{HAttMatting++}), which can achieve high-quality alpha mattes with only RGB images. The \emph{HAttMatting++} can process variant opacity with different types of objects and has no dependency on auxiliary input. \emph{HAttMatting++} can achieve state-of-the-art performance compared to the existing single-input matting methods.
\item We present a hierarchical and progressive attention mechanism that can aggregate advanced pyramidal features and different-level appearance cues to produce adaptive semantics and variant foreground details for alpha perception. We resort to a hybrid loss consisting of Mean Square Error~\emph{MSE}, Structural SIMilarity, Adversarial Loss, and sentry supervision to provide efficient guidance for model training.
\item We build a large-scale and challenging matting dataset consisting of 59,600 training images and 1000 test images, 646 distinct foreground alpha mattes in total. To the best of our knowledge, this is the largest matting dataset with diverse foreground objects, which can further improve the robustness of \emph{HAttMatting++}. Meanwhile, we have made our dataset publicly available, which can promote further research and evaluation.
\end{itemize}

\begin{figure*}[t]
	\centering
	\includegraphics[width=\linewidth]{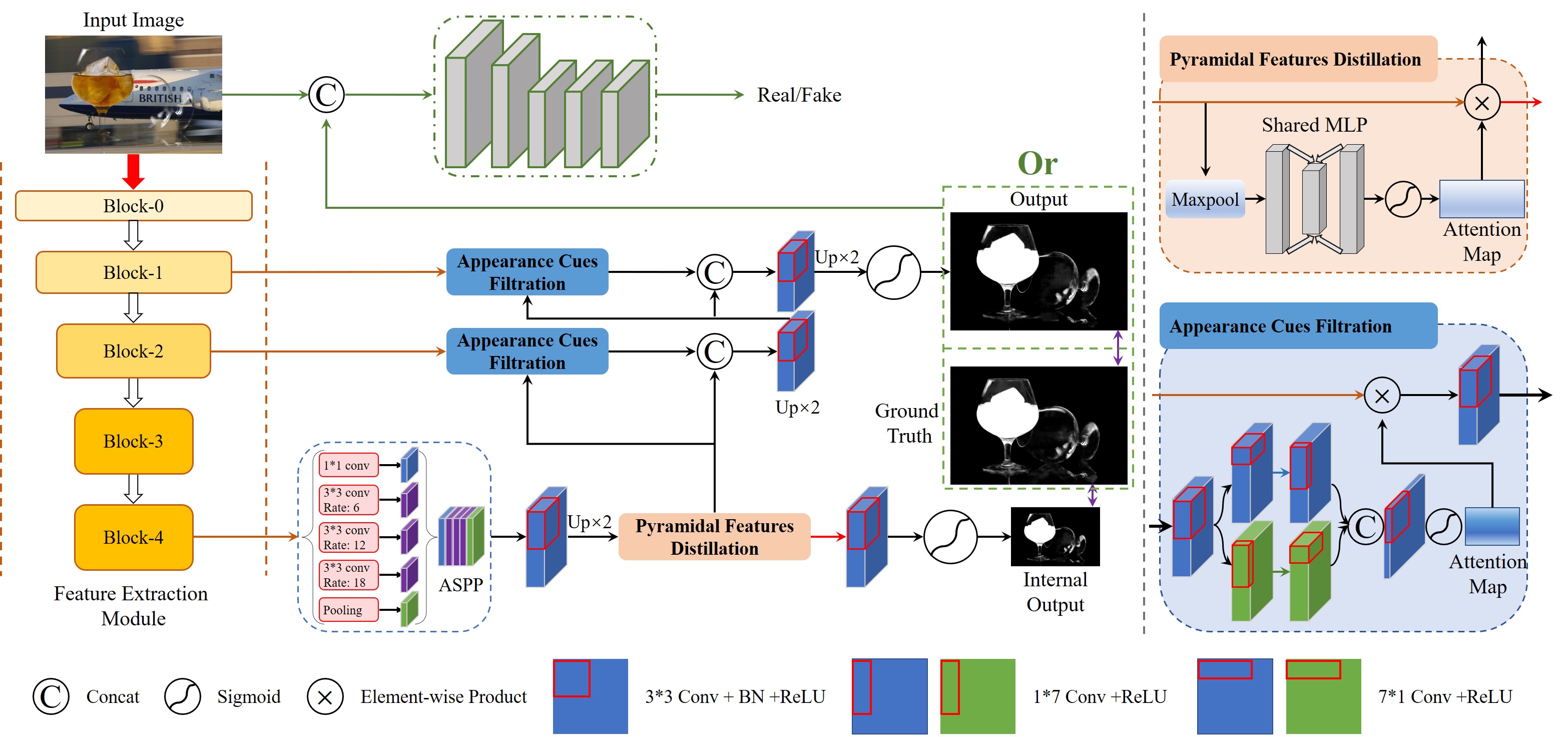}
	\caption{Pipeline of our \emph{HAttMatting++}. The orange box (Pyramidal Features Distillation) indicates channel-wise attention to distill pyramidal information extracted from \emph{ASPP}~\cite{Chen2018DeepLab}. The blue box (Appearance Cues Filtration) represents spatial attention to filter appearance cues, which are extracted from block1 and block2 in the feature extraction module.}
	\label{fig:pipeline}
\end{figure*}

A preliminary version of this work was presented earlier in~\cite{Qiao_2020_CVPR}. In this paper, we basically expand our previous work and summarize the main differences as follows:

\textbf{Progressive structure aggregation for multi-scale appearance cues exploration.} As shown in Fig.~\ref{fig:pipeline}, instead of directly extracting  appearance cues from block1 as in~\cite{Qiao_2020_CVPR}, we employ a top-down spatial attention mechanism to aggregate pyramidal features with progressive appearance cues. Specifically, we consider the low-level feature maps from block2 in ResNeXt~\cite{Xie2017Aggregated} as secondary appearance cues, and they are attended by our proposed appearance cues filtration model, then aggregated with pyramidal features. We argue that such secondary appearance cues can be used as a complement to improve some middle-level image texture or details (the leaves, hands, etc).

\textbf{Sentry supervision for advanced semantics enhancement.} We import sentry supervision after the pyramidal features distillation module. As we advocated in our previous conference paper, the aggregation between matting-adapted semantics and appearance cues can be primarily devoted to the alpha matte generation. The advanced matting-adapted semantics play a significant role in the structure aggregation process since the appearance cues filtration module also requires them to suppress redundant \emph{BG} information. Hence, we add sentry supervision after the pyramidal features as the auxiliary guidance to enhance adaptive semantics. On the one hand, it can further improve pyramidal features distillation and select matting-adapted semantics. On the other hand, it can provide better guidance for subsequent spatial attention and final aggregation. Additionally, the whole training procedure and convergence progress can be optimized due to the backward propagation of the sentry supervision.

\textbf{More extensive experiments for further analysis and evaluation.} Compared to our previous conference paper, we conduct more experiments to demonstrate our model. Comprehensive ablation study experiments are also performed to compare the potential variation of the network architecture and prove the effectiveness of different components on the baseline network. Besides, we evaluate the proposed \emph{HAttMatting++} on more real-world images, and the high-quality alpha mattes can prove the versatility of our model.

\section{Related work}
\label{sec:related}
Deep learning brings a major evolution for natural image matting with a high representation of \emph{FG} structure, and we briefly review image matting from two categories: traditional and deep-learning methods.

\textbf{Traditional matting.} Existing matting methods mostly achieve \emph{FG} opacity by virtue of additional input: trimap or scribbles. The trimap is composed of \emph{FG}, \emph{BG}, and transition to partition the input image, while the scribbles indicate these three category regions by several user-specified scribbles. The transition region signifies \emph{FG} boundaries, which is the key point for image matting. Although scribbles approaches~\cite{Levin2007A,Levin2008Spectral,Guan2006Easy} are convenient for novice users, they significantly deteriorate the alpha matte since there is insufficient information that can be referenced. Therefore, a majority of methods exploit trimaps as essential assistance to perceiving \emph{FG} structure.

Traditional matting methods primarily resort to color features extracted from the input image to confine transition regions. According to the different ways of using color features, they can be roughly divided into two categories: sampling-based and affinity-based methods. Sampling-based methods~\cite{Karacan2015Image,Rhemann2011A,Shahrian2013Improving,Wang2007Optimized} solve alpha mattes by representing each pixel inside transition regions with a pair of certain \emph{FG/BG} pixels. According to the local smoothness assumption on the image statistics, most of these methods require that these sampling colors are close to a certain foreground or background. Affinity-based methods~\cite{Levin2007A,Levin2008Spectral,Aksoy2017Designing,Chen2013KNN,Lee2011Nonlocal} utilize the affinities of neighboring pixels between certain labels and transition regions to perceive \emph{FG} boundaries, which often suffer from high computational complexity. One of the most representative methods is Closed-form Matting~\cite{Levin2007A}, which is derived from the matting Laplacian and can produce alpha matte by solving a sparse linear system of equations.

Both sampling and affinity methods primarily harness color similarity among pixels to predict alpha matte, incapable of describing the advanced structure of \emph{FG}. As a result, they usually produce obvious artifacts when \emph{FG} and \emph{BG} share similar colors.


\textbf{Deep-learning matting.} Compared with traditional methods that use low-level features, learning-based methods have refreshed all the previous state-of-the-art records in image and video matting. Similar to other computer vision tasks, matting objects also process a general structure that can be represented by high-level semantic features. Shen~\et~\cite{Shen2016Deep} proposed the first automatic matting system for portrait photos in an end-to-end fashion. Cho~\et~\cite{cho2019deep} combined the results of \cite{Levin2007A} and \cite{Chen2013KNN} with normalized RGB images as input, and learn an end-to-end deep network to predict better alpha mattes. Xu~\et~\cite{Xu2017Deep} proposed deep image matting (\emph{DIM}), which integrated \emph{RGB} images with trimaps as conjunct input, utilizing advanced semantics to estimate alpha mattes. Lutz~\et~\cite{lutz2018alphagan} explored generative adversarial networks (GAN)~\cite{Goodfellow2014Generative} to generate alpha mattes with pleasing visual effects, which demonstrated the effectiveness of the discriminator model for pixel-wise matting perception. Tang~\et~\cite{Tang_2019_CVPR} proposed a hybrid sampling and learning-based approach to image matting. Cai~\et~\cite{cai2019disentangled} and Hou~\et~\cite{hou2019context} both established two branches to perceive alpha mattes and these two branches mutually reinforced each other to refine the final results. Hao~\et~\cite{hao2019indexnet} and Dai~\cite{dai2021learning} unified upsampling operators with the index function to improve the encoder-decoder network. Inspired by image inpainting~\cite{yu2018generative}, Li~\et~\cite{li2020natural} utilize guided contextual attention to improve the transmission of opacity information. Yu~\et~\cite{Yu2021AAAI} explore patch-based crop and stitch to implement the context dependency in high-resolution matting. Though the trimap expansions in SIM~\cite{sun2021semantic} can extend the range of transition capture, they have a restriction on alpha categories. However, all these matting methods rely on trimap as additional input to enhance their semantic distillation, while producing trimap is difficult for common users.

Some matting frameworks \cite{Chen2018SHM,Cho2016Automatic,yu2021mask} resort to semantic segmentation to generate pseudo trimap, which requires intermediate masks and sometimes can cause \emph{FG} profiles or boundaries incomplete. Wei~\et~\cite{wei2021improved}, Yang~\et~\cite{Yang2018Active}, and Yang~\et~\cite{Yang2020Smart} sacrificed user interactions and extra feedback time to improve alpha mattes. While the combination of the multi-scale features in~\cite{sss} can generate alpha mattes automatically, it has a very slow execution. Zhang~\et~\cite{Zhang2019CVPR} investigated semantic segmentation variants for \emph{FG} and \emph{BG} weight map fusion to obtain alpha mattes. Although they can produce competent results without trimaps, failure cases occur when segmentation is inapplicable. Qiao~\et~\cite{Qiao2020multi} exploit information integration to regress alpha mattes. Sengupta~\et~\cite{sengupta2020background} and Lin~\et~\cite{lin2021real} employ additional \emph{BG} images to predict alpha mattes, which must share the environment as default. Trimap-free methods are similar in nature to saliency-based methods~\cite{tian2020weakly,tian2021learning,tian2022bi}. Compared to existing matting networks, we perform a profound excavation of the matting information in the input image, and inspired by the recent success of attention mechanism and contextual correlation~\cite{mei2020don,mei2021exploring}, we propose a hierarchical and progressive aggregation framework to integrate adaptive semantics and refined appearance cues. These improvements enable us to produce high-quality alpha mattes without trimaps.

\section{Methodology}
\label{sec:methodology}
\subsection{Overview}
\label{ssec:overview}
We can conclude from Eq.~\ref{image_synthesis} that the complete object \emph{FG} should consist of two parts: 1) the main body indicating \emph{FG} category and profiles ($\alpha_z=1$), and 2) the internal texture and boundary details located in the transition region ($\alpha_{z} \in (0,1)$). The Former can be obtained by advanced semantic information, while the latter usually comes from input images or low-level CNN features, termed appearance cues, and they can be combined to achieve alpha mattes. Although the low-level features and advanced semantic information can complement each other, their vanilla combination may produce some interference in alpha mattes. On the one hand, the low-level features usually contain texture details outside \emph{FG}. On the other hand, the highly-abstracted advanced semantics typically contain global information and may be too sensitive to object categories, which means they contribute unequally to alpha mattes that mostly consider \emph{FG} context. Consequently, we argue that the advanced semantics and appearance cues need essential processing before combination. First, natural image matting is supposed to handle different types of \emph{FG} objects, which suggests that we should distill advanced semantics to attend to \emph{FG} information, and appropriately suppress them to reduce their sensitivity to object class. Second, as shown in Fig.~\ref{fig:feature_maps}, appearance cues involve unnecessary \emph{BG} details, which need to be erased in alpha mattes.

Based on these observations, the core idea of our approach is to select matting-adapted semantic information and eliminate irrelevant \emph{BG} texture in appearance cues, then aggregate them to predict alpha mattes. For this purpose, we adopt channel-wise attention to distill advanced semantics extracted from Atrous Spatial Pyramid Pooling (\emph{ASPP})~\cite{Chen2018DeepLab}, and perform spatial attention on different-levels appearance cues to eliminate image texture details outside \emph{FG} simultaneously. Our well-designed hierarchical and progressive attention mechanism can perceive \emph{FG} structure from adaptive semantics and refined boundaries, and their combination can achieve better results. Moreover, we design a hybrid loss to further guide the network training by combining Mean Square Error (\emph{MSE}), Structural SIMilarity (\emph{SSIM}), and Adversial loss~\cite{Goodfellow2014Generative}, which are respectively responsible for pixel-wise precision, structure consistency, and visual quality. We also import sentry supervision after the pyramidal features distillation to improve adaptive semantics.

\subsection{Network Architecture}
\label{ssec:network atchitecture}
\subsubsection{Overall Network Design} 
The pipeline of our proposed \emph{HAttMatting++} is unfolded in Fig.~\ref{fig:pipeline}. We harness ResNeXt\cite{Xie2017Aggregated} as the backbone in consideration of their powerful ability to extract high-level semantic information. In order to obtain a larger receptive field while retaining high-resolution feature maps, we remove the max pooling layers, set the stride of block1 and block4 as 1, and employ dilation convolutions in block4. The advanced feature maps from block4 are then fed to \emph{ASPP}~\cite{Chen2018DeepLab} module for multi-level semantics capture. 

\begin{figure}[htbp]
\centering
\arrayrulecolor{tabcolor}
\setlength{\tabcolsep}{3pt}\small{
	\begin{tabular}{c|c|c}
		\includegraphics[scale=0.0896]{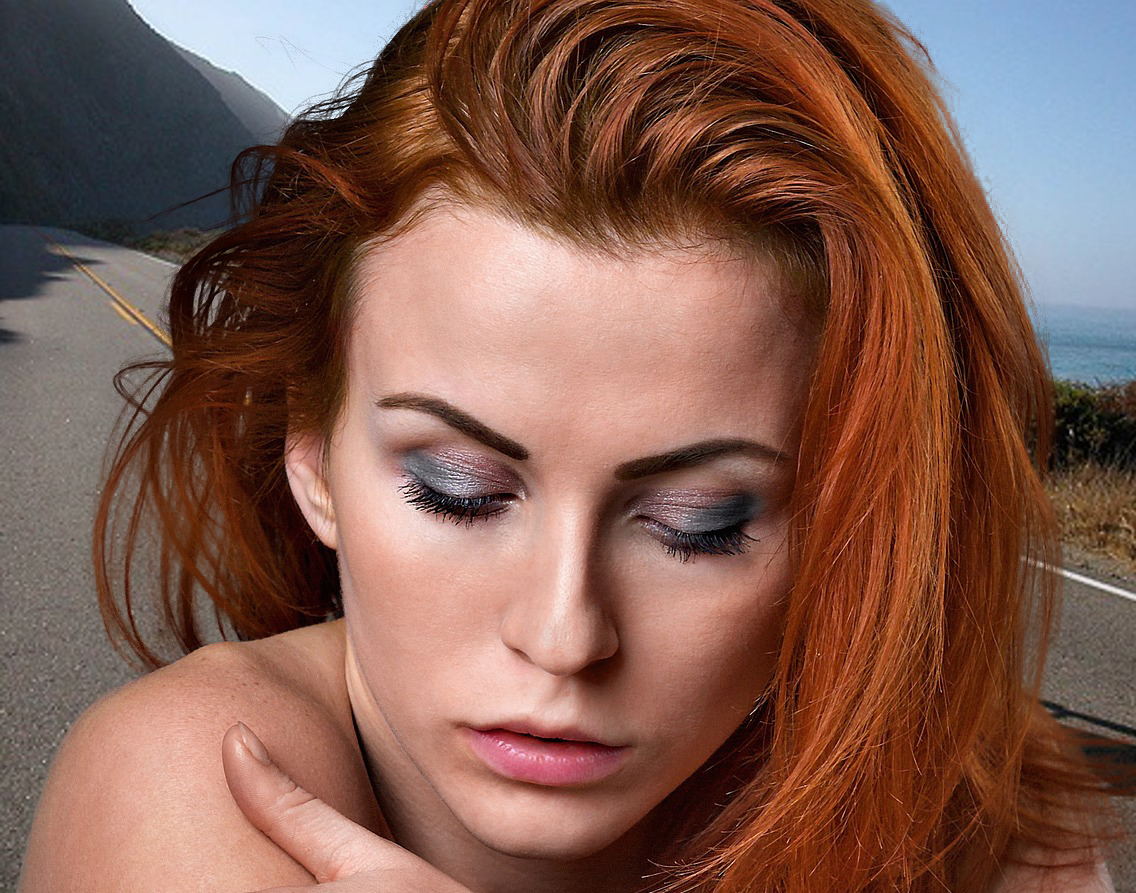} &
		\includegraphics[scale=0.0625]{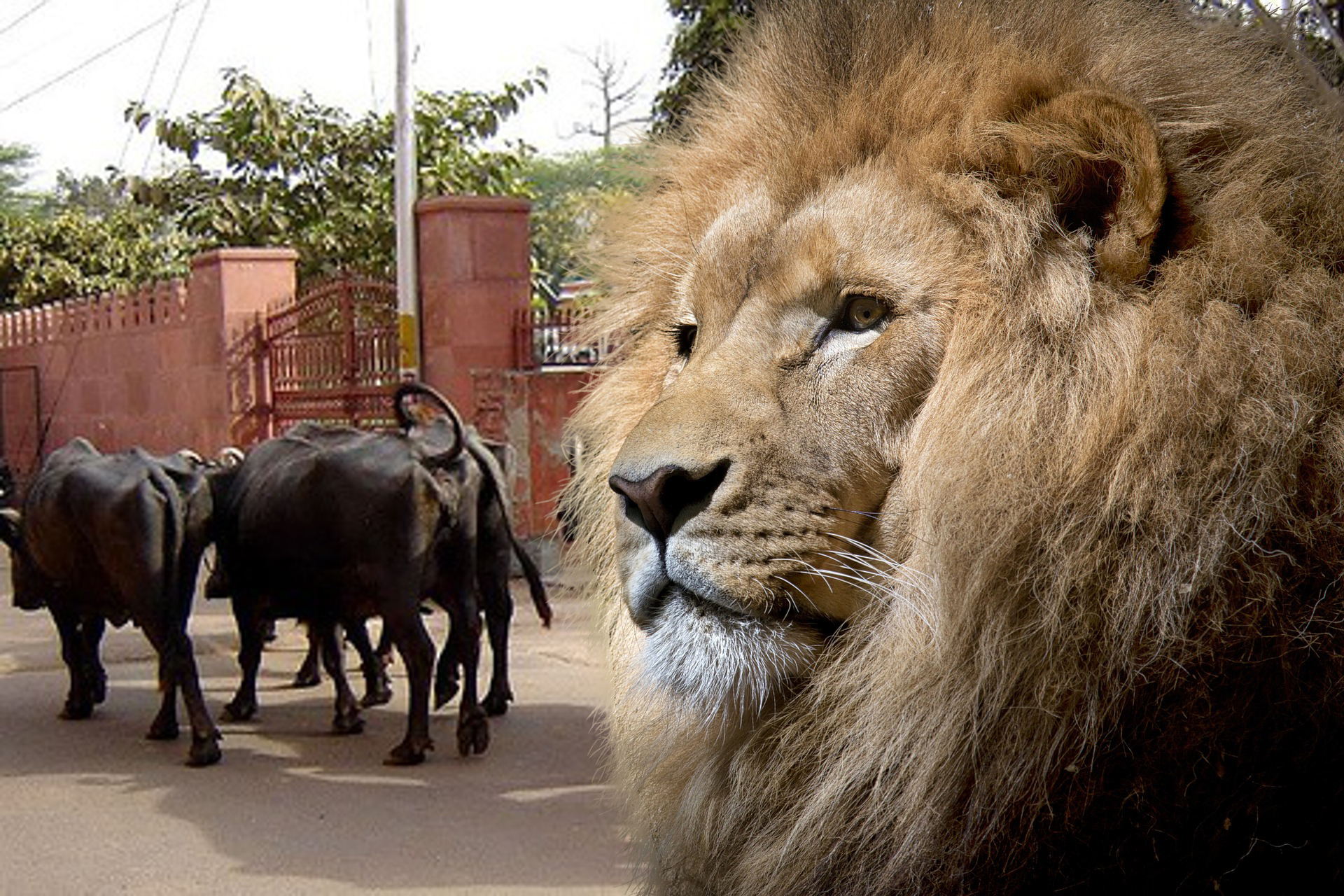} &
		\includegraphics[scale=0.0625]{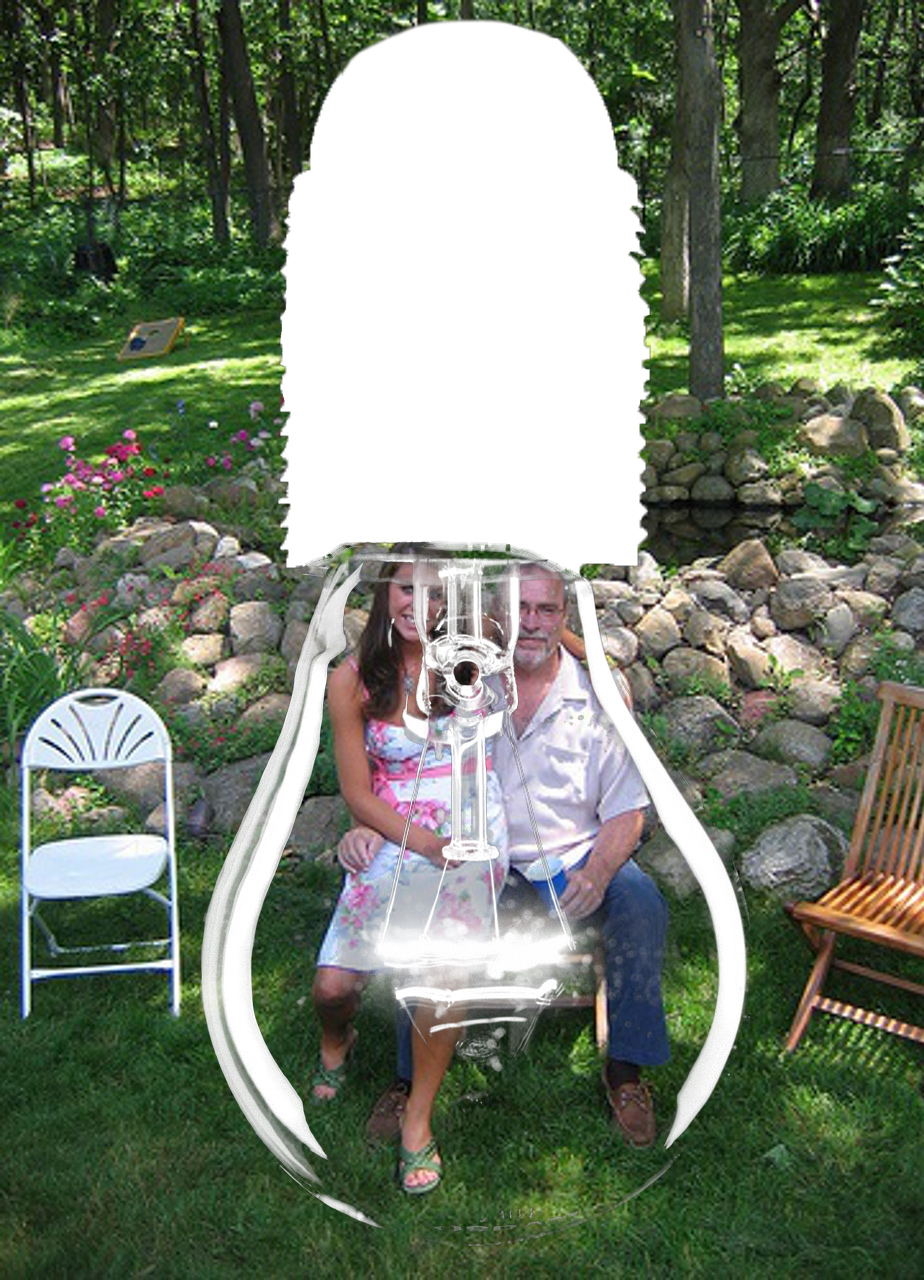} \\
		
		\includegraphics[scale=0.1792]{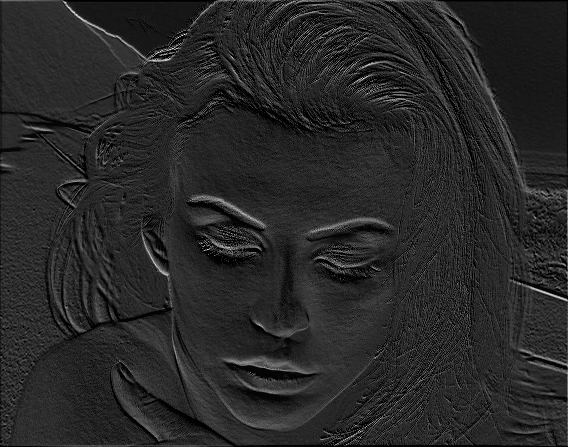} &
		\includegraphics[scale=0.125]{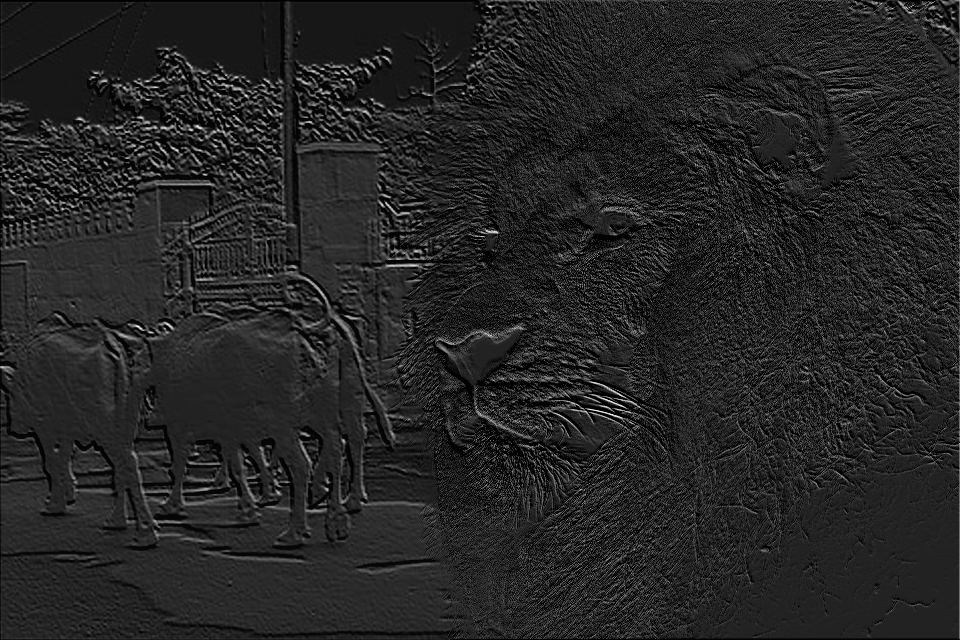} &
		\includegraphics[scale=0.125]{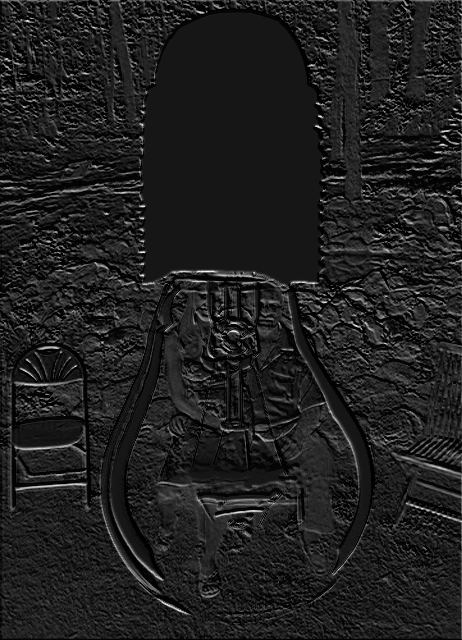} \\
\end{tabular}}
\caption{The input images and corresponding appearance cues.}
\label{fig:feature_maps}
\end{figure}

There are five blocks in total in our feature extraction network, and we regard the first three layers (block0 to block2) as shallow layers and the rest (block3 and block4) as deep layers. The low-level features at shallow layers include more detailed information (see Fig.~\ref{fig:feature_maps}), such as texture, boundary, and internal structure, but they also contain more noises outside \emph{FG}. By contrast, high-level features at deep layers can provide abstract semantic information (see Fig.~\ref{fig:pipeline}), which is beneficial to locating most parts of FG and suppressing the noises.

Different from previous works~\cite{cai2019disentangled, Zhang2019CVPR} that simply fuse the upsampled deep-layer features with the shallow-layer features by concatenation or element-wise summation operation, we adopt a more aggressive yet efficient form called attention-guided hierarchical and progressive structure aggregation. As shown in Fig.~\ref{fig:pipeline}, for the high-level features at deep layers, we apply a pyramidal features distillation module to select matting-adapted semantics and locate the \emph{FG} regions with strong semantic responses. For the low-level features at shallow layers, we present an appearance cues filtration module to refine the boundaries of \emph{FG} and restrain the noises in \emph{BG}. Then, we combine the distilled pyramidal features with the filtered progressive appearance cues in a top-down manner, and these hierarchical features can jointly contribute to the resolution of alpha mattes. Besides, we utilize the discriminator network refer to PatchGAN~\cite{CycleGAN2017,isola2017image} to enhance the visual quality of alpha mattes.

\subsubsection{Pyramidal Features Distillation}
\label{sssec:CA}
The extracted pyramidal features devote unequally to \emph{FG} structure regression, hence we perform channel-wise attention on pyramidal features to distill adaptive semantic attributes. As the orange box is shown in Fig.~\ref{fig:pipeline}, we present the pyramidal features after upsampling as $\boldsymbol{F}^{i}_{p} \in \mathbb{R}^{C \times H/4 \times W/4}$, and here $H$ and $W$ are equal to the height and width of the input image. Similarly, the attended pyramidal features are $\boldsymbol{F}^{o}_{p} \in \mathbb{R}^{C \times H/4 \times W/4}$. First, the upsampled $\boldsymbol{F}^{i}_{p} \in \mathbb{R}^{C \times H/4 \times W/4}$ are feed to our pyramidal features distillation module, and then we utilize global pooling to generalize the feature maps for aggregating spatial information. Subsequently, a shared multi-layer perception (MLP) which consists of two consecutive fully connected (FC) layers is employed to further distill semantic attributes. Then, through using sigmoid operation, the range of the final channel-wise attention maps were mapped to $[0,1]$. The above process can be described as:
\begin{equation}
\label{equ:eq_CA}
\boldsymbol{F}^{o}_{p}
=\sigma\left(MLP\left(Pool\left({\boldsymbol{F}^{i}_{p}}, w_{c}^{0}\right)\right), w_{c}^{1}\right) \otimes \boldsymbol{F}^{i}_{p}
\end{equation}
where $w_{c}$ refers to the parameters in our channel-wise attention block, $\sigma$ refers to the sigmoid operation and the $Pool$ represents the global pooling operation. The channel-wise attention can select pyramidal features adapted to image matting, and retain \emph{FG} profiles and category attributes. The pyramidal features are learned from deep ResNext block, which are highly abstract semantic information, thus we need appearance cues to restore \emph{FG} details in alpha mattes.

\subsubsection{Appearance Cues Filtration}
\label{sssec:SA}
Image matting requests precise \emph{FG} boundaries, while high-level pyramidal features are incapable of catching such texture details. Therefor, we bridge a skip connection between the appearance cues and the advanced pyramidal semantics, which can transport low-level structure information for alpha matte generation. The representative feature maps extracted from block1 are illustrated in the second row of Fig.~\ref{fig:feature_maps}, we take these features as our initial appearance cues. These appearance cues can depict sophisticated image texture, compatible with the boundary accuracy required by alpha matte perception. 

The proposed \emph{HAttMatting++} can leverage appearance cues to enhance \emph{FG} boundaries in the results. Correspondingly, we import spatial attention to filter appearance cues located in \emph{BG} and emphasize the ones inside \emph{FG} simultaneously. Specifically, we use kernel size 1 * 7 and 7 * 1 to execute horizontal and vertical direction attention respectively. The blue box in Fig.~\ref{fig:pipeline} shows our spatial attention. The input of the black arrow indicates the guidance information from the previous stage, which can be depicted as $\boldsymbol{F}_{gda} \in \mathbb{R}^{C \times h \times w}$ ($h$ and $w$ correspond to different resolutions on different-levels appearance cues). They are further disposed of via two parallel convolutions with the above two filter kernel. Their combination are then concatenated and convolved by a standard convolution layer, followed by a sigmoid layer, producing our 2D spatial attention map. This spatial attention map was utilized to handle the appearance cues ($\boldsymbol{F}^{i}_{a} \in \mathbb{R}^{C \times h \times w}$) by the operation of element-wise product, to remove the textures and details that belong to \emph{BG}. The above process can be described as:
\begin{equation}
\label{eq:s1}
\left.S_{1}=f^{7\times 1}\left(f^{1\times 7}\left(f^{3\times 3}\left(\boldsymbol{F}_{gda}, w_{S}^{0}\right), w_{S}^{h1}\right), w_{S}^{v1}\right)\right)
\end{equation}
\begin{equation}
\label{eq:s2}
\left.S_{2}=f^{1\times 7}\left(f^{7\times 1}\left(f^{3\times 3}\left(\boldsymbol{F}_{gda}, w_{S}^{0}\right), w_{S}^{v2}\right), w_{S}^{h2}\right)\right)
\end{equation}
\begin{equation}
\label{equ:eq_SA}
\boldsymbol{F}^{o}_{a}=\sigma \left(f^{1 \times 1} \left( Cat\left(S_{1}, S_{2}\right) \right) ,w_{S}^{1}\right) \otimes \boldsymbol{F}^{i}_{a}
\end{equation}
where $\sigma$ denotes the sigmoid function and $f^{n \times m}$ represents a convolution operation with the filter size of $n \times m$ and $w_{S}$ represents the parameters in our spatial attention. $\boldsymbol{F}^{o}_{a} \in \mathbb{R}^{C \times h \times w}$ is the output of our appearance cues filtration module. Despite the appearance cues involve sufficient image texture, only the regions inside or surrounding \emph{FG} can contribute to alpha mattes. Therefor, we use distilled pyramidal features as the guidance, combined with low-level information at different levels to jointly improve alpha mattes.

There are many researches~\cite{woo2018cbam,Long2016SCA} that combine spatial and channel-wise attention, and we have three main differences based compared to them. (1) We use the high-level features as spatial attention maps to filter low-level features, which can suppress background appearance cues via attended foreground semantics. (2) The channel-wise attention is only imported to distill the pyramidal features, and we use MaxPool to capture the most representative features. Thus, the dominated semantics can be extracted and enhanced for the FG profiles. (3) The attention mechanism is employed in the multiple decoder phases, guiding the encoder features progressively.

\subsubsection{Top-down Progressive Features Aggregation} 
Since the feature maps of block0 are originally derived from the input images, which are highly crude for alpha perception, we ignore it and just consider the low-level features from block1 and block2 as our appearance cues (denoted as initial appearance cues and secondary appearance cues separately in the following). First, we perform channel-wise attention on the advanced pyramidal features extracted from ASPP~\cite{Chen2018DeepLab} to select adaptive semantics. Then, we consider the distilled semantics as guidance and execute the first-stage spatial attention on secondary appearance cues (the $h$ and $w$ in Eq.~(\ref{equ:eq_SA}) are $H/4$ and $W/4$ separately). The filtered secondary appearance cues and the distilled semantics are aggregated as the subsequent guidance. Then we carry out another spatial attention on initial appearance cues to further remove redundant \emph{BG} information (here the $h$ and $w$ in Eq.~(\ref{equ:eq_SA}) are $H/2$ and $W/2$ separately). After this, we perform the final hierarchical features aggregation, and the alpha mattes can be generated by another upsampling operation. It is worth noting that even if the second spatial attention does not explicitly use the distilled pyramidal features as guidance, it has been implicitly included in the result of the first aggregation. Additionally, we add sentry supervision after pyramidal features distillation module to enhance matting-adapted semantics. It can also provide better guidance for appearance cues filtration module and promote the convergence of the spatial attention during backward propagation.

The channel-wise and spatial attention can jointly optimize the alpha matte generation: one responsible for pyramidal features selection and the other responsible for appearance cues filtration. This well-designed top-down hierarchical attention mechanism can efficiently attend low-level features and advanced semantics, and their progressive aggregation can produce high-quality alpha mattes with fine-grained details.

\subsection{Loss Function}
\label{sec:loss function}
Pixel regression related loss funtion ($\mathcal{L}{1}$ or \emph{MSE} loss) are usually adopted as the object function for alpha matte prediction~\cite{Xu2017Deep,cai2019disentangled}. They can generate competent alpha mattes via pixel-wise supervision. However, such regression loss only measures the difference in absolute pixel space, without consideration of \emph{FG} structure. Therefor, we introduce \emph{SSIM} loss ($\mathcal{L}_{ssim}$) to calculate structure similarity between the predicted alpha mattes and ground truth. Structural SIMilarity (\emph{SSIM})~\cite{Zhou2004Image} has demonstrated a striking ability to boost structure consistency in the predicted images~\cite{Qin_2019_CVPR,Wan2018CRRN}. Apart from the aforementioned loss function, we add adversarial loss ($\mathcal{L}_{adv}$) to promote the visual quality of the predicted alpha mattes. In the proposed \emph{HAttMatting++}, we employ this hybrid loss function to guide the network training, achieving effective alpha matte optimization. Our loss function is defined as follows:
\begin{equation}
\label{eq:total_loss}
\mathcal{L}_{total} = \lambda_{1}\mathcal{L}_{adv} + \lambda_{2}\mathcal{L}
_{MSE} + \lambda_{3}\mathcal{L}_{SSIM} + \lambda_{4}\mathcal{L}_{sentry}, 
\end{equation}
$\mathcal{L}_{adv}, \mathcal{L}_{MSE} \text{ and } \mathcal{L}_{SSIM}$ can improve alpha mattes from visual quality, pixel-wise accuracy and \emph{FG} structure similarity separately. While the $\mathcal{L}_{sentry}$ refer to the sentry supervision after the pyramidal features distillation module. And it was designed to improve the adaptive semantics and promote the convergence of the spatial attention. $\lambda_{1}, \lambda_{2}, \lambda_{3} \text{ and } \lambda_{4}$ represent balance coefficients for loss function. $\mathcal{L}_{adv}$ is defined as:
\begin{equation}
\label{eq:adv_loss}
\mathcal{L}_{adv} = E_(I,A)[log(D(I,A)+log(1-D(I,G(I))))],
\end{equation}
where \emph{I} represents the input image and \emph{A} is the predicted alpha matte. $\mathcal{L}_{MSE}$ is expressed as:
\begin{equation}
\label{eq:mse_loss}
\mathcal{L}_{MSE} = \frac{1}{|\Omega|}\sum_{i}^{\Omega}(\alpha_{p}^{i} - 
\alpha_{g}^{i})^{2},\qquad \alpha_{p}^{i},\alpha_{g}^{i} \in[0,1],
\end{equation}
where $\Omega$ represents pixels set and $|\Omega|$ is the number of pixels(\ie the size of the input image). $\alpha_{p}^{i}$ and $\alpha_{g}^{i}$ are the predicted and ground truth alpha values at pixel i respectively. $\mathcal{L}_{MSE}$ can ensure the pixel-wise accuracy of alpha matte estimation. We establish \emph{FG} structure optimization via $\mathcal{L}_{SSIM}$ as:
\begin{equation}
\label{eq:ssim_loss}
\mathcal{L}_{SSIM} = 1 - \frac{(2\mu_{p}\mu_{g} + c_{1})(2\sigma_{pg} + c_{2})}
{(\mu_{p}^{2} + \mu_{g}^{2} + c_{1})(\sigma_{p}^{2} + \sigma_{g}^{2} + c_{2})}.
\end{equation}
here $\mu_{p}$, $\mu_{g}$ and $\sigma_{p}$, $\sigma_{g}$ are the mean and standard deviations of $\alpha_{p}^{i}$ and $\alpha_{g}^{i}$. With $\mathcal{L}_{SSIM}$ as guidance, our method can further improve \emph{FG} structure. 
\begin{equation}
\label{eq:sentry_loss}
\mathcal{L}_{sentry} = \frac{1}{|\Omega_{\sim4}|}\sum_{i}^{\Omega_{\sim4}}(\alpha_{p}^{i} - 
\alpha_{g}^{i})^{2},\qquad \alpha_{p}^{i},\alpha_{g}^{i} \in[0,1],
\end{equation}
where $\Omega_{\sim4}$ represents a quarter of the full pixels set. The $\mathcal{L}_{sentry}$ is partly motivated by the side output~\cite{hou2017deeply}, we take the same loss function as $\mathcal{L}_{MSE}$ to supervise the pyramidal distillation module. The only difference is that the resolution of $\mathcal{L}_{sentry}$ is decreased by a factor of 4 than the original resolution of $\mathcal{L}_{MSE}$. $\mathcal{L}_{sentry}$ can assist the spatial attention to restrict the noises out of \emph{BG} and further improve the pixel-wise accuracy of the final alpha matte.

\subsection{Implementation Details}
\label{ssec:implementation details}
We implement \emph{HAttMatting++} using PyTorch. During the training process, for the data augmentation, we crop the image and GT pairs centered on random pixels in the transition regions indicated by trimaps, and then they are randomly cropped to $512\times512$ and $640\times640$, and $800\times800$. Then, they were resized to a resolution of $512\times512$ and augmented by horizontally random flipping. In order to accelerate the training process and prevent over-fitting, we use the pre-trained ResNeXt-101 network~\cite{Xie2017Aggregated} as the feature extraction backbone, while the other layers are randomly initialized from a Gaussian distribution. For loss optimization, we use the stochastic gradient descent(\emph{SGD}) optimizer with the momentum of 0.9 and a weight of 0.007, adjusted by the ``poly`` policy\cite{Wei2015ParseNet} with the power of 0.9 for 20 epochs. The balance coefficients $\lambda_{1}$, $\lambda_{2}$ , $\lambda_{3}$, and $\lambda_{4}$ in Eq.~\ref{eq:total_loss} are 0.05, 1, 0.1, and 0.3 during the first epoch and are revised as 0.05, 1, 0.025 and 0.1 for the subsequent 19 epochs. Our \emph{HAttMatting++} is trained on 3 joint Tesla V100 graphics cards with a mini-batch size of 4, and it takes about 24 hours for the network to finish 20 epochs.

\section{Experiments and Analyses}
\label{sec:experiments}
In this section, we evaluate \emph{HAttMatting++} on two datasets: the public Adobe Composition-1K~\cite{Xu2017Deep} and our Distinctions-646. We first compare \emph{HAttMatting++} with state-of-the-art methods both quantitatively and qualitatively. Then we perform ablation study for \emph{HAttMatting++} and the previous version \emph{HAttMatting}~\cite{Qiao_2020_CVPR} on both datasets to demonstrate the significance of several crucial components. Finally, we execute \emph{HAttMatting++} on real scenarios to generate alpha mattes.

\begin{figure}[hb]
	\arrayrulecolor{tabcolor}
	\setlength{\tabcolsep}{2pt}\small{
		\begin{tabular}{cccc}
			\includegraphics[scale=0.0514]{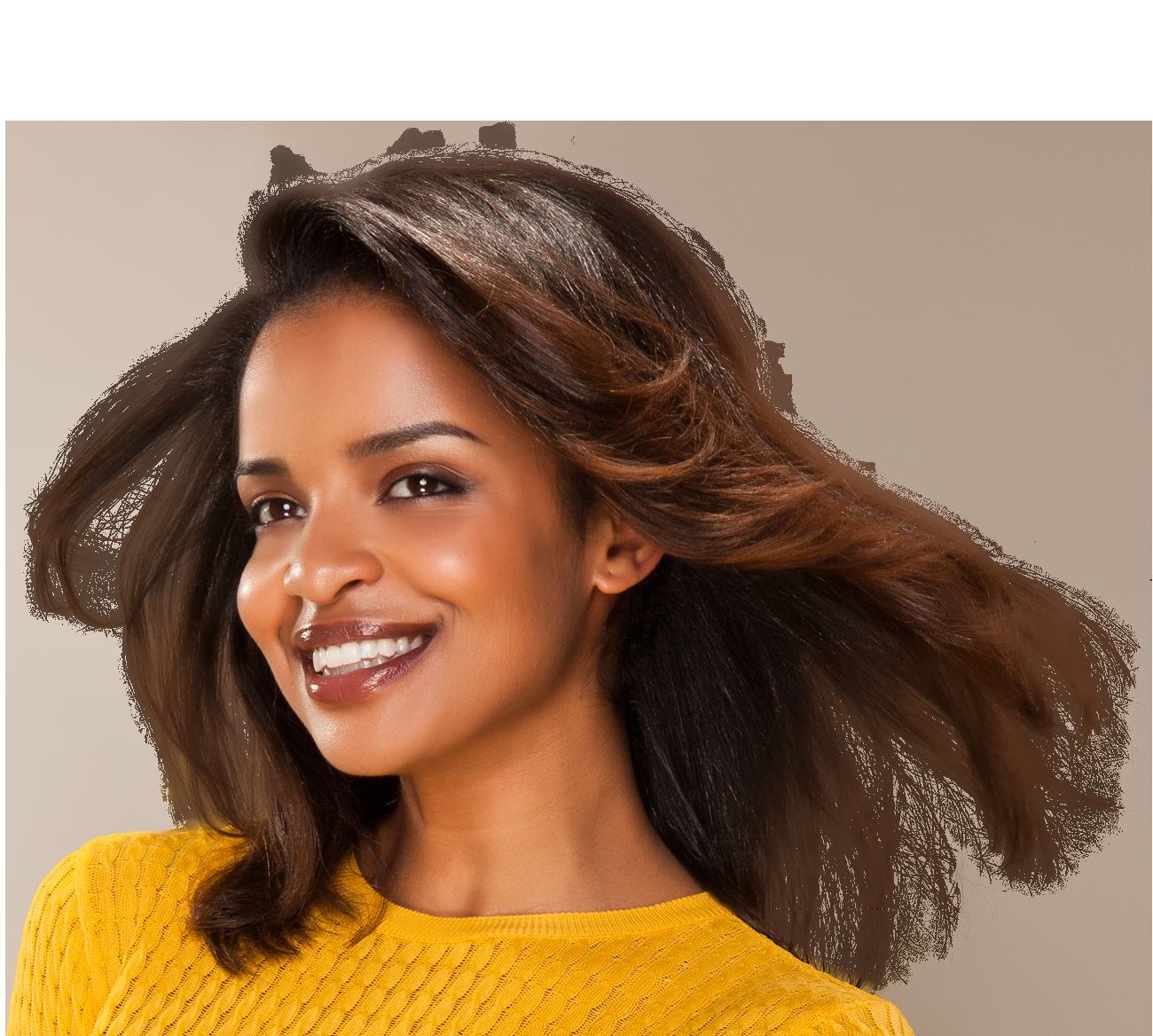}  &
			\includegraphics[scale=0.088]{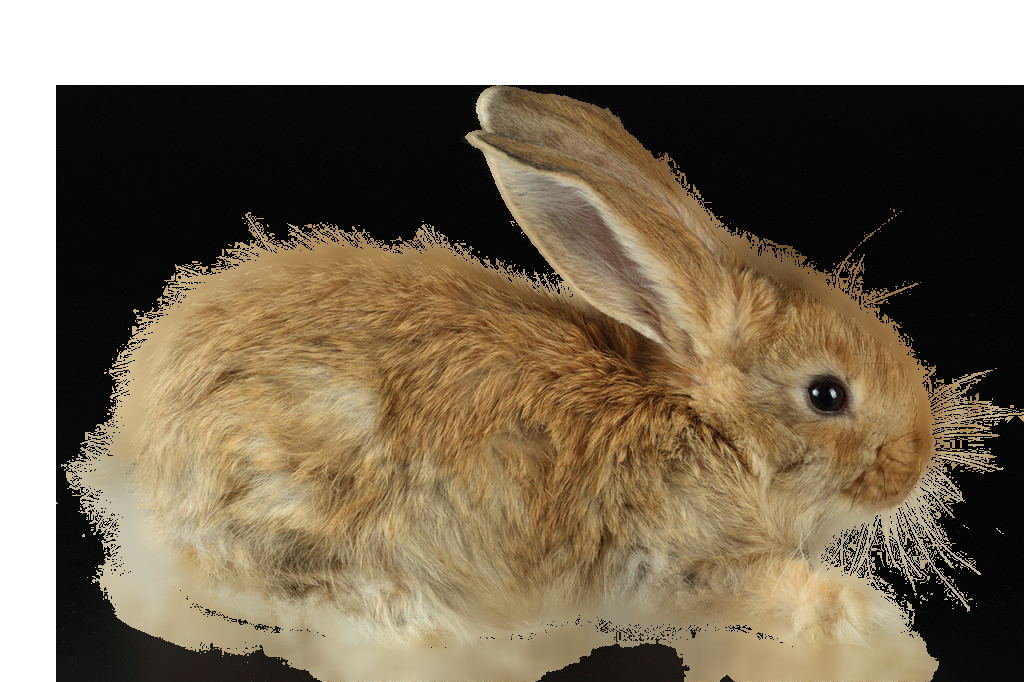} &
			\includegraphics[scale=0.0469]{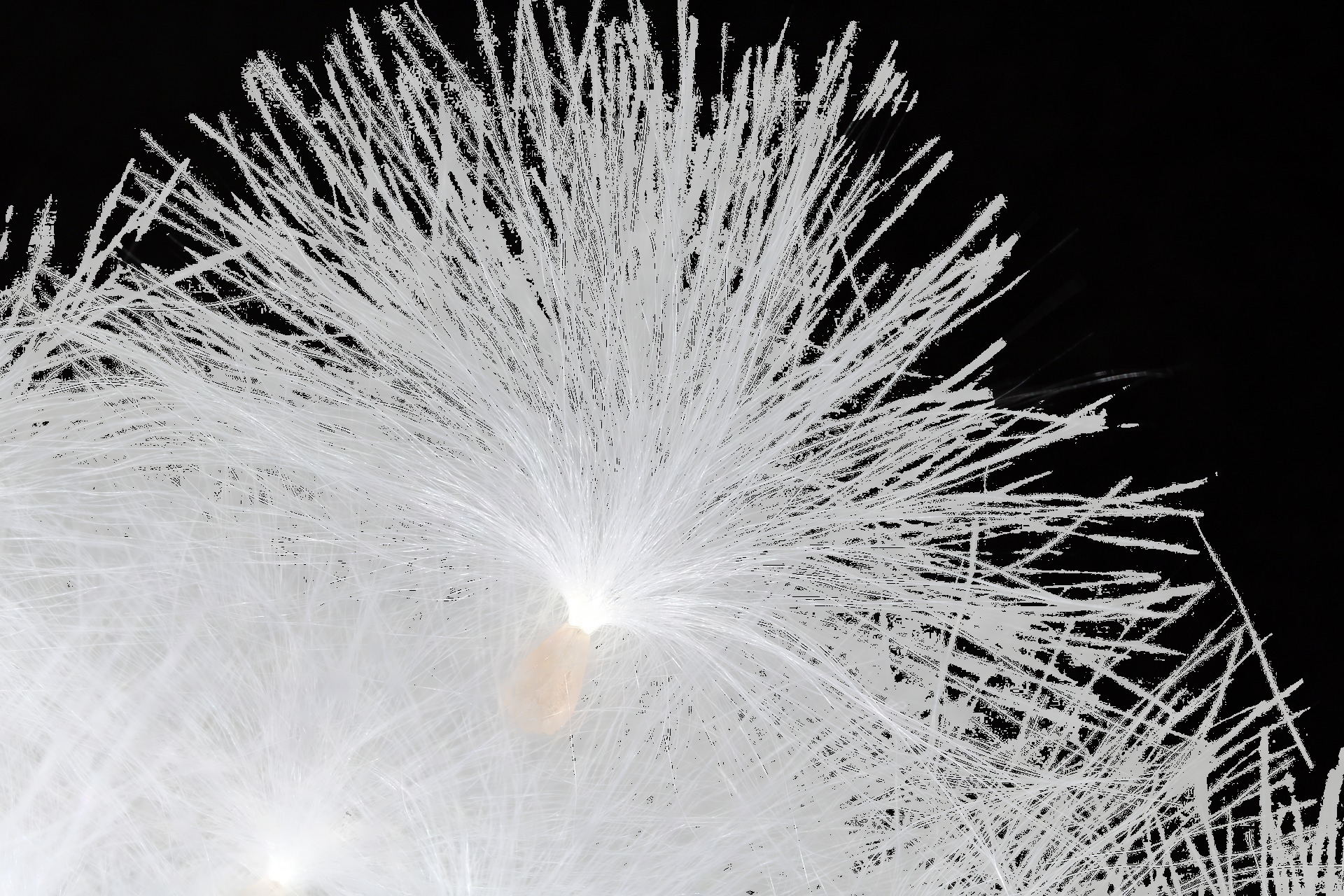} &
			\includegraphics[scale=0.0659]{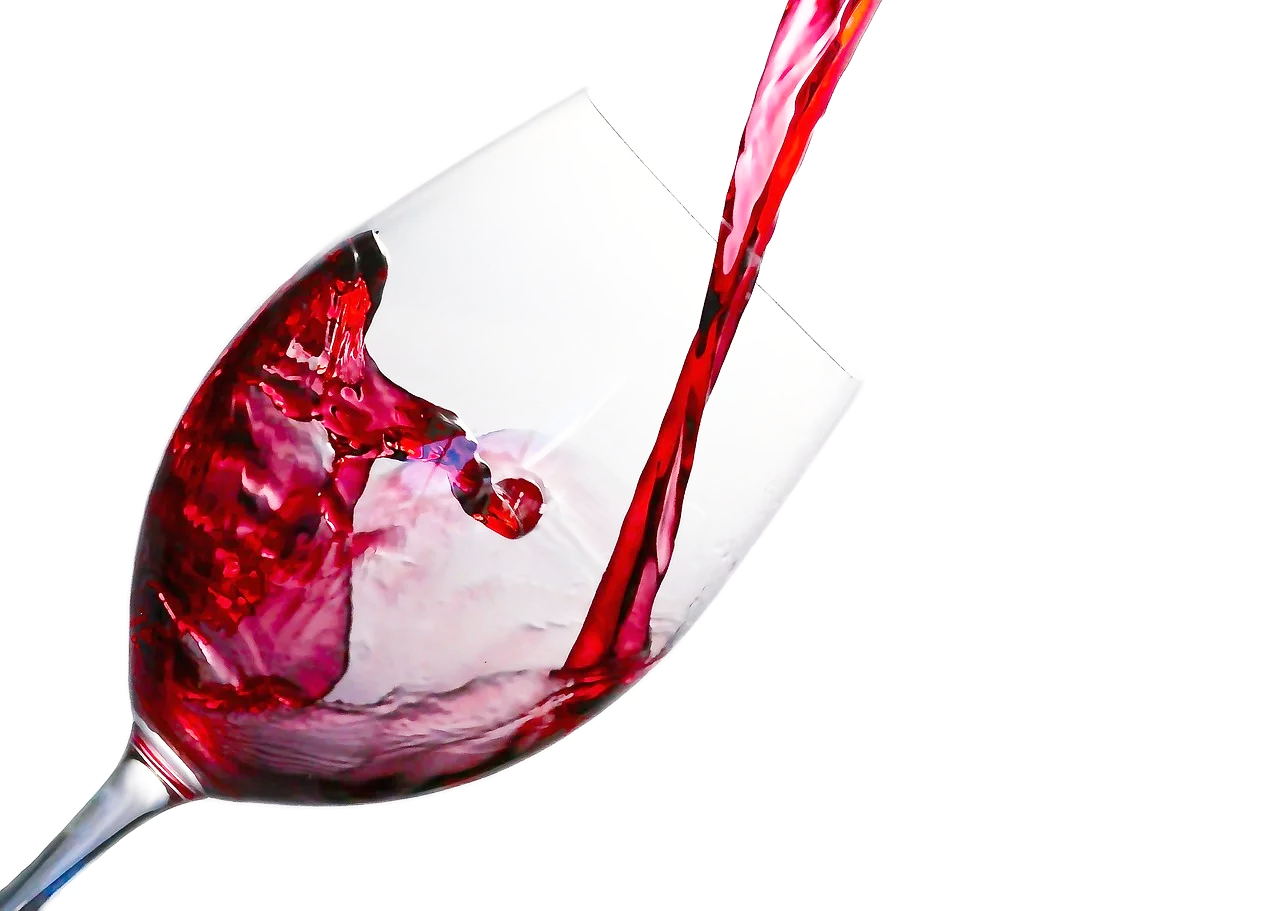} \\
			
			\includegraphics[scale=0.0514]{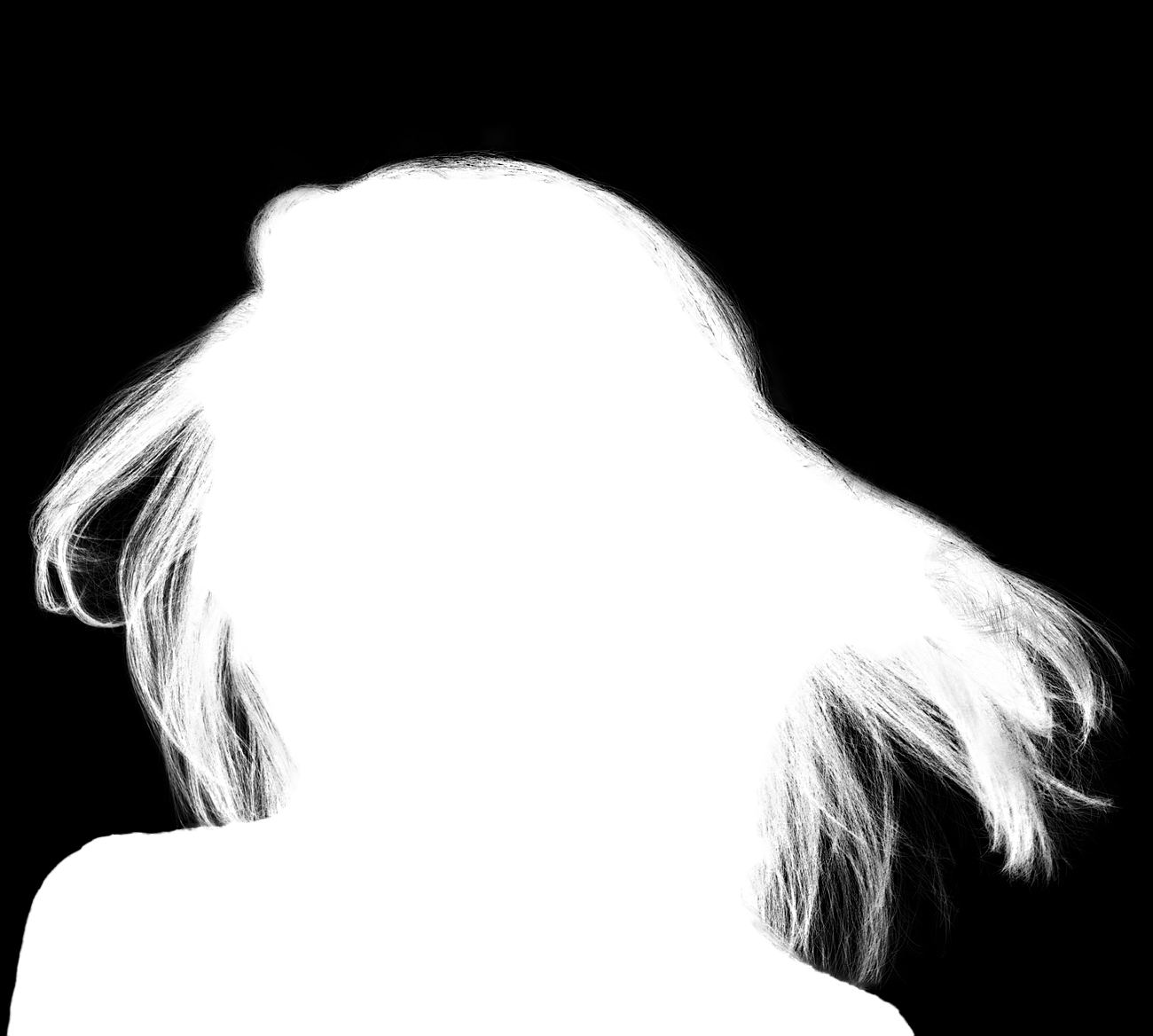} &
			\includegraphics[scale=0.088]{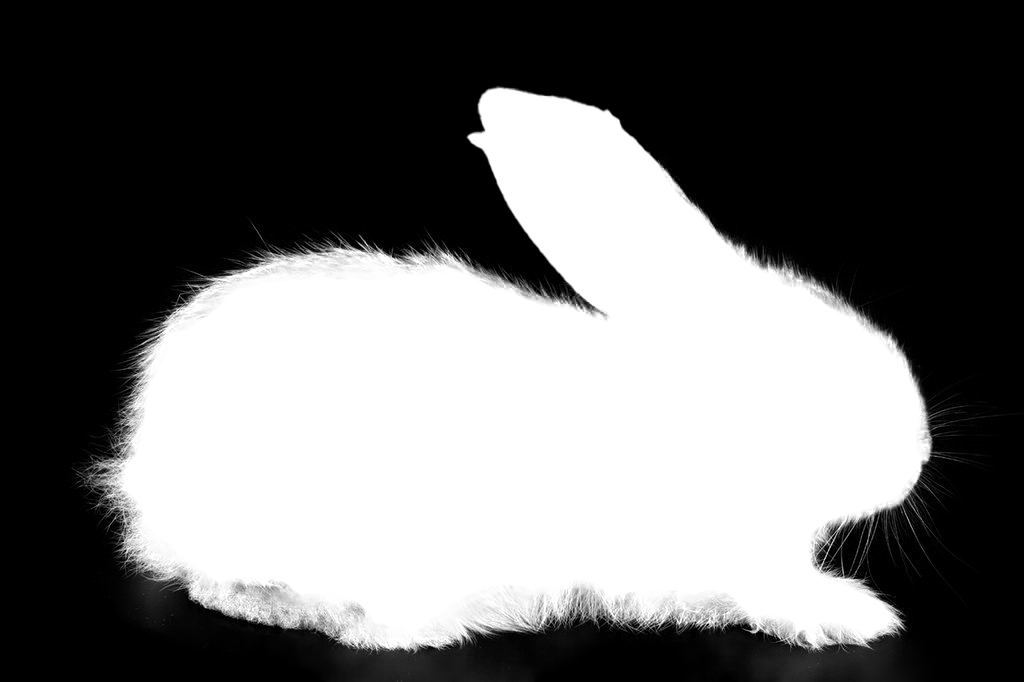} &
			\includegraphics[scale=0.0469]{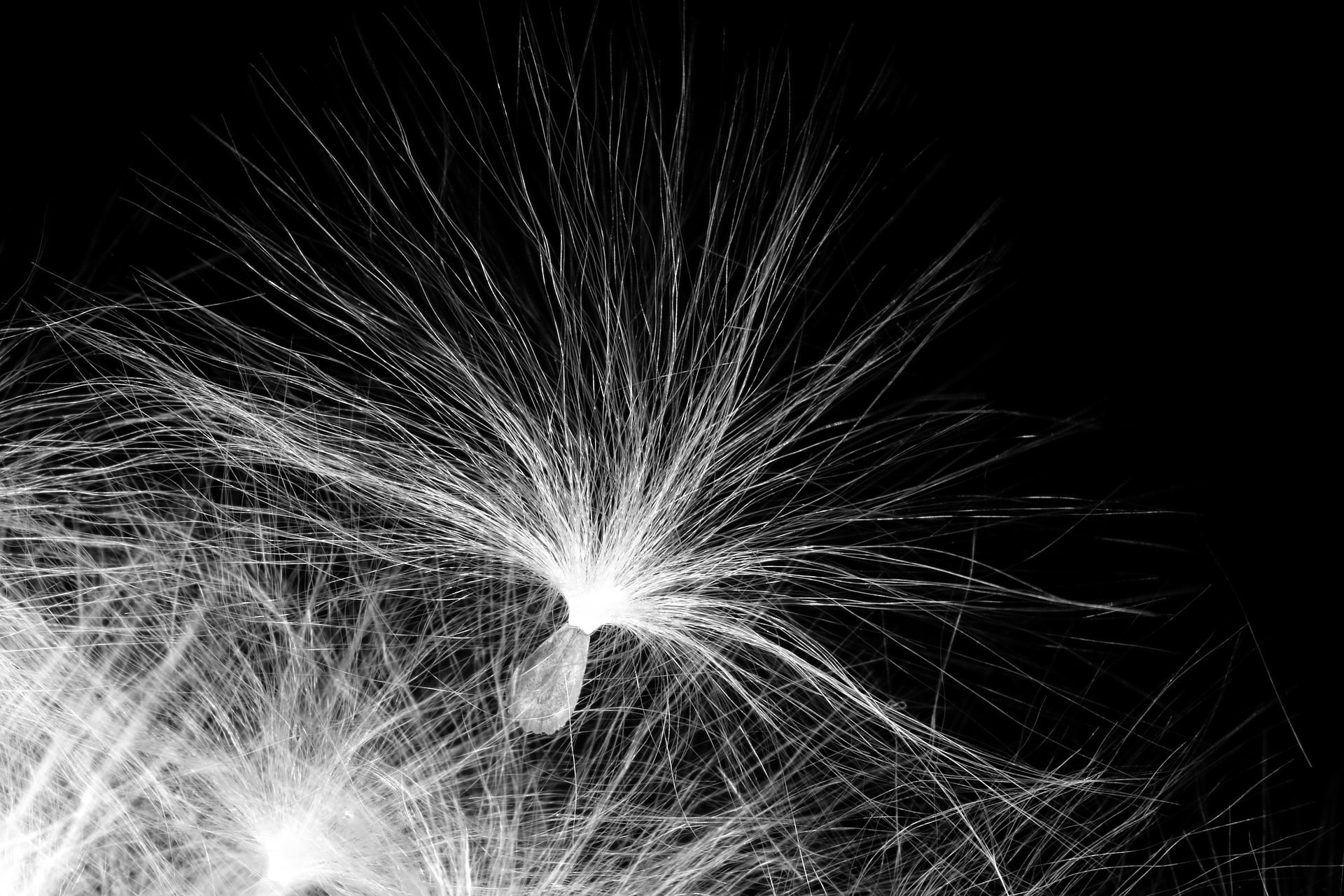} &
			\includegraphics[scale=0.0659]{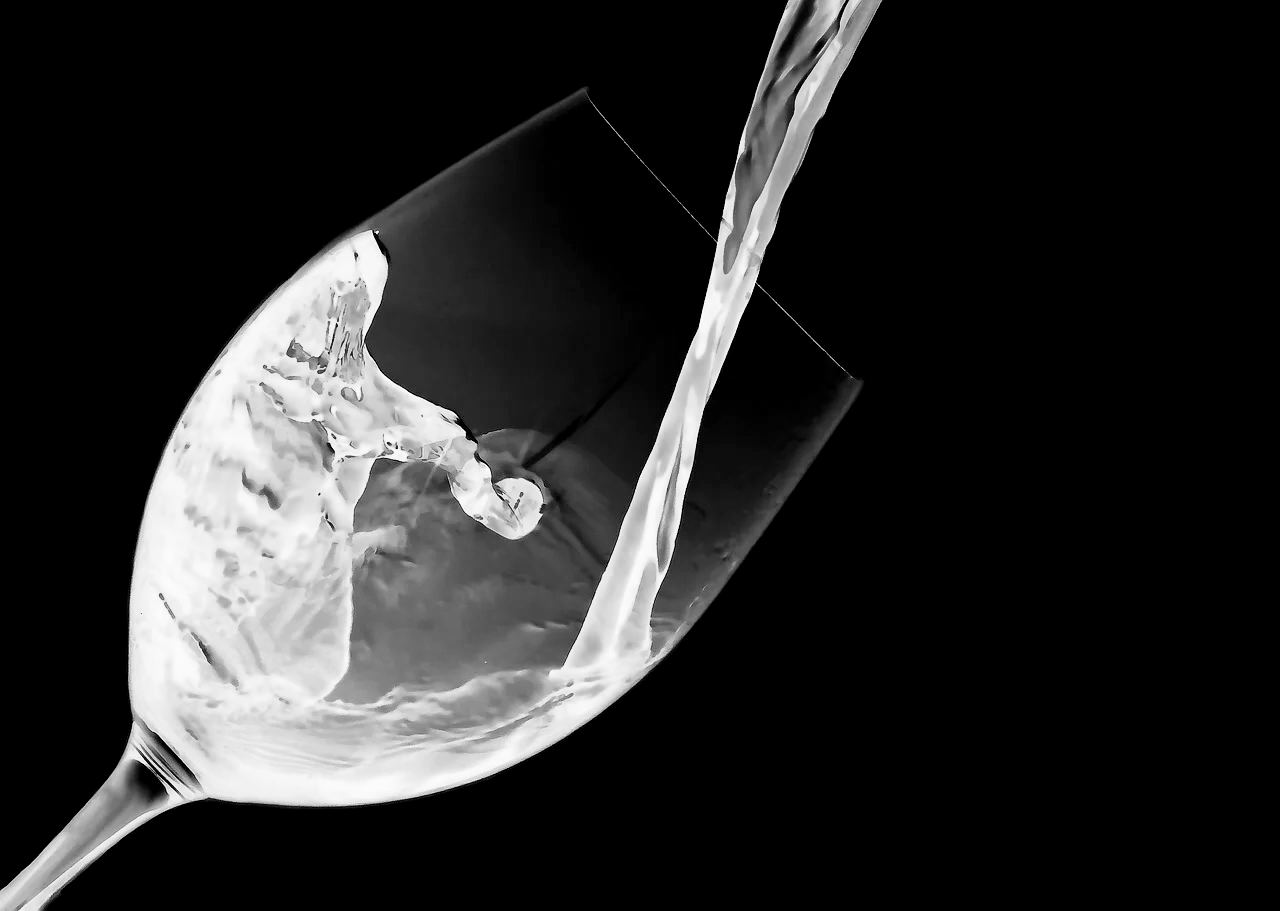} \\
			
			\includegraphics[scale=0.0514]{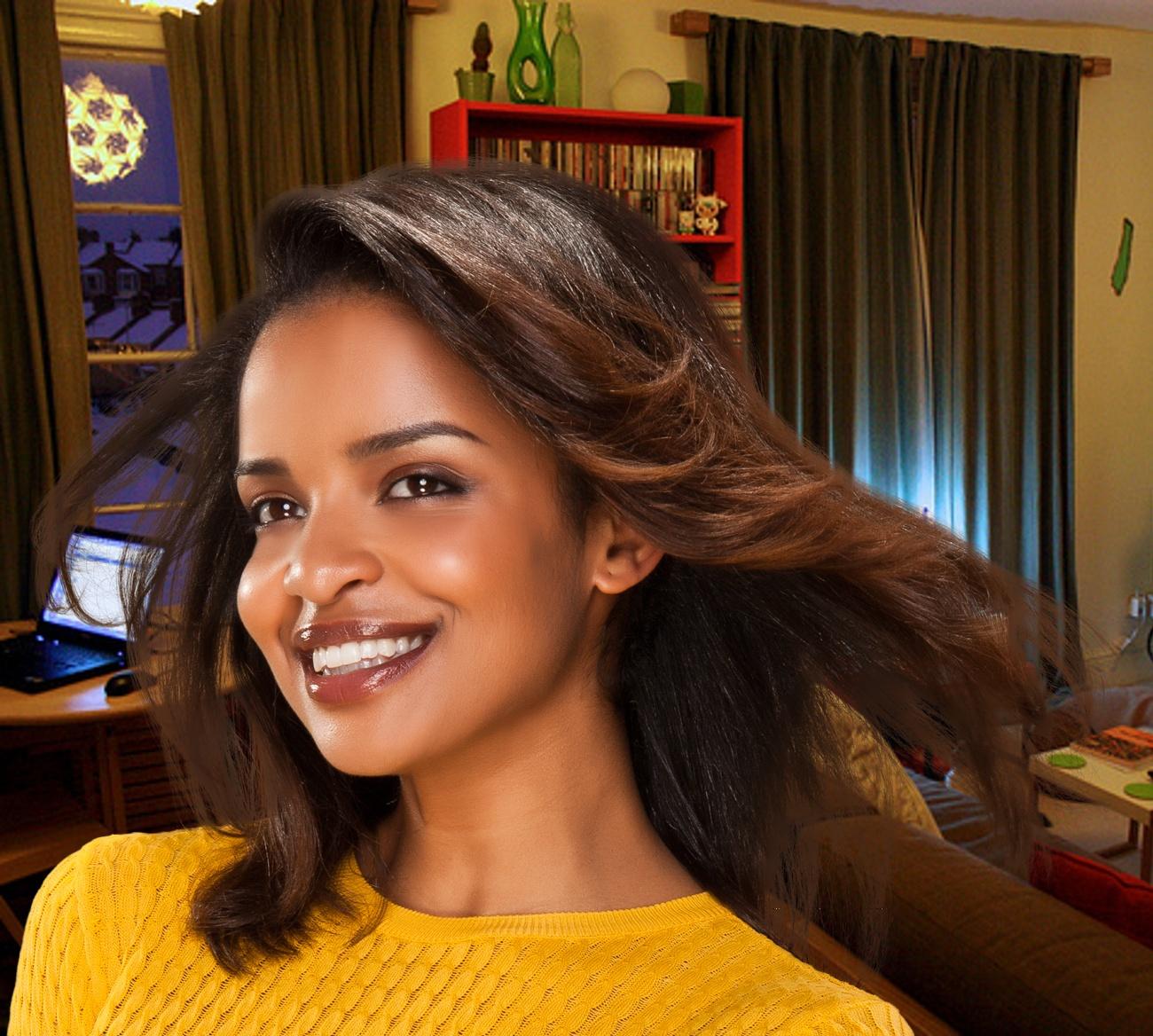} &
			\includegraphics[scale=0.088]{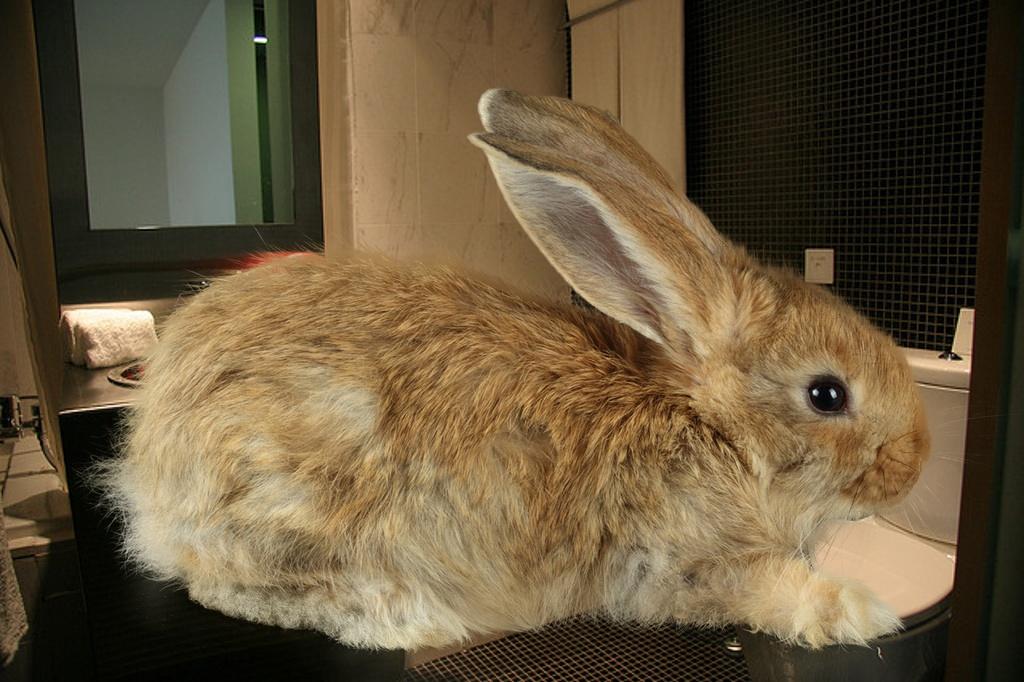} &
			\includegraphics[scale=0.0469]{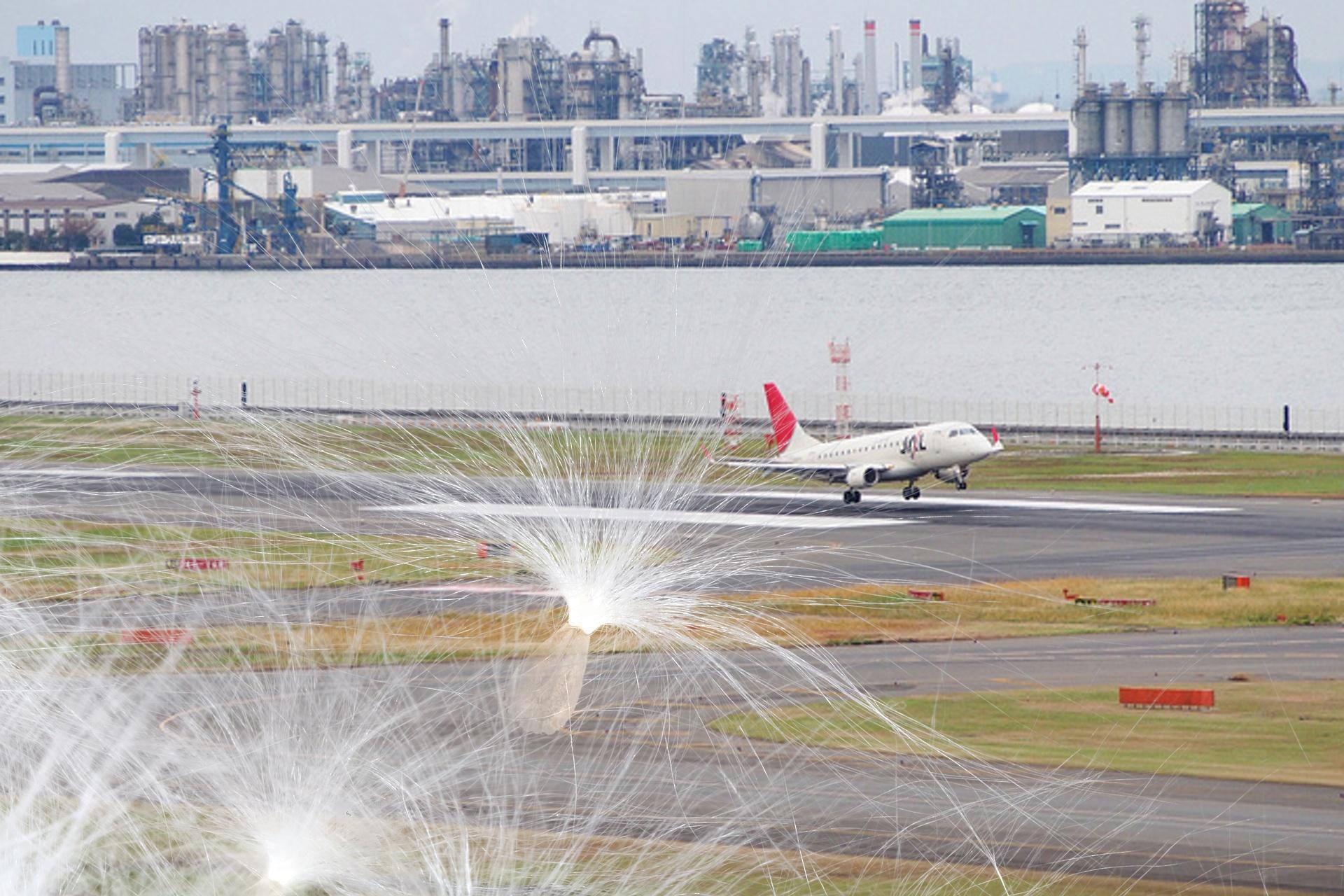} &
			\includegraphics[scale=0.0659]{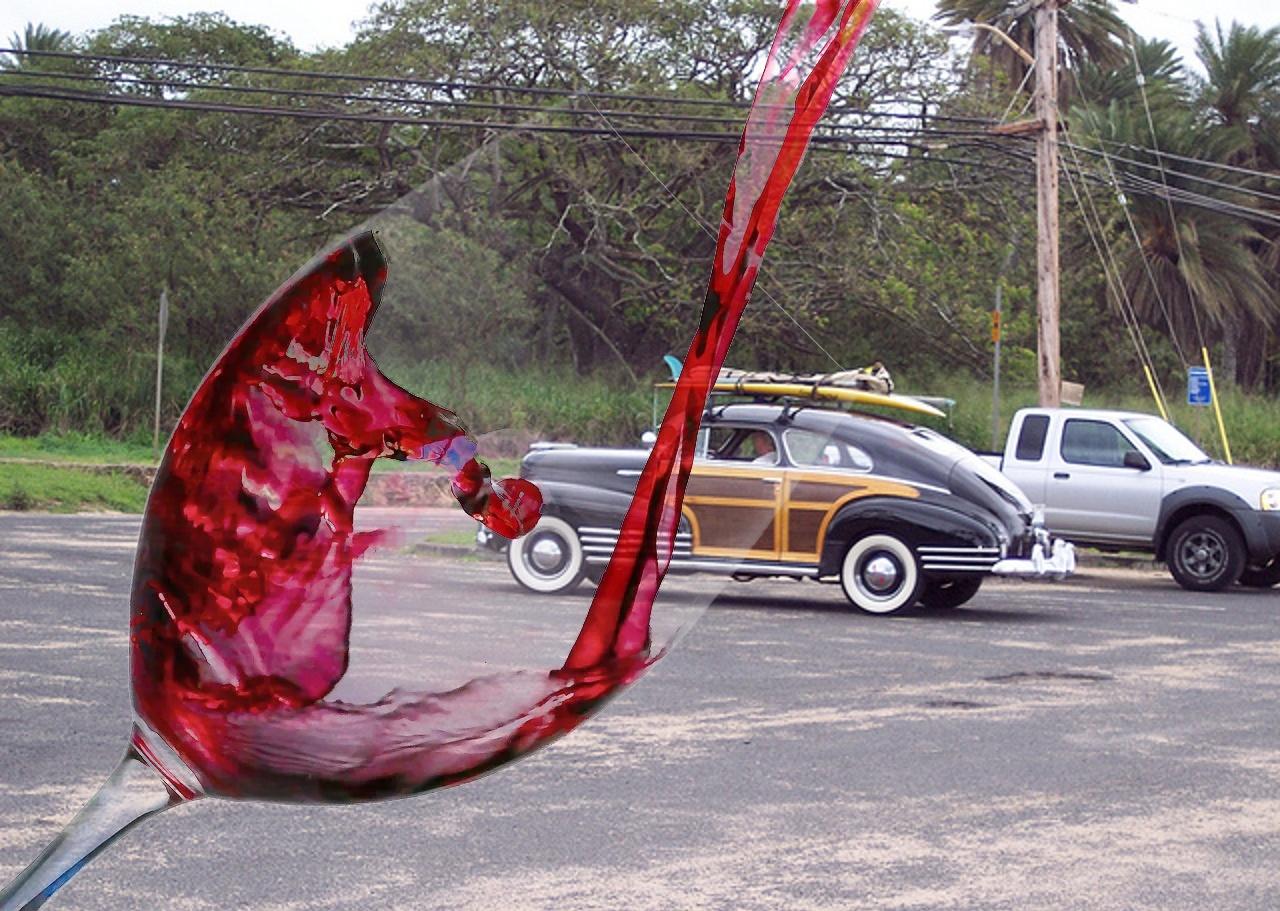} \\
	\end{tabular}}
	\caption{Some examples of our Distinctions-646 dataset. The first row-FG images, the second row-the ground truth alpha mattes, the third row-the corresponding composited results on new \emph{BG}. The FG images display irregularly due to truncated boundaries. Please zoom in to see the finer details and texture. }
	\label{fig:dataset}
\end{figure}
\subsection{Datasets and Evaluation Metrics}
\label{ssec:data_metrics}
\textbf{Datasets.} The first dataset is the public Adobe Composition-1K~\cite{Xu2017Deep}. The training set consists of 431 \emph{FG} objects with the corresponding ground truth alpha mattes. Each \emph{FG} image is combined with 100 \emph{BG} images from MS COCO dataset~\cite{Lin2014Microsoft} to composite the input images. For the test set, the Composition-1K contains 50 \emph{FG} images as well as the corresponding alpha mattes, and $1000$ \emph{BG} images from PASCAL VOC2012 dataset~\cite{pascal-voc-2007}. The training and test sets were synthesized through the algorithm provided by~\cite{Xu2017Deep}.

The second is our Distinctions-646 dataset. The Adobe Composition-1K contains many consecutive video frames, and cropped patches from the same image, and there are only 250 dissimilar \emph{FG} objects in their training set. To improve the versatility and robustness of the matting during training, we construct our Distinctions-646 dataset composed of 646 distinct \emph{FG} images. As shown in Fig.~\ref{fig:dataset}, the dataset we proposed has a high degree of alpha precision and high freedom of \emph{FG} categories and shapes. For training and test, we divide these \emph{FG} examples into $596$ and $50$ and then produce $59,600$ training images and $1000$ test examples according to the composition rules in \cite{Xu2017Deep}. We have released the download right of our dataset. 

\begin{figure*}[b]
	\setlength{\tabcolsep}{0.5pt}\small{
		\begin{tabular}{cccccc}
			\includegraphics[scale=0.0333]{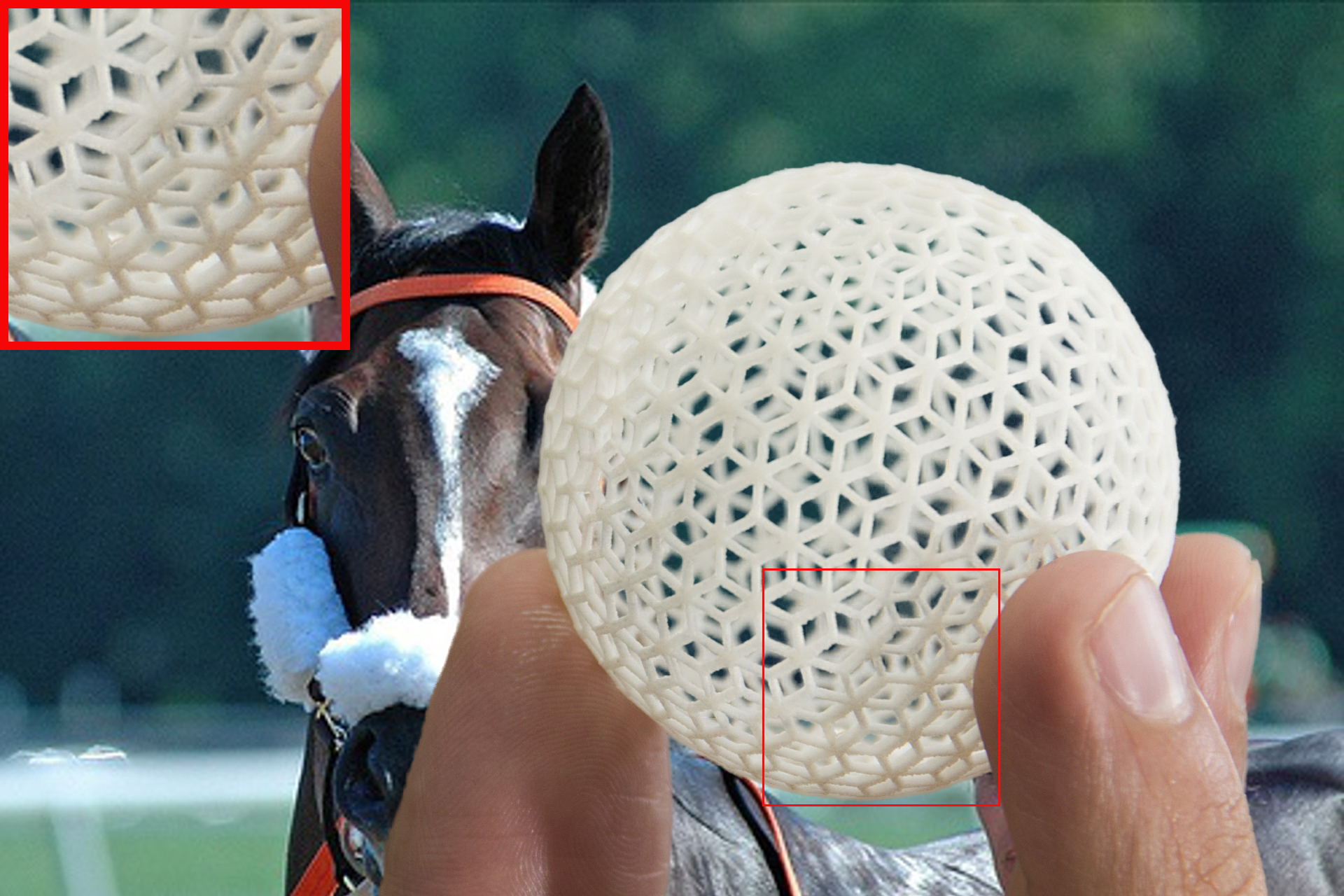} &
			\includegraphics[scale=0.0333]{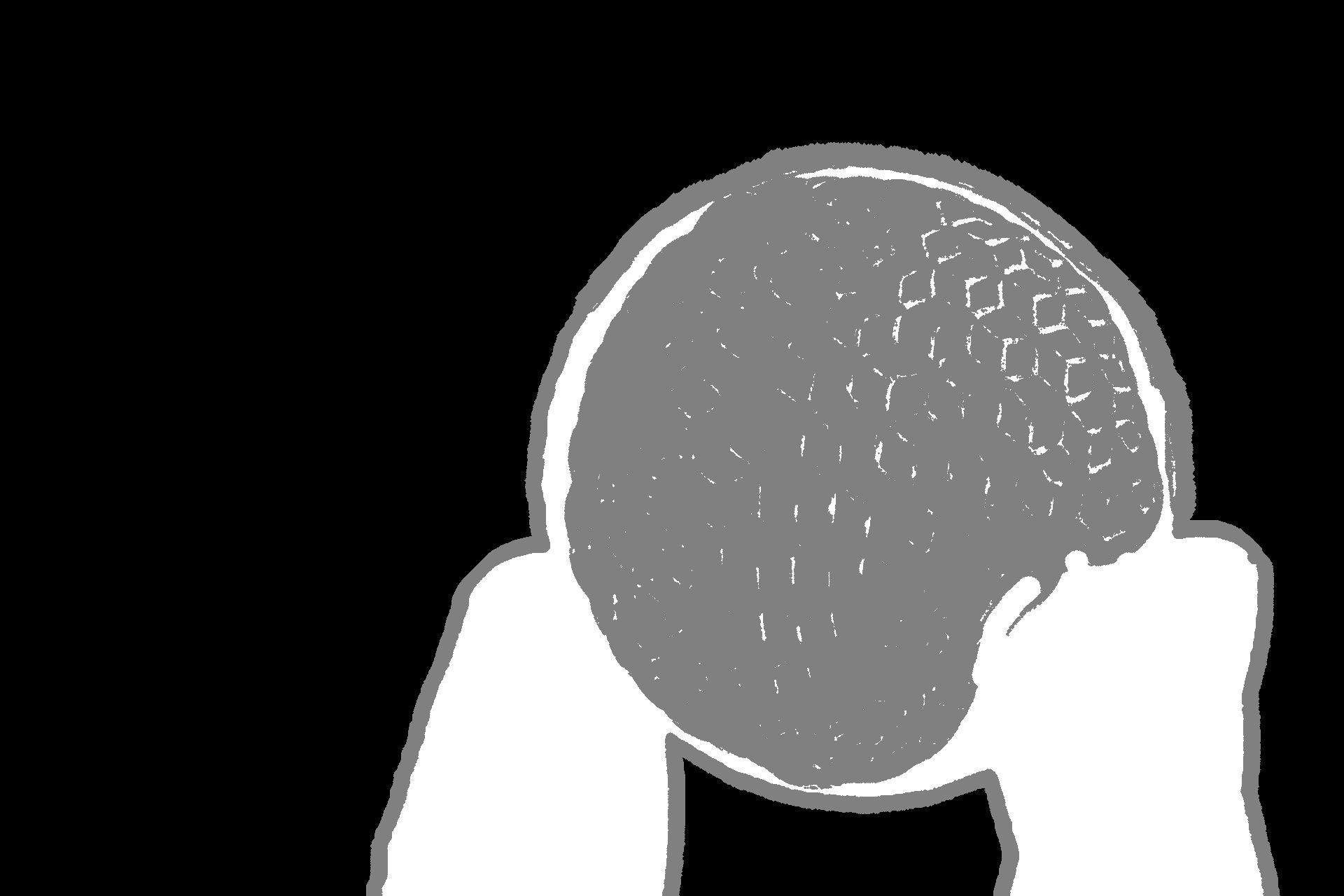} &
			\includegraphics[scale=0.0333]{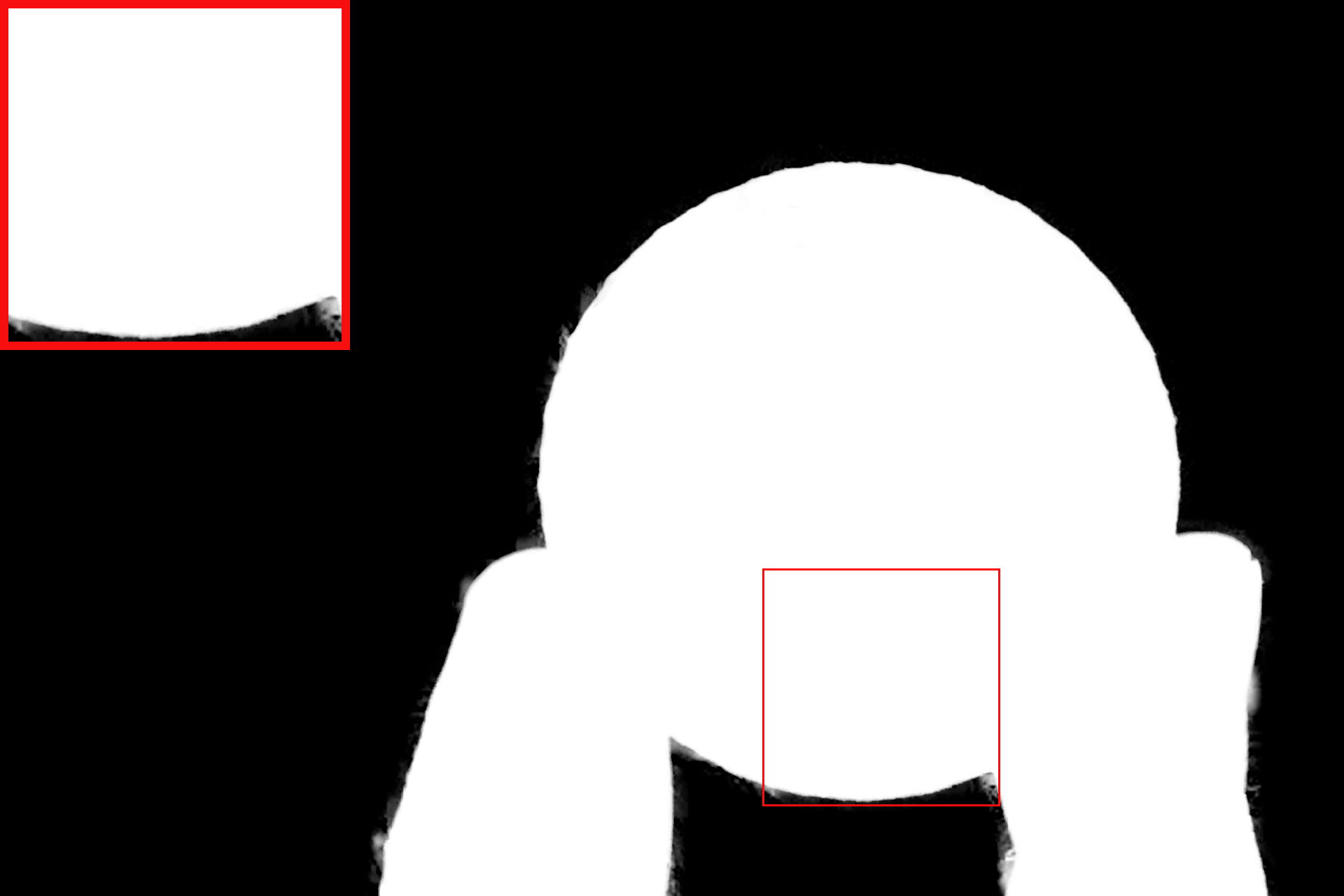} &
			\includegraphics[scale=0.0333]{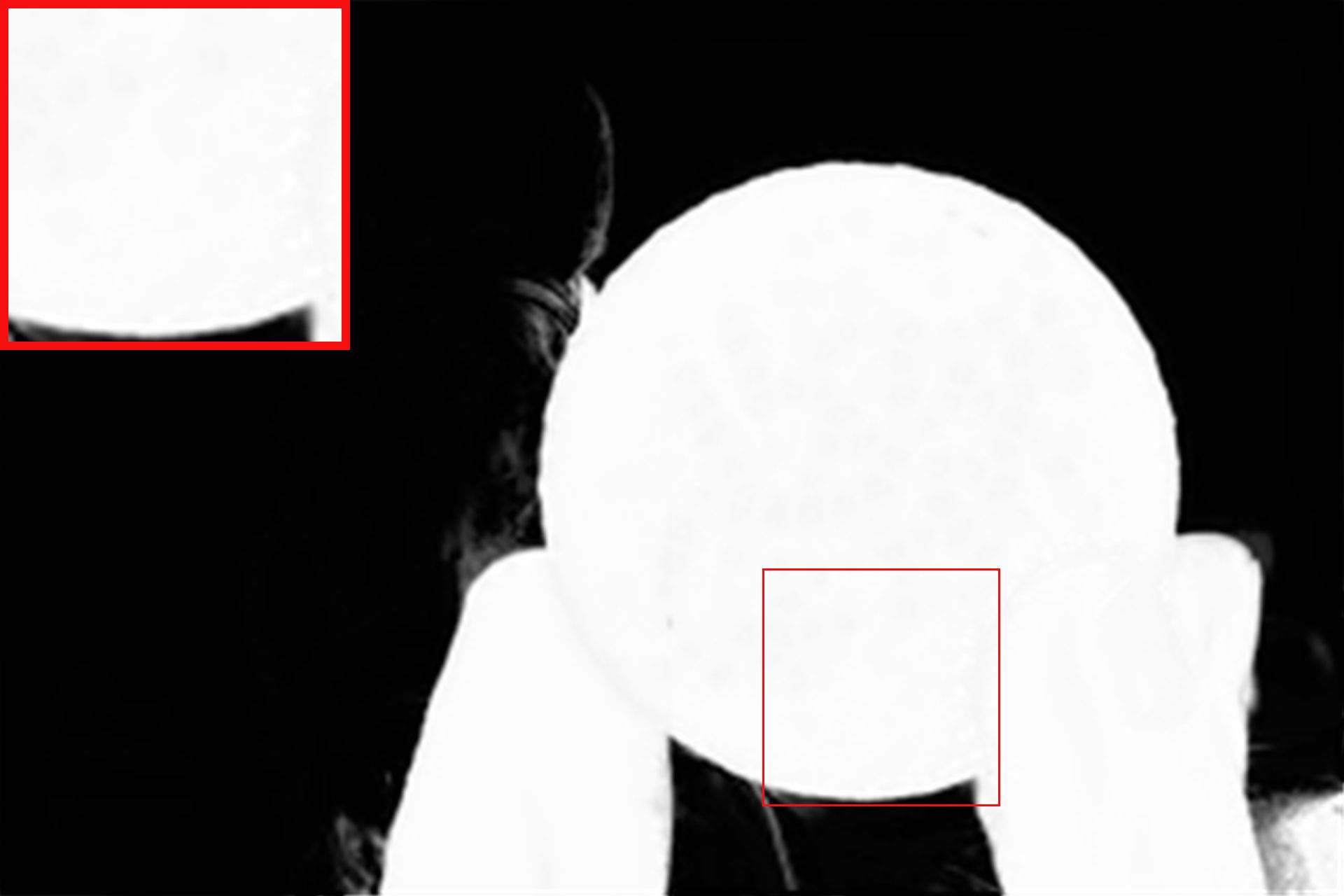} &
			\includegraphics[scale=0.0333]{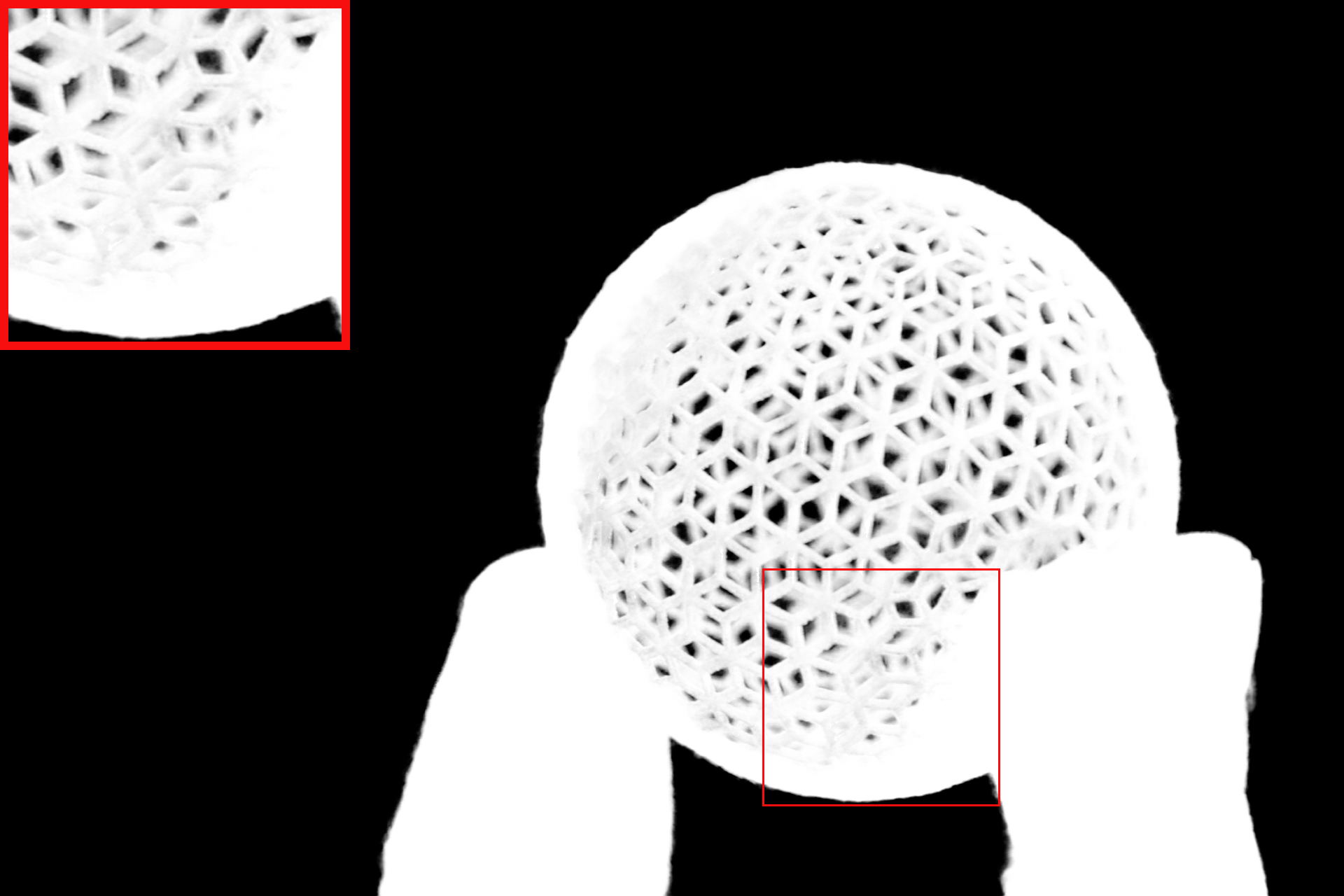} &
			\includegraphics[scale=0.0333]{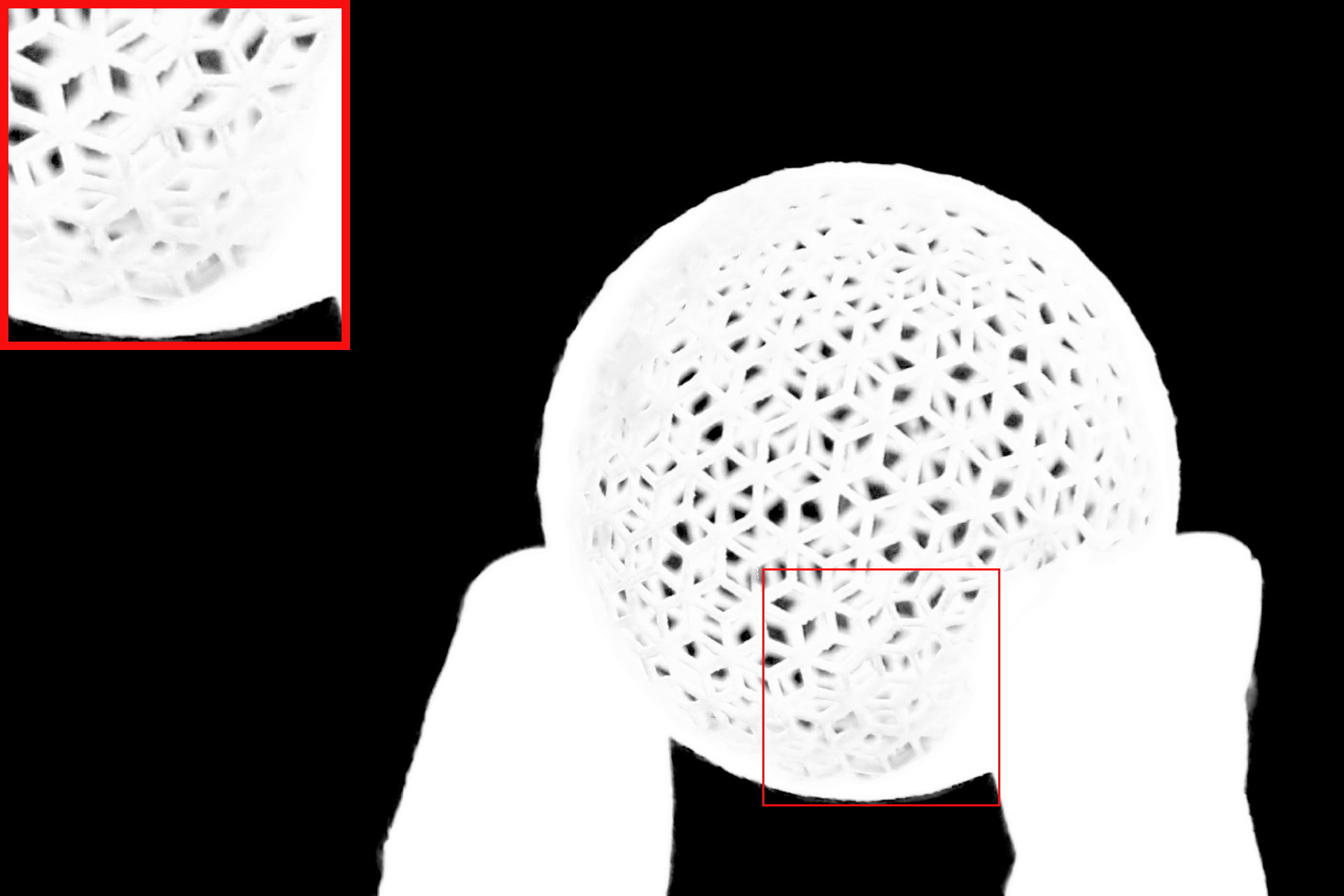} \\
			
			Input Image & Trimap & Closed Form~\cite{Levin2007A} & 
			SSS~\cite{sss} & DIM~\cite{Xu2017Deep} & IndexNet~\cite{hao2019indexnet} \\
			
			\includegraphics[scale=0.0333]{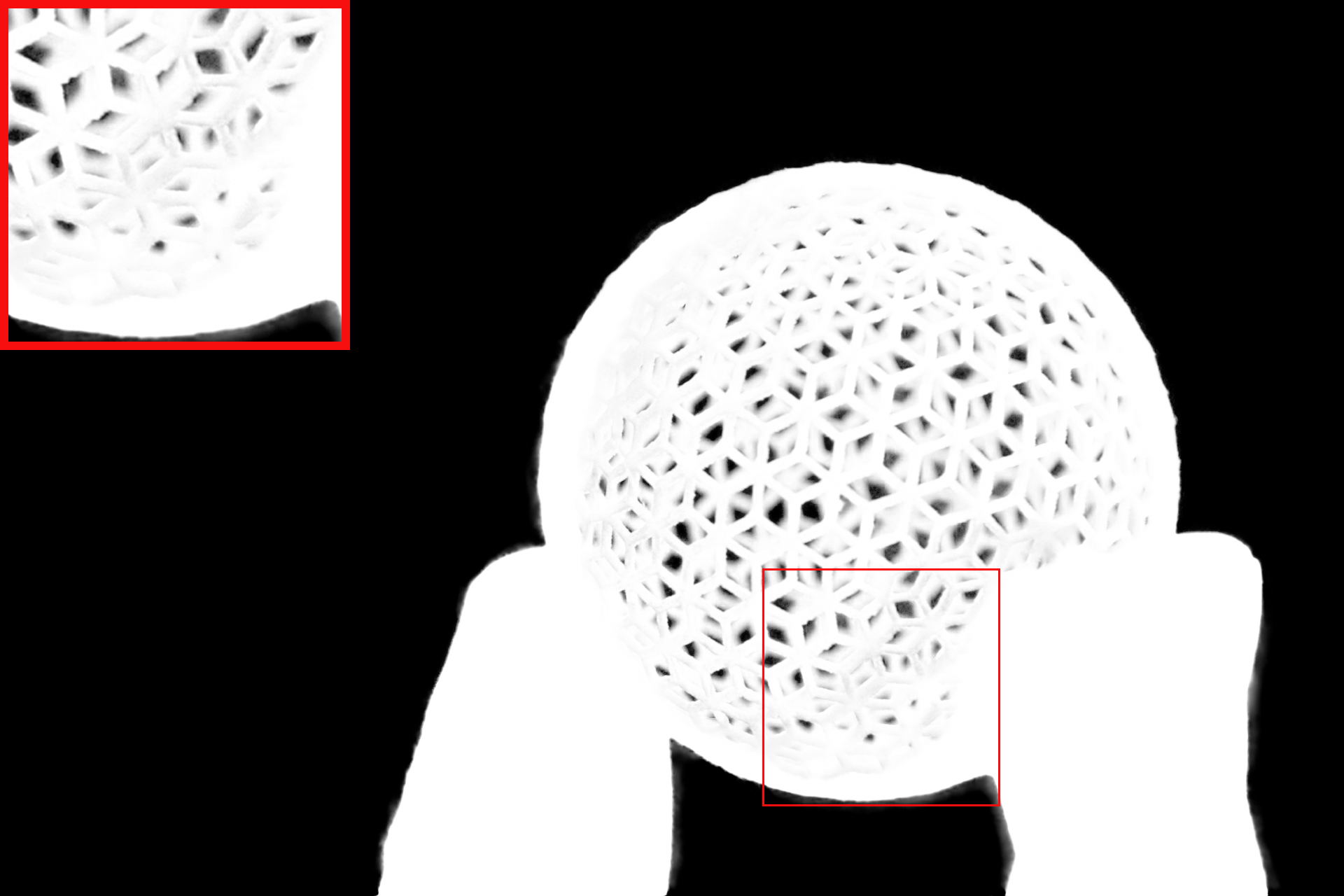} &
			\includegraphics[scale=0.0444]{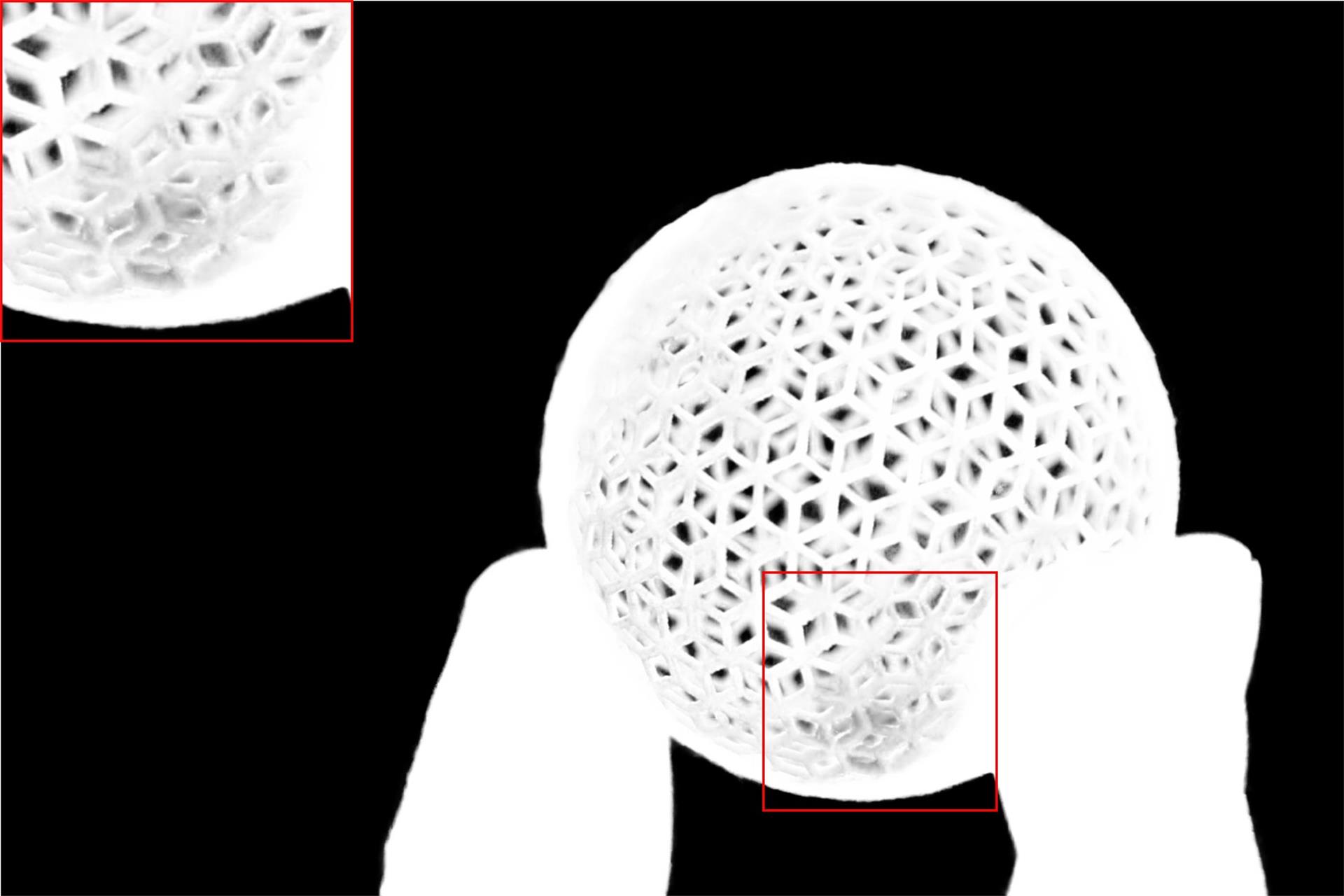} &
			\includegraphics[scale=0.0444]{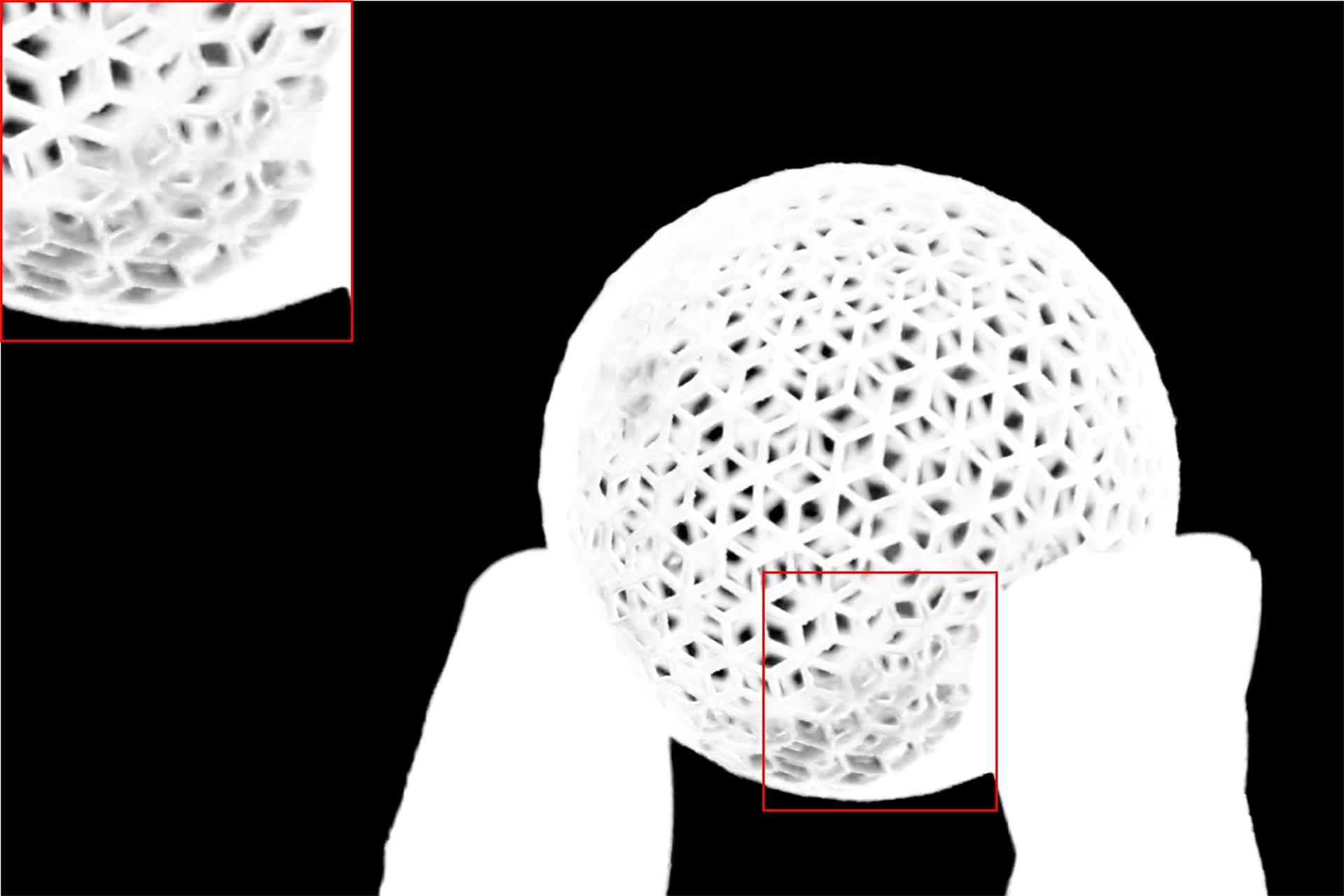} &
			\includegraphics[scale=0.0333]{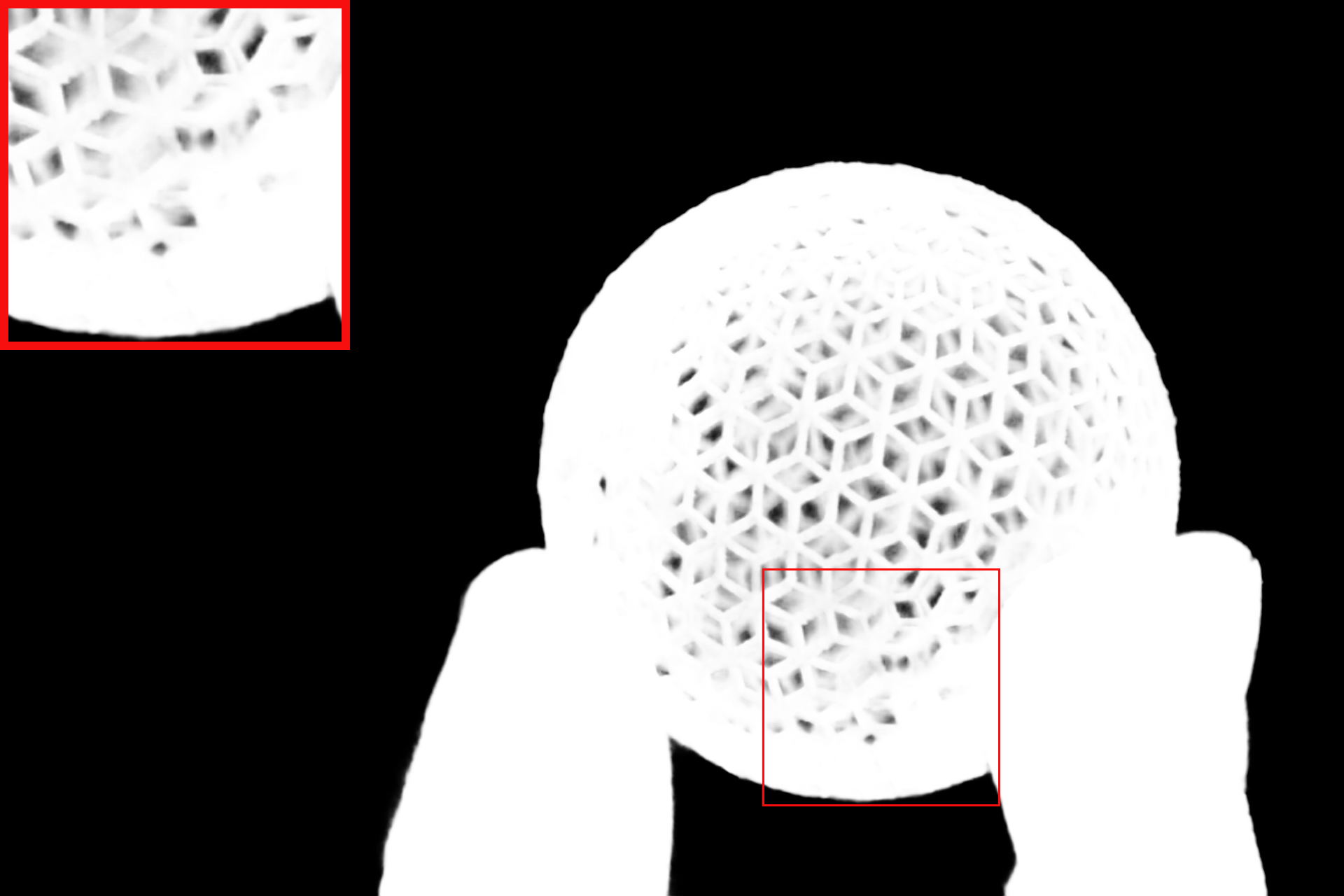} &
			\includegraphics[scale=0.0333]{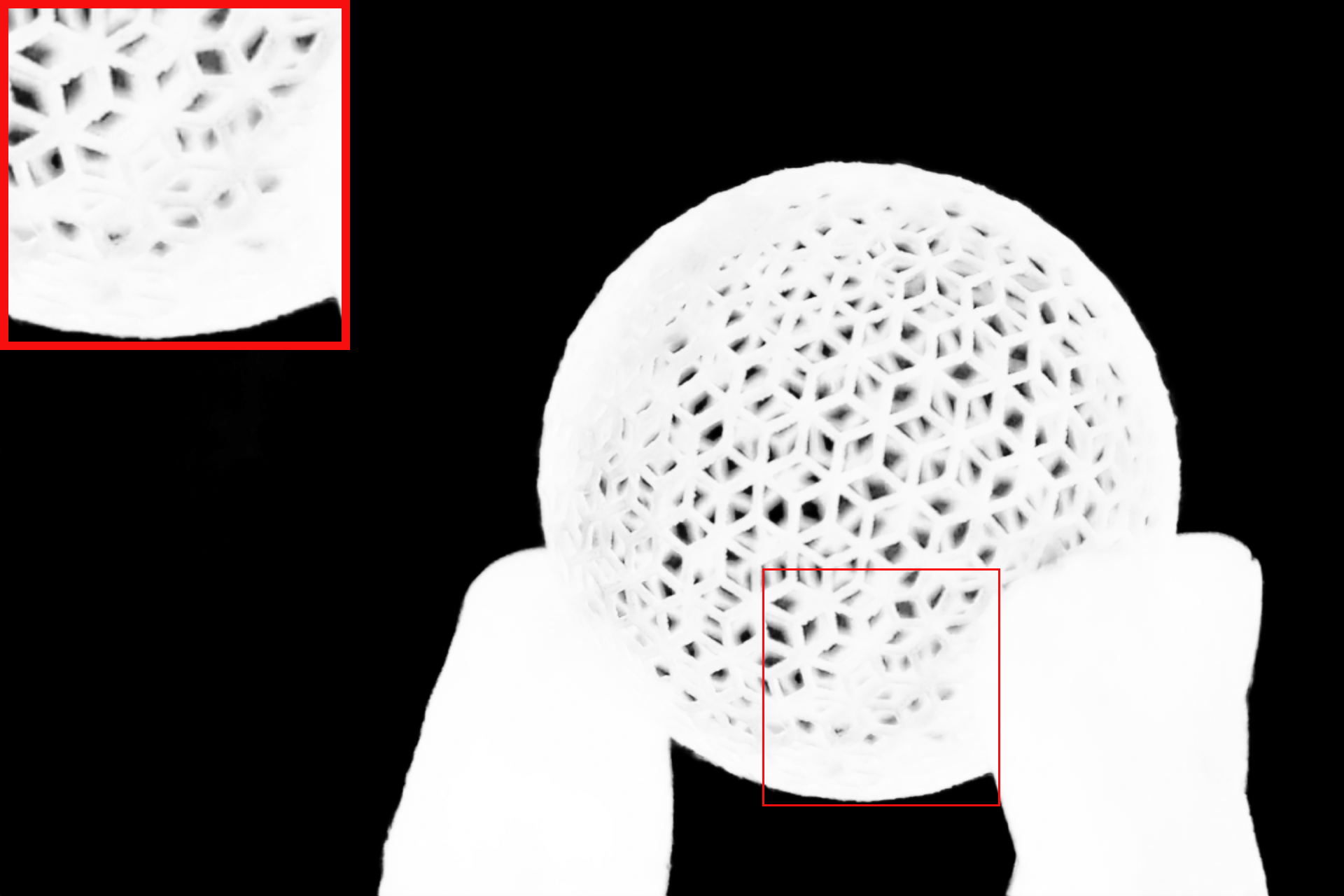} &
			\includegraphics[scale=0.0333]{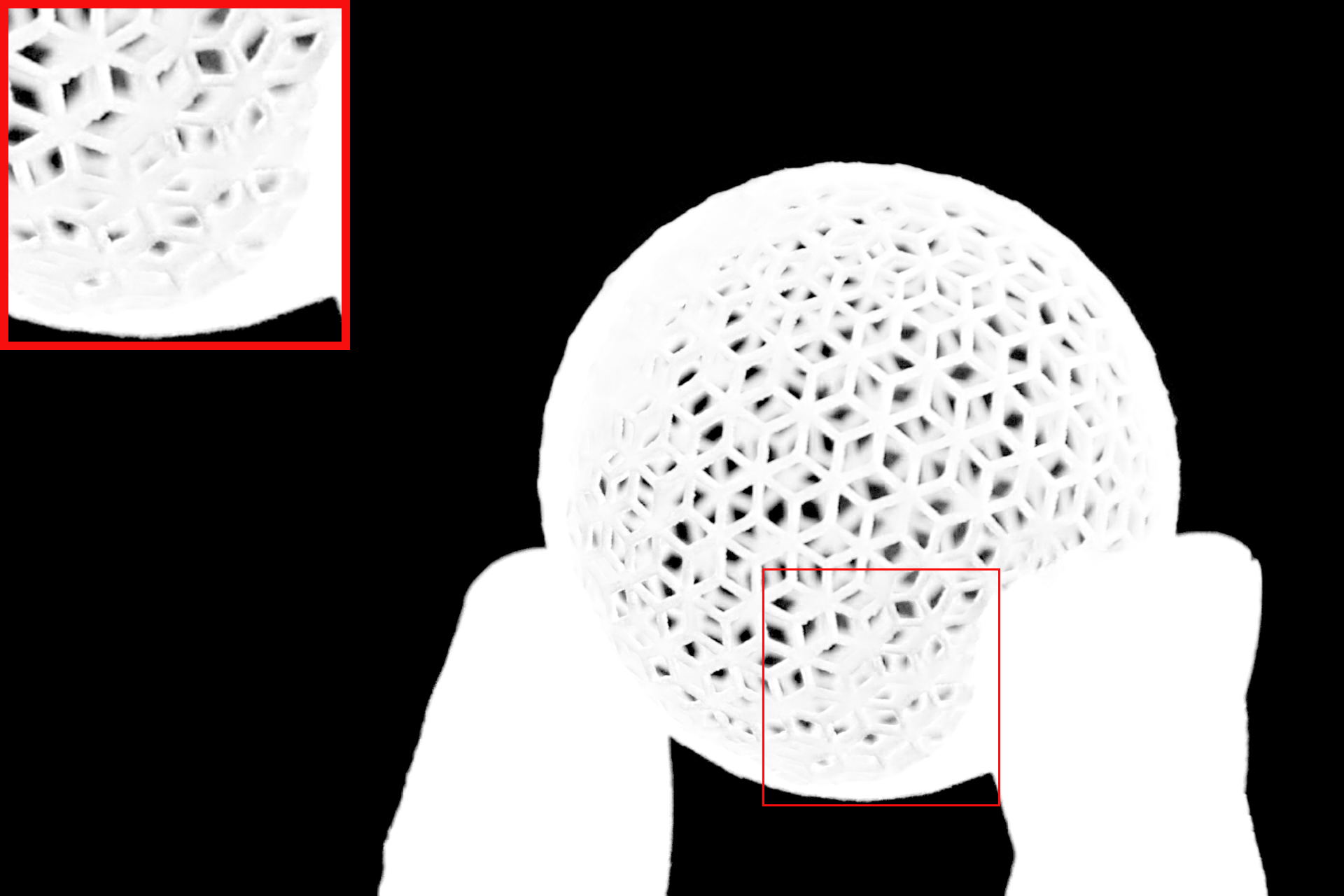} \\
			
			CA~\cite{hou2019context} & GCA~\cite{li2020natural} & A${^2}$U~\cite{dai2021learning} &  Late Fusion~\cite{Zhang2019CVPR} 
			& Ours & Ground Truth \\
			\includegraphics[scale=0.05]{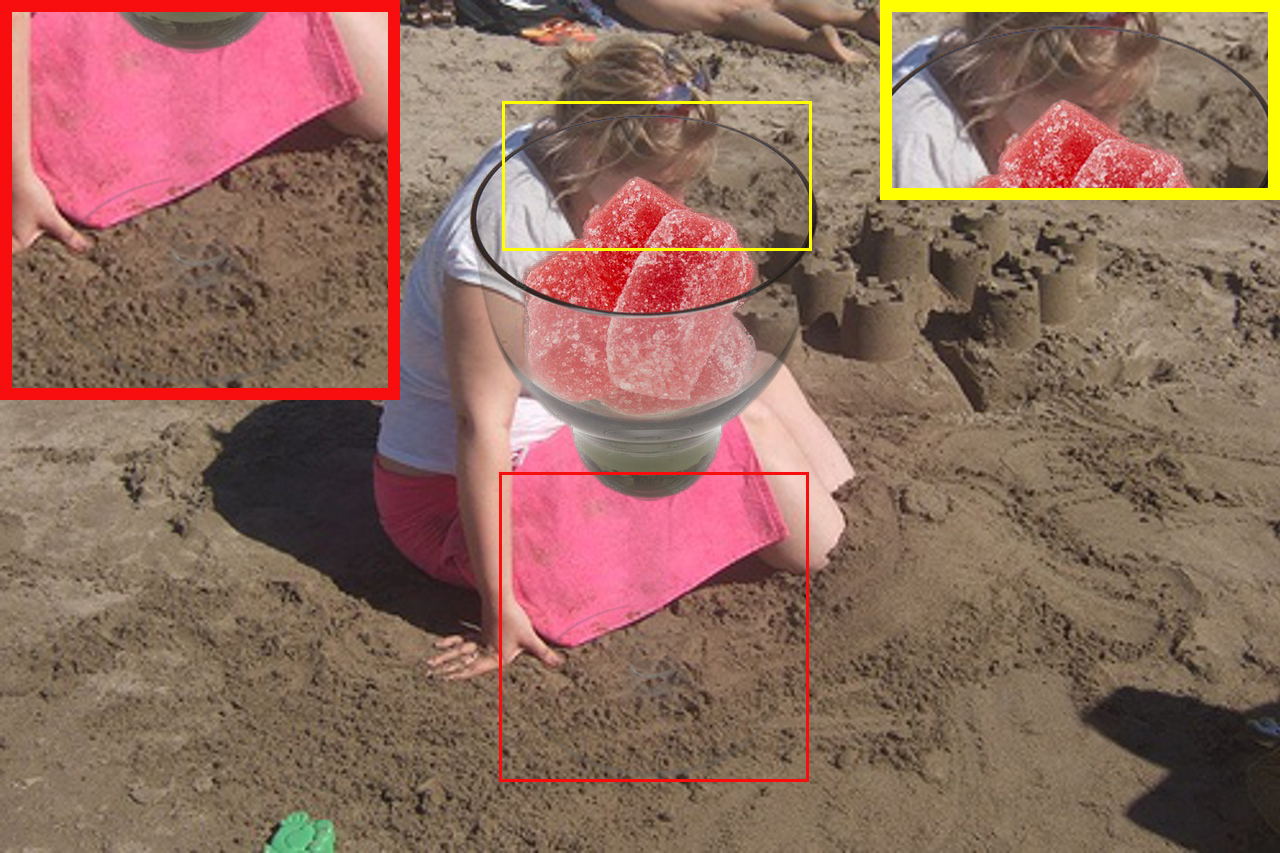} &
			\includegraphics[scale=0.05]{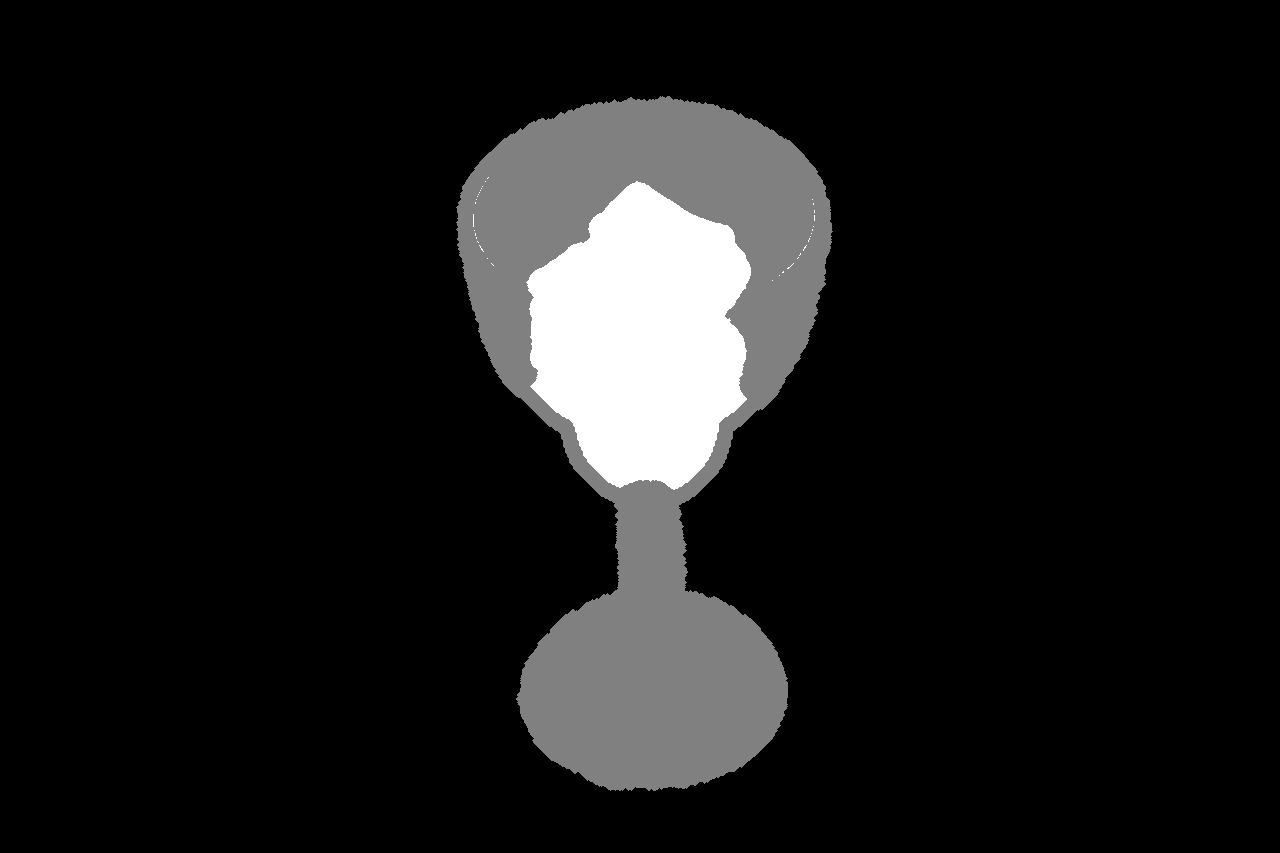} &
			\includegraphics[scale=0.05]{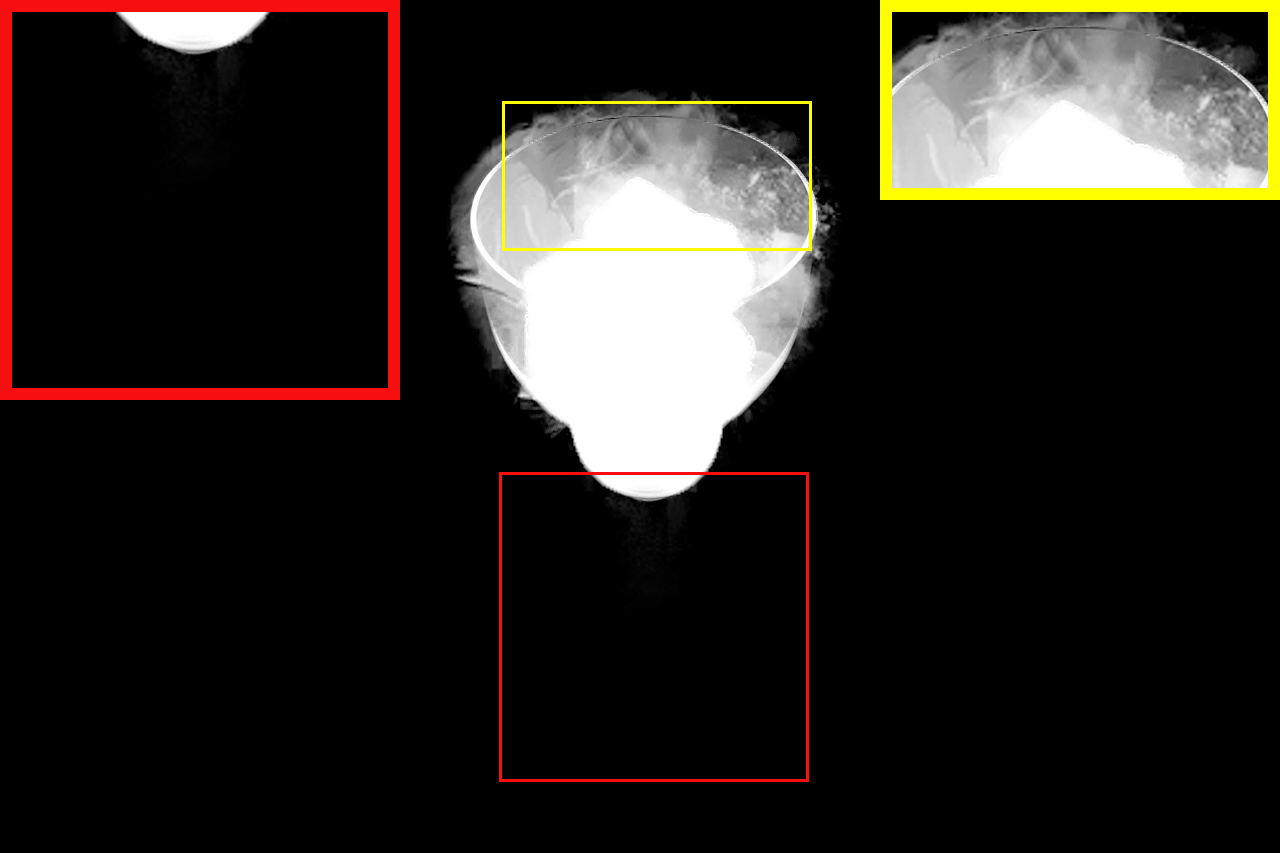} &
			\includegraphics[scale=0.05]{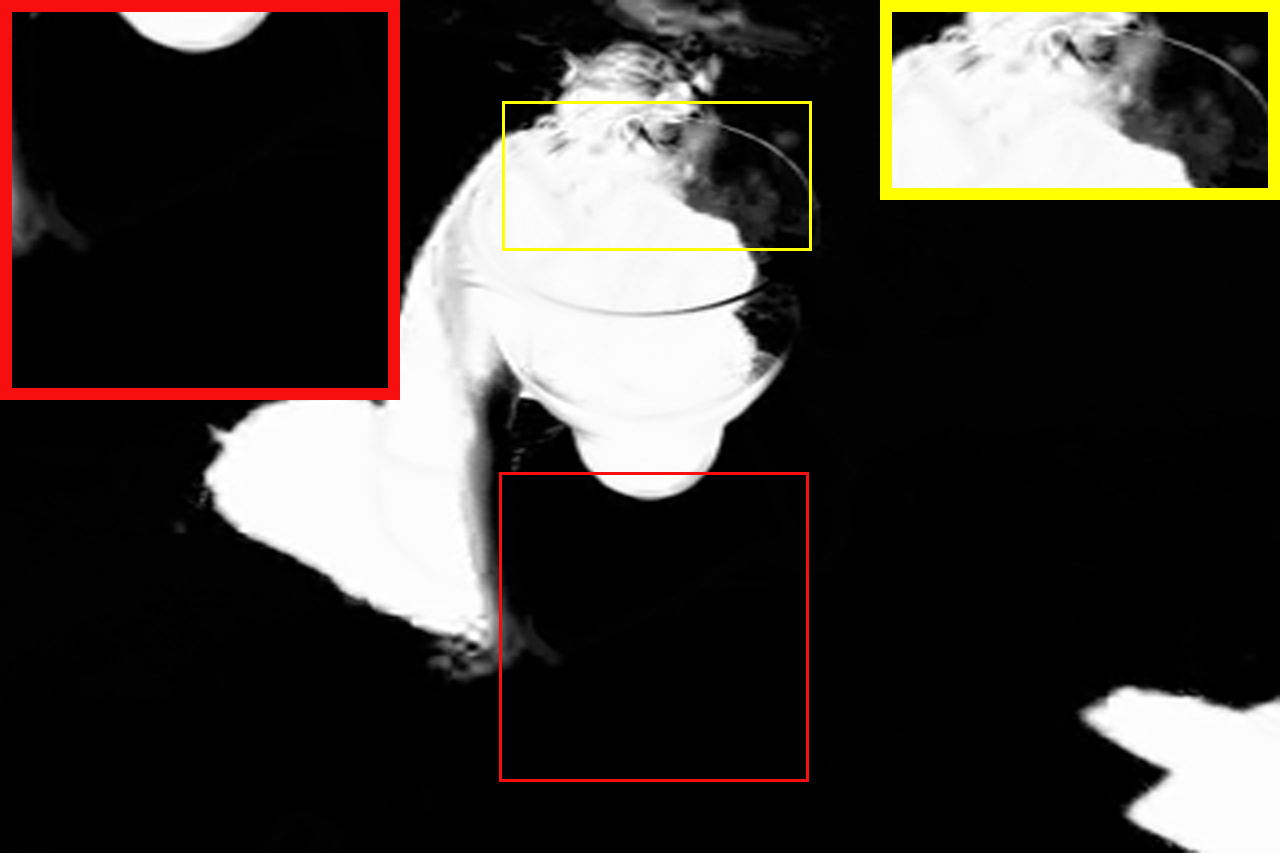} &
			\includegraphics[scale=0.05]{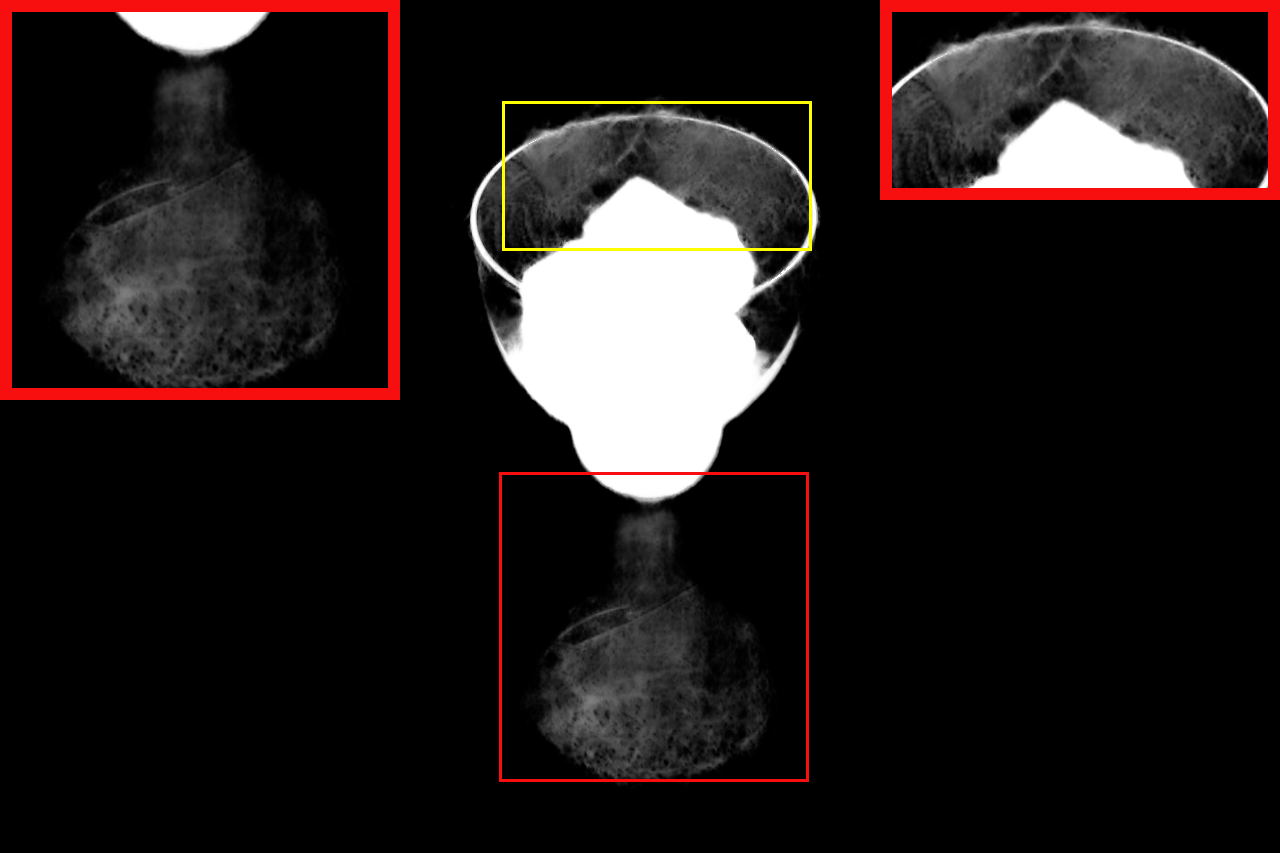} &
			\includegraphics[scale=0.05]{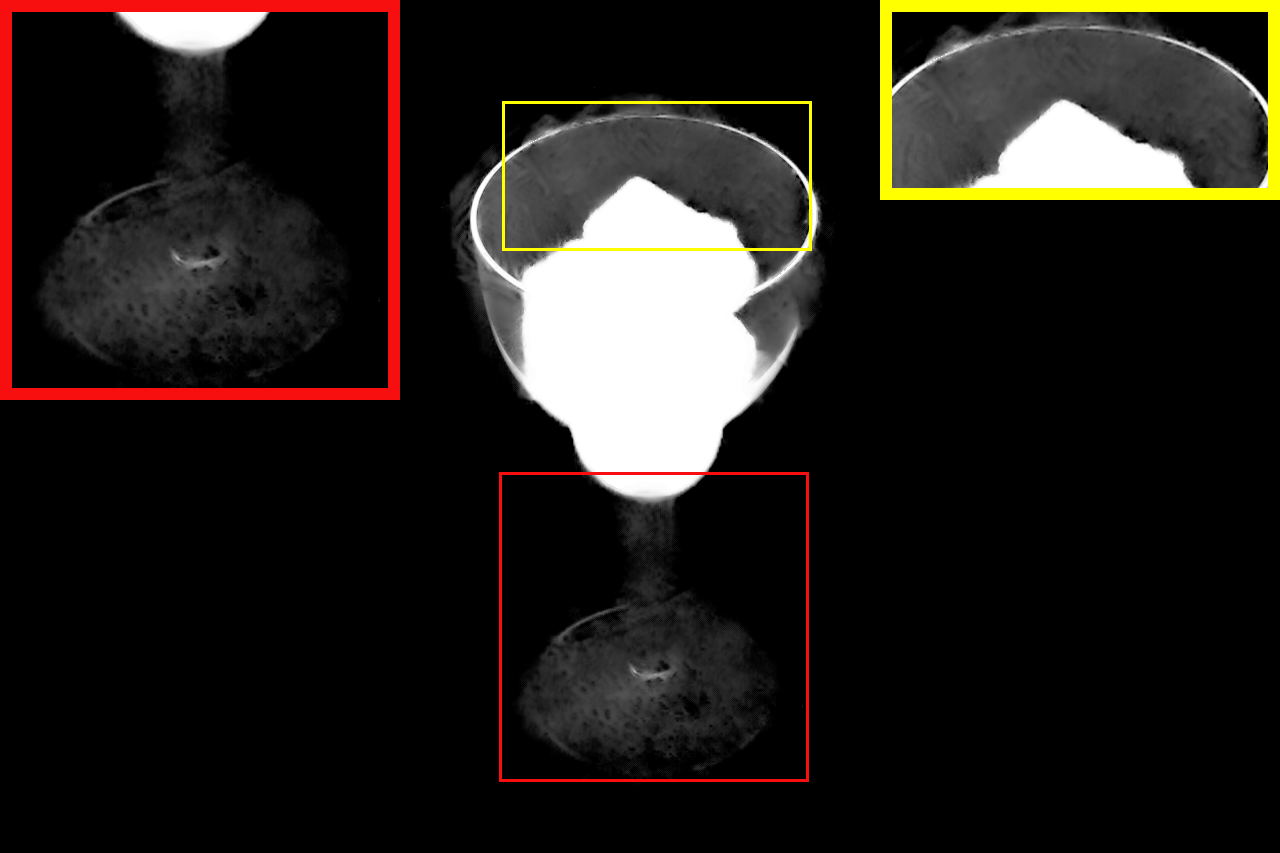} \\
			
			Input Image & Trimap & Closed Form~\cite{Levin2007A} & 
			SSS~\cite{sss} & DIM~\cite{Xu2017Deep} & IndexNet~\cite{hao2019indexnet} \\
			
			\includegraphics[scale=0.05]{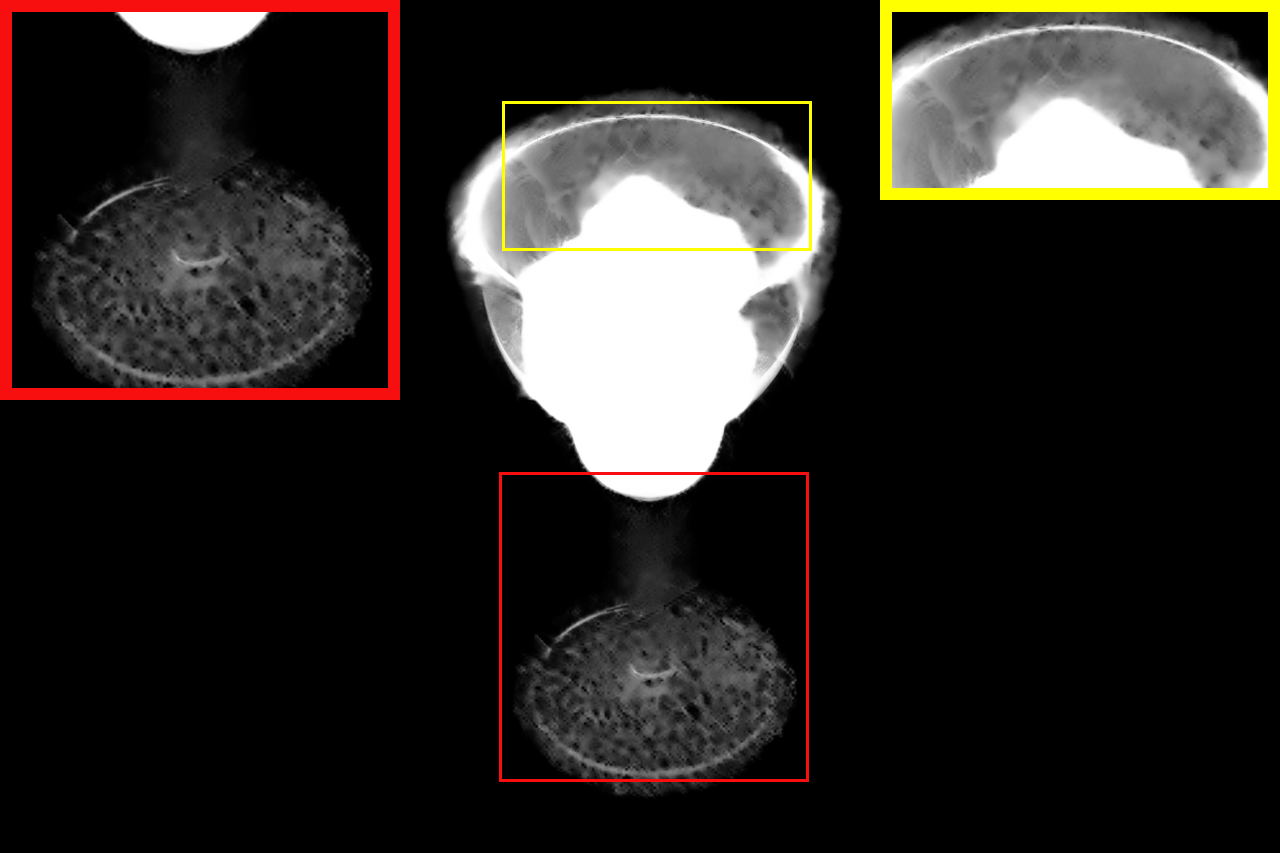} &
			\includegraphics[scale=0.066]{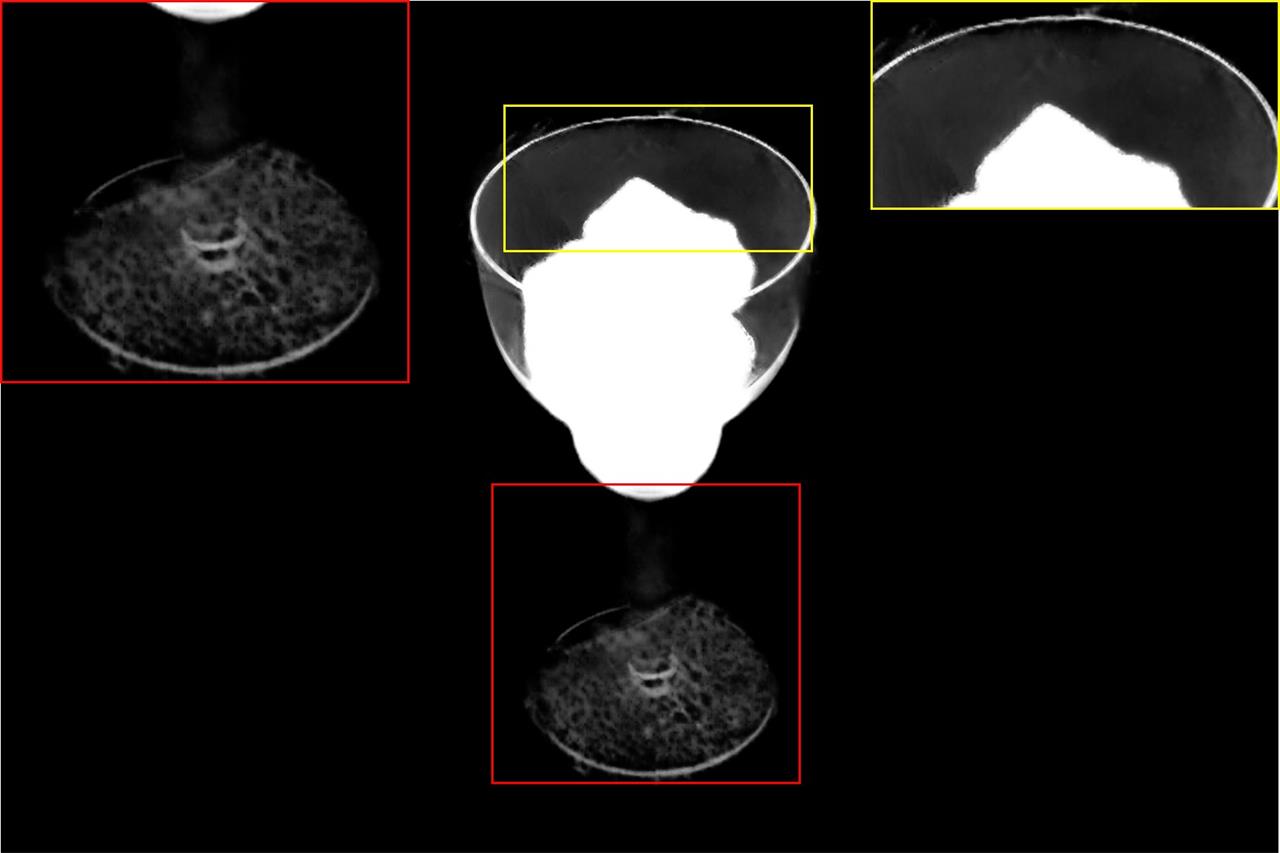} &
			\includegraphics[scale=0.066]{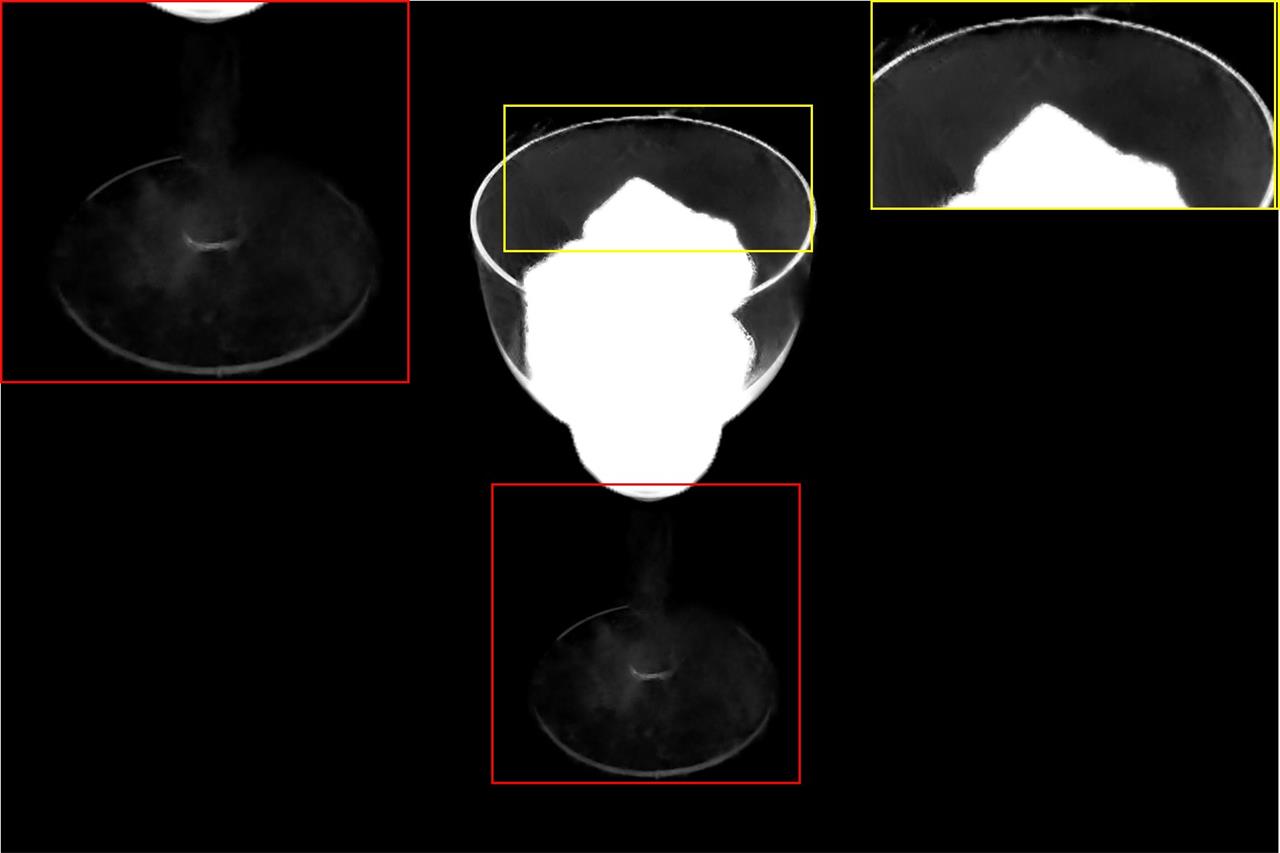} &
			\includegraphics[scale=0.05]{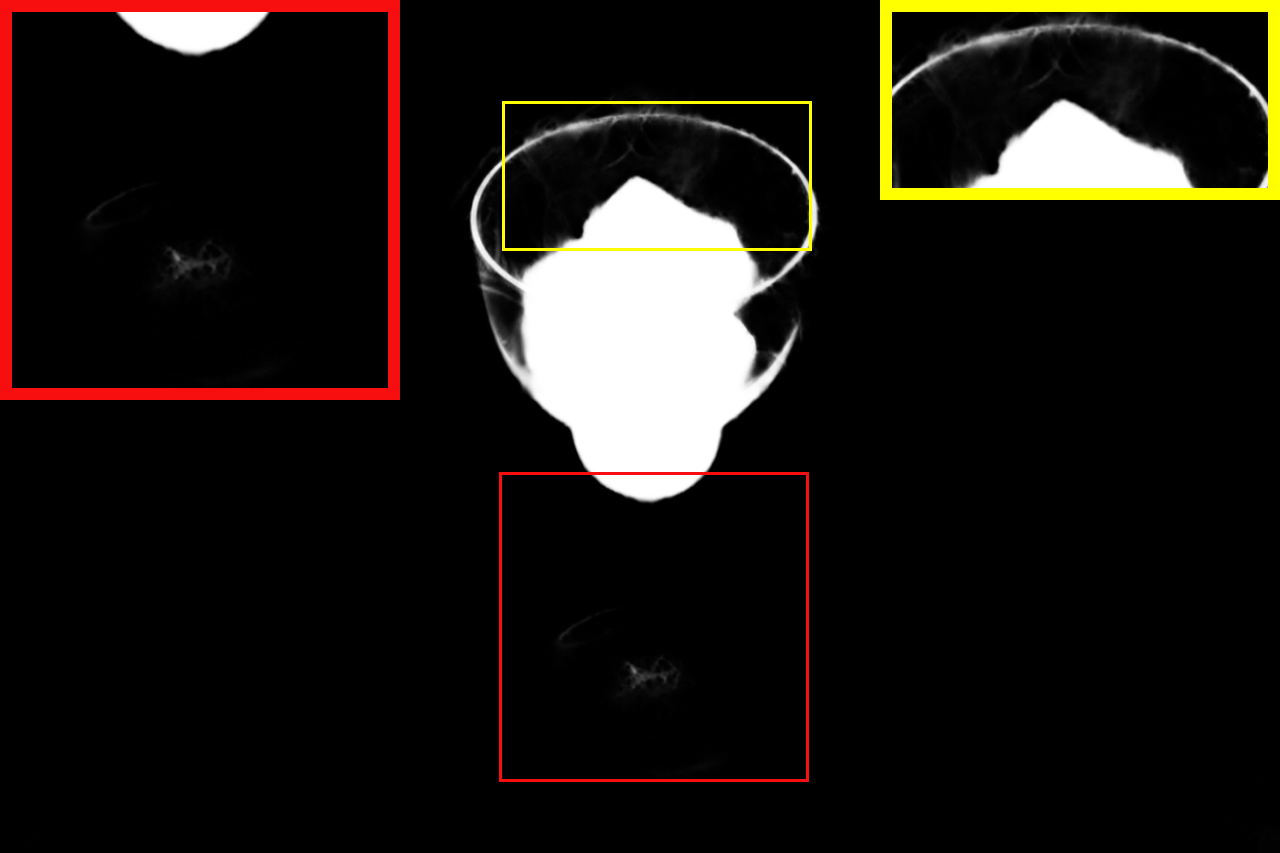} &
			\includegraphics[scale=0.05]{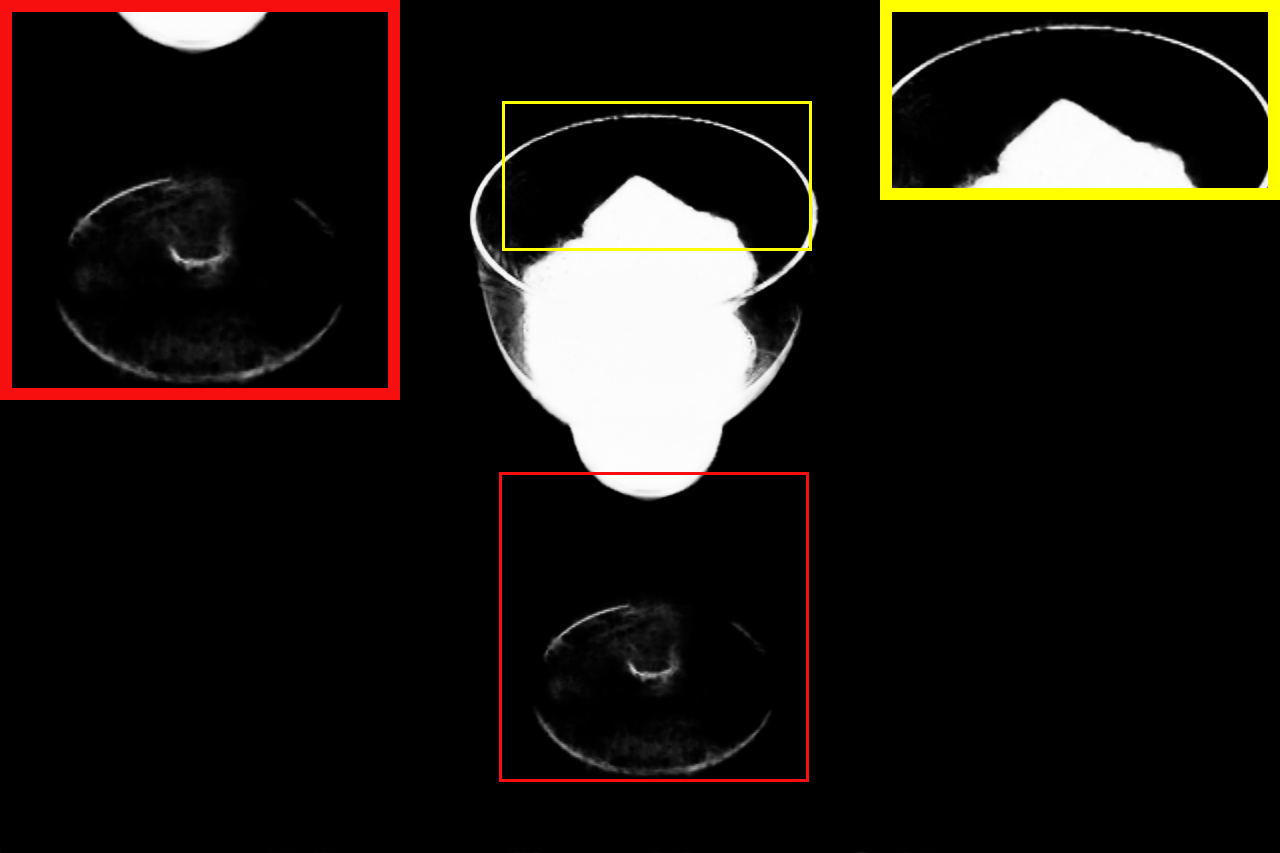} &
			\includegraphics[scale=0.05]{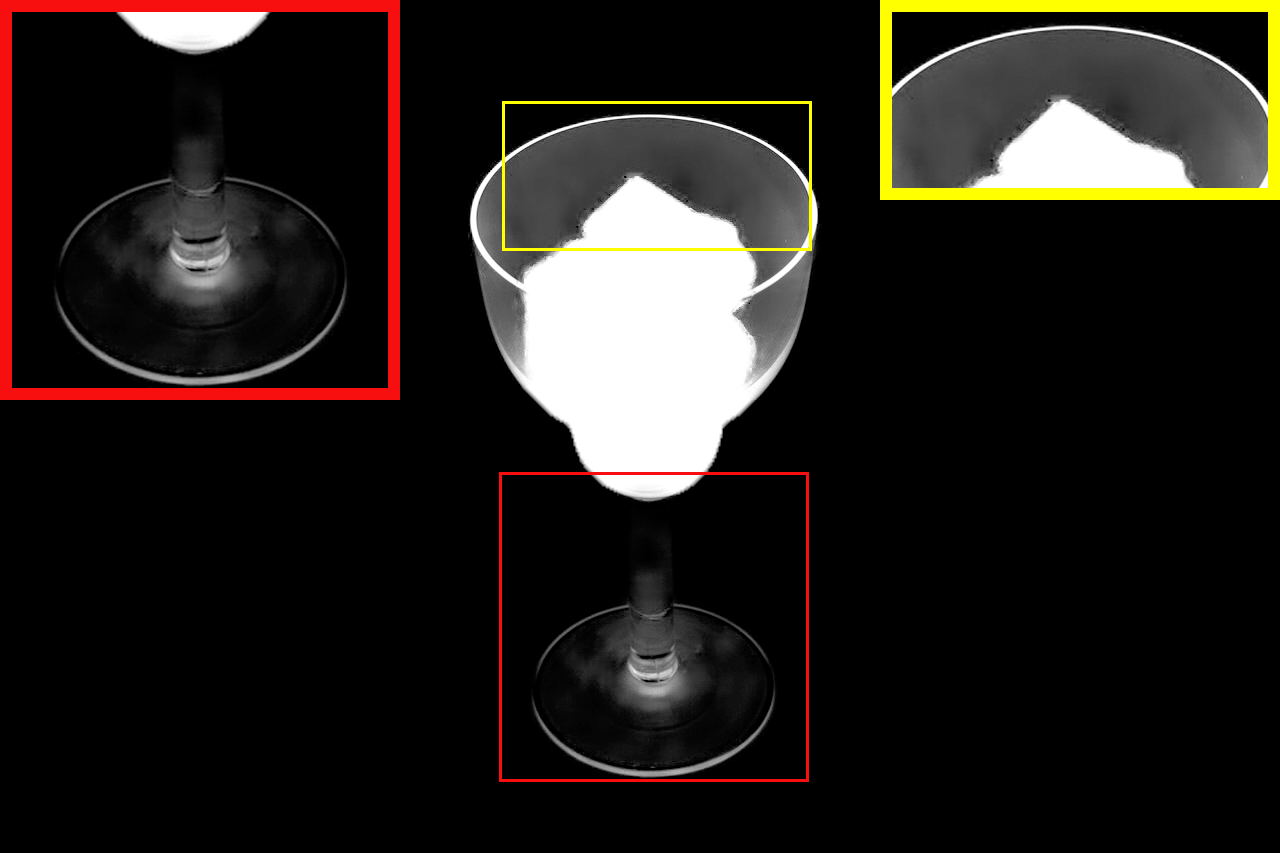} \\
			
			CA~\cite{hou2019context} & GCA~\cite{li2020natural} & A${^2}$U~\cite{dai2021learning} &  Late Fusion~\cite{Zhang2019CVPR} 
			& Ours & Ground Truth \\
			
			\includegraphics[scale=0.0333]{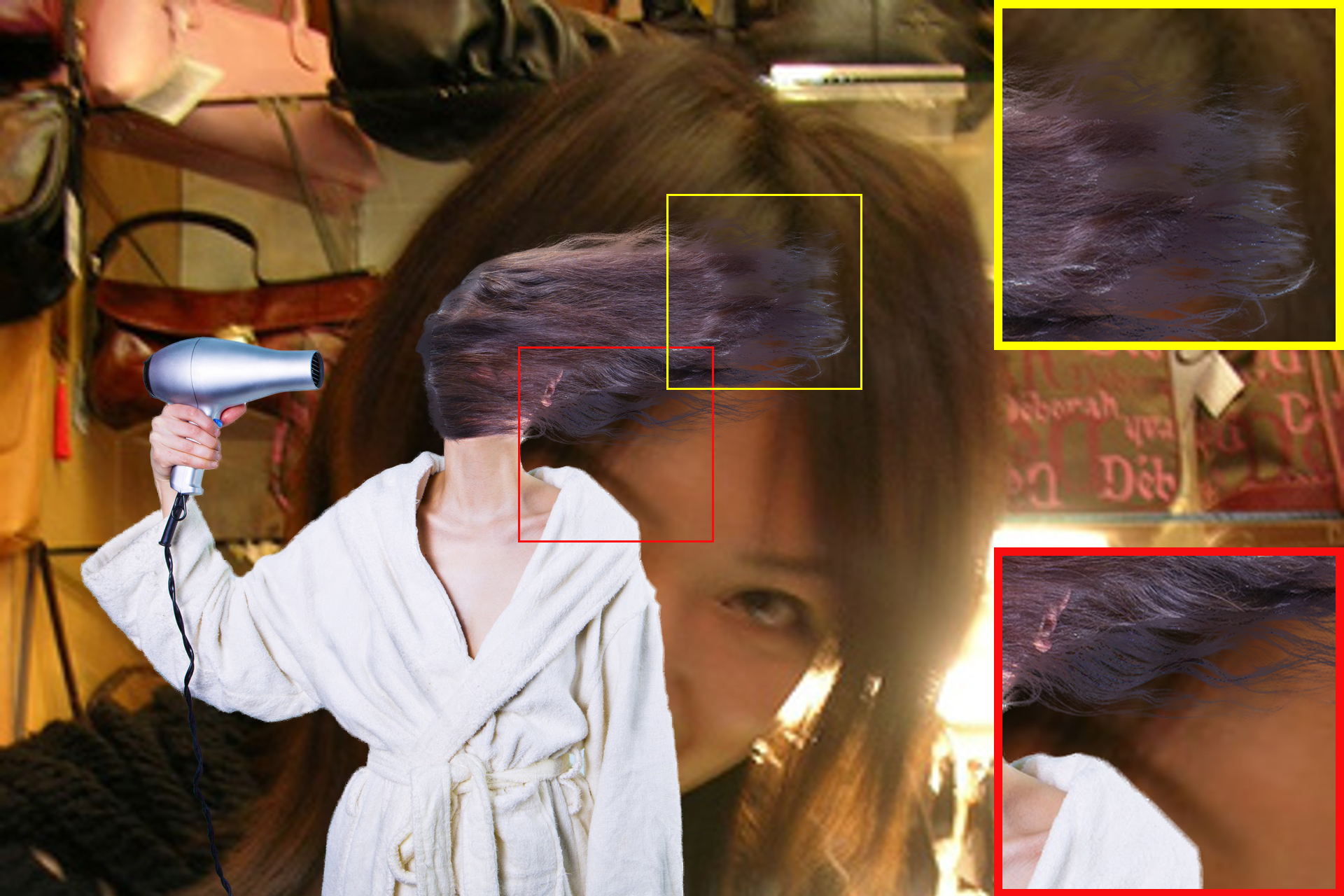} &
			\includegraphics[scale=0.0333]{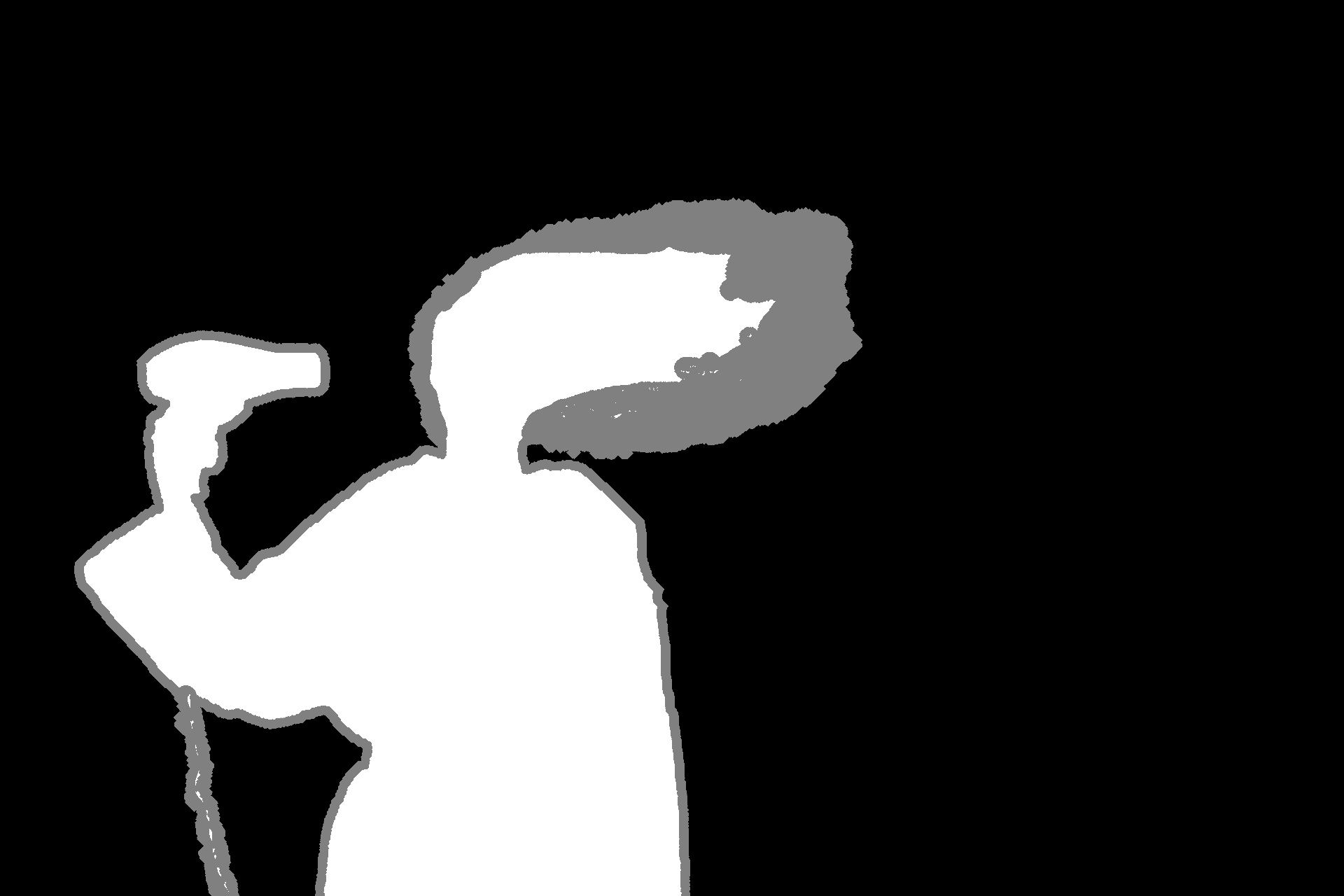} &
			\includegraphics[scale=0.0333]{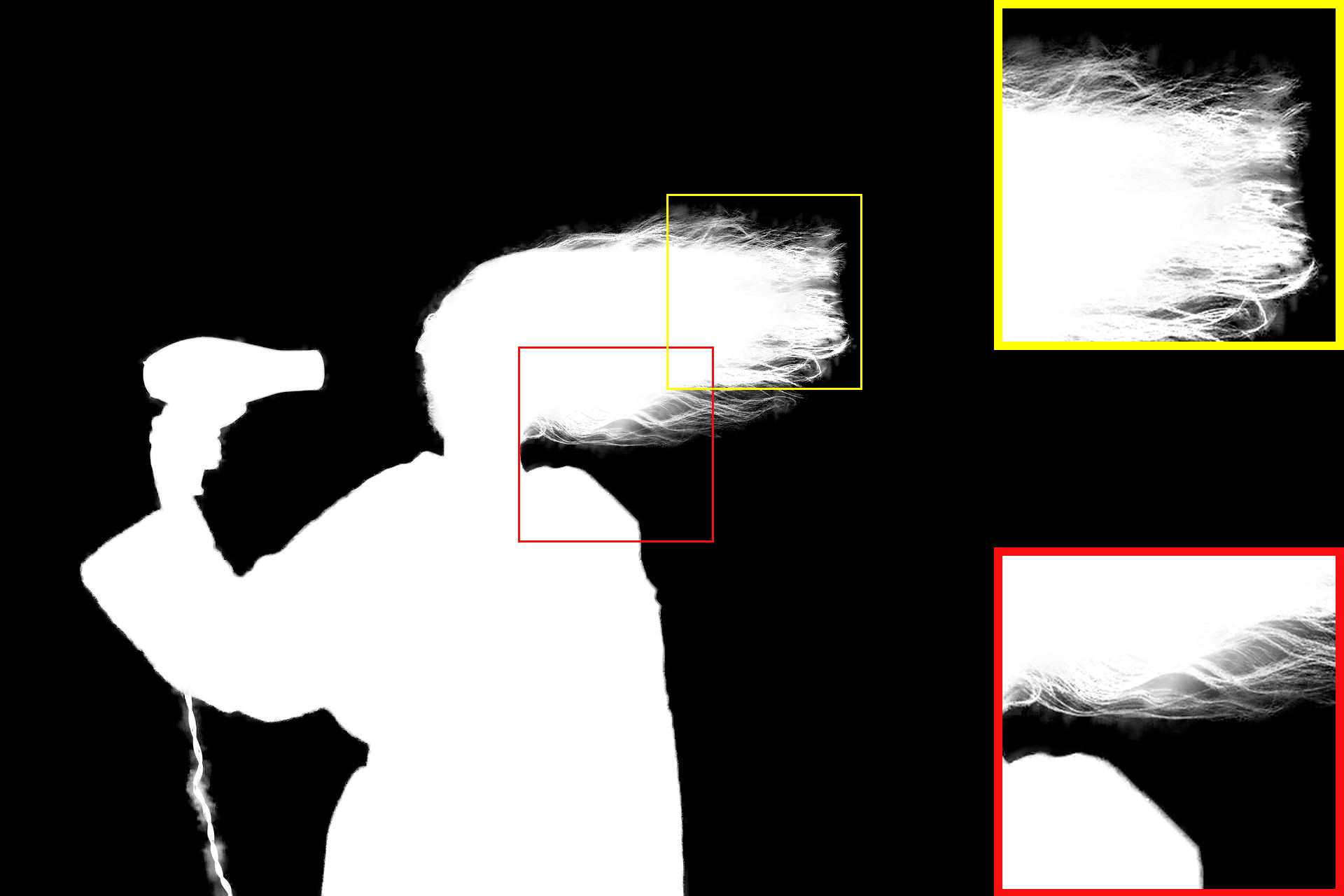} &
			\includegraphics[scale=0.0333]{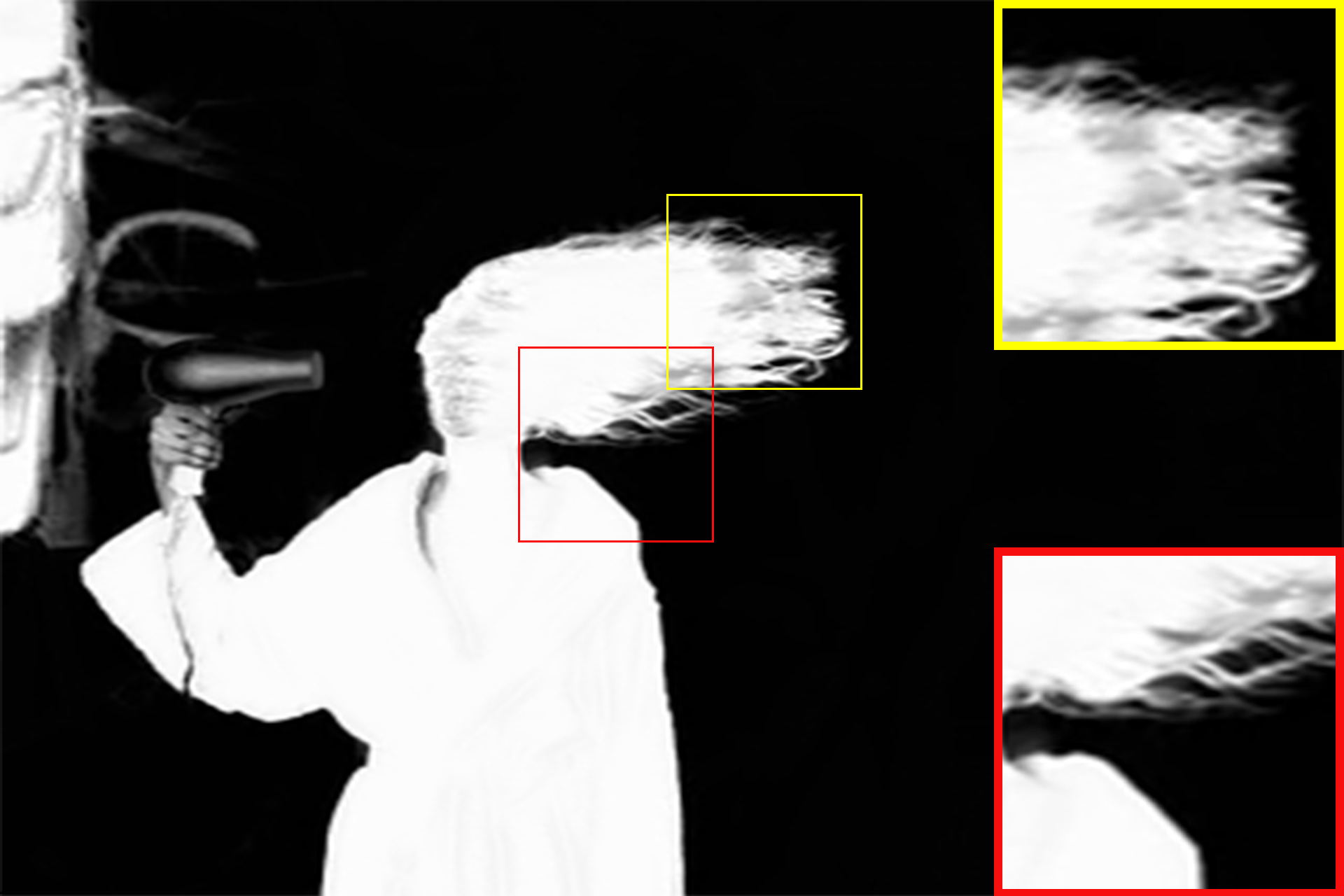} &
			\includegraphics[scale=0.0333]{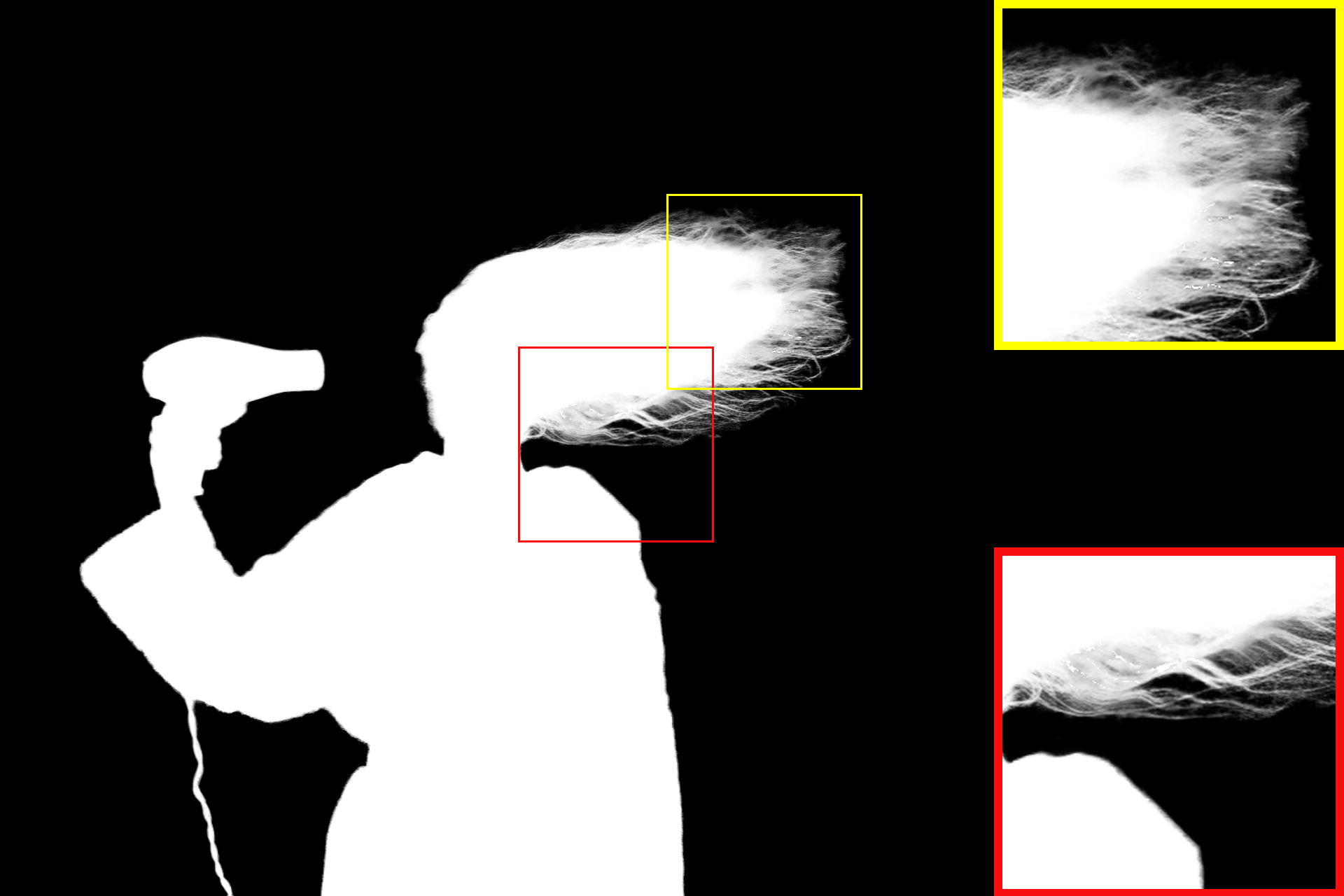} &
			\includegraphics[scale=0.0333]{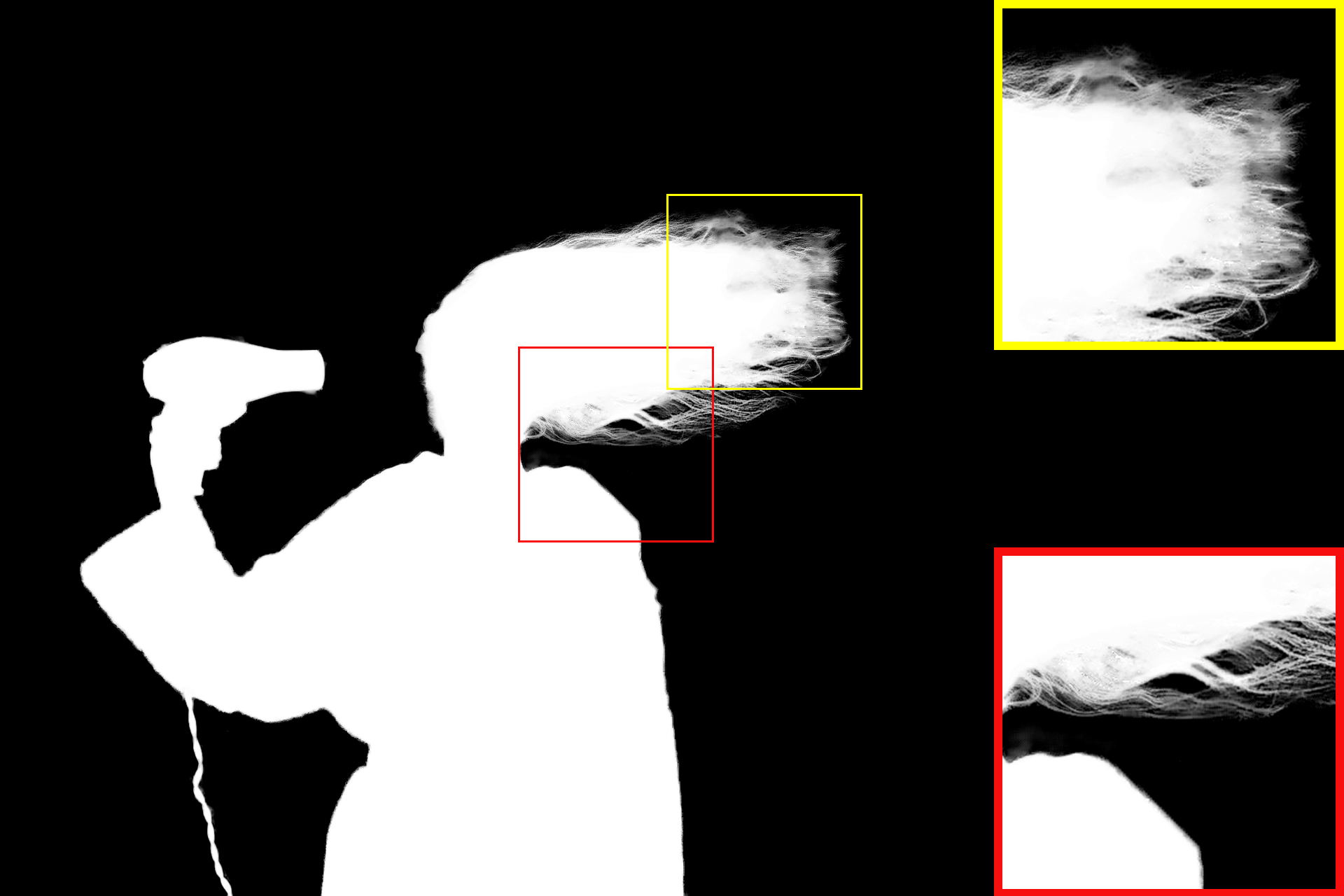} \\
			
			Input Image & Trimap & Closed Form~\cite{Levin2007A} & 
			SSS~\cite{sss} & DIM~\cite{Xu2017Deep} & IndexNet~\cite{hao2019indexnet} \\
			
			\includegraphics[scale=0.0333]{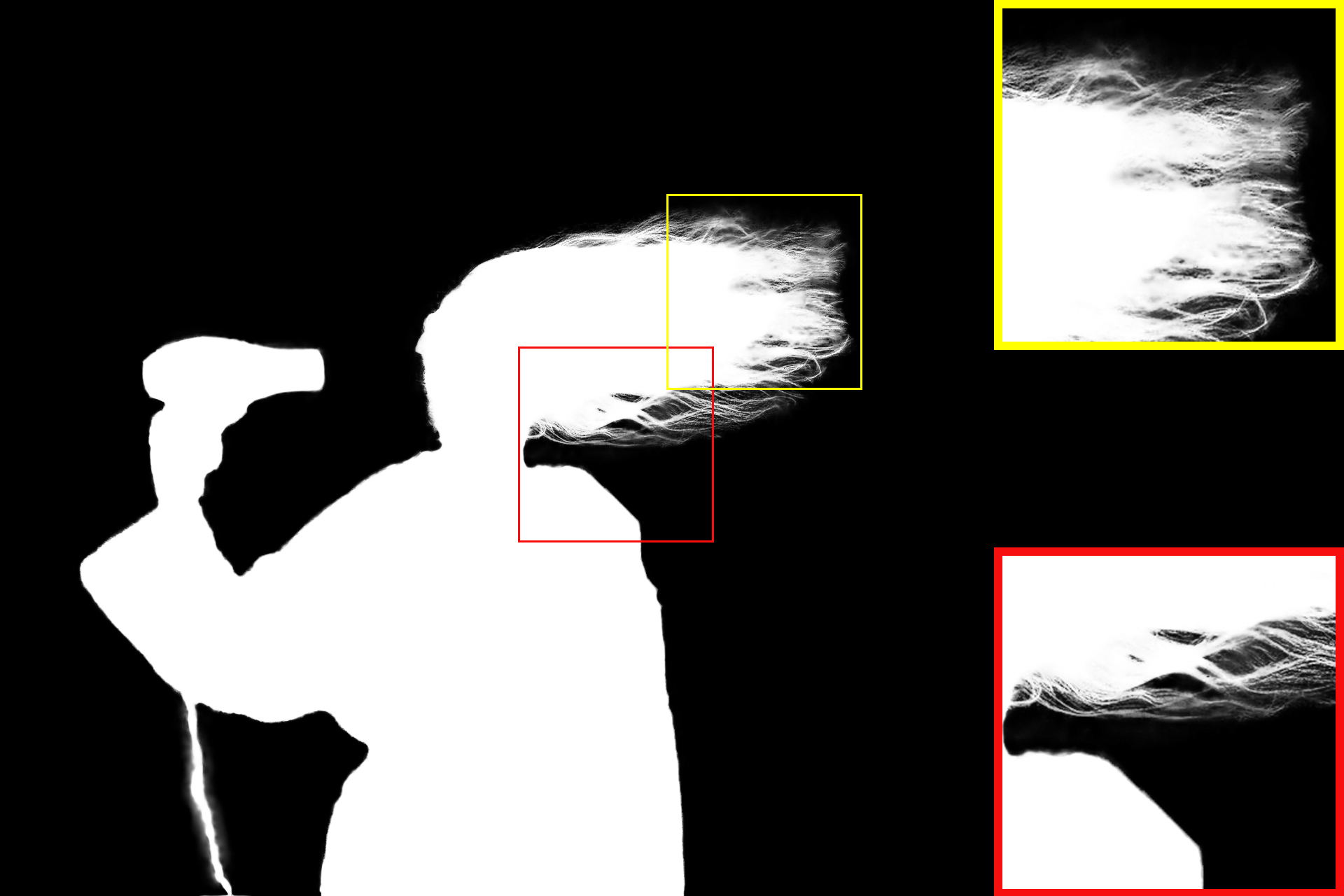} &
			\includegraphics[scale=0.0444]{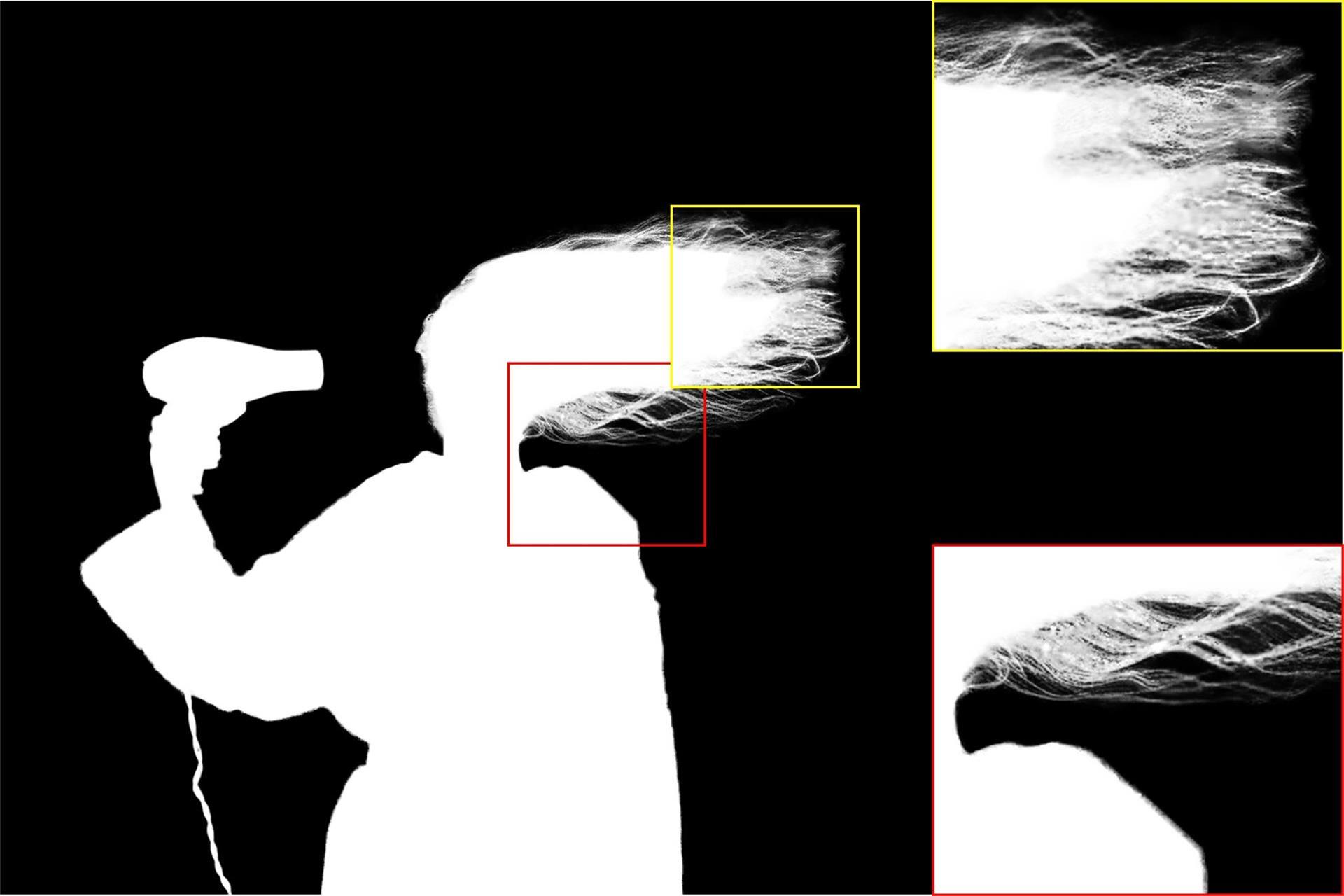} &
			\includegraphics[scale=0.0444]{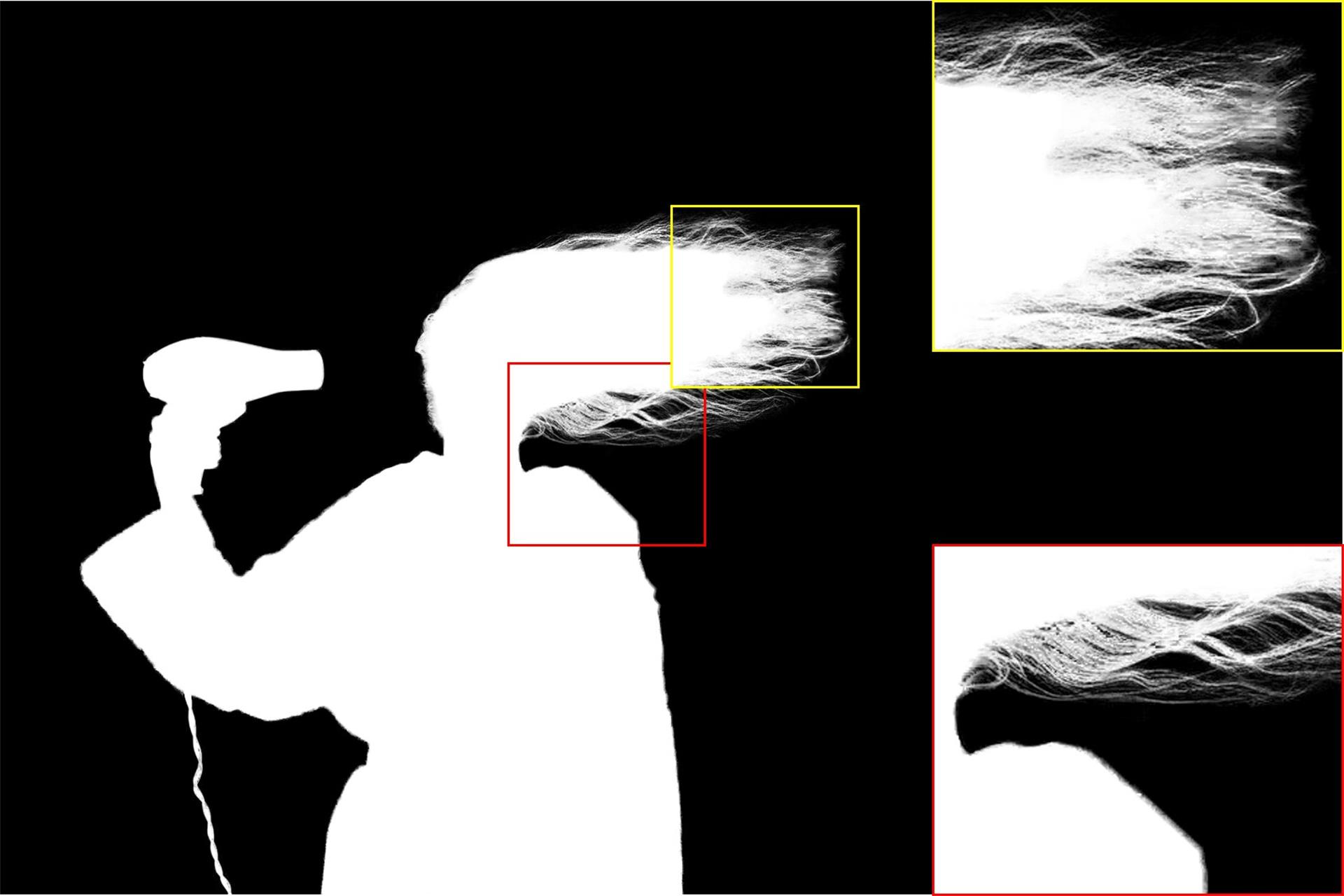} &
			\includegraphics[scale=0.0333]{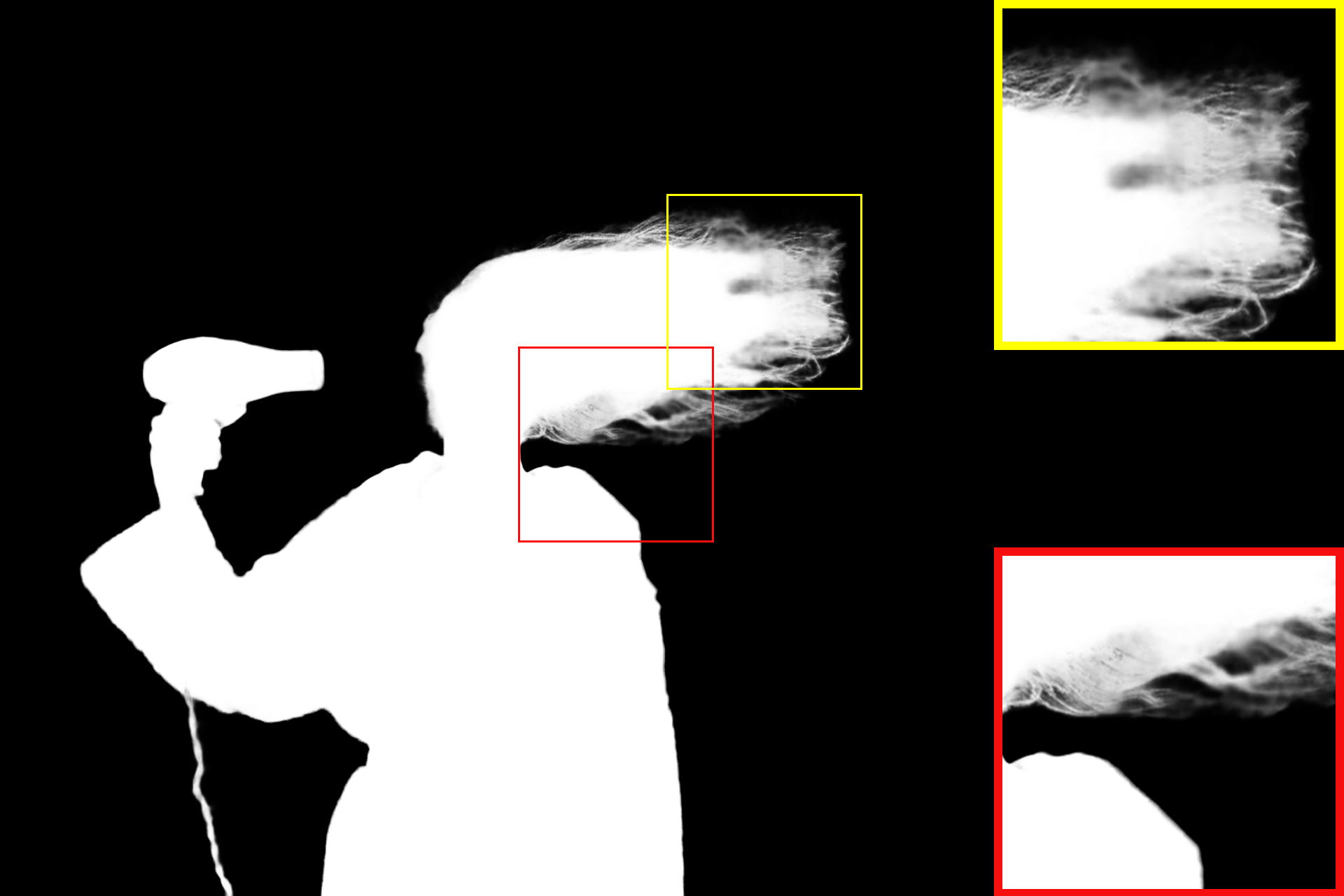} &
			\includegraphics[scale=0.0333]{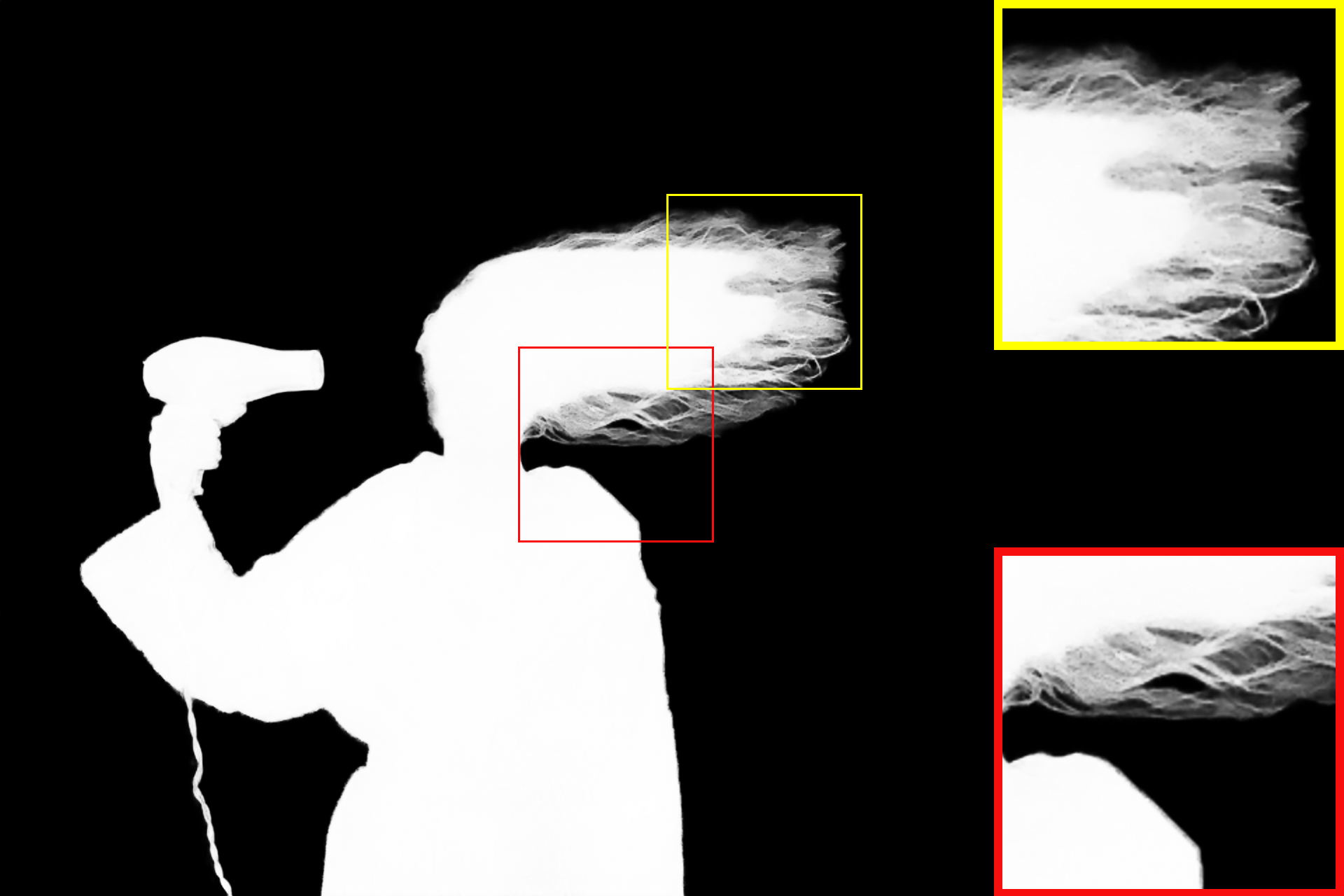} &
			\includegraphics[scale=0.0333]{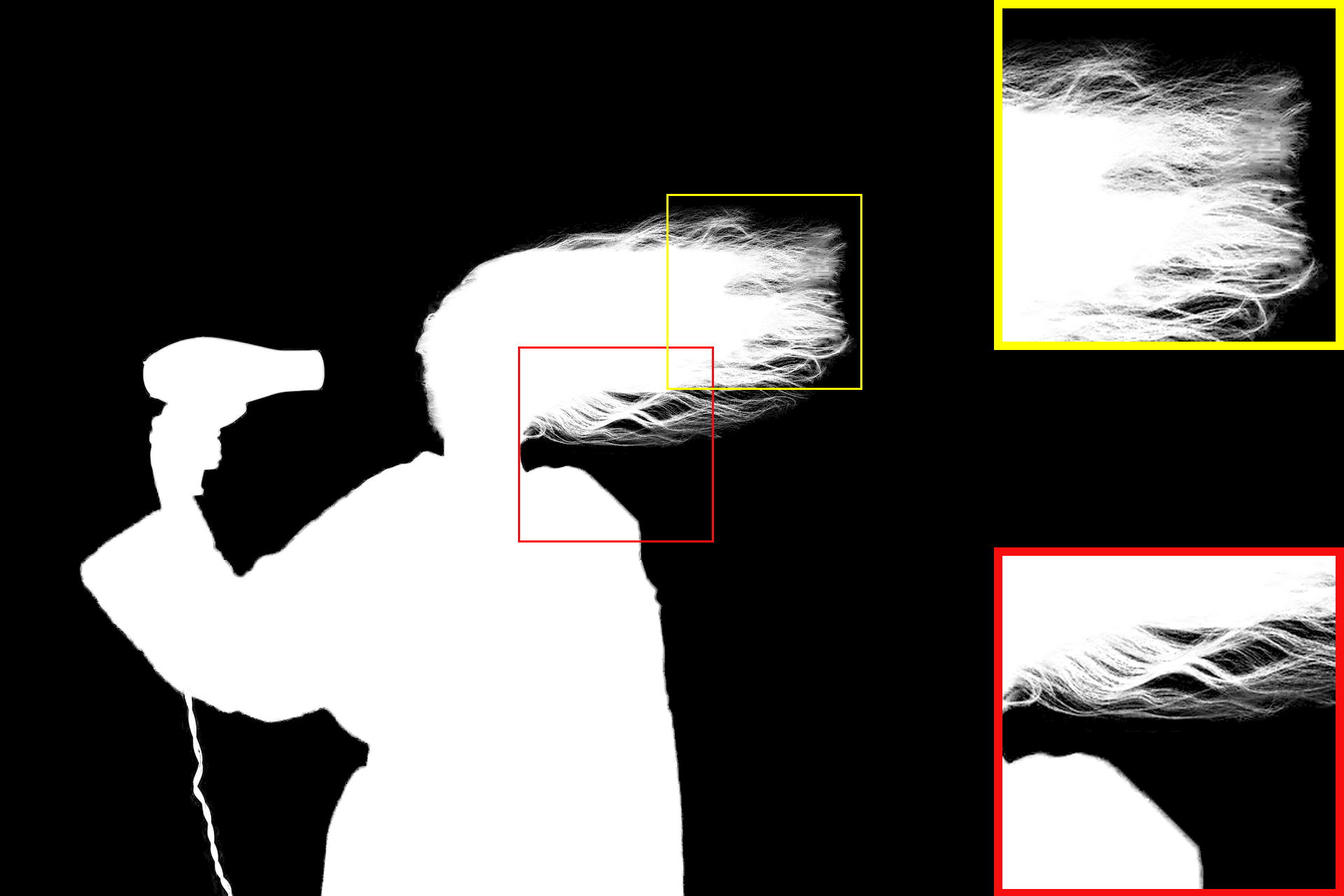} \\
			
			CA~\cite{hou2019context} & GCA~\cite{li2020natural} & A${^2}$U~\cite{dai2021learning} &  Late Fusion~\cite{Zhang2019CVPR} 
			& Ours & Ground Truth \\
			
	\end{tabular}}
	\caption{The visual comparisons on the Composition-1k test set. The segments in SSS~\cite{sss} are hand-picked.}
	\label{fig:visual_adobe}
\end{figure*}
\textbf{Evaluation metrics.} We evaluate the alpha mattes produced by our \emph{HAttMatting++} following the four common quantitative metrics: the sum of absolute difference(SAD), mean square error(MSE), the gradient(Grad), and connectivity(Conn) proposed by\cite{rhemann2009perceptually}. A better image matting method can mostly produce high-quality alpha mattes, thus reducing the values of the above four metrics.

\subsection{Comparison to the State-of-the-art}
\label{ssec:comparisons}
\textbf{Evaluation on the Composition-1k test set.} Here we compare \emph{HAttMatting++} with $6$ traditional hand-crafted algorithms: Shared Matting~\cite{gastal2010shared}, Learning Based~\cite{zheng2009learning}, Global Matting~\cite{Rhemann2011A}, ClosedForm~\cite{Levin2007A}, KNN Matting~\cite{Chen2013KNN}, Information-Flow~\cite{Aksoy2017Designing} as well as $8$ deep learning-based methods:
\begin{table}[t]
	\renewcommand{\arraystretch}{1.3}
	\setlength{\belowcaptionskip}{2pt}
	\caption{The Quantitative Comparisons on Composition-1k Testing Set. ``human" means the metrics are only calculated on the human images.}
	\label{tab:quantitative_adobe}
	\small
	\centering
	\setlength{\tabcolsep}{0.6mm}{
		\begin{tabular}{l|ccccc}
			\hline
			Methods & SAD$\downarrow$ & MSE$\downarrow$ & Gradient$\downarrow$ & Connectivity$\downarrow$ & Additional input \\
			\hline
			Shared Matting~\cite{gastal2010shared} & 125.37 & 0.029 & 144.28 & 123.53 & Trimap \\
			Learning Based~\cite{zheng2009learning} & 95.04 & 0.018 & 76.63 & 98.92 & Trimap \\
			Global Matting~\cite{Rhemann2011A} & 156.88 & 0.042 & 112.28 & 155.08 & Trimap \\
			ClosedForm~\cite{Levin2007A} & 124.68 & 0.025 & 115.31 & 106.06 & Trimap \\
			KNN Matting~\cite{Chen2013KNN} & 126.24 & 0.025 & 117.17 & 131.05 & Trimap \\
			DCNN~\cite{cho2019deep} & 115.82 & 0.023 & 107.36 & 111.23 & Trimap \\
			Information-Flow~\cite{Aksoy2017Designing} & 70.36 & 0.013 & 42.79 & 70.66 & Trimap \\
			DIM~\cite{Xu2017Deep} & 48.87 & 0.008 & 31.04 & 50.36 & Trimap \\
			AlphaGAN~\cite{lutz2018alphagan} & 90.94 & 0.018 & 93.92 & 95.29 & Trimap \\
			SampleNet~\cite{Tang_2019_CVPR} & 48.03 & 0.008 & 35.19 & 56.55 & Trimap \\
			IndexNet~\cite{hao2019indexnet} & 44.52 & 0.005 & 29.88 & 42.37 & Trimap \\
			CA Matting~\cite{hou2019context} & 38.73 & 0.004 & 26.13 & 35.89 & Trimap \\
			GCA Matting~\cite{li2020natural} & 35.27 & 0.0034 & 19.72 & 31.93 & Trimap \\
			A${^2}$U~\cite{dai2021learning} & 33.78 & 0.0037 & 18.04 & 31.00 & Trimap \\
			\hline
			\hline
			\rowcolor{mygray}
			Late Fusion~\cite{Zhang2019CVPR} & 58.34 & 0.011 & 41.63 & 59.74 & No \\
			\rowcolor{mygray}
			MSIA~\cite{Qiao2020multi} & 47.86 & 0.007 & 28.61 & 43.39 & No \\
			\rowcolor{mygray}
			HAttMatting~\cite{Qiao_2020_CVPR} & 44.01 & 0.007 & 29.26 & 46.41 & No \\
			\hline
			\hline
			Basic & 126.31 & 0.025 & 111.35 & 118.71 & No \\
			Basic + SSIM & 102.79 & 0.021 & 88.04 & 110.14 & No \\
			Basic + Low & 89.39 & 0.016 & 56.67 & 90.03 & No \\
			Basic + CA  & 96.67 & 0.018 & 73.94 & 95.08 & No \\
			Basic + Low + CA & 72.73 & 0.013 & 49.53 & 65.92 & No \\
			Basic + Low + SA & 54.91 & 0.011 & 46.21 & 60.40 & No \\
			Basic + Low + CA + SA & 49.67 & 0.009 & 41.11 & 53.76 & No \\
			\hline
			\hline
			\rowcolor{mygray}
			HAttMatting + multi-scale & 45.69 & 0.0065 & 28.61 & 45.83 & No \\
			\rowcolor{mygray}
			HAttMatting + Sentry & 43.46 & 0.0064 & 28.17 & 46.11 & No \\
			\rowcolor{mygray}
			\emph{HAttMatting++} & \textbf{43.27} & \textbf{0.0059} & \textbf{27.91} & \textbf{44.09} & No \\
			\hline
			BGMv2 & 15.56 & 0.012 & 11.46 & 13.94 & Background images \\
			HAttMatting++\_human & 18.40 & 0.011 & 8.59 & 16.13 & No \\
			\hline	
	\end{tabular}}
\end{table}
DCNN~\cite{cho2019deep}, DIM~\cite{Xu2017Deep}, AlphaGAN~\cite{lutz2018alphagan}, SSS~\cite{sss}, SampleNet~\cite{Tang_2019_CVPR}, Context-aware~\cite{hou2019context}, IndexNet~\cite{hao2019indexnet}, MSIA~\cite{Qiao2020multi}, GCA~\cite{li2020natural}, A${^2}$U~\cite{dai2021learning}, Late Fusion~\cite{Zhang2019CVPR} and BGMv2~\cite{lin2021real}. The evaluation codes of the above methods are from original papers or our implementations using recommended parameters.

SSS~\cite{sss}, MSIA~\cite{Qiao2020multi}, Late Fusion~\cite{Zhang2019CVPR}, and our \emph{HAttMatting} series can generate alpha mattes without trimap. For the other methods that rely on trimaps as input, we can generate transition regions along with the areas where alpha values are between 0 and 1, using random dilation in the range of [1, 25]. We harness full-resolution input images for fair contrast and the visual results are illustrated in Fig.~\ref{fig:visual_adobe}. The quantitative comparisons are reported in Table~\ref{tab:quantitative_adobe}, and the four metrics are all calculated on the whole image. In Table~\ref{tab:quantitative_adobe}, the methods in gray (the Late Fusion and our \emph{HAttMatting} series) only take \emph{RGB} images as input, while the others require trimap as assistance to guarantee the accuracy of alpha mattes. ``Basic'' means our baseline network, and the corresponding ``Basic +'' represents that we assemble different components on the baseline to generate alpha mattes. Here ``HAttMatting'' means our previous published conference paper. ``HAttMatting + multi-scale'' and ``HAttMatting + Sentry'' indicate that we add the secondary appearance cues and sentry supervision respectively. \emph{HAttMatting++} denotes the current full model (with multi-scale appearance cues and sentry supervision).

Compared to the conventional hand-crafted methods, the \emph{HAttMatting++} exhibits significant superiority over them, which can be clearly observed in Fig.~\ref{fig:visual_adobe} and Table~\ref{tab:quantitative_adobe}. As for the deep learning-based approaches, we divide them into two categories: trimap-free methods and trimap-based ones. As shown in Table~\ref{tab:quantitative_adobe} and Fig.~\ref{fig:visual_adobe}, compared with the trimap-free methods of SSS~\cite{sss}, MSIA~\cite{Qiao2020multi}, and Late Fusion~\cite{Zhang2019CVPR}, our method can surpass them by a large margin both quantitatively and qualitatively. Since the results of SSS are separate semantic blocks, and the inference process is very time-consuming, we segment some sample images, then manually select and splice different semantic blocks to form the final alpha mattes. Similar results could be found in~\cite{sss}. 

\begin{figure*}[b]
	\setlength{\tabcolsep}{1pt}\small{
		\begin{tabular}{cccccc}
			\includegraphics[scale=0.032]{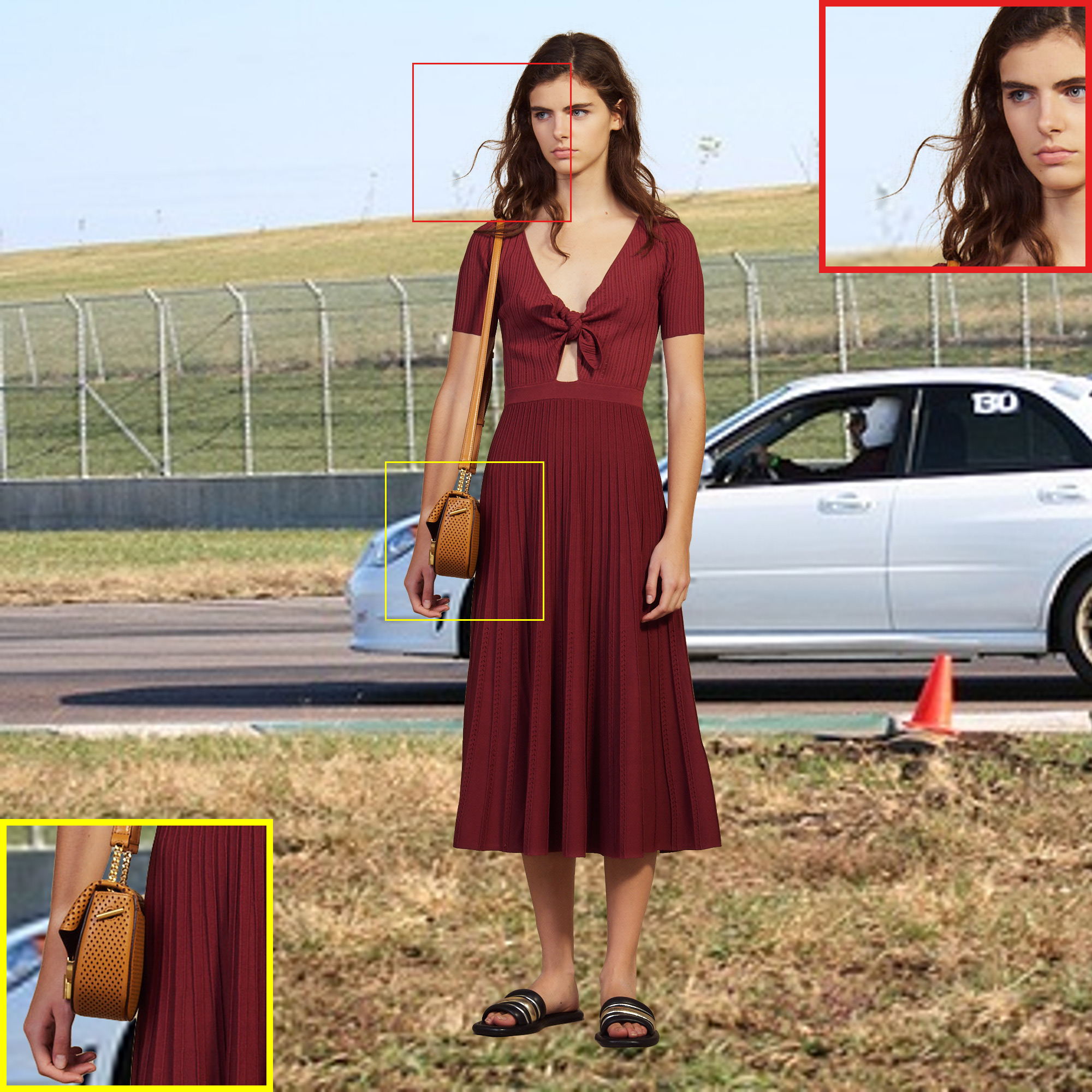} &
			\includegraphics[scale=0.032]{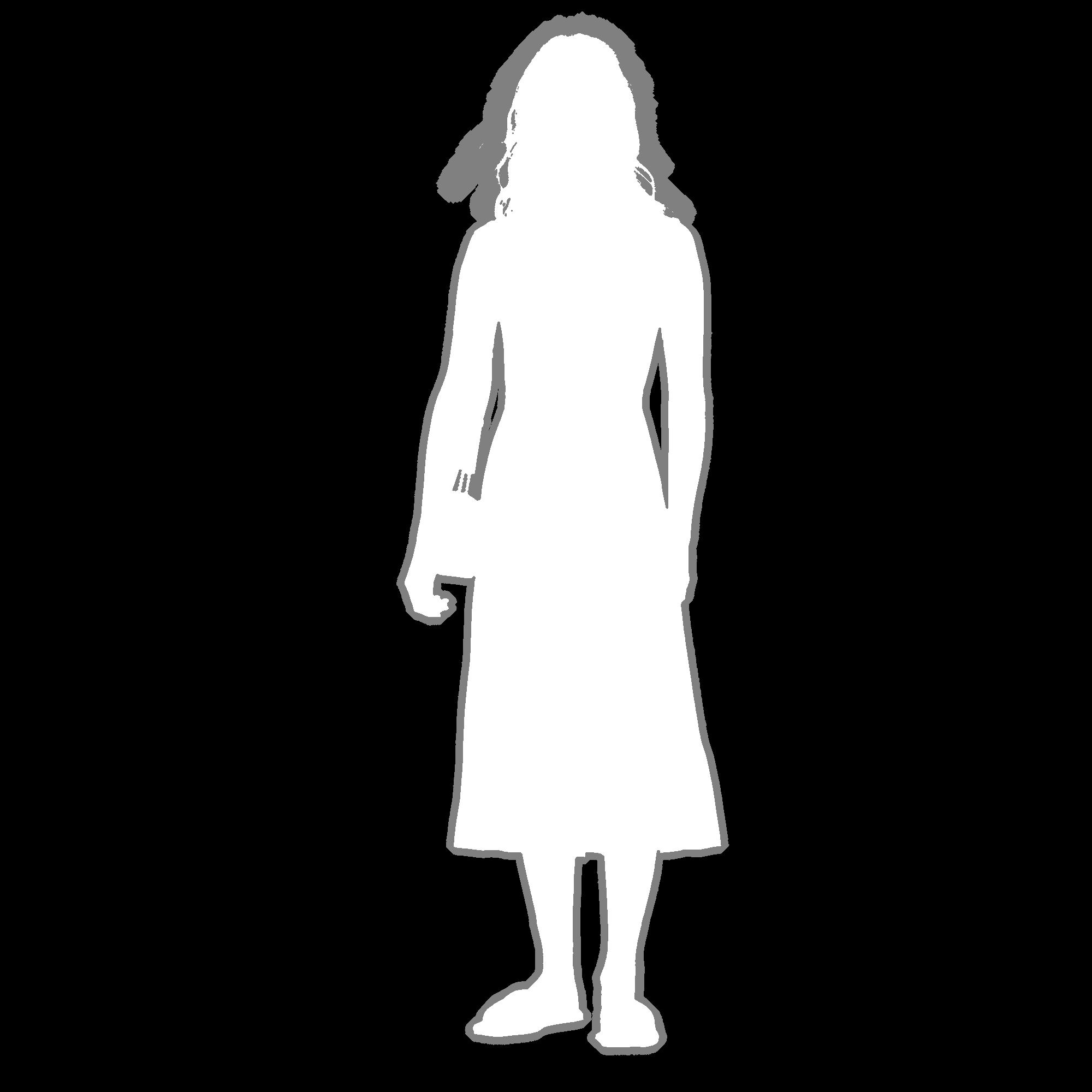} &
			\includegraphics[scale=0.032]{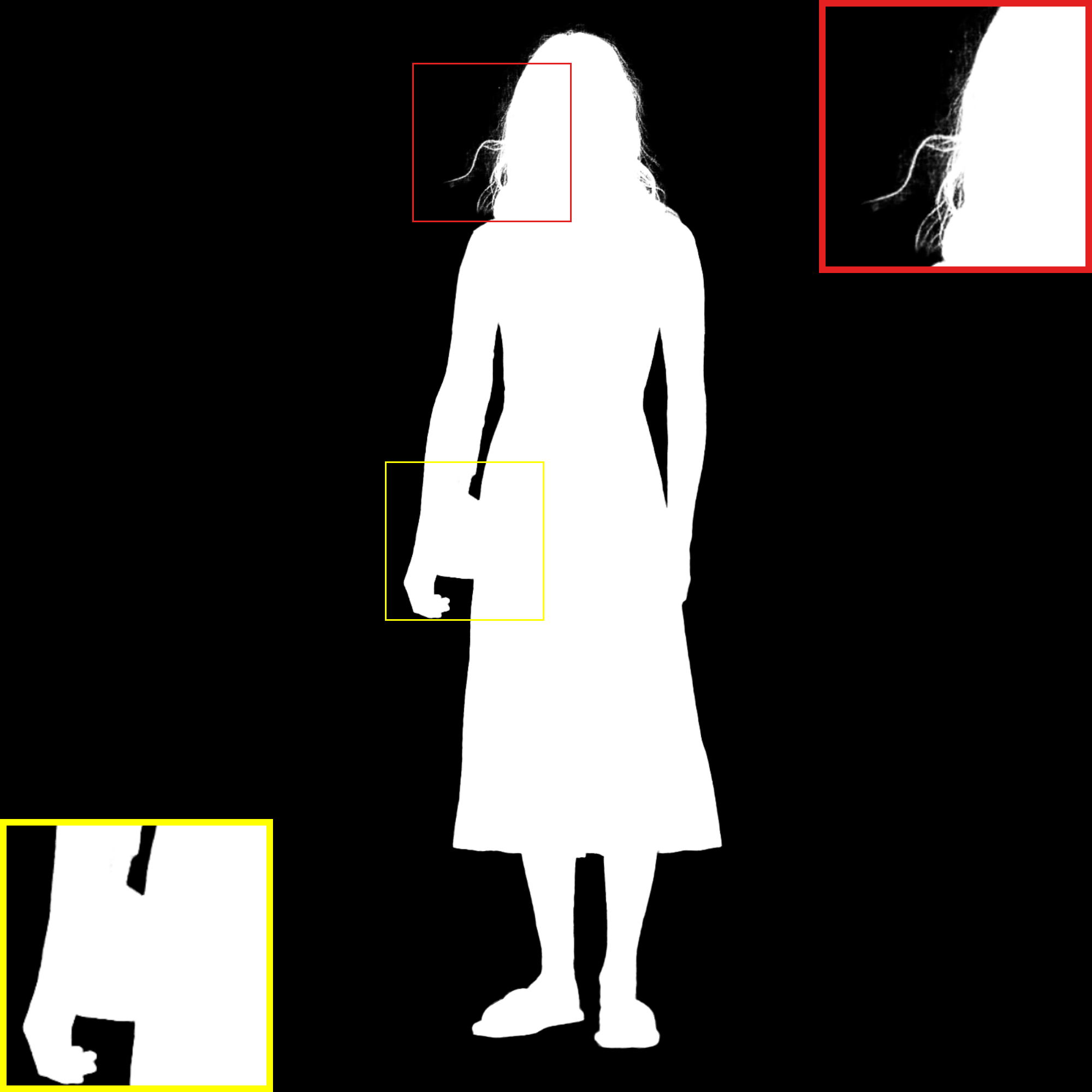} &
			\includegraphics[scale=0.032]{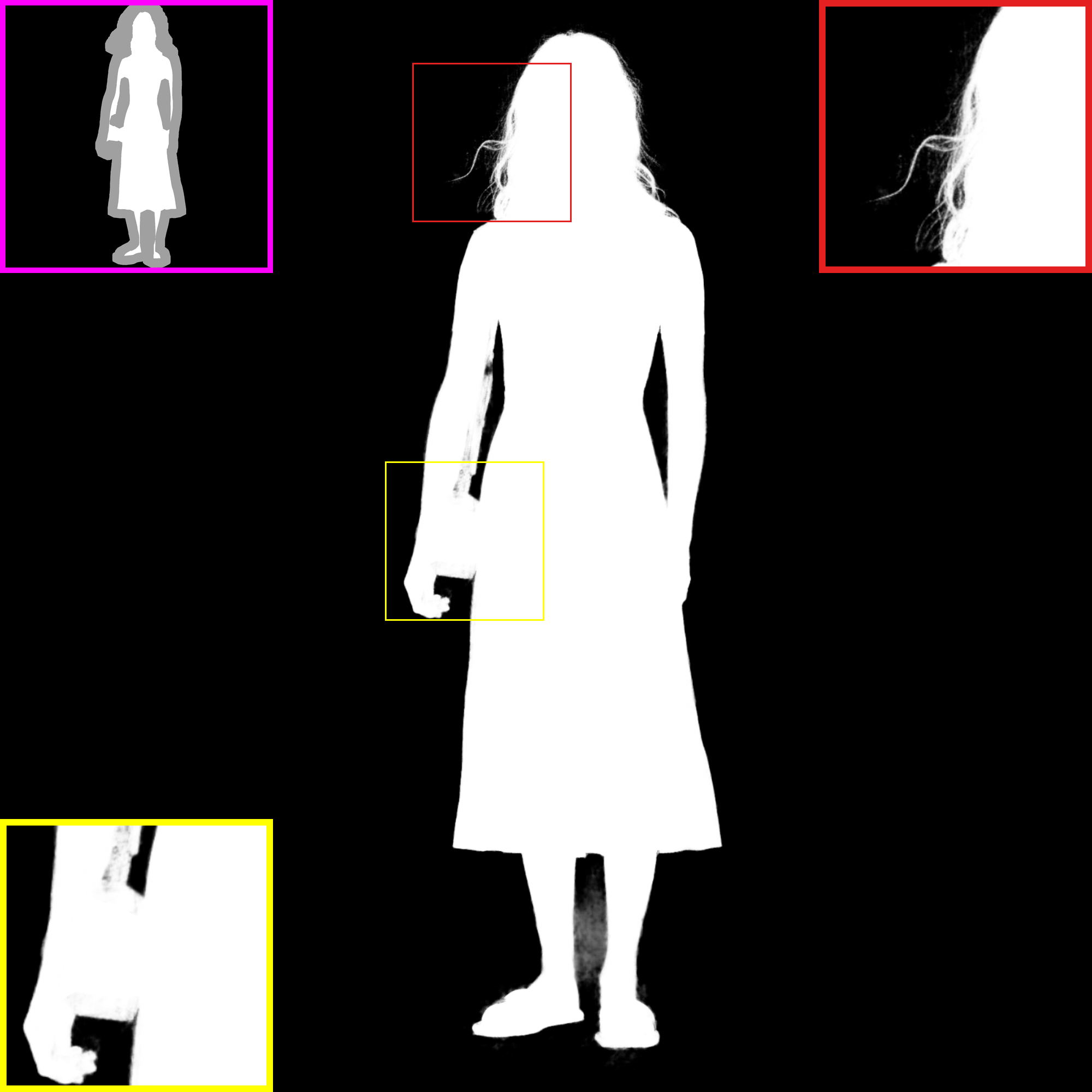} &
			\includegraphics[scale=0.032]{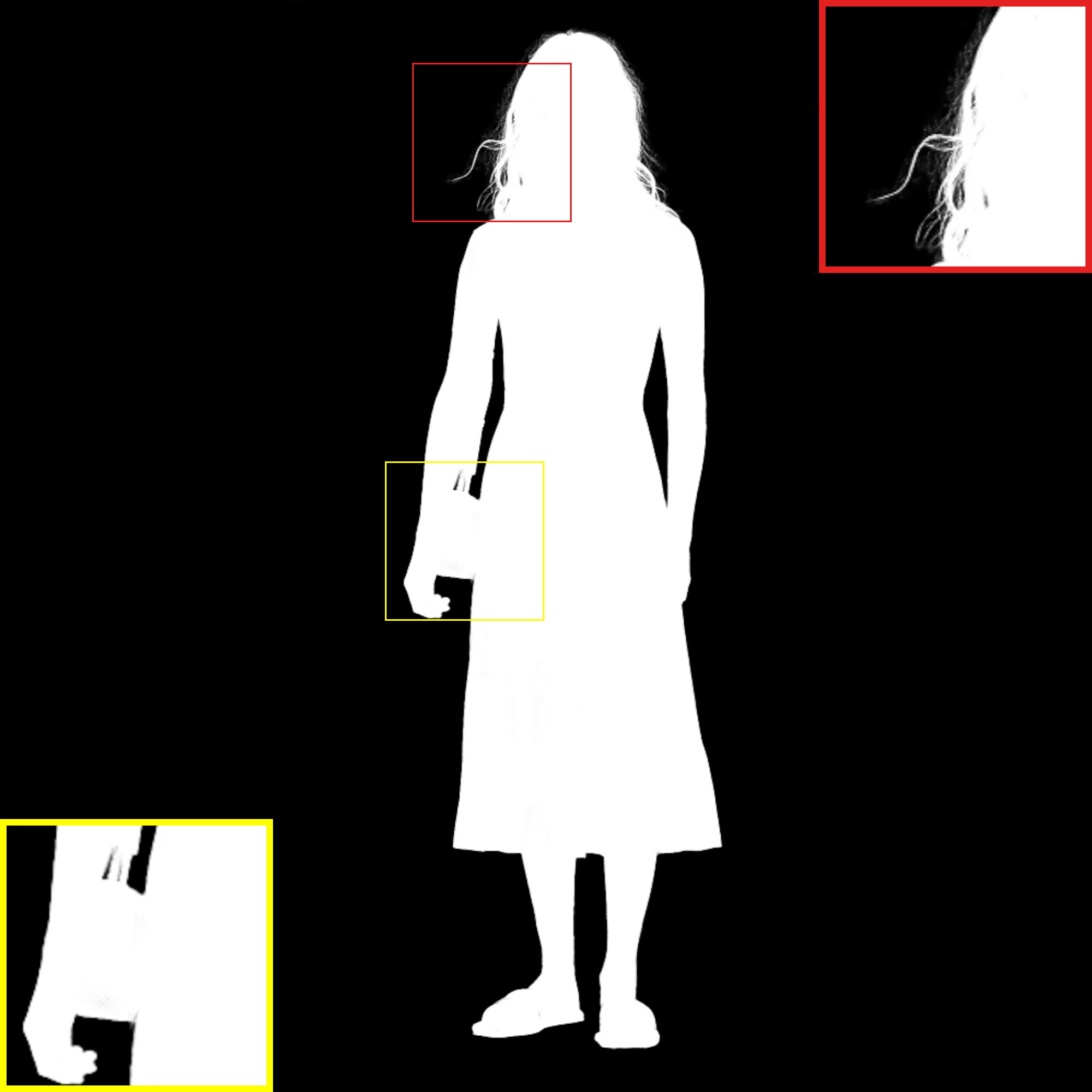} &
			\includegraphics[scale=0.032]{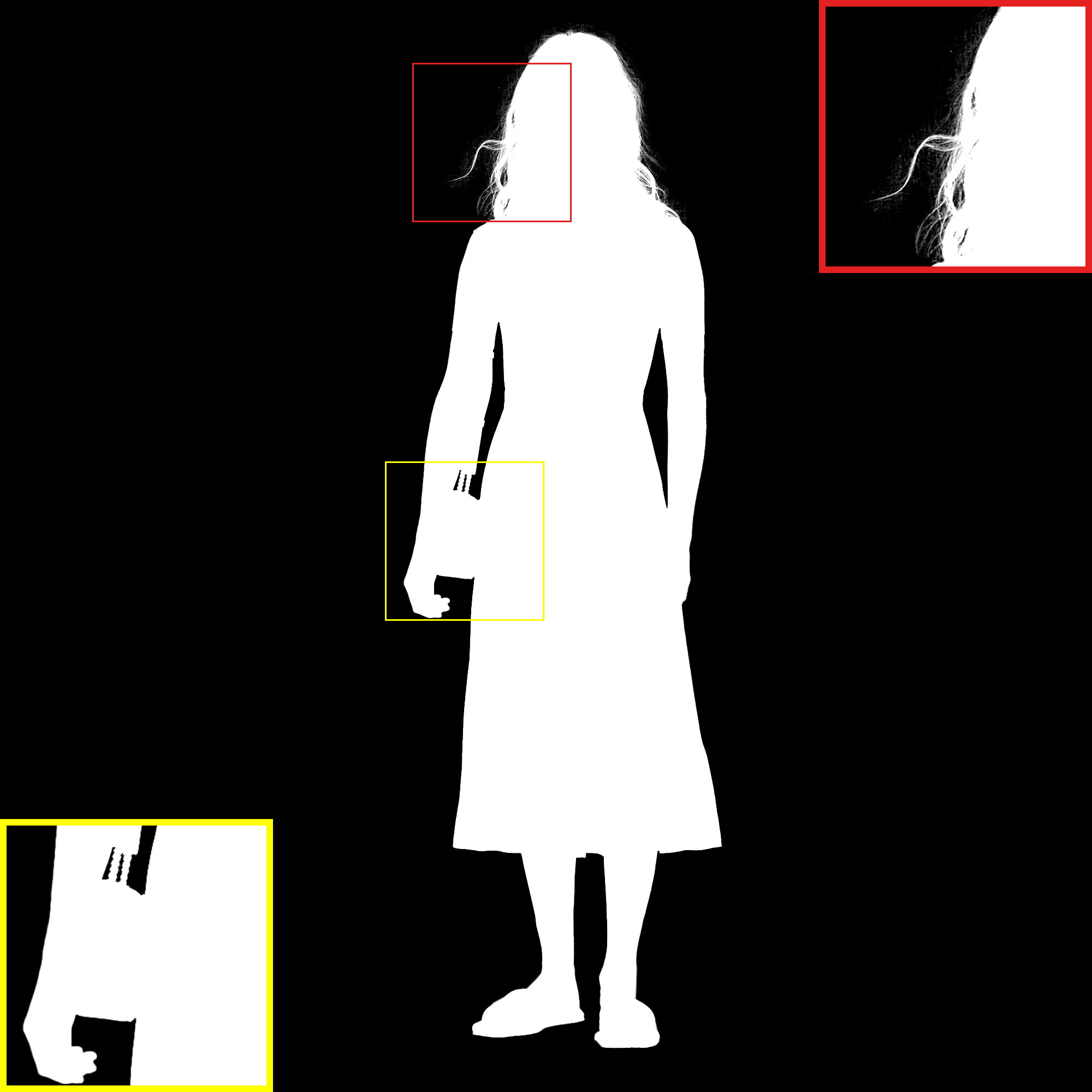} \\
			
			\includegraphics[scale=0.034]{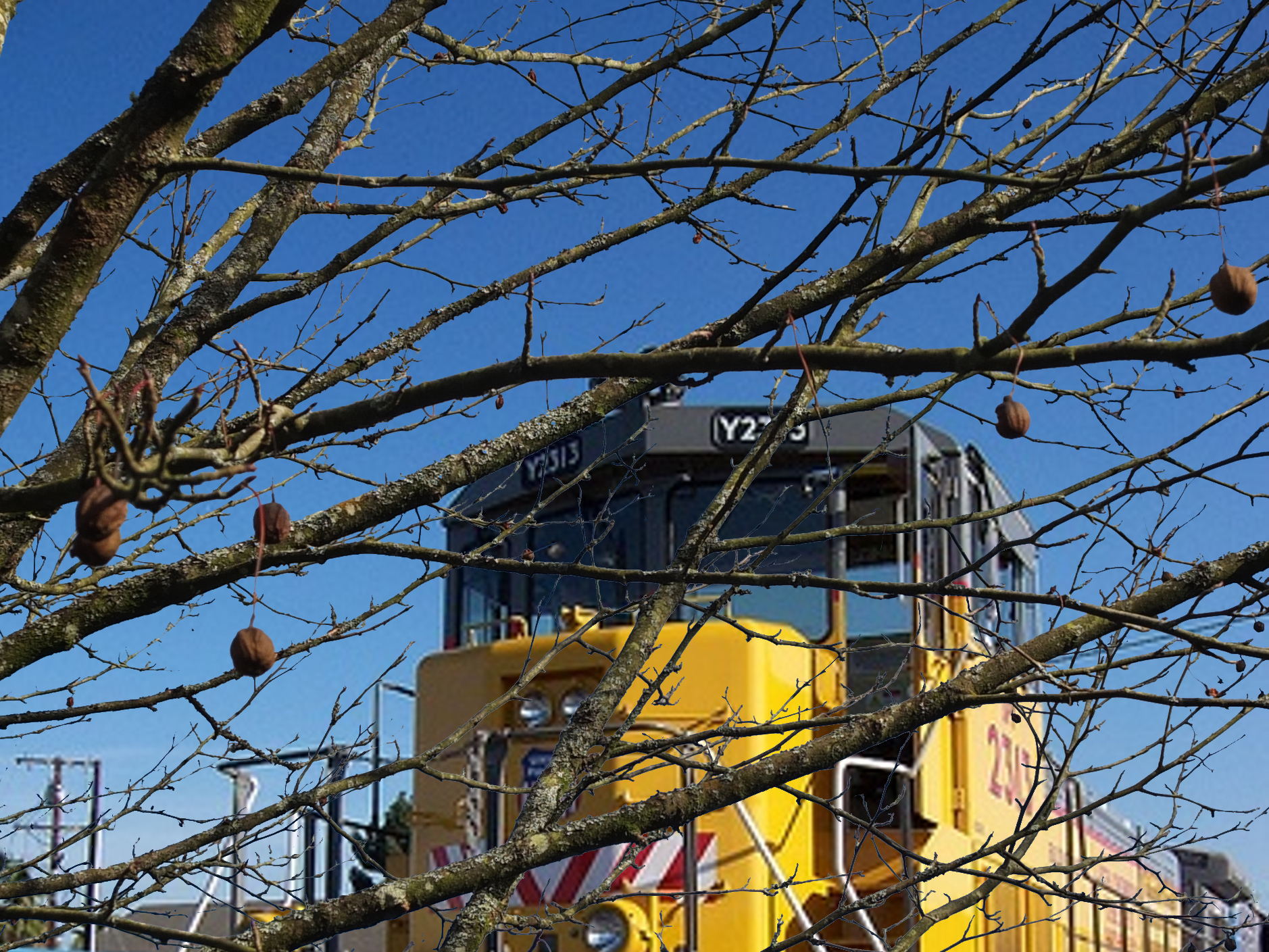} &
			\includegraphics[scale=0.034]{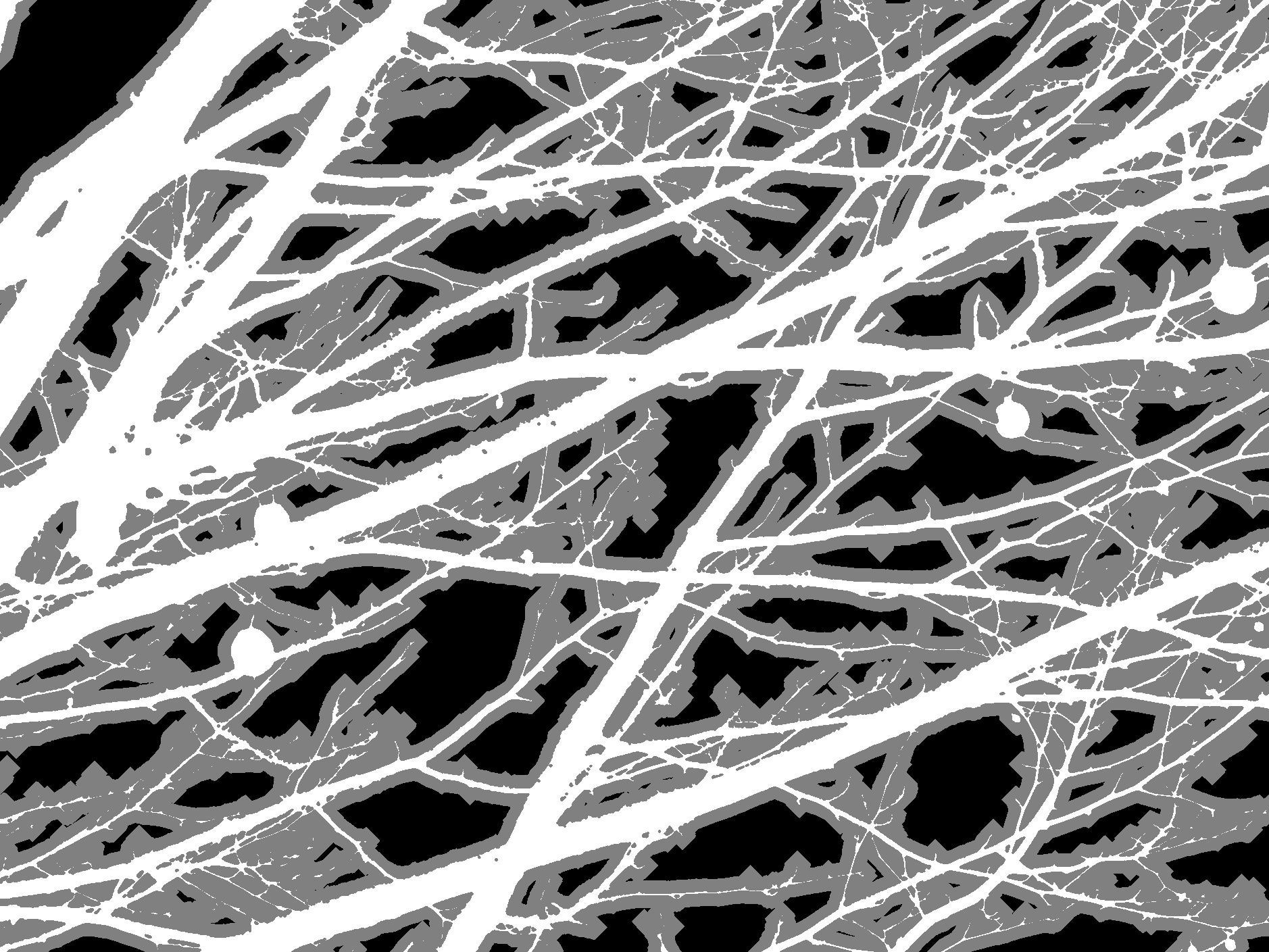} &
			\includegraphics[scale=0.034]{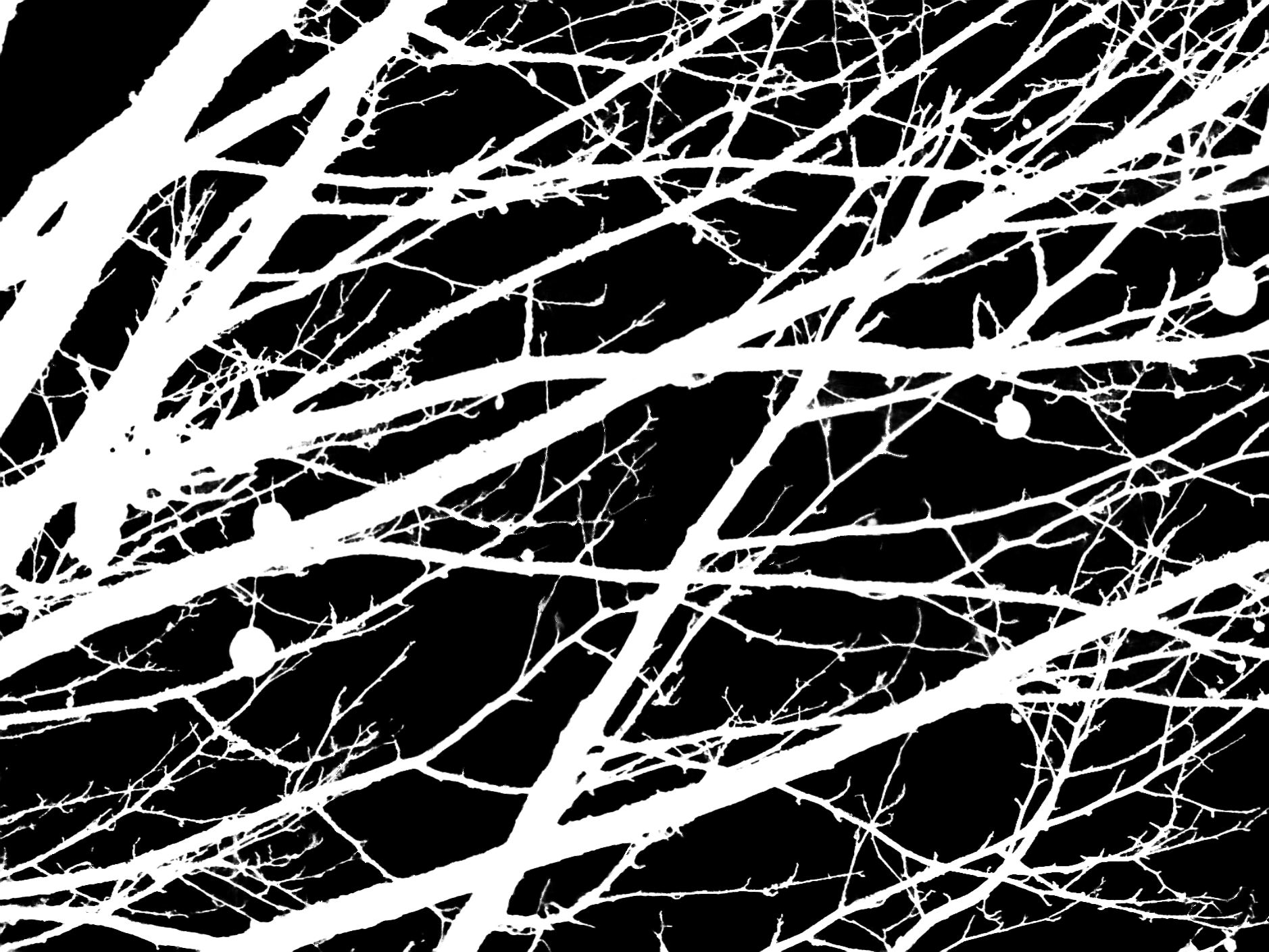} &
			\includegraphics[scale=0.034]{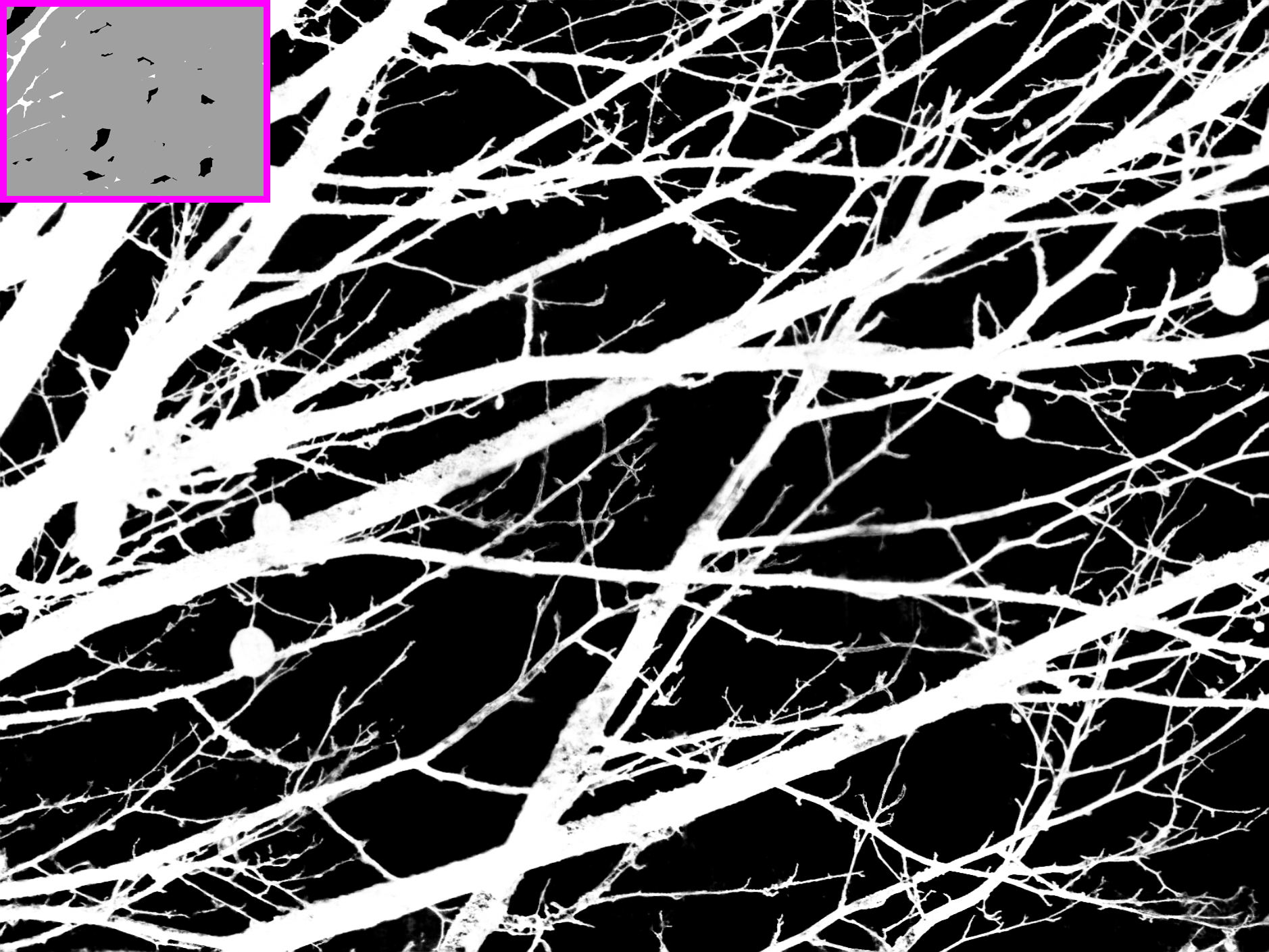} &
			\includegraphics[scale=0.034]{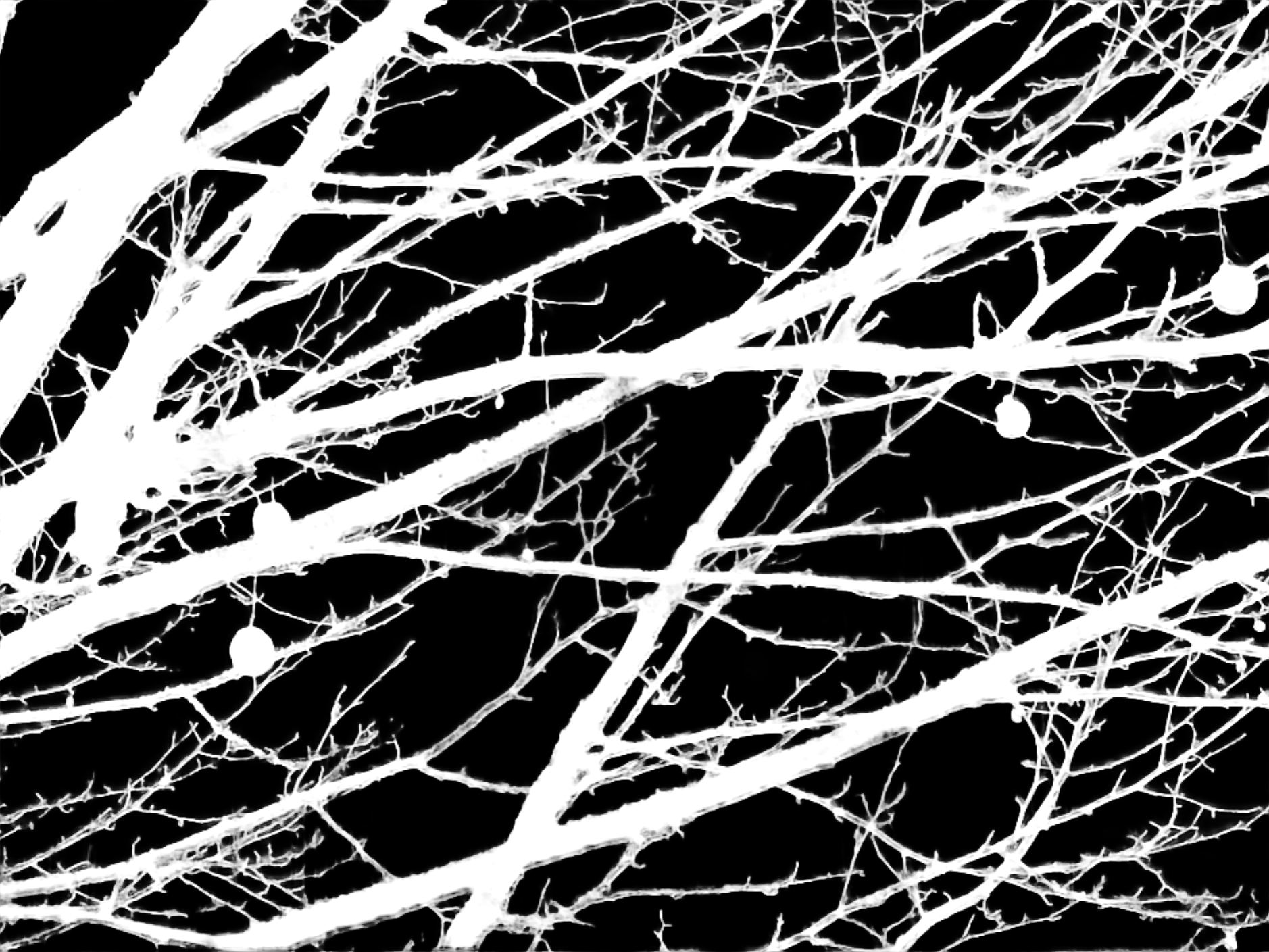} &
			\includegraphics[scale=0.034]{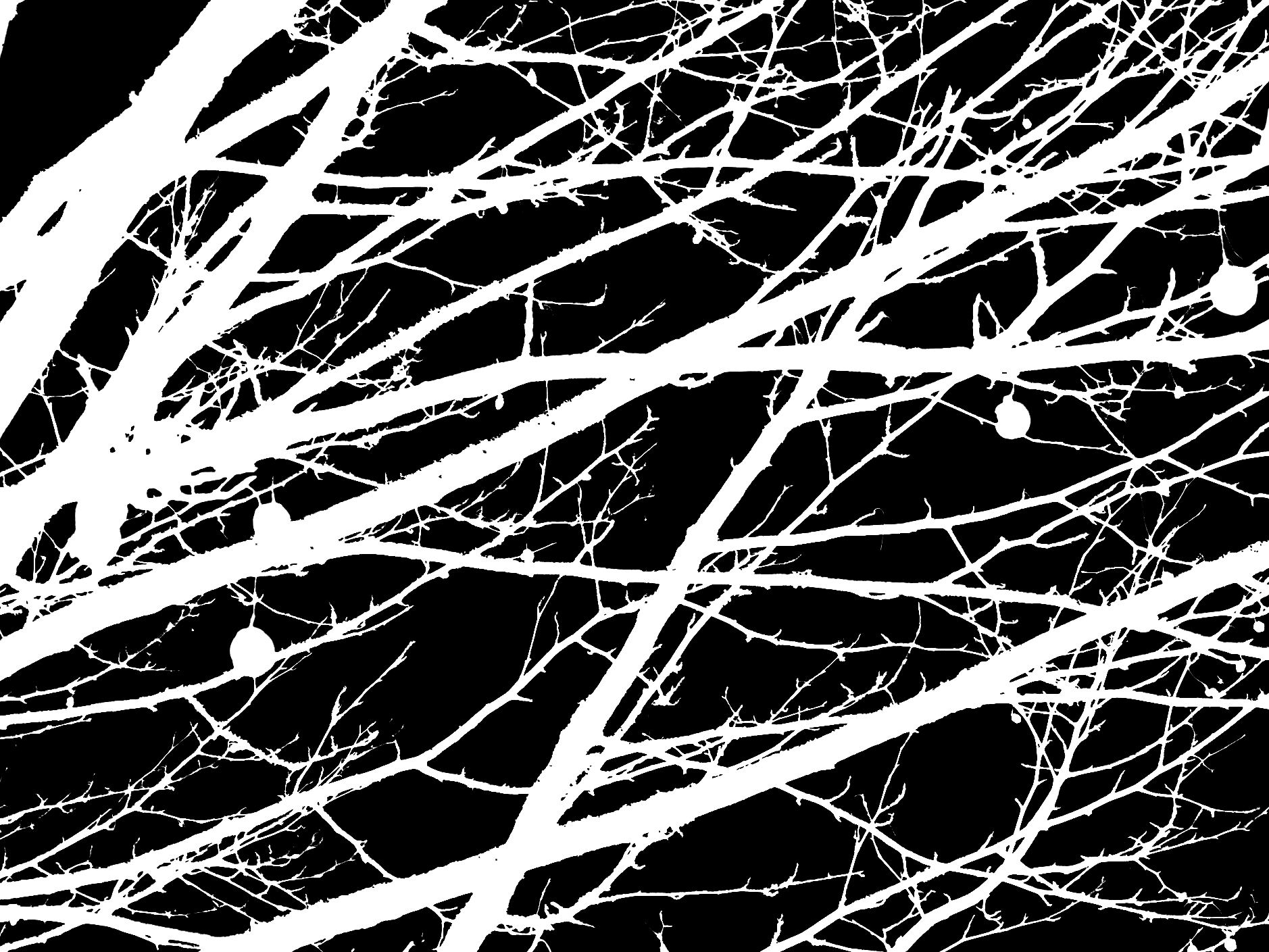} \\
			
			\includegraphics[scale=0.0333]{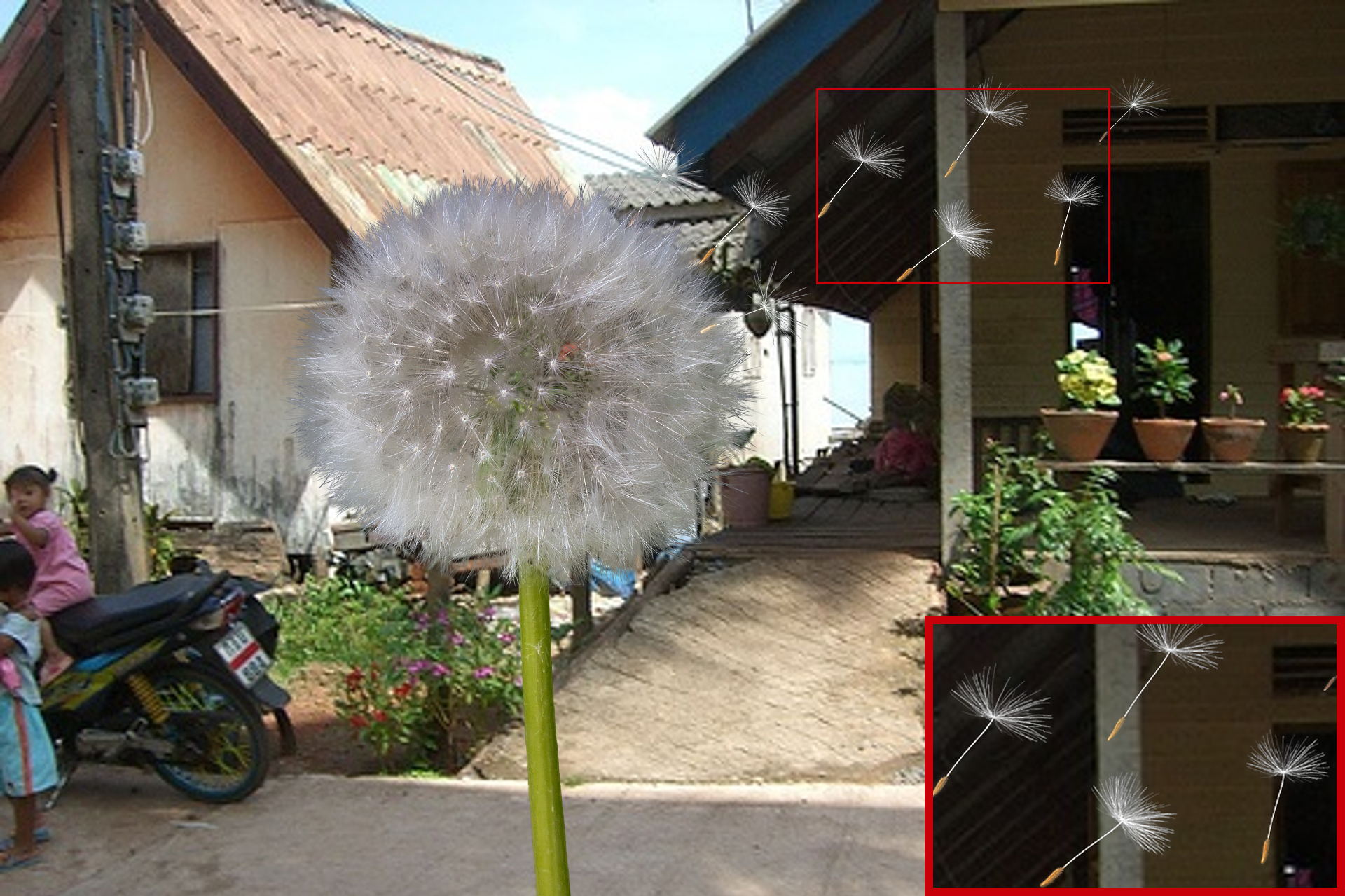} &
			\includegraphics[scale=0.0333]{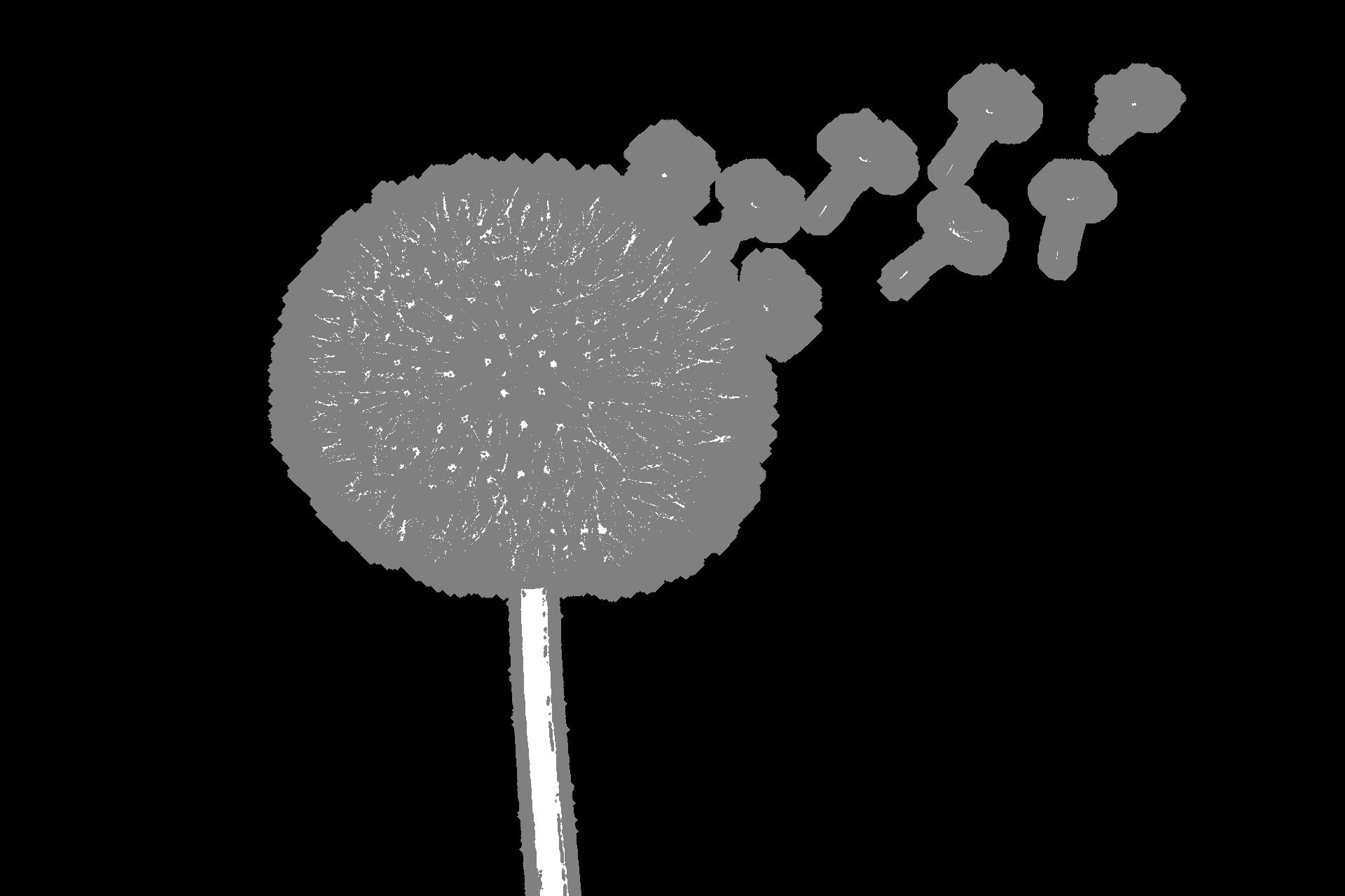} &
			\includegraphics[scale=0.0333]{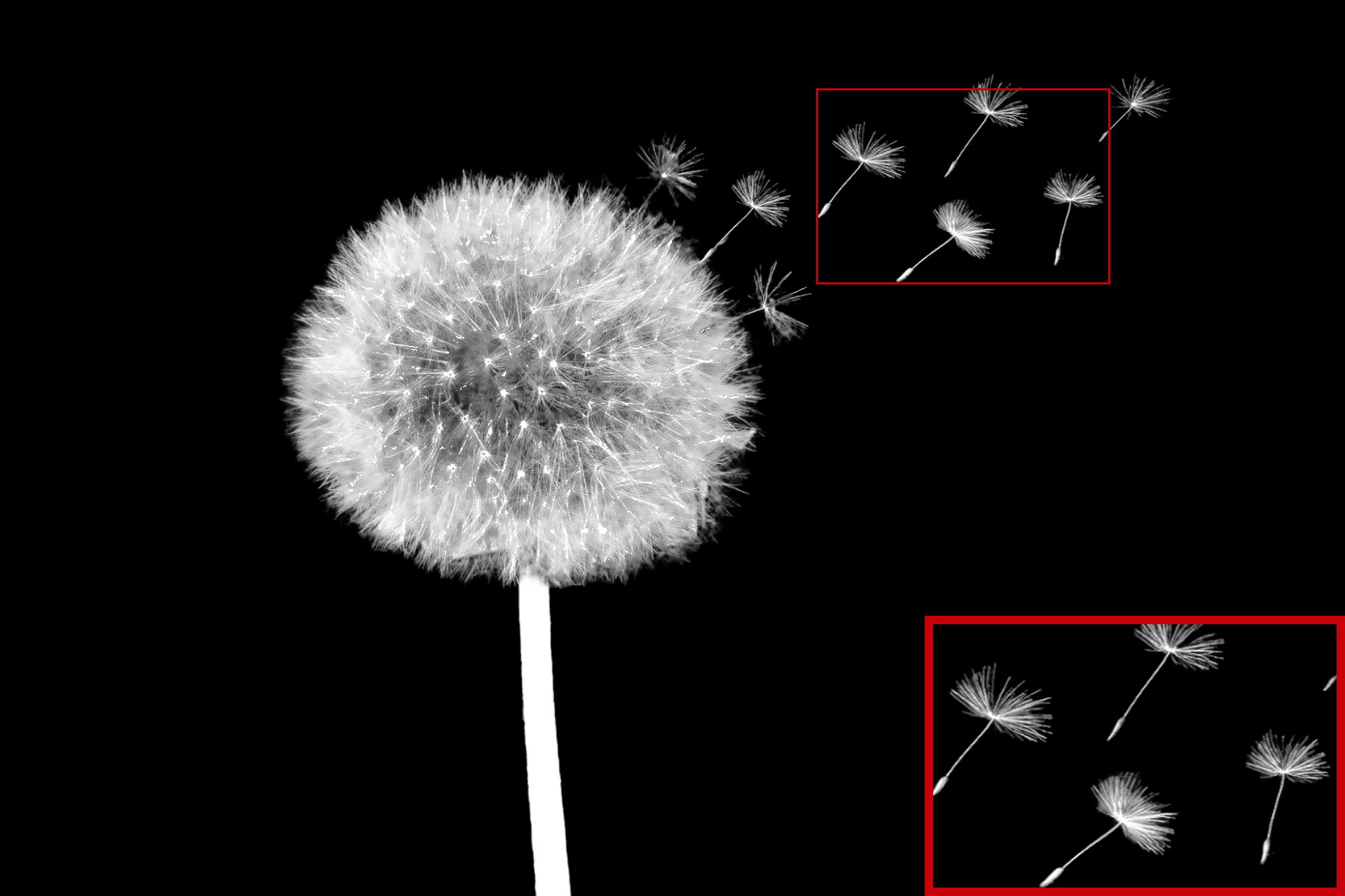} &
			\includegraphics[scale=0.0333]{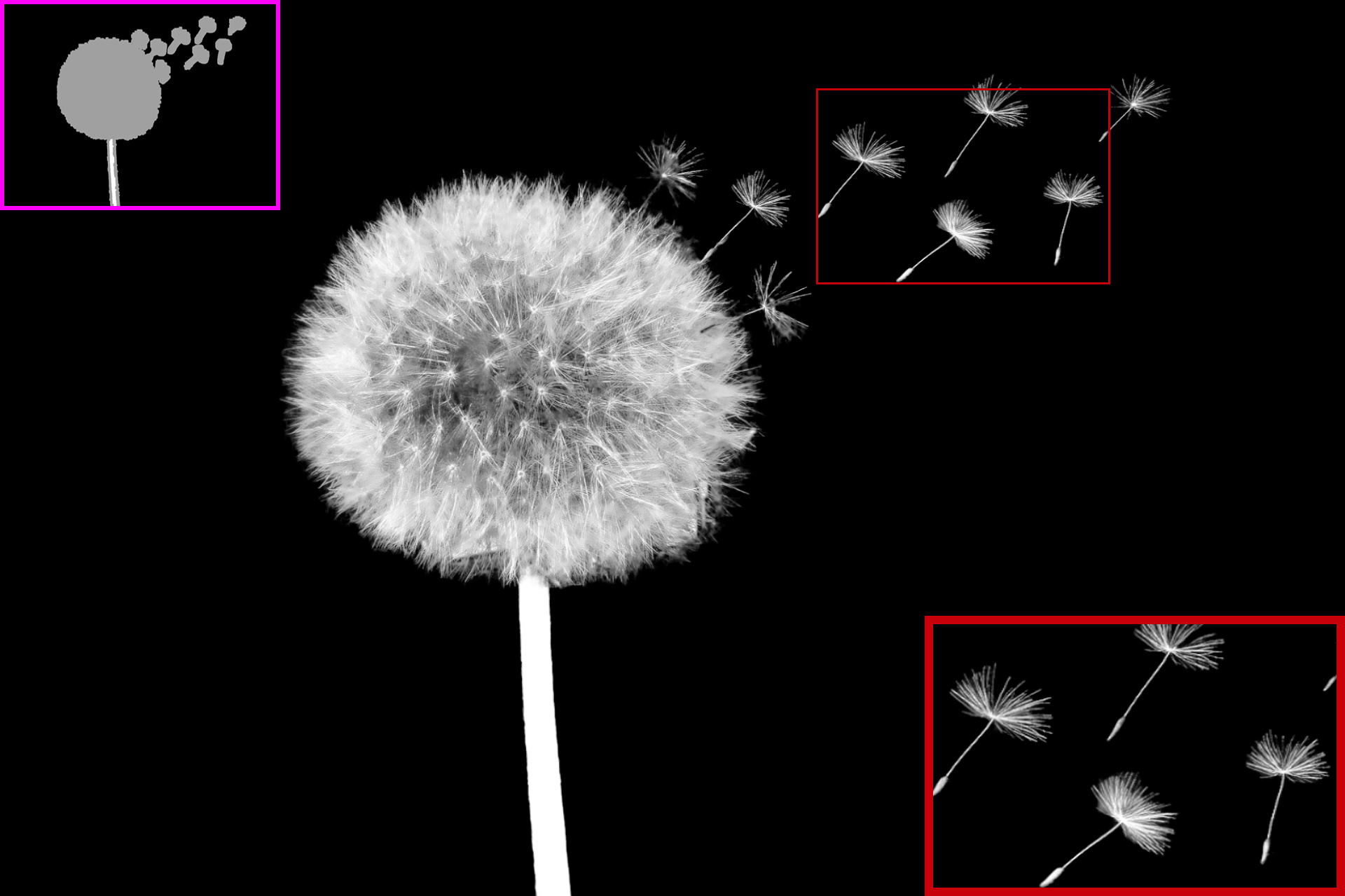} &
			\includegraphics[scale=0.0333]{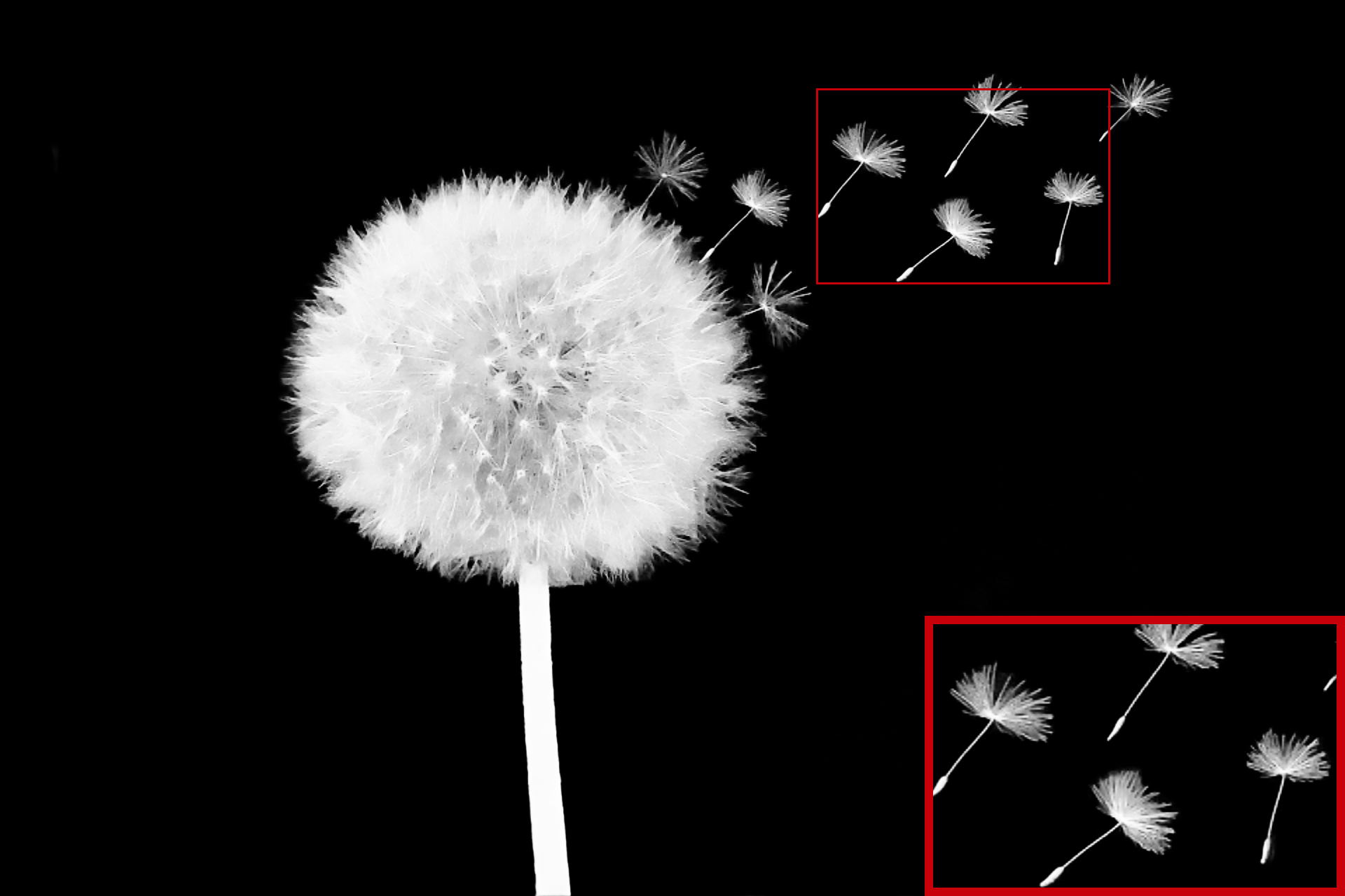} &
			\includegraphics[scale=0.0333]{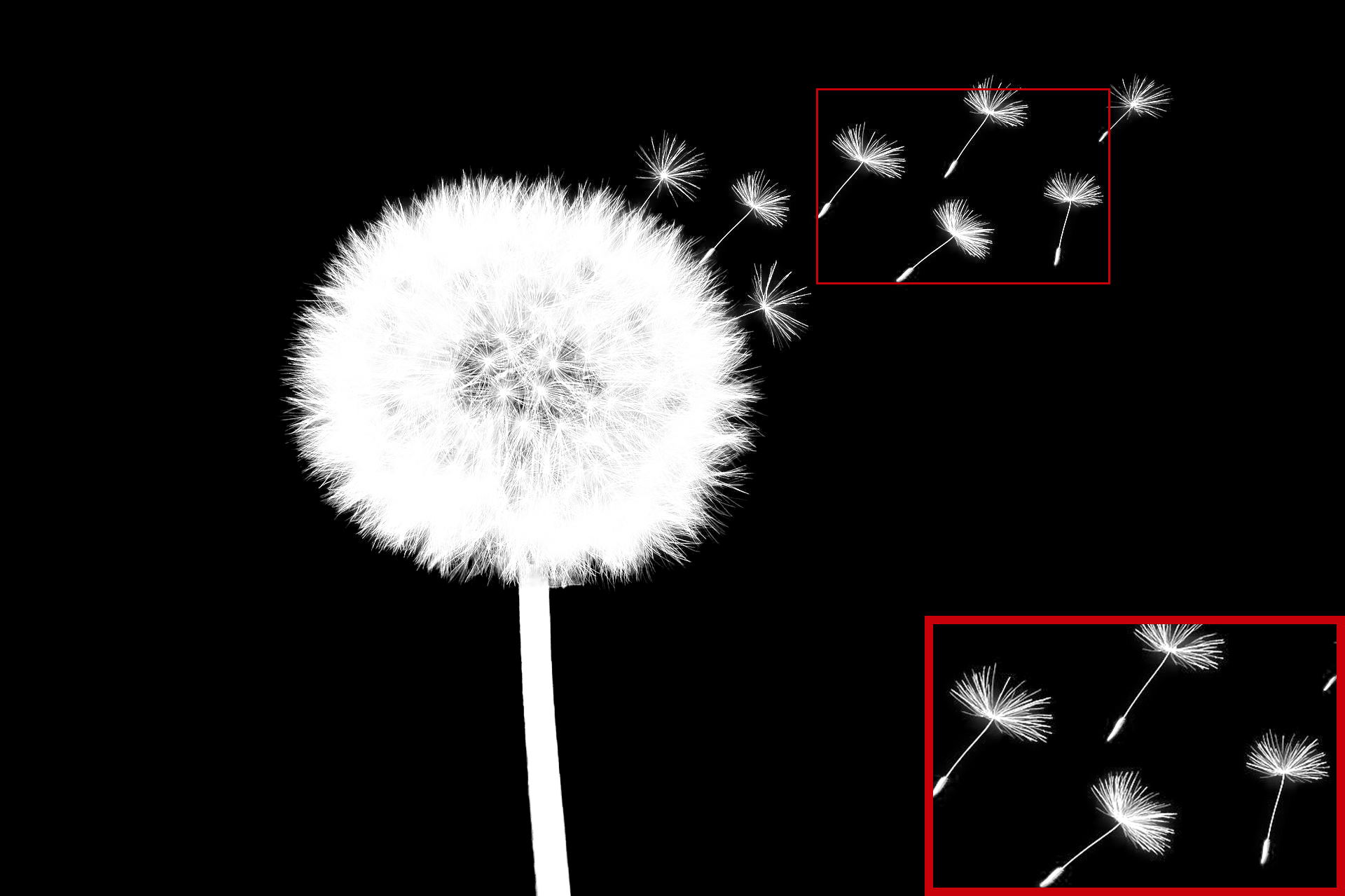} \\
			
			Input Image & Trimap & DIM~\cite{Xu2017Deep} & DIM + Large & HAttMatting++ & Ground Truth \\
	\end{tabular}}
	\caption{The visual comparisons on our Distinctions-646 test set. The ``DIM+Large'' means that we feed DIM with trimaps that have larger transition region, while our method can generate high-quality alpha mattes without trimaps. }
	\label{fig:visual_our}
\end{figure*}

\begin{figure}[htbp]
	\centering
	\arrayrulecolor{tabcolor}
	\setlength{\tabcolsep}{0.6pt}\small{
		\begin{tabular}{ccccc}
			\includegraphics[width=0.18\linewidth]{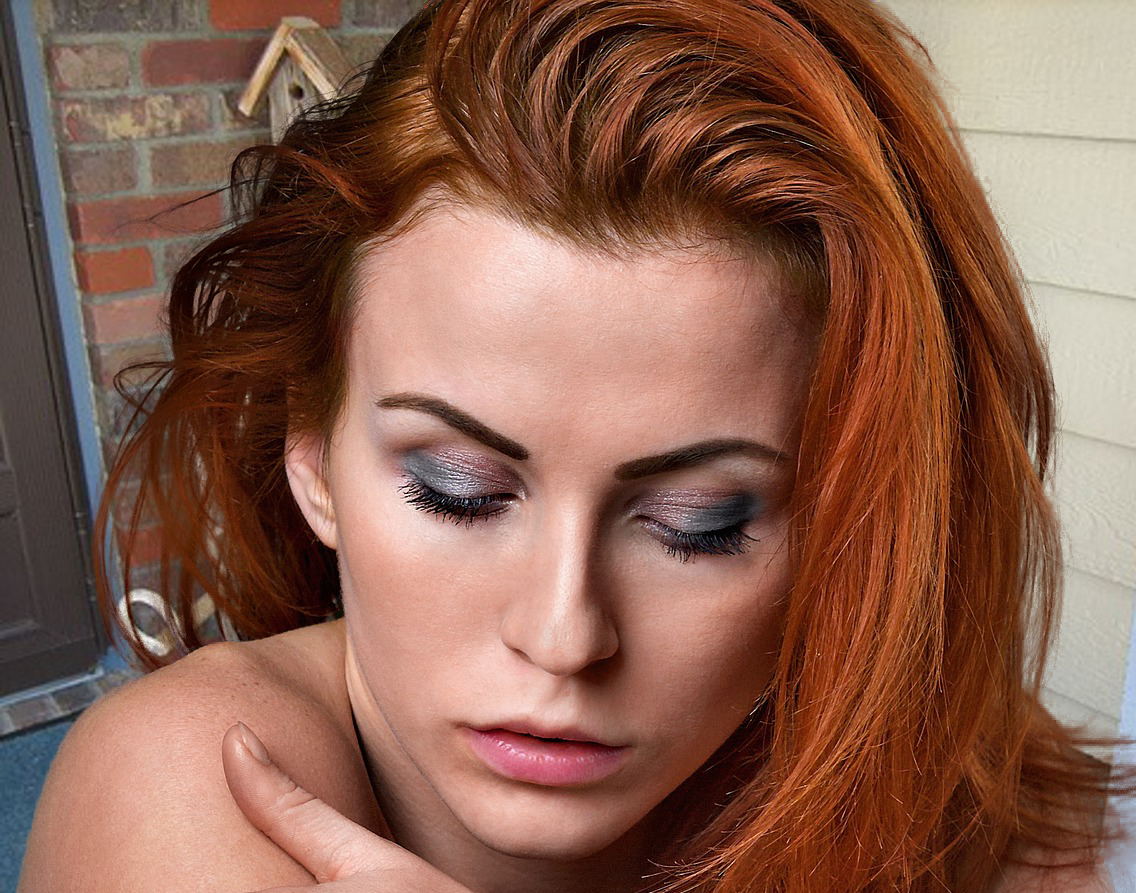} &
			\includegraphics[width=0.18\linewidth]{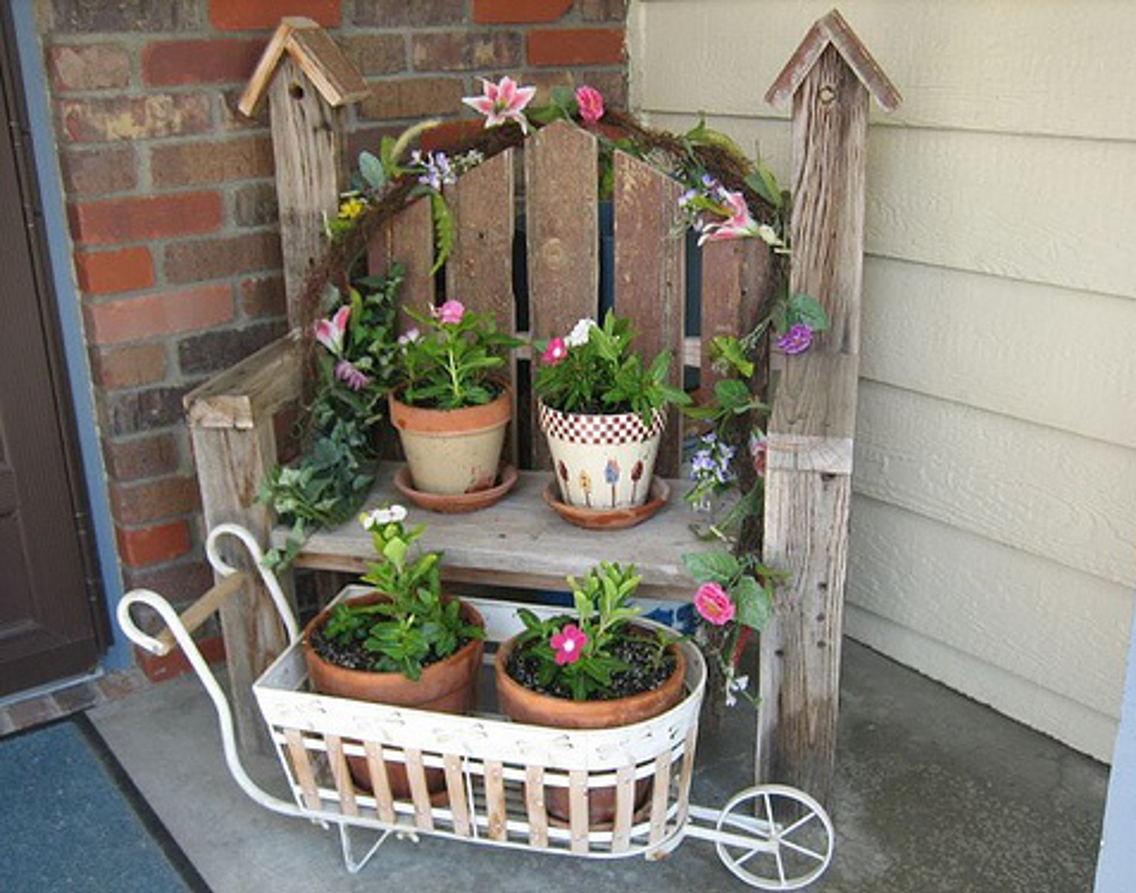} &
			\includegraphics[width=0.18\linewidth]{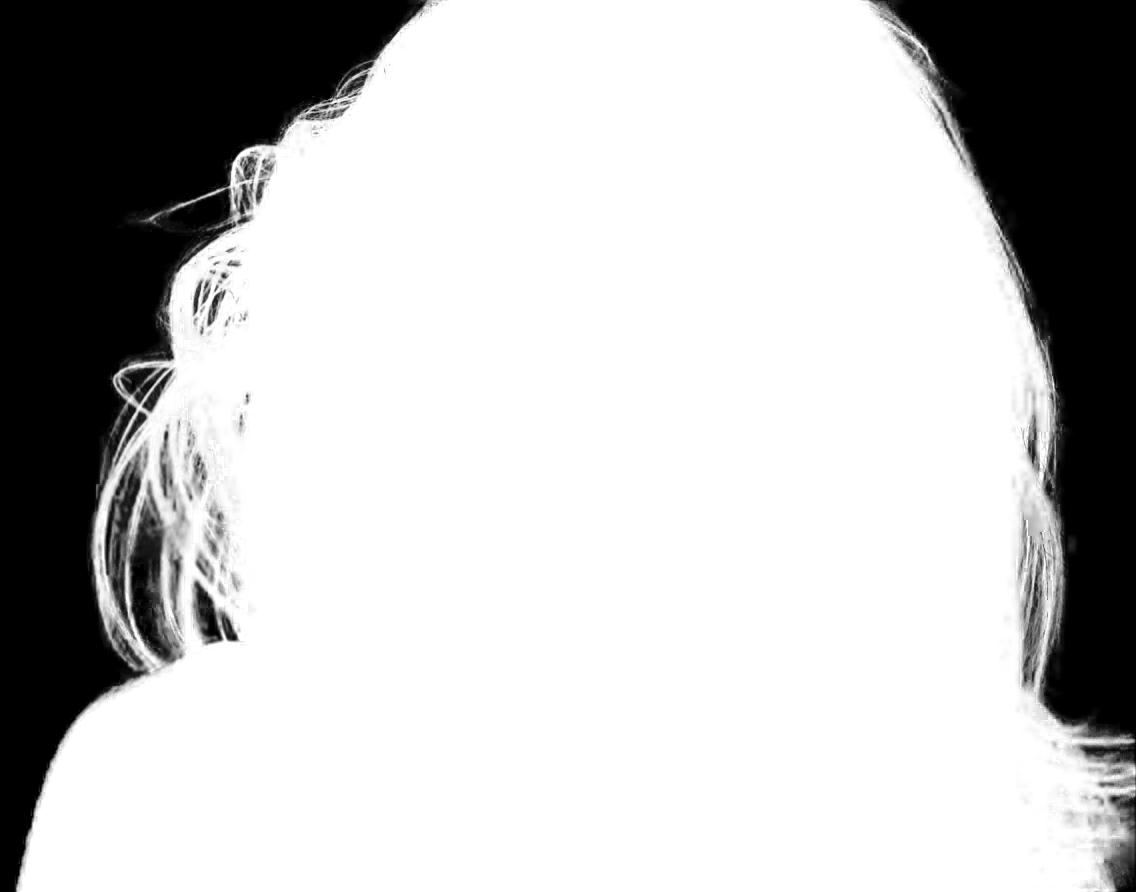} &
			\includegraphics[width=0.18\linewidth]{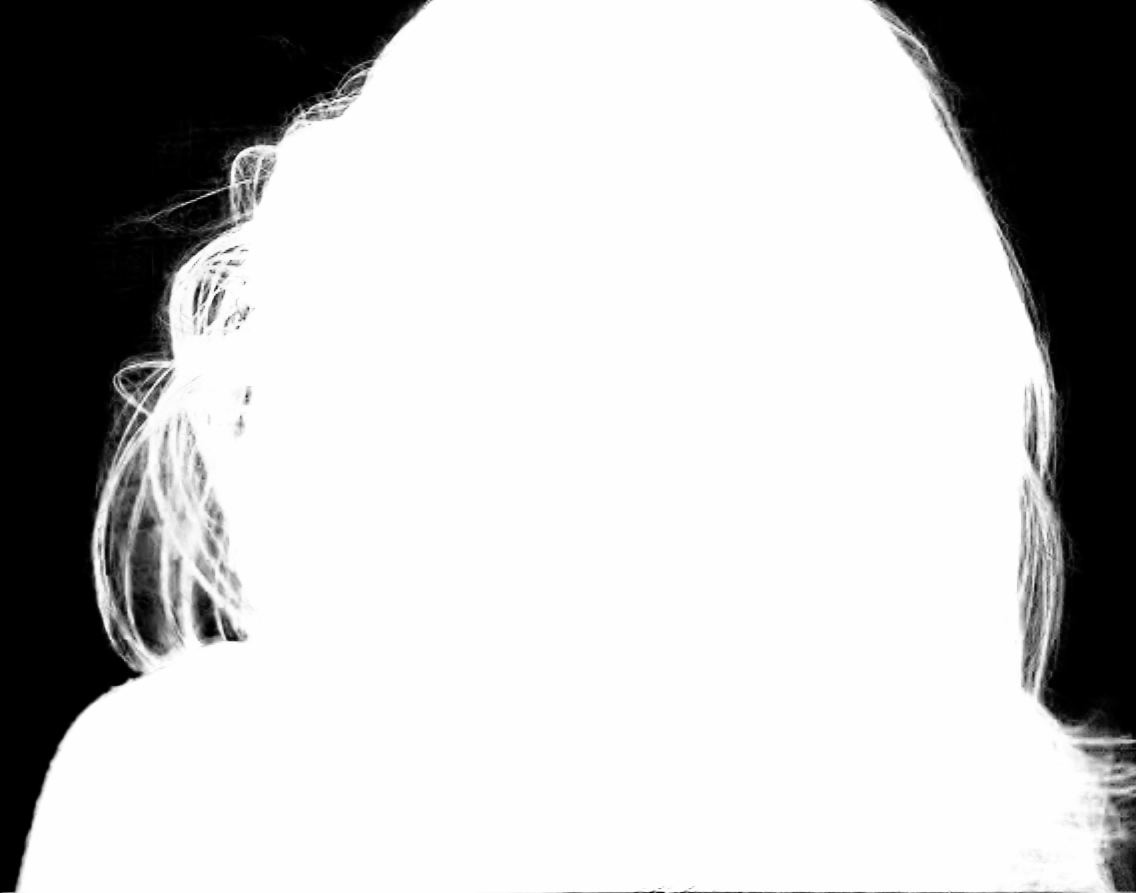} &
			\includegraphics[width=0.18\linewidth]{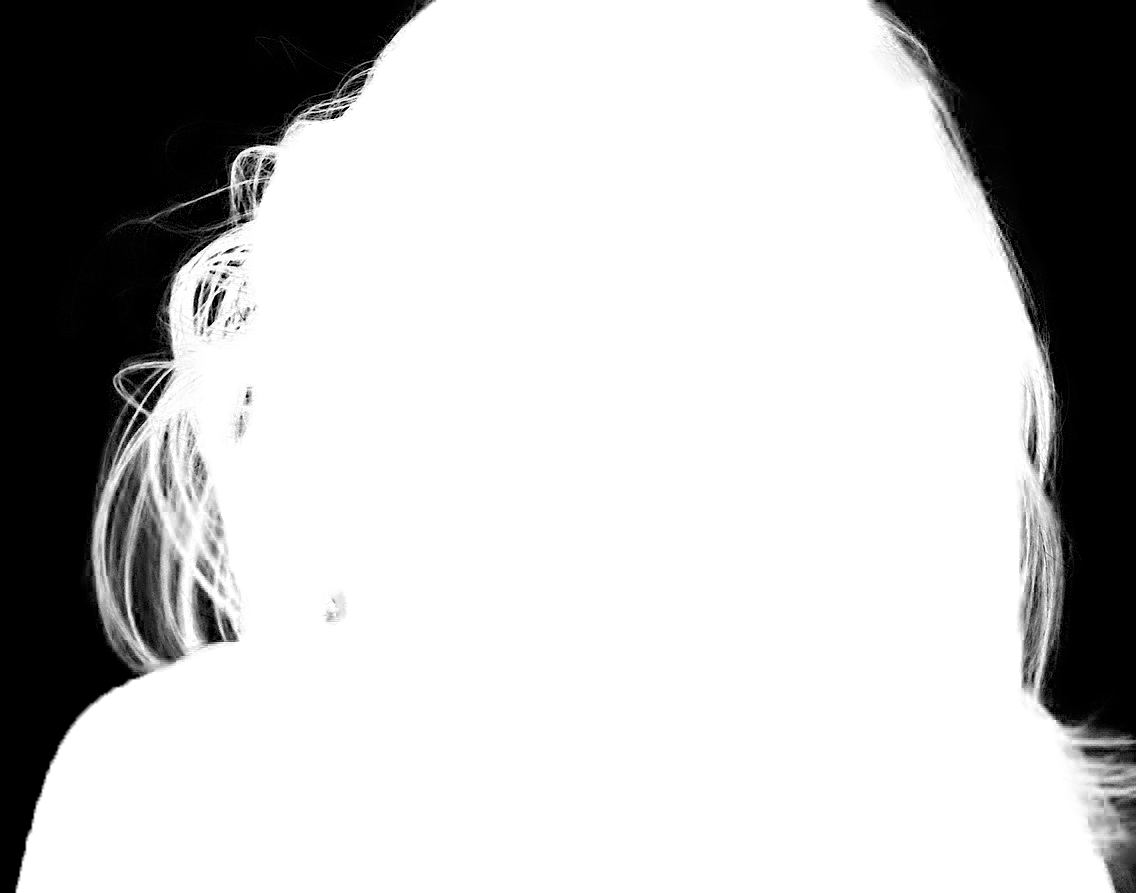} \\
			
			\includegraphics[width=0.18\linewidth]{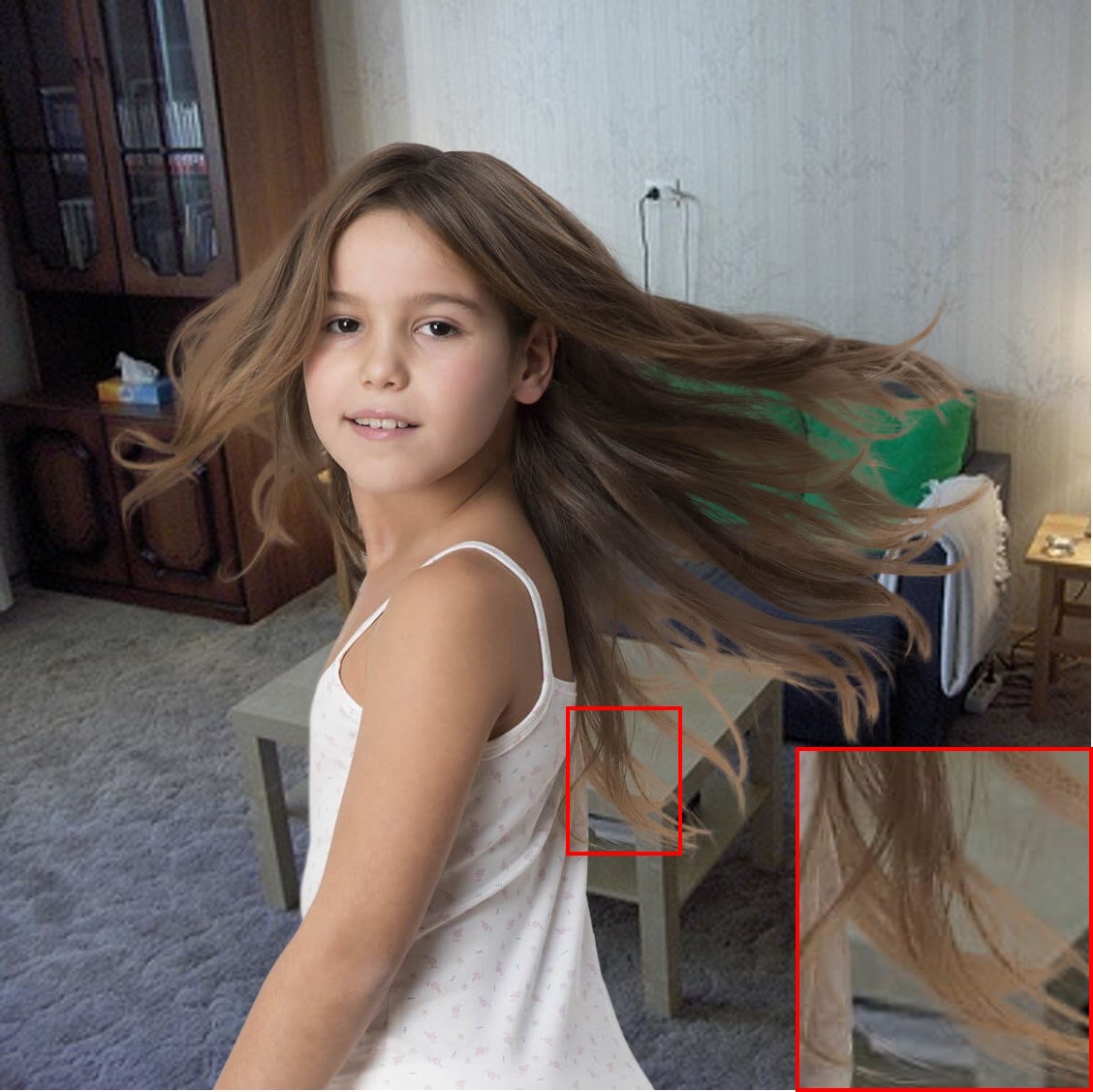} &
			\includegraphics[width=0.18\linewidth]{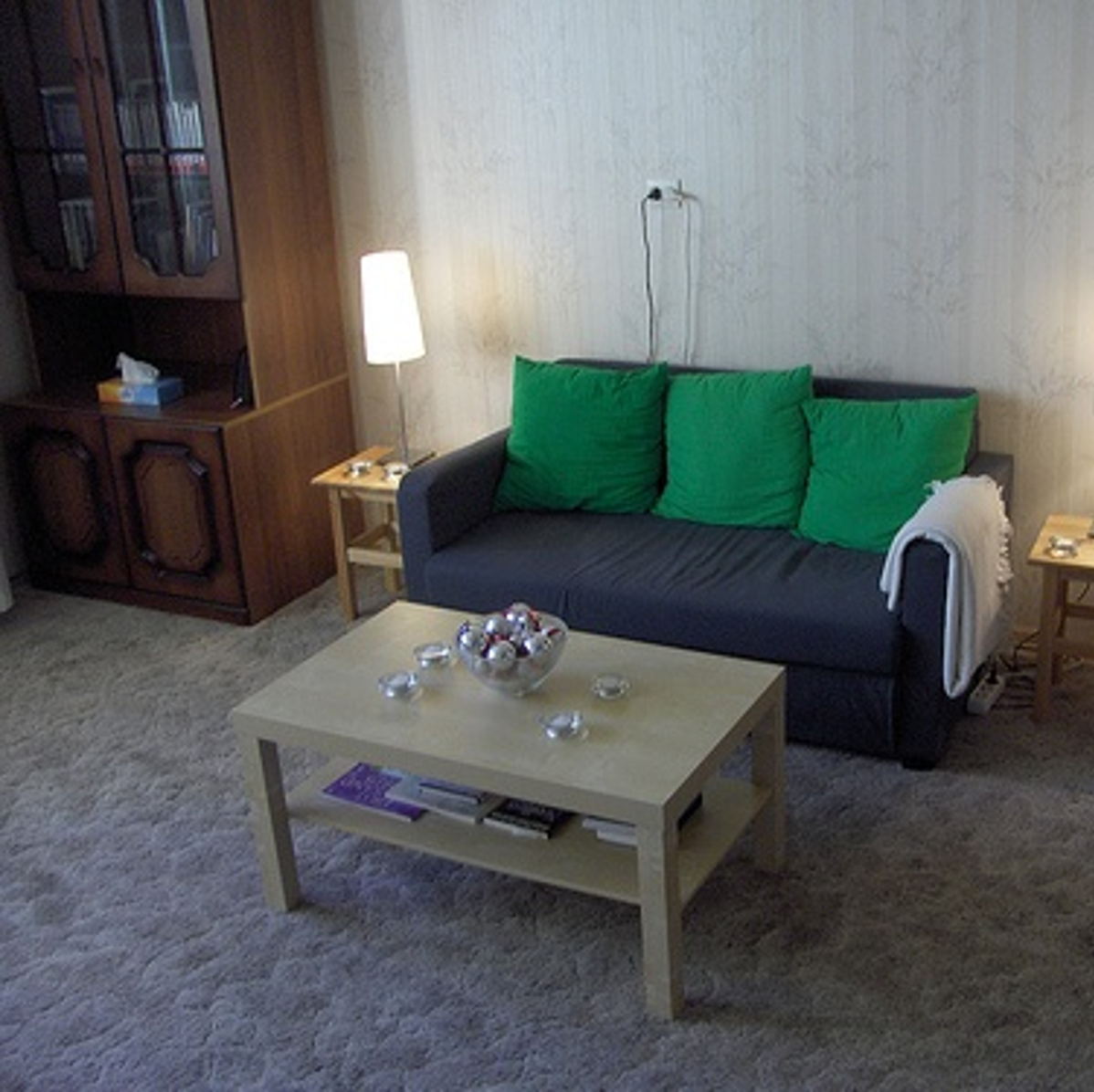} &
			\includegraphics[width=0.18\linewidth]{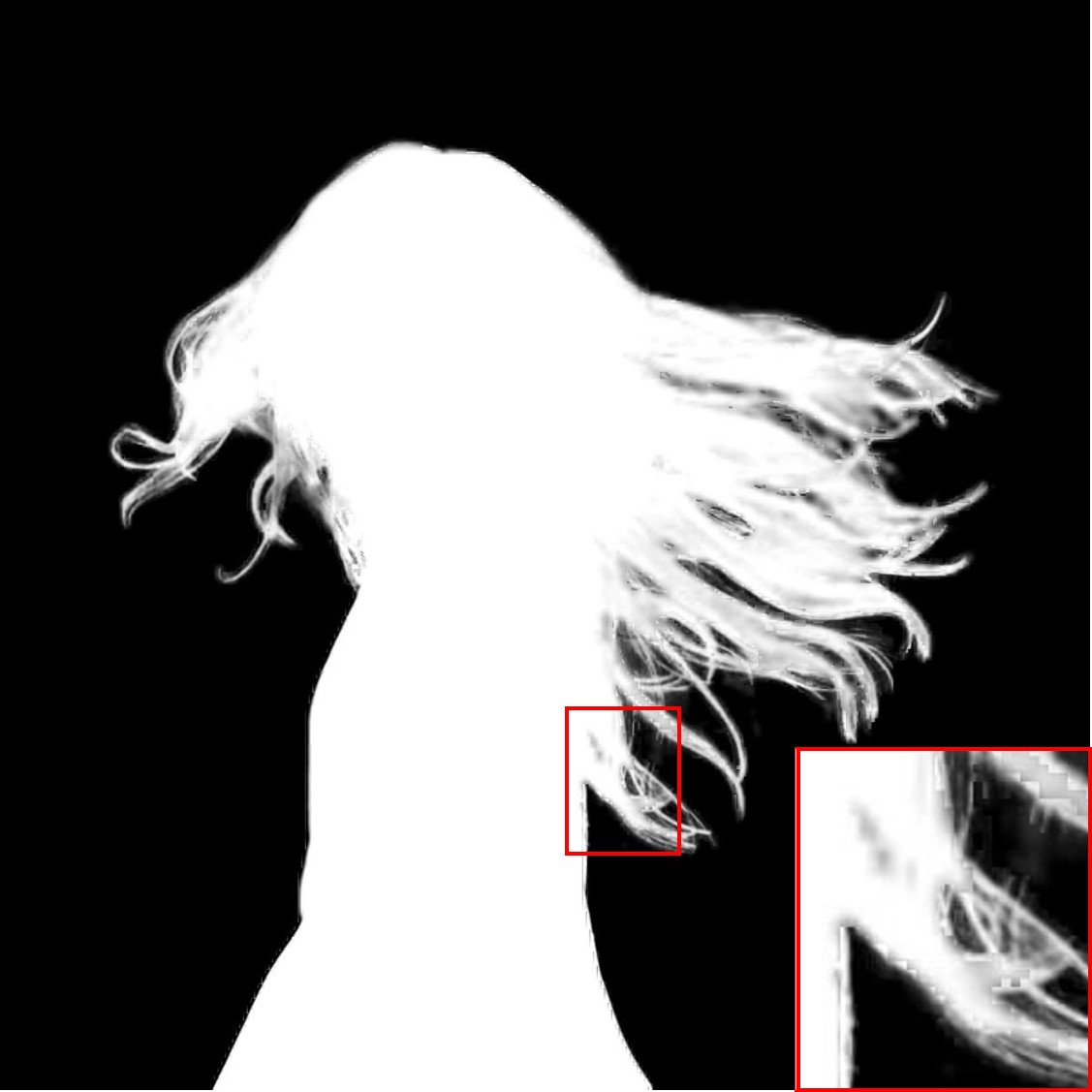} &
			\includegraphics[width=0.18\linewidth]{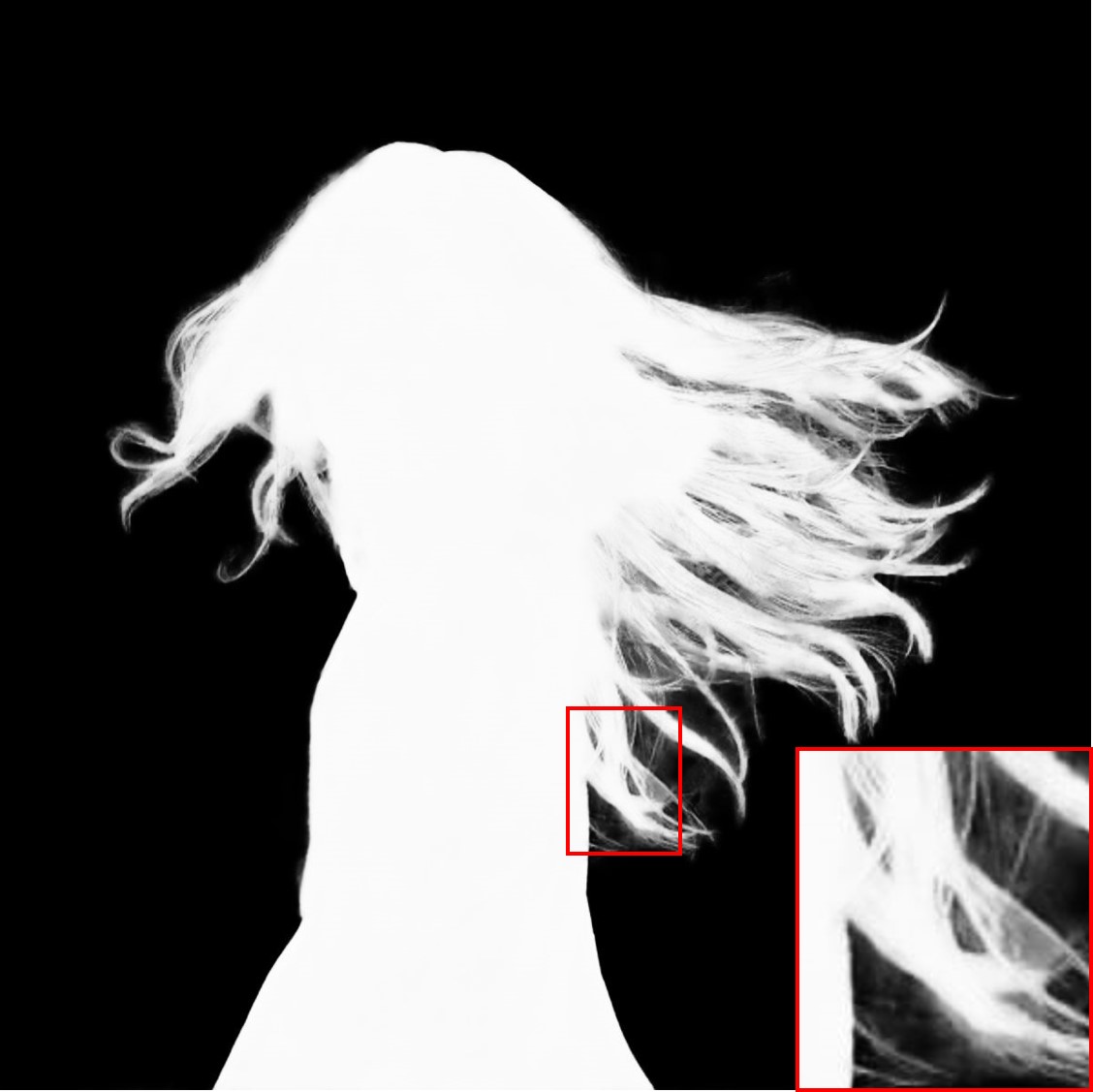} &
			\includegraphics[width=0.18\linewidth]{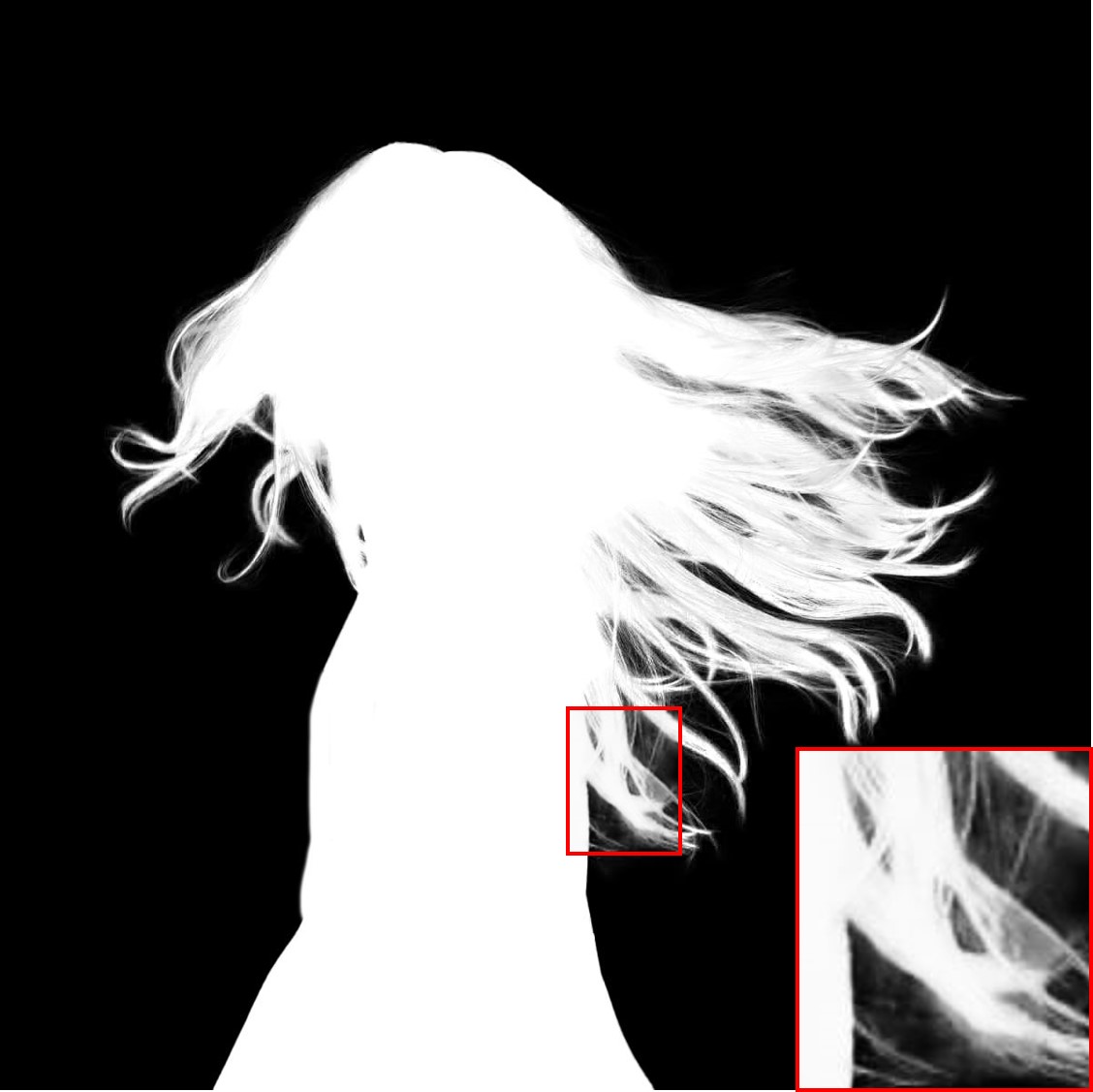} \\
			
			Input Image & Background Image & BGMv2~\cite{lin2021real} & Ours & Ground Truth \\
	\end{tabular}}
	\caption{The visual comparisons with BGMv2~\cite{lin2021real} on human samples.}
	\label{fig:com_bgmv2}
\end{figure}
Compared with trimap-based methods, our \emph{HAttMatting++} has more sophisticated details than DCNN~\cite{cho2019deep}, DIM~\cite{Xu2017Deep}, and is better than SampleNet~\cite{Tang_2019_CVPR}, since we employ hierarchical and progressive attention mechanism to distill advanced semantics and different-levels appearance cues, and their aggregation achieves relatively complete \emph{FG} profiles and boundaries. Our \emph{HAttMatting++} is slightly inferior to Context-Aware~\cite{hou2019context} and IndexNet~\cite{hao2019indexnet} on some metrics. The former establishes two branches and resorts to \emph{FG} image supervision to predict alpha mattes, while the latter learns index functions to capture texture and boundary details. Both the cutting-edge GCA~\cite{li2020natural} and A${^2}$U~\cite{dai2021learning} matting can achieve better alpha mattes by profound exploration of trimaps. Although they both generate high-quality alpha mattes, trimaps are strongly required during their training and inference phase, which restricts their effectiveness in practical applications. Our \emph{HAttMatting++} only need single\emph{RGB} images as input, which is very convenient for novice users. The comparative results with BGMv2~\cite{lin2021real} are reported at the bottom of Table~\ref{tab:quantitative_adobe}. The BGMv2 implementation is the same as the original paper (trained only on human images). Here we use the $220$ human samples in the Composition-1k testing set, rather than $5$ random background images. Independent of the background images, we achieve better MSE and Gradient metrics than BGMv2. The visual results are shown in Fig.~\ref{fig:com_bgmv2} (the first row).

\begin{table}[t]
\renewcommand{\arraystretch}{1.3}
\setlength{\belowcaptionskip}{2pt}
\caption{The Quantitative Comparisons on our Distinctions-646 Test Set. }
\label{tab:quantitative our}
\small
\centering
\setlength{\tabcolsep}{0.6mm}{
	\begin{tabular}{l|ccccc}
		\hline
		Methods & SAD$\downarrow$ & MSE$\downarrow$  & Gradient$\downarrow$ & Connectivity$\downarrow$ & Additional input \\
		\hline
		Shared Matting~\cite{gastal2010shared} & 119.56 & 0.026 & 129.61 & 114.37 & Trimap \\
		Learning Based~\cite{zheng2009learning} & 105.04 & 0.021 & 94.16 & 110.41 & Trimap \\
		Global Matting~\cite{Rhemann2011A} & 135.56 & 0.039 & 119.53 & 136.44 & Trimap \\
		ClosedForm~\cite{Levin2007A} & 105.73 & 0.023 & 91.76 & 114.55 & Trimap \\
		KNN Matting~\cite{Chen2013KNN} & 116.68 & 0.025 & 103.15 & 121.45 & Trimap \\
		DCNN~\cite{cho2019deep} & 103.81 & 0.020 & 82.45 & 99.96 & Trimap \\
		Information-Flow~\cite{Aksoy2017Designing} & 78.89 & 0.016 & 58.72 & 80.47 & Trimap \\
		DIM~\cite{Xu2017Deep} & 47.56 & 0.009 & 43.29 & 55.90 & Trimap \\
		\hline
		\hline
		Basic & 129.94 & 0.028 & 124.57 & 120.22 & No \\
		Basic + SSIM & 121.79 & 0.025 & 110.21 & 117.41 & No \\
		Basic + Low & 98.88 & 0.020 & 84.11 & 92.88 & No \\
		Basic + CA  & 104.23 & 0.022 & 90.87 & 101.9 & No \\
		Basic + Low + CA & 85.57 & 0.015 & 79.16 & 88.38 & No \\
		Basic + Low + SA & 78.14 & 0.014 & 60.87 & 71.90 & No \\
		Basic + Low + CA + SA & 57.31 & 0.011 & 52.14 & 63.02 & No \\
		\hline
		\rowcolor{mygray}
		HAttMatting~\cite{Qiao_2020_CVPR} & 48.98 & 0.0094 & 41.57 & 49.93 & No \\
		\hline
		\hline
		\rowcolor{mygray}
		HAttMatting + multi-scale & 48.32 & 0.0091 & 41.15 & 48.01 & No \\
		\rowcolor{mygray}
		HAttMatting + Sentry & 47.53 & 0.0090 & 40.91 & 46.34 & No \\
		\rowcolor{mygray}
		\emph{HAttMatting++} & \textbf{47.38} & \textbf{0.0088} & \textbf{40.09} & \textbf{45.60} & No \\
		\hline
		BGMv2 & 11.53 & 0.010 & 9.69 & 9.70 & Background images \\
		HAttMatting++\_human & 11.46 & 0.010 & 8.91 & 9.88 & No \\
		\hline
\end{tabular}}
\end{table}
\textbf{Evaluation on our Distinctions-646.} For our Distinctions-646 dataset, we compare \emph{HAttMatting++} with 8 recent matting methods, including Shared Matting~\cite{gastal2010shared}, Learning Based~\cite{zheng2009learning}, Global Matting~\cite{Rhemann2011A}, ClosedForm~\cite{Levin2007A}, KNN Matting~\cite{Chen2013KNN}, DCNN~\cite{cho2019deep}, Information-Flow~\cite{Aksoy2017Designing} and DIM~\cite{Xu2017Deep} and BGMv2~\cite{lin2021real}. We also use random dilation to generate high-quality trimaps~\cite{Xu2017Deep} and relevant metrics are computed on the whole image.

The quantitative comparisons are displayed in Table~\ref{tab:quantitative our}, where the definition of ``Basic'', ``Basic +'', and ``HAttMatting +'' are the same with Table~\ref{tab:quantitative_adobe}. Our \emph{HAttMatting++} shows a clear advantage on all four metrics compared to all the traditional methods, and is slightly better than DIM~\cite{Xu2017Deep} on SAD and MSE metrics, while obviously better than it in Grad and Conn metrics, which indicates that our model can achieve improved visual quality. It is noted that only our method can generate alpha mattes without trimaps, and all the other methods demand trimaps to confine the transition region, which effectively improves the performance of these methods. Fig.~\ref{fig:visual_our} illustrates the visual comparison with DIM~\cite{Xu2017Deep} network. Here we enlarge the transition region to reduce the accuracy of trimap, and the corresponding alpha mattes with DIM are shown in the fourth column. The deterioration in visual quality is evident with the transition region expanded, which can verify that DIM has a strong dependence on the quality of trimaps. The alpha mattes produced by \emph{HAttMatting++} exhibit sophisticated texture details, which mainly benefits from the aggregation of adaptive semantics and valid appearance cues in our model. The comparisons of human samples with BGMv2 are demonstrated in Table~\ref{tab:quantitative our} and Fig.~\ref{fig:com_bgmv2} (the second row). We also employ the official BGMv2 weights that are trained on human images, and the proposed \emph{HAttMatting} achieves better SAD and MSE metrics.

\subsection{Ablation Study}
\label{ssec:ablation}
The core idea of our \emph{HAttMatting++} is to extract adaptive pyramidal features and filter progressive appearance cues, and then aggregate them to generate alpha mattes. To accomplish this goal, we employ channel-wise attention (CA) and spatial attention (SA) to re-weight pyramidal features and different-levels appearance cues separately. We also introduce \emph{SSIM} in our loss function to further improve the \emph{FG} structure. 

In this section, we first discuss the potential variation of the proposed network architecture, and then analyze the properties of some important components in the conference paper~\cite{Qiao_2020_CVPR} by combining them on the baseline network, and finally expound the presented multi-level information fusion and sentry supervision. All combinations of different components below are trained and tested on the Composition-1k dataset and our Distinctions-646 dataset. The correlated evaluation values are summarized in Table~\ref{tab:quantitative_adobe} and Table~\ref{tab:quantitative our}.

\subsubsection{Alternative Semantics Extraction Strategies}
\label{sec:strategies}
In the pipeline of our \emph{HAttMatting++}, we extract high-level semantic features from the block4 of ResNeXt~\cite{Xie2017Aggregated}, then feed them to ASPP~\cite{Chen2018DeepLab} module to capture pyramidal features. The original intention of such architecture is based on the consideration that \emph{FG} objects always occupy the majority of the input image in matting, and there is no need to design multi-scale framework to capture objects of various sizes.

\begin{figure}[t]
\centering
\includegraphics[width=0.92\linewidth]{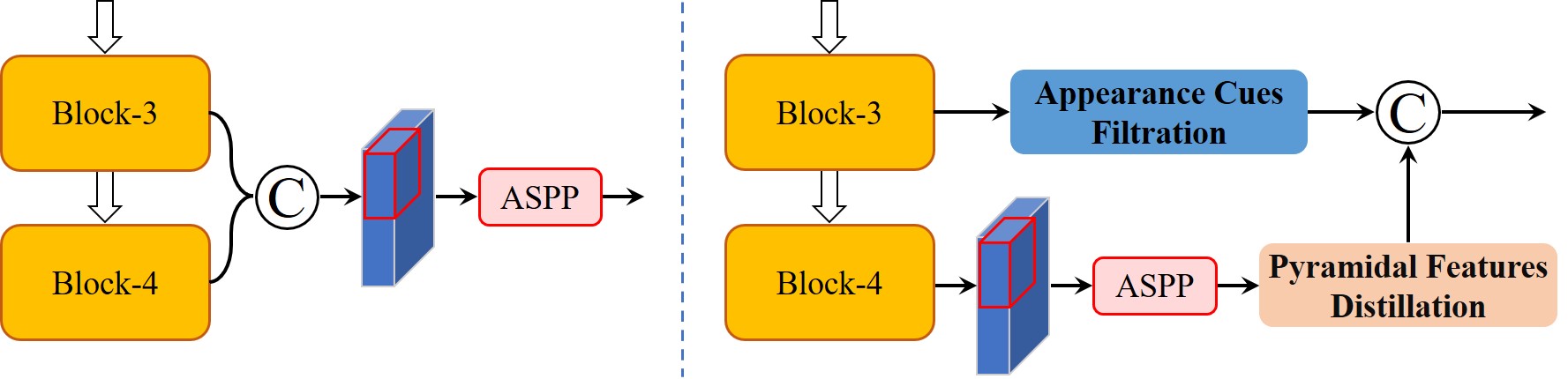}
\caption{Some potential pyramidal features extraction architectures.}
\label{fig:multi_scale}
\end{figure}

Actually, we have tried other network architectures to obtain pyramidal features. For the fair comparison with different extraction strategies at the architecture level, we remove the sentry supervision and take the previous conference paper as the baseline. As shown in the left of Fig.~\ref{fig:multi_scale}, we attempt to concatenate the feature maps from block3 and block4, and assume that the fusion between them can lead to more abundant high-level semantics for locating most part of \emph{FG}.  Theoretically speaking, we can give the block3 branch a weight map $0$ to degenerate the architectures in Fig.~\ref{fig:multi_scale} into our pipeline. However, it is difficult to achieve the same accuracy as our pipeline by combining the semantic features of block3 in the training process, possibly because that the coarse semantics from block3 may result in a diversion of the pyramidal features. Besides, another form of semantic fusion we tried is shown in the right of Fig.~\ref{fig:multi_scale}. The channel-wise attention and spatial attention were performed on the features extracted from ASPP~\cite{Chen2018DeepLab} and block3, and then their combination associated with the low-level features extracted from block1 were sent to execute further spatial attention. However, the feature maps from block3 are most highly abstract and only contain limited appearance cues for exploration, which may contrary disturb the semantic extraction of the backbone.

For low-level features, we also attempt to utilize the input image as appearance cues, like previous matting works~\cite{Yu2018Deep_Prop,Xu2017Deep,Zhang2019CVPR}, but the complex color distribution will obscure texture details and disturb the refinement of \emph{FG} boundaries.
\begin{table}[h]
\setlength{\belowcaptionskip}{2pt}
\caption{The Comparisons with Different Semantic Features Extraction Strategies. }
\label{tab:com_strategies}
\small
\centering
\setlength{\tabcolsep}{2.4mm}{
	\begin{tabular}{l|ccccc}
		\hline
		Methods & SAD$\downarrow$ & MSE$\downarrow$  & Gradient$\downarrow$ & Connectivity$\downarrow$ \\
		\hline
		Input Image & 115.23 & 0.0157 & 51.26 & 106.06 \\
		Strategy-1 & 77.29 & 0.0098 & 44.93 & 78.63 \\
		Strategy-2 & 64.37 & 0.0072 & 32.03 & 56.99 \\
		\hline
		HAttMatting + MS & \textbf{45.69} & \textbf{0.0065} & \textbf{28.61}  
		& \textbf{45.83}\\
		\hline
\end{tabular}}
\end{table}

The quantitative comparisons are shown in Table~\ref{tab:com_strategies}, the implementation details of all models are the same as \emph{HAttMatting++} and all the results are evaluated on the Composition-1k test set. The ``Input Image'' means we only extract spatial cues from input image, and ``Strategy-1'' and ``Strategy-2'' refer to the alternative pyramidal features capture strategies on the left and right of Fig.~\ref{fig:multi_scale} respectively. The alpha mattes produced by ``HAttMatting + MS''  are better than the other features extraction strategies. In our analysis, we argue that \emph{FG} object occupies most of the input image and the top layer of the network can represent the most important semantic information. The top semantics can suggest the \emph{FG} object in image matting, and the external branches like Fig.~\ref{fig:multi_scale} can affect the convergence of the training process on the \emph{FG} to some extent.

\begin{figure}[t]
\setlength{\tabcolsep}{1pt}\small{
	\begin{tabular}{ccc}
		\includegraphics[scale=0.042]{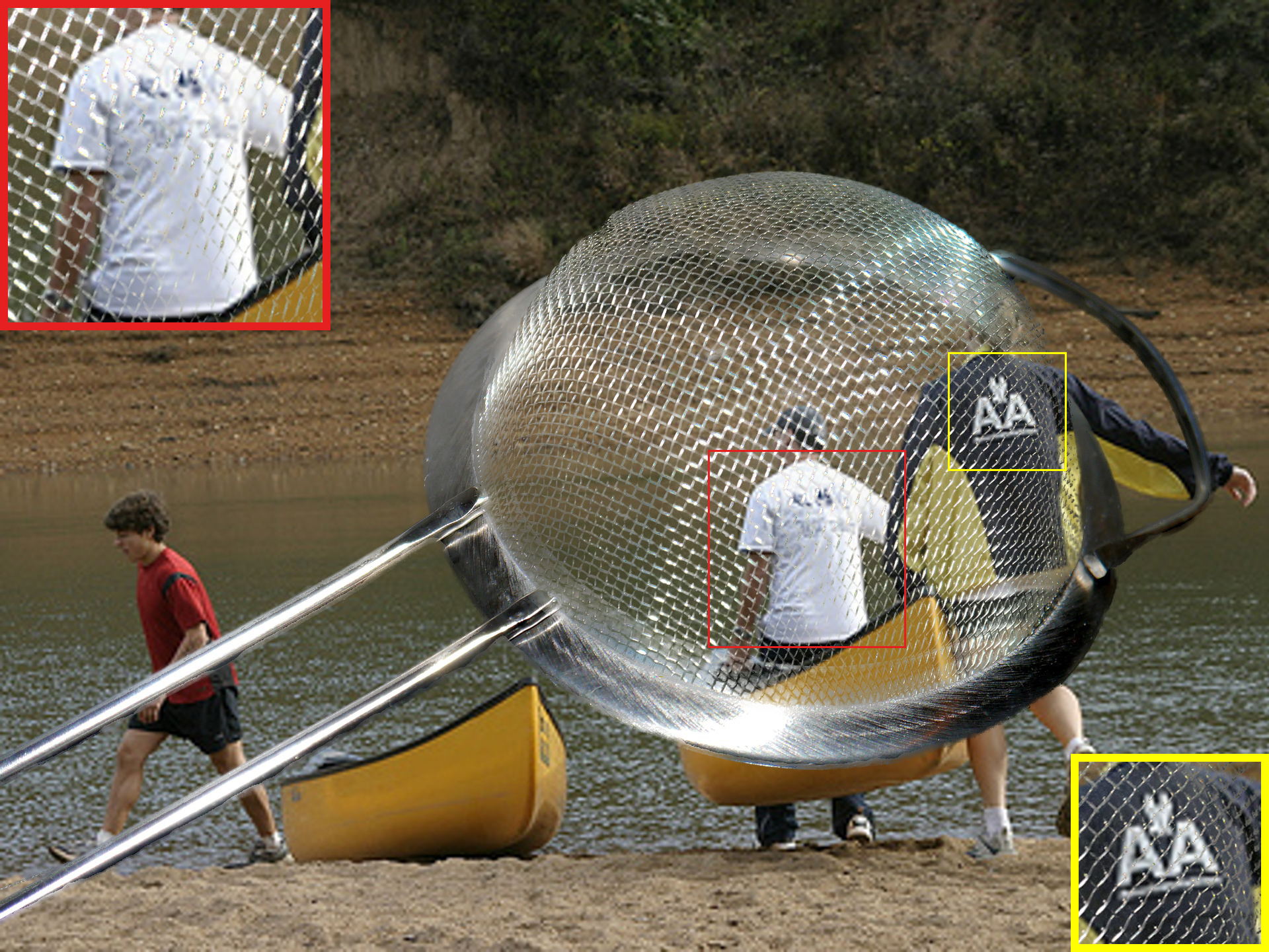} &
		\includegraphics[scale=0.042]{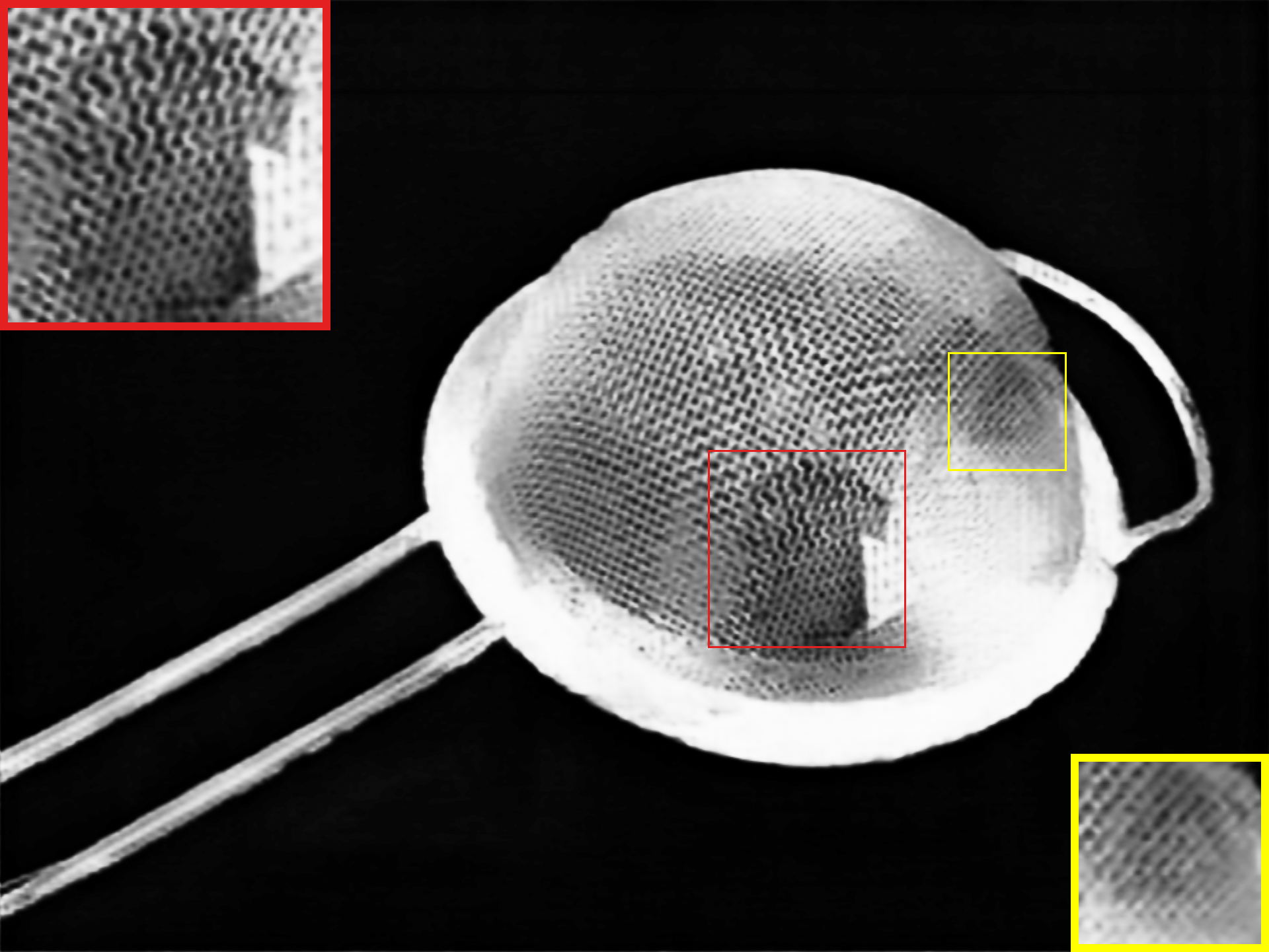} &
		\includegraphics[scale=0.042]{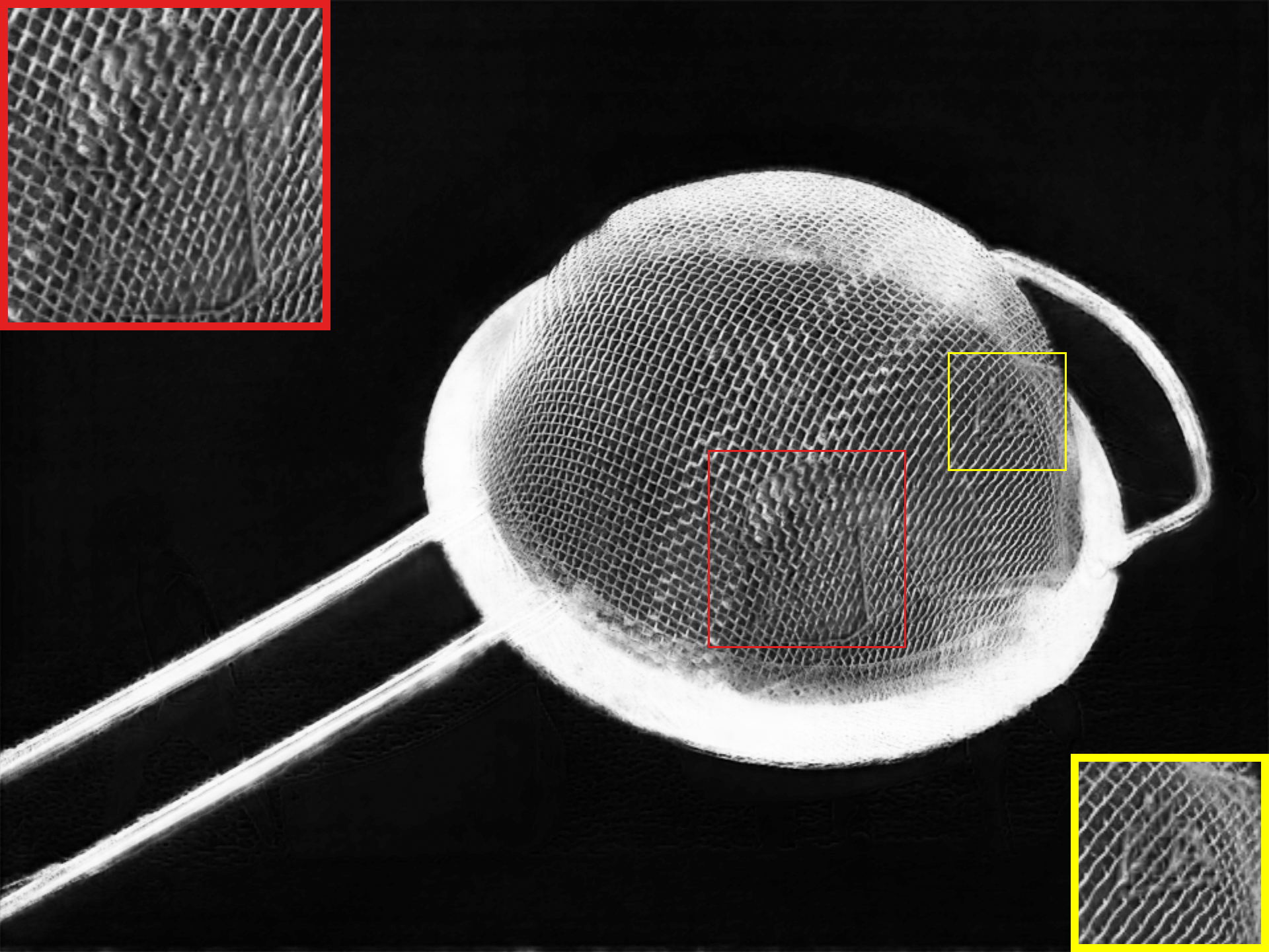} \\
		
		Input Image & Basic & Basic + Low + CA \\
		
		\includegraphics[scale=0.042]{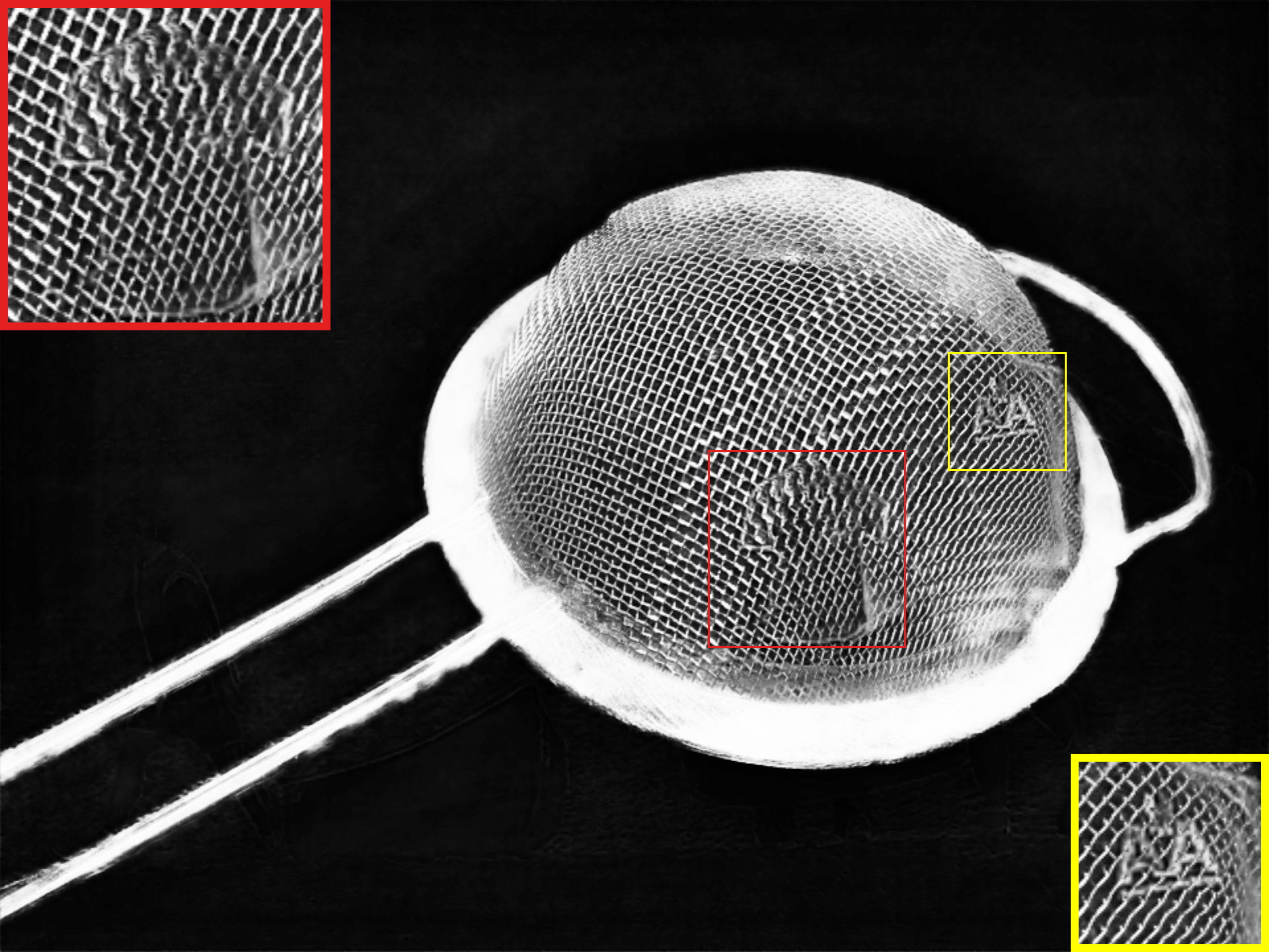} &
		\includegraphics[scale=0.042]{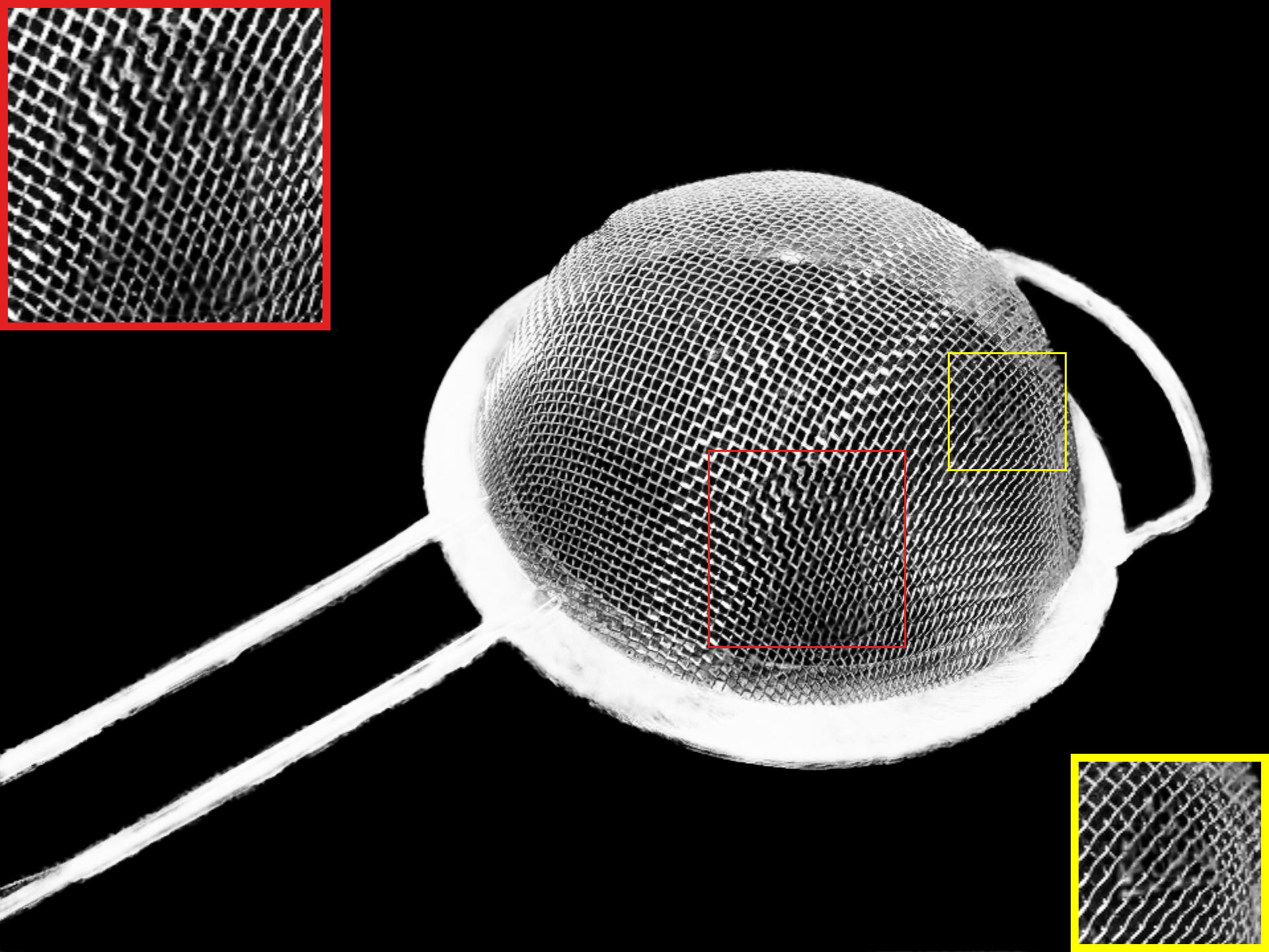} &
		\includegraphics[scale=0.042]{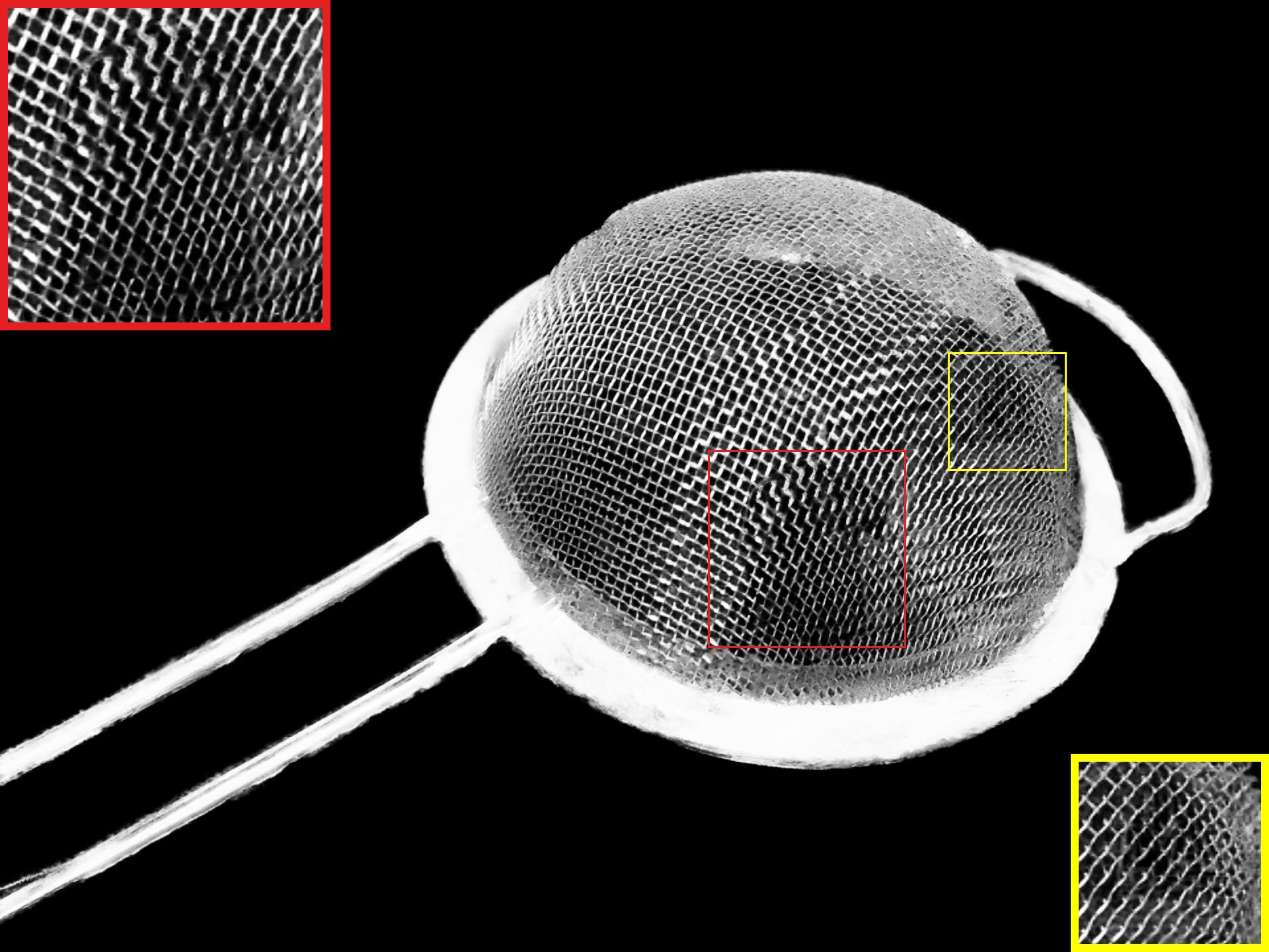} \\
		
		Basic + Low + SA & HAttMatting~\cite{Qiao_2020_CVPR} & HAttMatting + MS \\
		
		\includegraphics[scale=0.042]{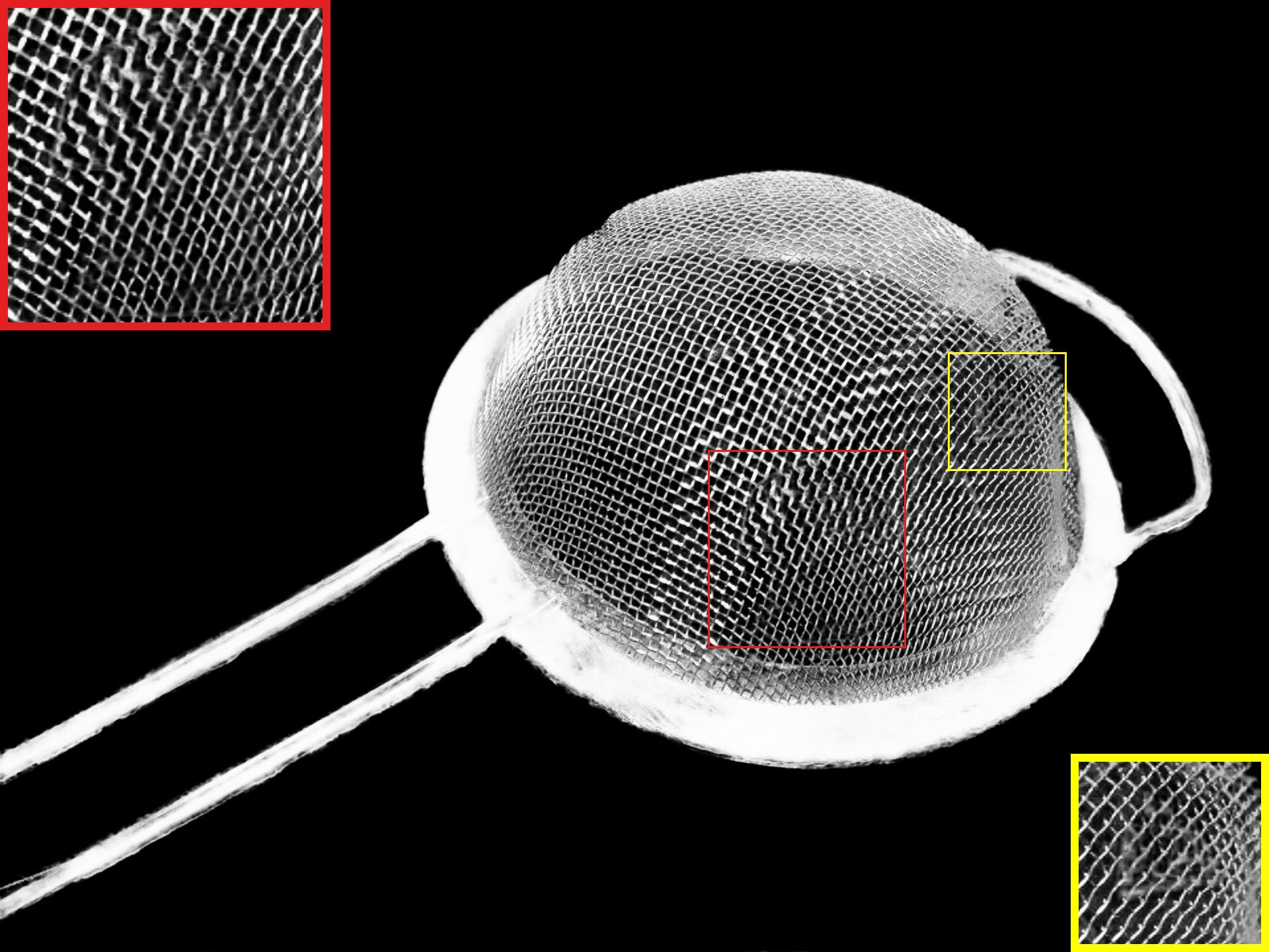} &
		\includegraphics[scale=0.042]{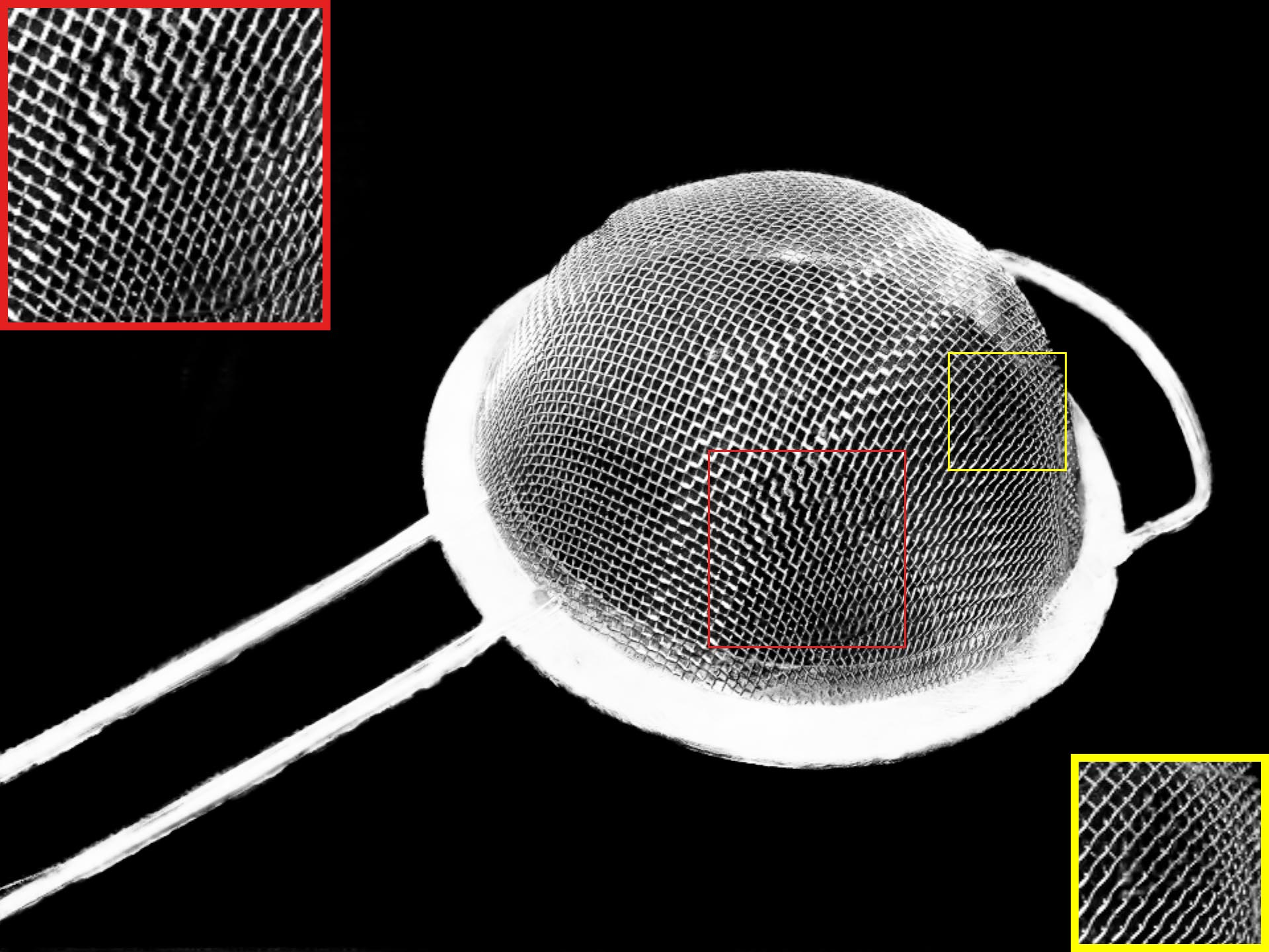} &
		\includegraphics[scale=0.042]{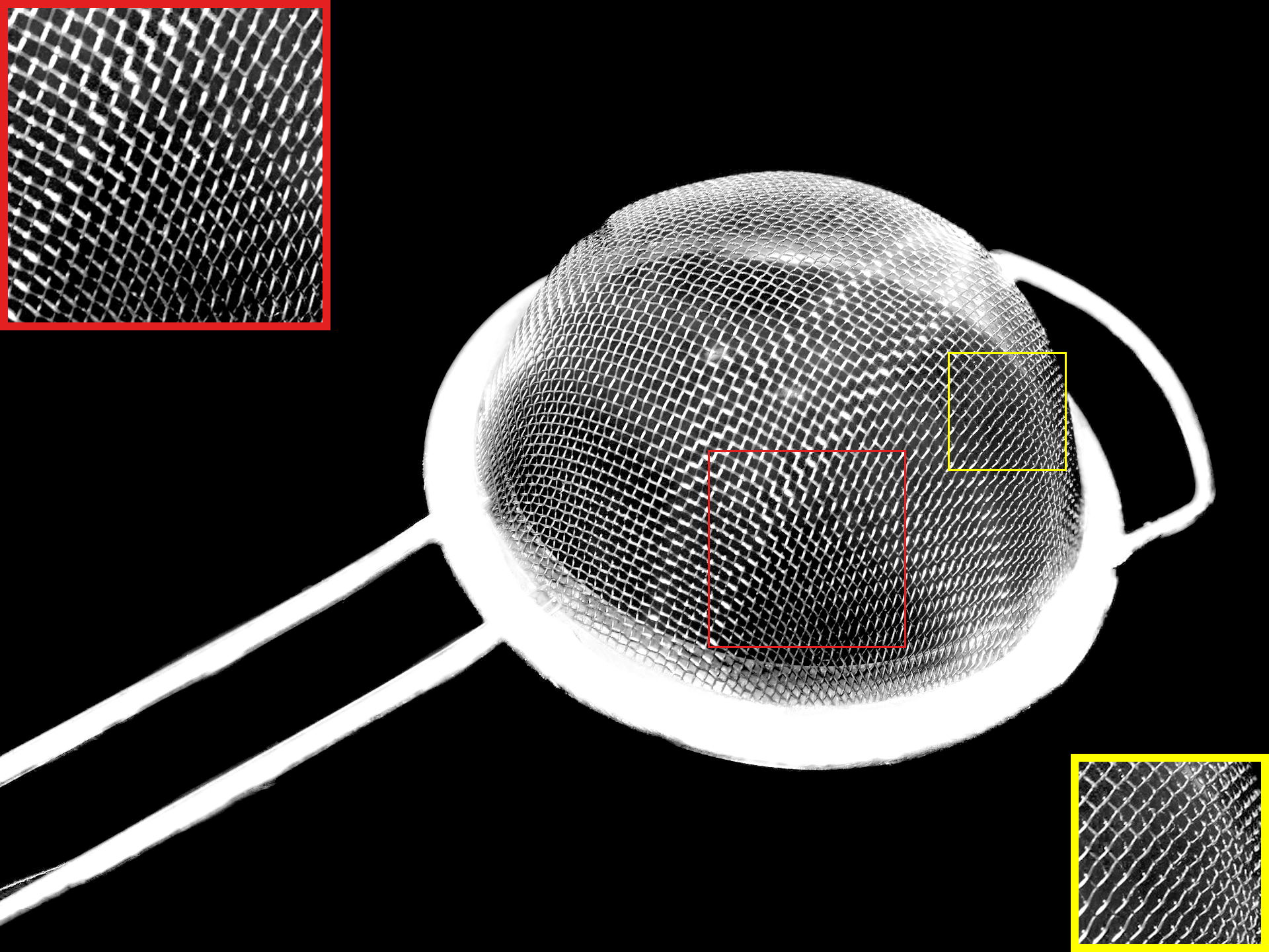} \\
		
		HAttMatting + Sen & \emph{HAttMatting++} & Ground Truth \\
\end{tabular}}
\caption{The visual comparison of different components. The ``HAttMatting + MS'' and ``HAttMatting + Sen'' means that we embed secondary appearance cues and sentry supervision in the previous version respectively, while \emph{HAttMatting++} represent our current full model. Each component has a significant improvement for alpha mattes. Please zoom in to see the finer difference.}
\label{fig:visual_basic}
\end{figure}

\subsubsection{Primary Combination}
\label{ssec:ori}
Here we analyze the effectiveness of different components in the previous version~\cite{Qiao_2020_CVPR}. We take the input image from Adobe Composition-1k dataset as an visual instance, and the corresponding alpha mattes produced by some following models are shown in the first and second rows of Fig.~\ref{fig:visual_basic}. The \textbf{Low} here denotes the appearance cues from the block1.

\textbf{Basic:} This is our baseline network, which only uses original pyramidal features to generate alpha mattes, and optimized by $\mathcal L_{MSE}$ and $\mathcal L_{adv}$.

\textbf{Basic + SSIM:} $\mathcal L_{SSIM}$ is involved in our loss function. It can be obviously seen from Table~\ref{tab:quantitative_adobe} and Table~\ref{tab:quantitative our} that even if only $\mathcal L_{SSIM}$ is added to optimize the baseline network, the results can surpass some traditional methods, such as Shared Matting\cite{gastal2010shared}, Global Matting~\cite{Rhemann2011A}, ClosedForm~\cite{Levin2007A} and KNN Matting~\cite{Chen2013KNN}. Meanwhile, since DCNN~\cite{cho2019deep} utilizes the results of closed-form matting~\cite{Levin2007A} and KNN matting~\cite{Chen2013KNN} to predict alpha mattes by relatively mediocre networks, our method is deservedly better than DCNN~\cite{cho2019deep}.

\textbf{Basic + Low:} Low-level appearance cues are directly aggregated with pyramidal features, which can furnish sophisticated texture and details for alpha mattes. Compared to the pyramidal features that own the majority of advanced semantics for locating the profiles of objects, the low-level features contain most of the texture and fine-grained structure information that can contribute to the boundary of \emph{FG}. After adding low-dimensional information, all the four metrics improve a lot, which is also reasonable.

\textbf{Basic + CA:} On the basis of baseline, we perform channel-wise attention to distill pyramidal features. CA can effectively suppress irrelevantly advanced semantics and reduce the sensitivity of the trained model to object classes, which means the network can handle diverse kinds of \emph{FG} objects and the model versatility is enhanced. We observe that the metric of MSE dropped 0.007 (from 0.025 to 0.018) by adding only one CA module. This can also corroborate the importance of CA for locating the profile of \emph{FG} and restraining the response of unrelated semantics.

\textbf{Basic + Low + CA:} This combination integrates the advantages of the above two modules to promote performance (Fig.~\ref{fig:visual_basic}). Both components can boost the performance above the baseline network, and their combination can effectively aggregate matting-adapted semantics and refined boundaries as expected.

\textbf{Basic + Low + SA:} Our modified SA can eliminate the \emph{BG} texture in appearance cues, improving subsequent aggregation process. It is noting that the spatial-attention map must exert on the low-level features, thus the``basic + SA'' comparison is unavailable in our ablation experiment. According to our quantitative and qualitative report, the Low and SA module can bring a huge enhancement to the final alpha mattes. Besides, we can observe from Fig.~\ref{fig:visual_basic} that although spatial attention can filter redundant \emph{BG} information, they are inapplicable for attending matting-adapted semantics without CA module (the black dress in \emph{BG} is also projected into the alpha mattes).

\textbf{Basic + Low + CA + SA:} We assemble CA, Low, and SA to achieve competent alpha mattes without SSIM. 

More detailed quantitative comparisons are shown in Table~\ref{tab:quantitative_adobe} and Table~\ref{tab:quantitative our}. It can be clearly seen that each component can significantly improve our results. The low-level structure information and advanced semantics are essentially complementary to each other in terms of image matting. CA can furnish \emph{FG} profiles by distilling pyramidal features, while SA can exhibit fine-grained internal texture and boundary details by filtrating initial appearance cues, and their aggregation can generate high-quality alpha mattes (Fig.~\ref{fig:visual_basic}).

\subsubsection{Expanded Combination}
\label{sssec:expanded}
We conduct this additional ablation study to verify the effectiveness of two newly-extended components on the basis of the previous conference version. The first is our progressive appearance cues extraction and aggregation, and the second is the sentry supervision. Similarly, these two models are trained and tested on both Adobe Composition-1K dataset and our Distinctions-646 dataset.

\textbf{HAttMatting + progressive structure aggregation:} Compared to the previous conference paper that only fuses the attended pyramidal features from block4 and appearance cues from block1, we import another information flow from block2 as the secondary appearance cues to execute structure aggregation, under the guidance of our attention mechanism progressively. We argue that this design can further integrate middle-levels information (the hands, leaves etc) to achieve multi-scale features aggregation. As shown in Fig.~\ref{fig:visual_basic} (this item is denoted as ``HAttMatting + MS''), the stripes of the sieve are further distilled (please zoom in to see the yellow box). Furthermore, the three metrics except the SAD improved slightly, which can also confirm the validity of secondary appearance cues for complementing necessary \emph{FG} texture and details.
\begin{figure*}[b]
	\centering
	\setlength{\tabcolsep}{1pt}\small{
		\begin{tabular}{ccccc}
			\includegraphics[width=0.19\linewidth]{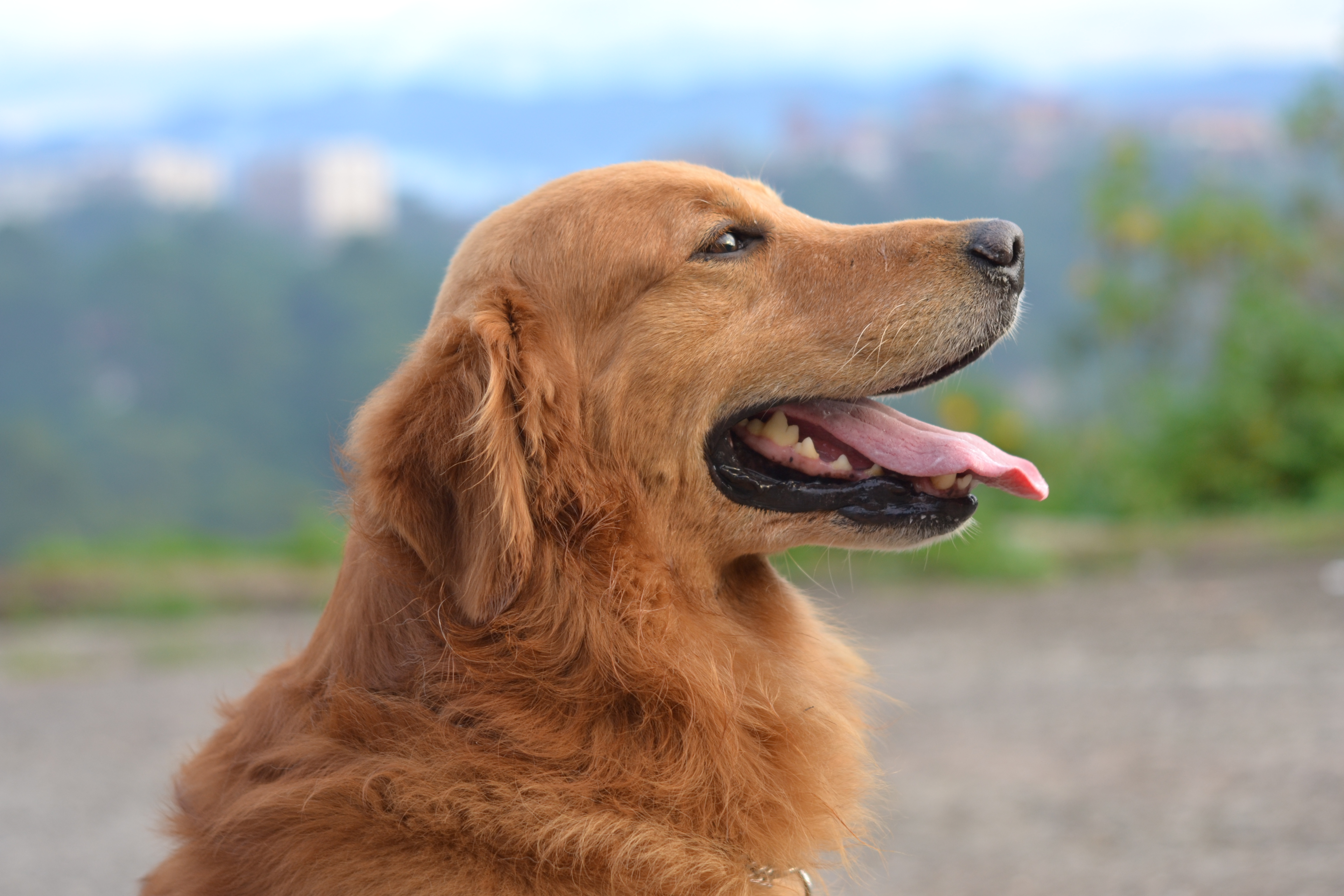} &
			\includegraphics[width=0.19\linewidth]{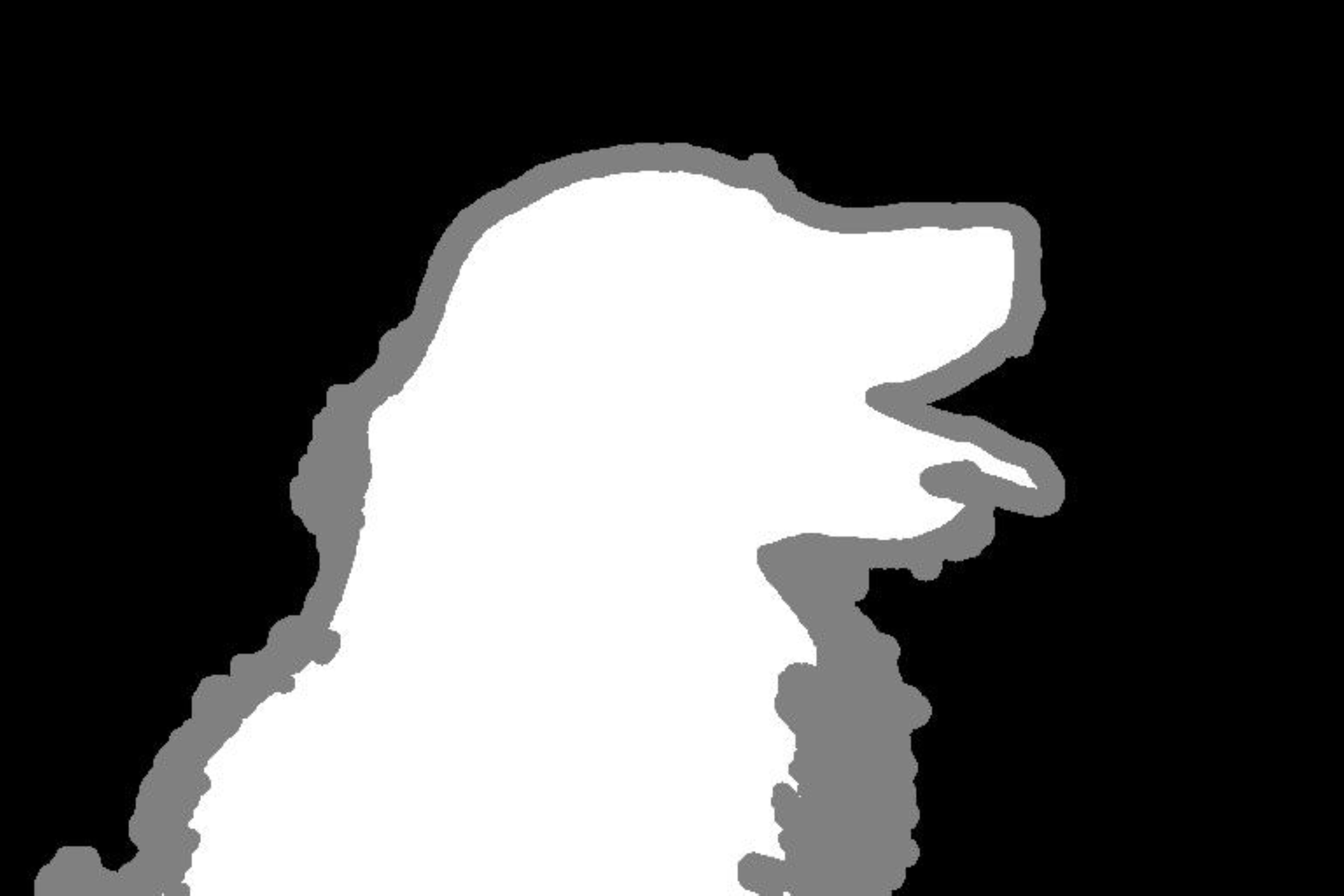} &
			\includegraphics[width=0.19\linewidth]{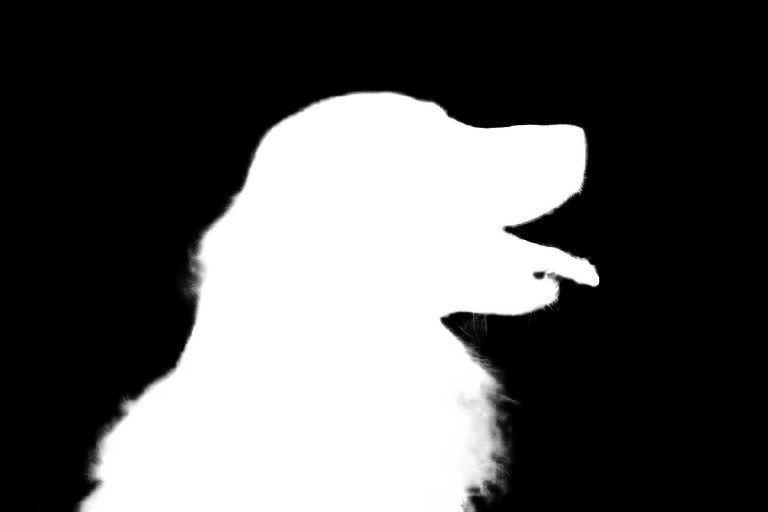} &
			\includegraphics[width=0.19\linewidth]{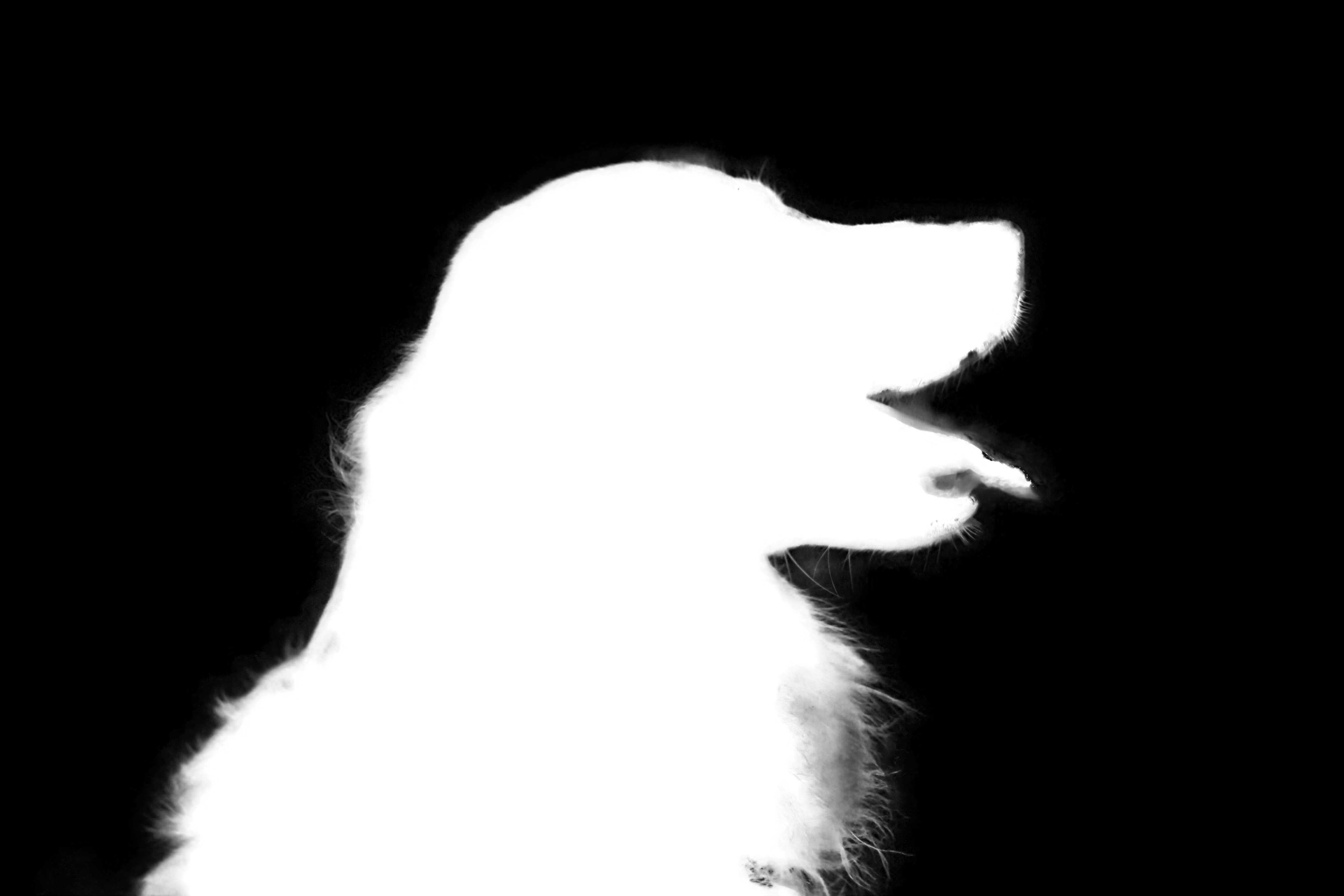} &
			\includegraphics[width=0.19\linewidth]{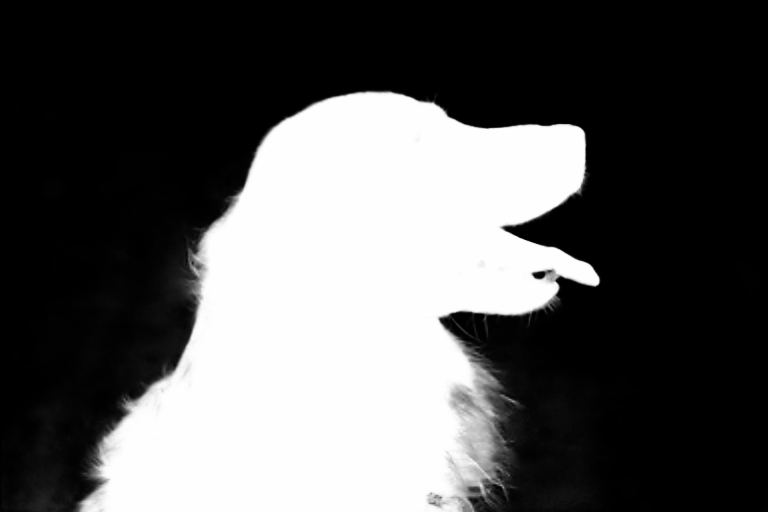} \\
			
			\includegraphics[width=0.19\linewidth]{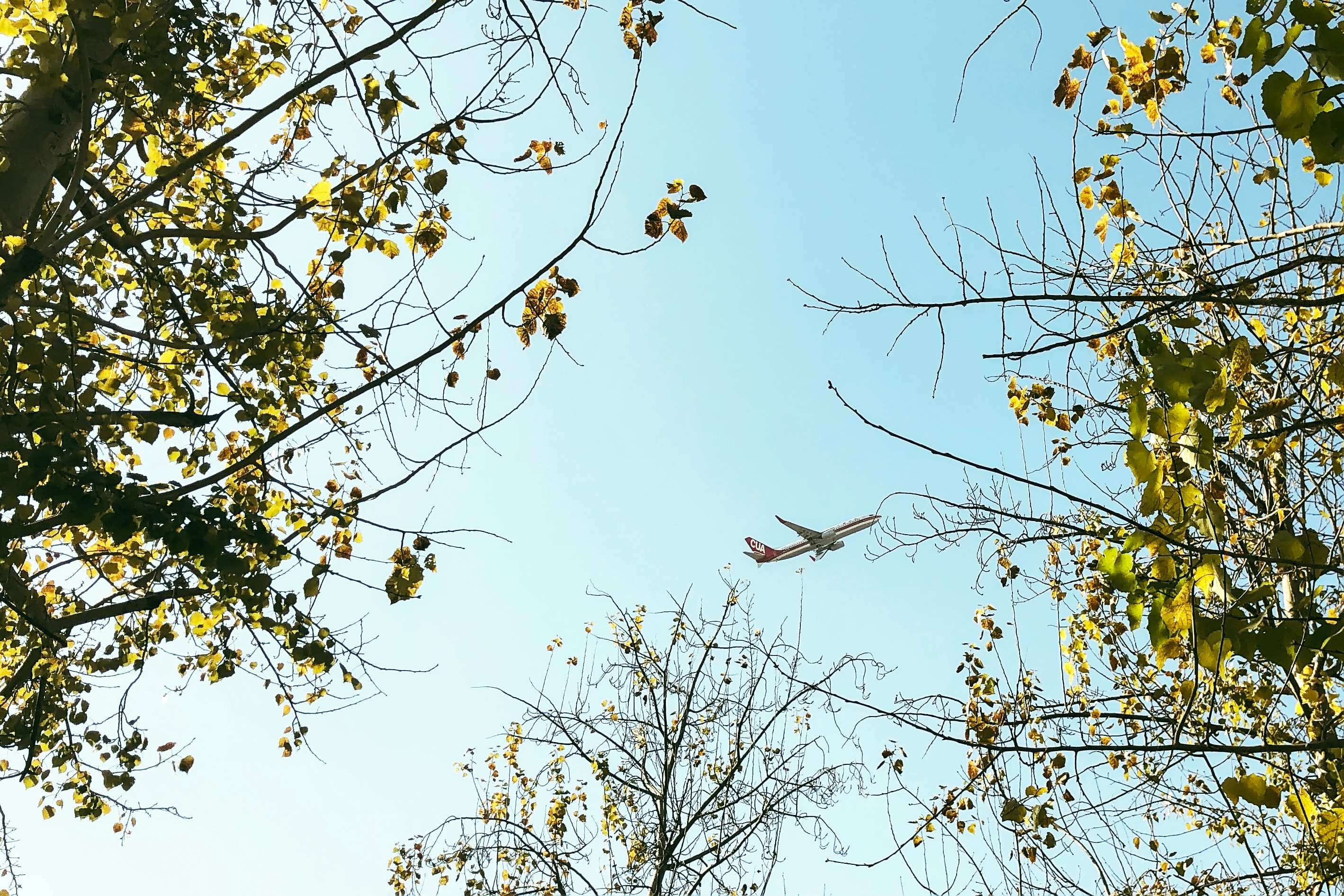} &
			\includegraphics[width=0.19\linewidth]{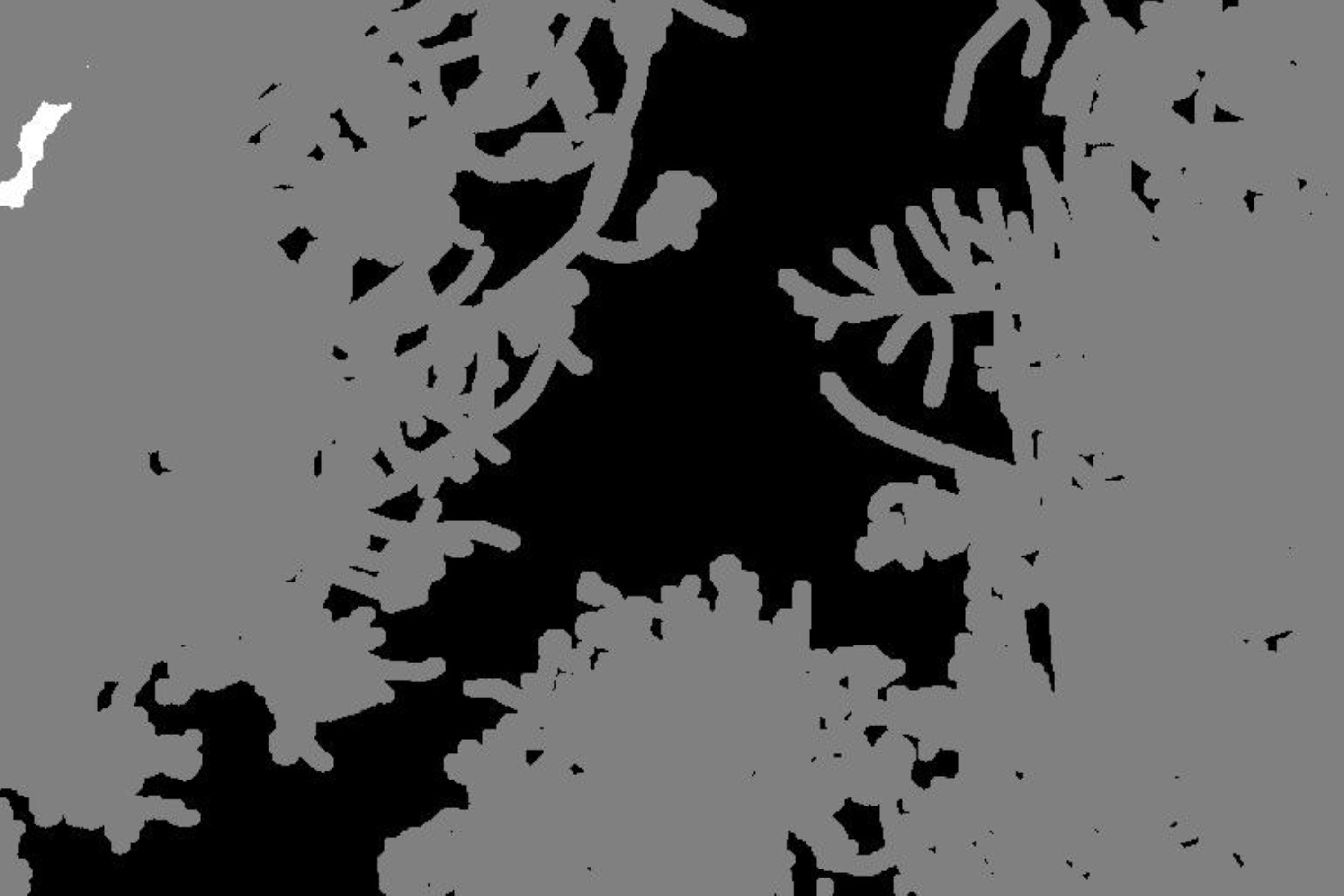} &
			\includegraphics[width=0.19\linewidth]{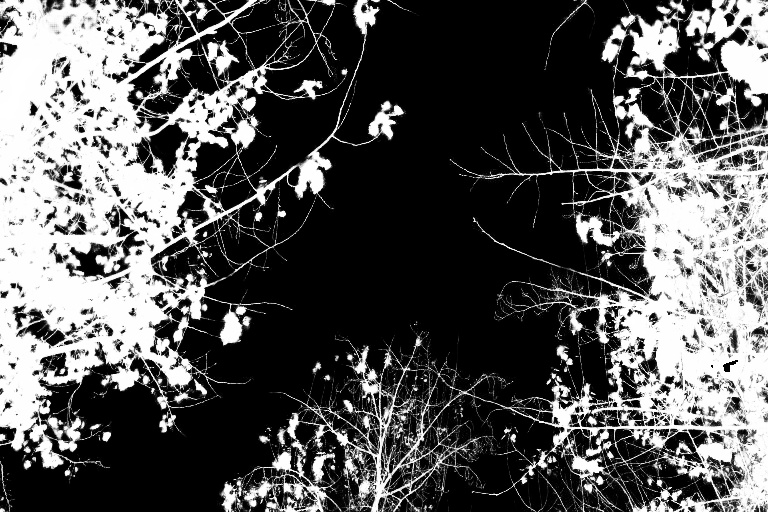} &
			\includegraphics[width=0.19\linewidth]{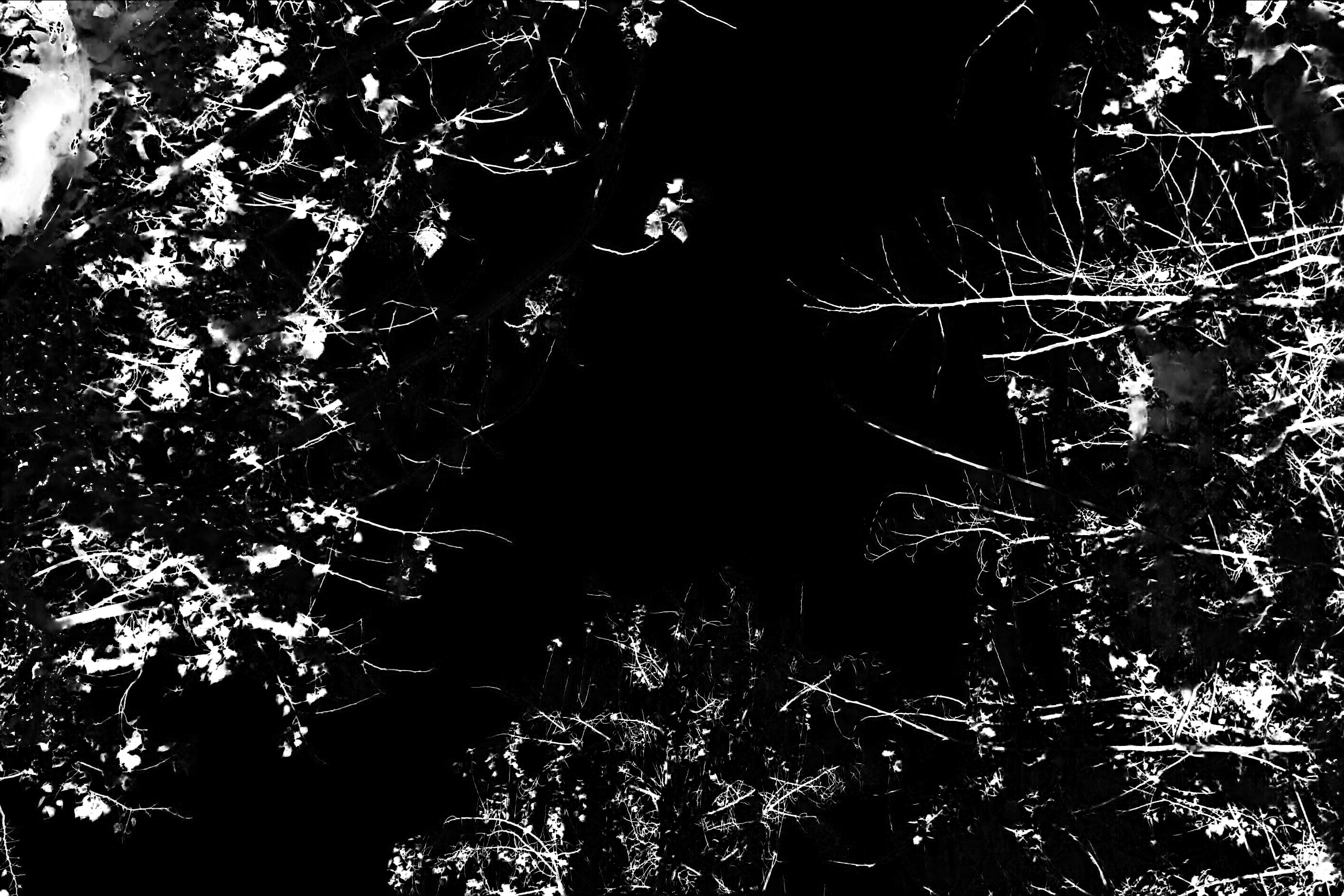} &
			\includegraphics[width=0.19\linewidth]{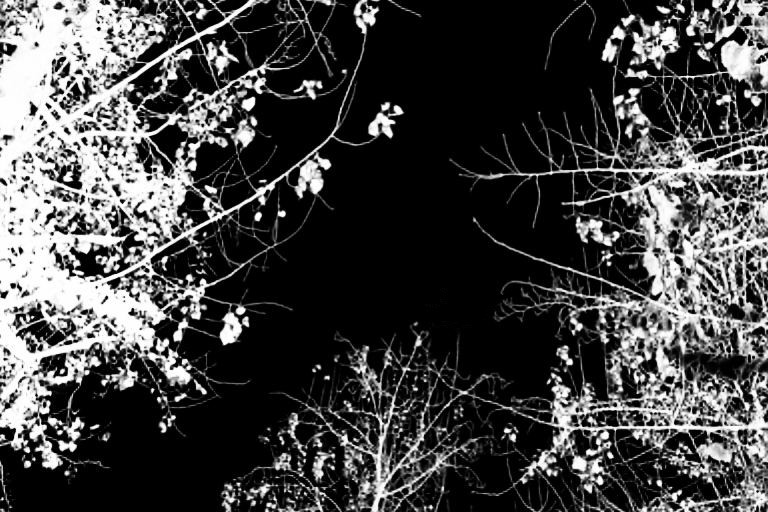} \\
			
			\includegraphics[width=0.19\linewidth]{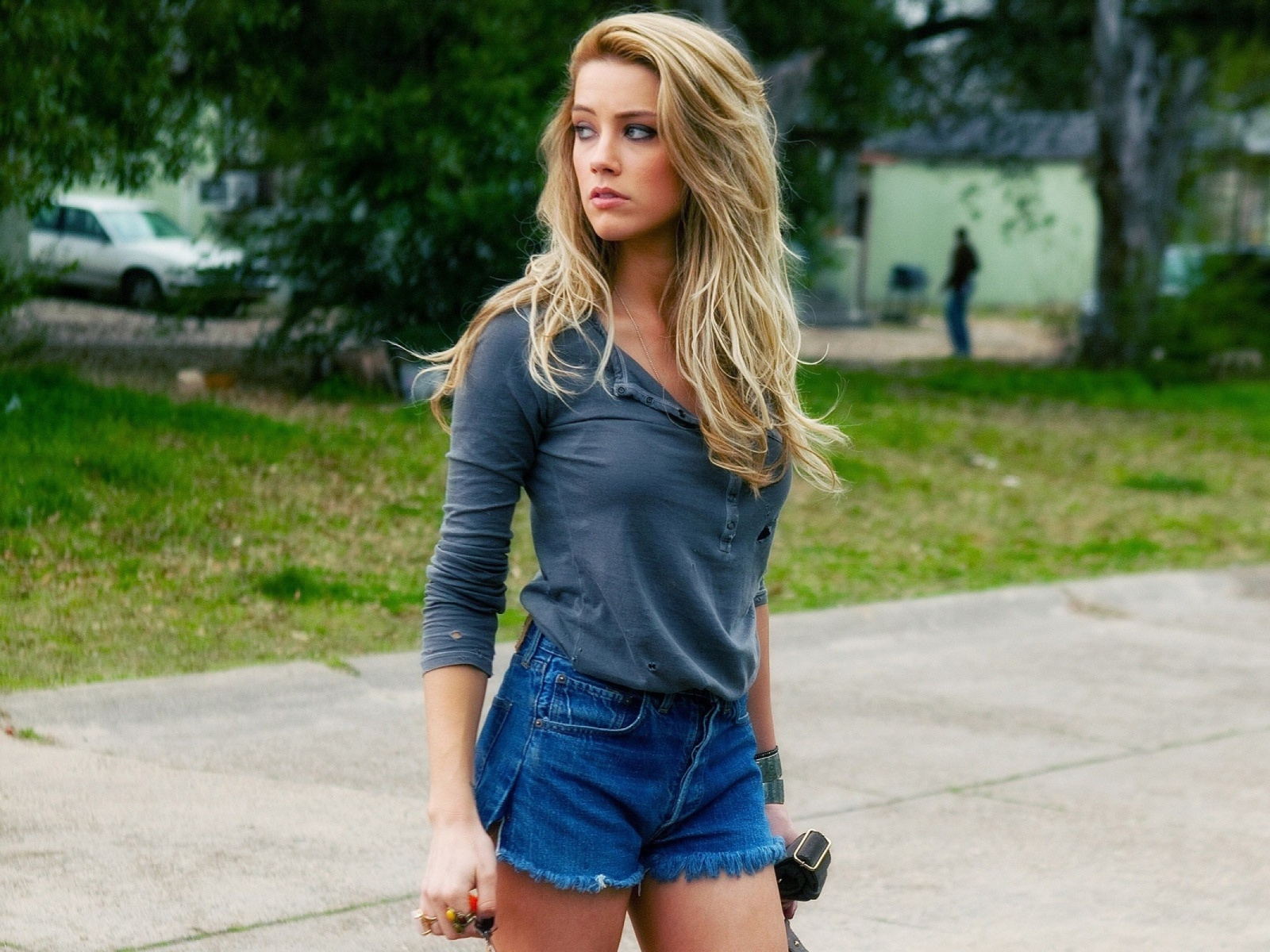} &
			\includegraphics[width=0.19\linewidth]{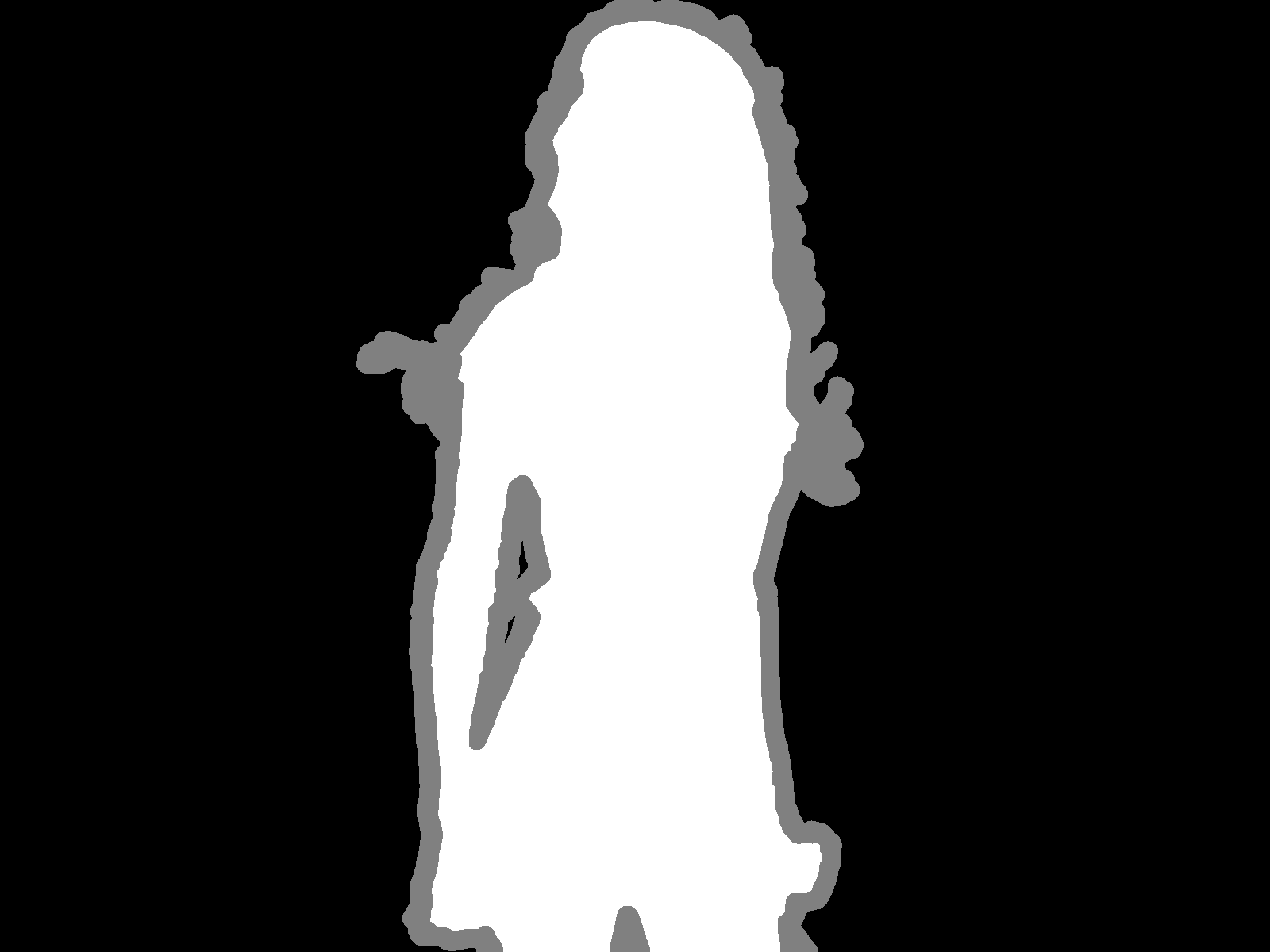} &
			\includegraphics[width=0.19\linewidth]{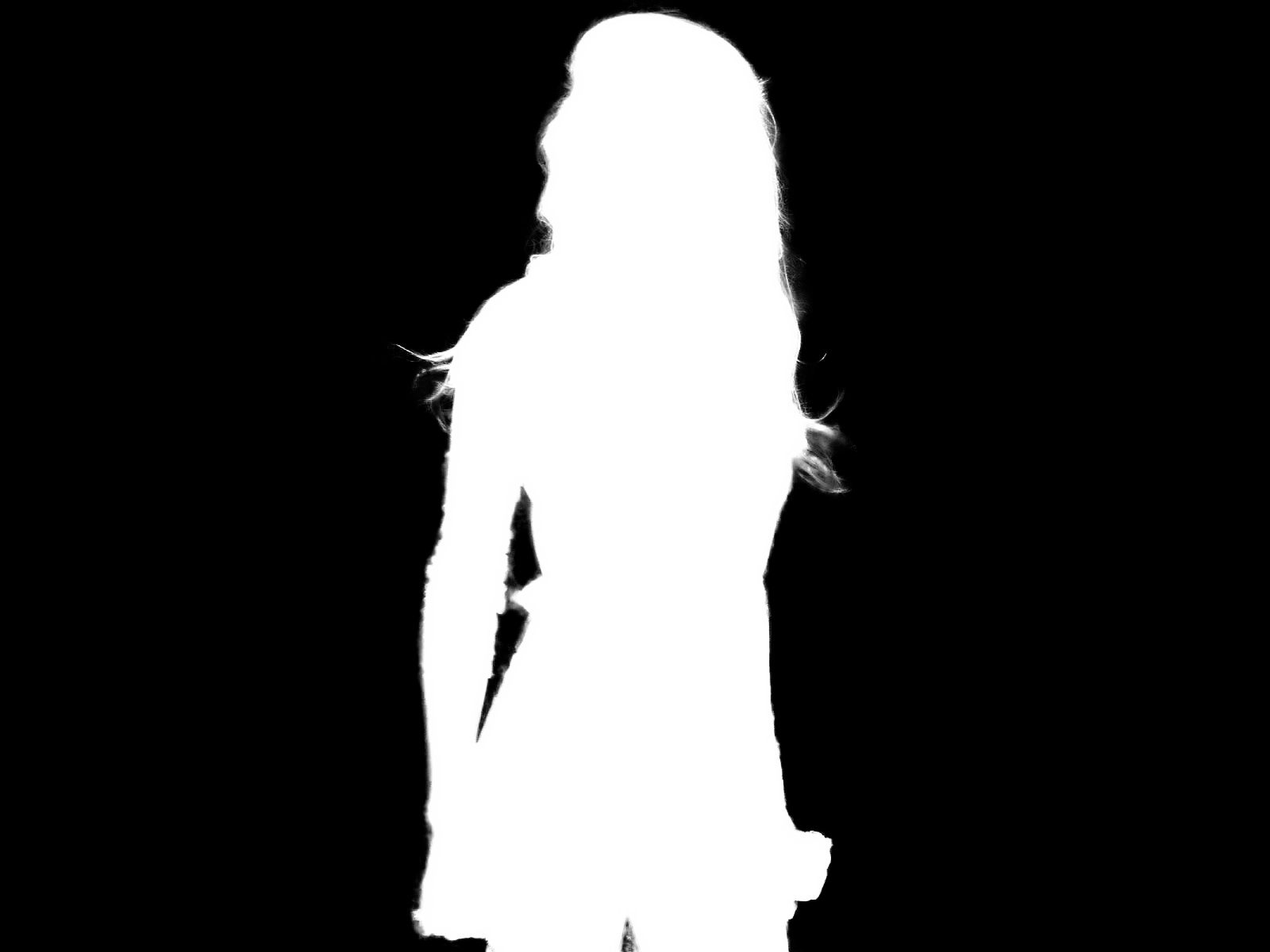} &
			\includegraphics[width=0.19\linewidth]{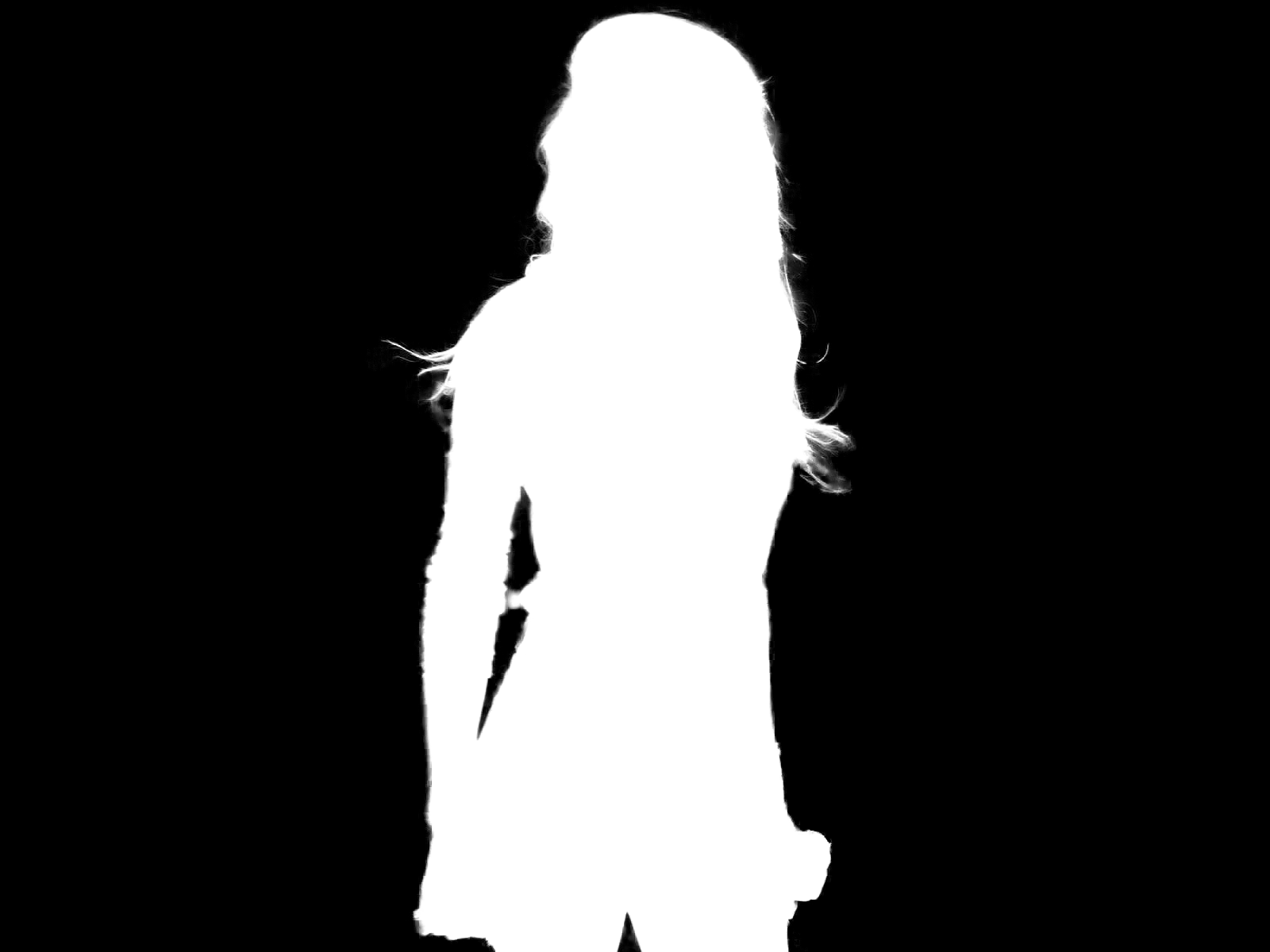} &
			\includegraphics[width=0.19\linewidth]{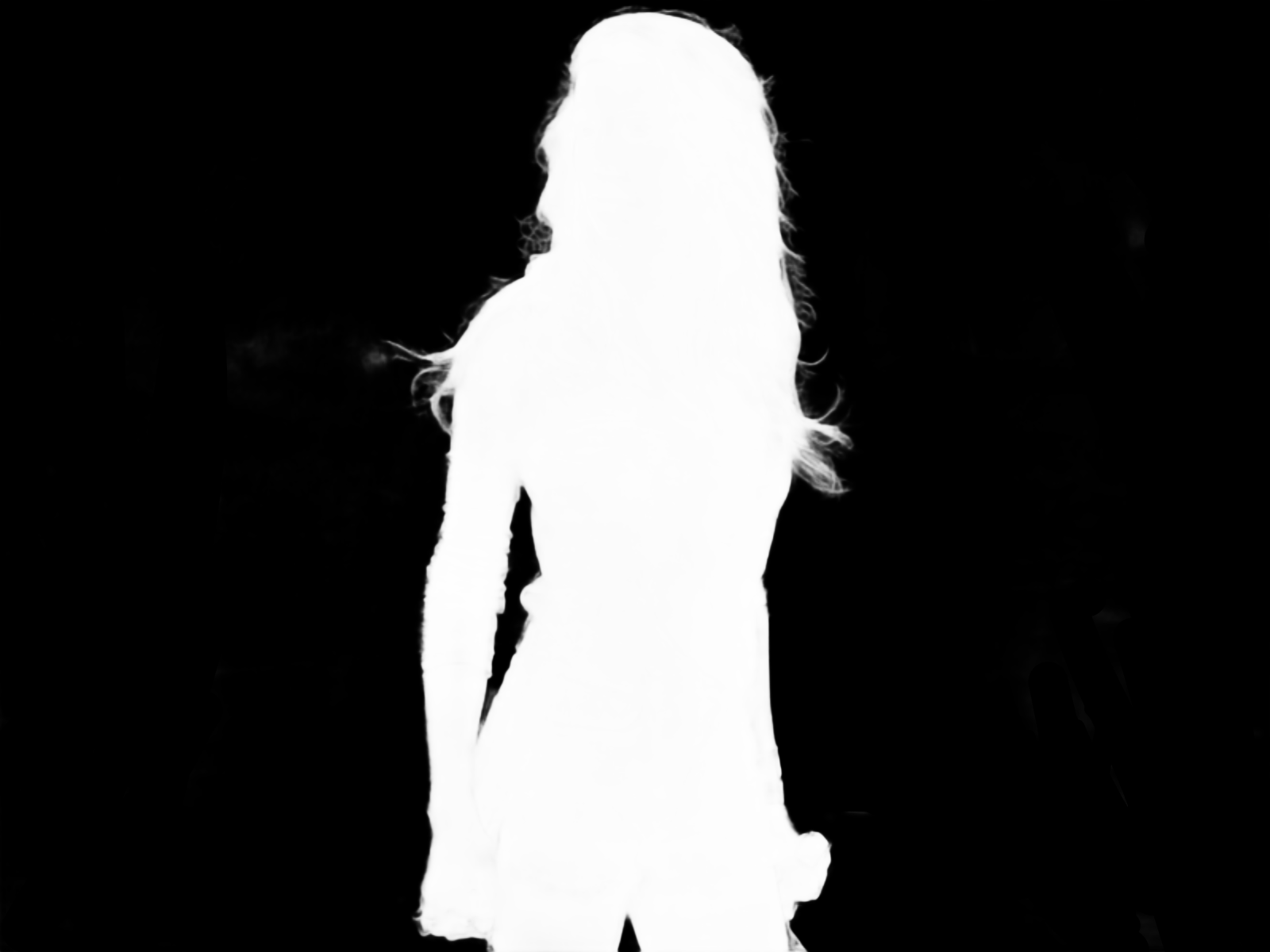} \\
			
			\includegraphics[width=0.19\linewidth]{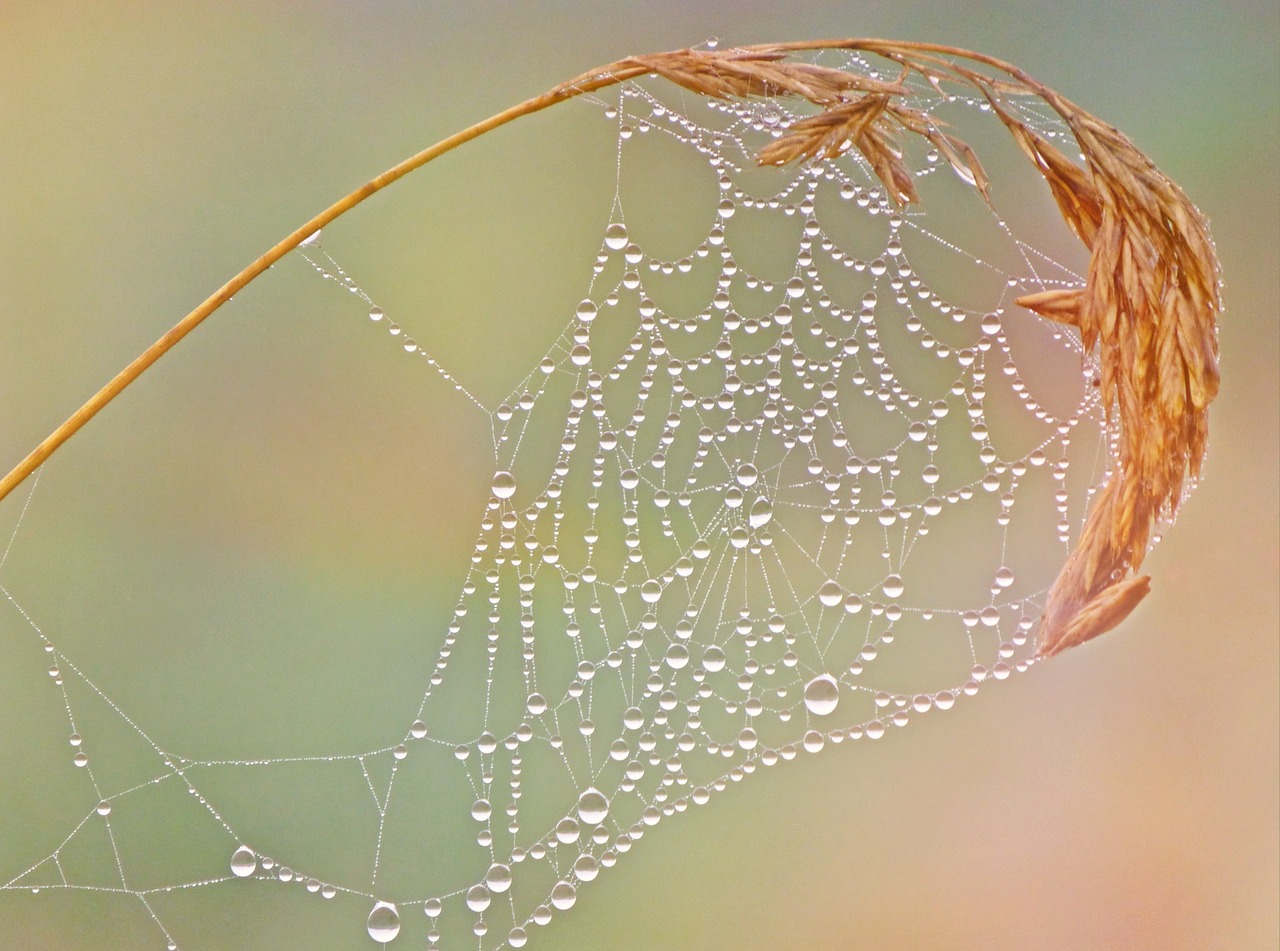} &
			\includegraphics[width=0.19\linewidth]{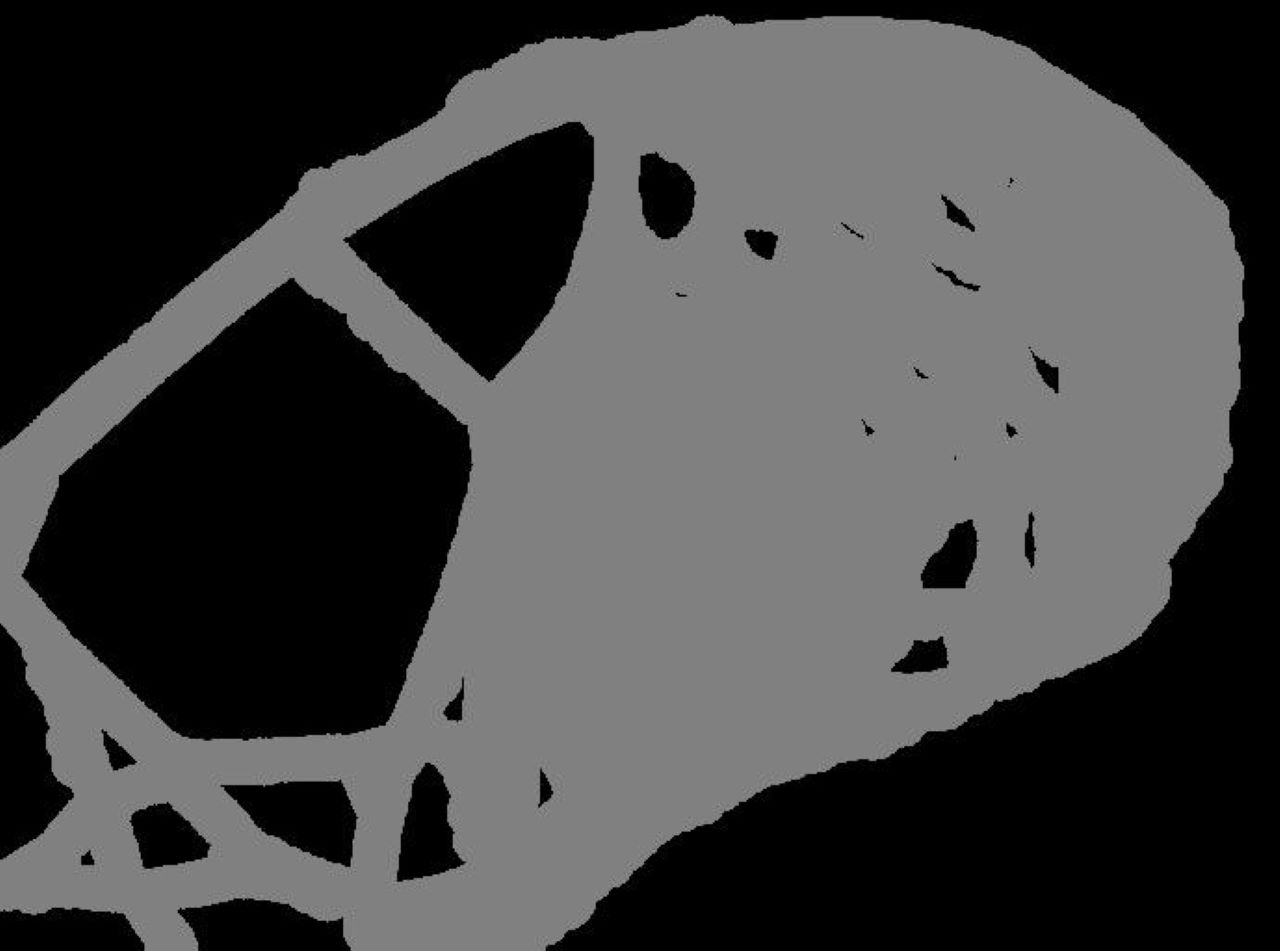} &
			\includegraphics[width=0.19\linewidth]{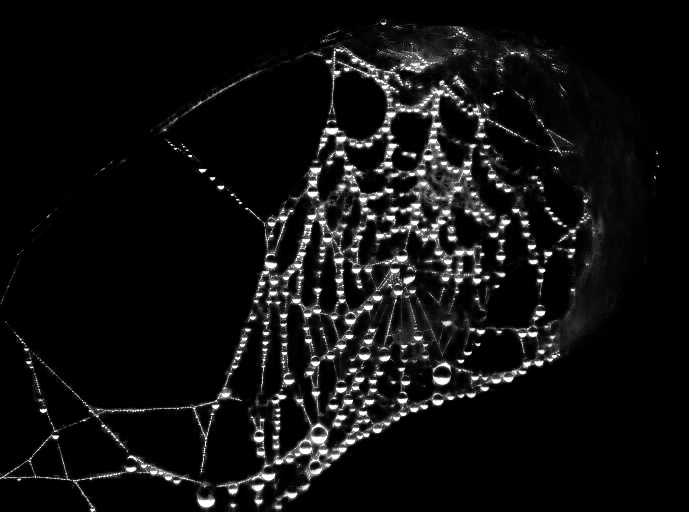} &
			\includegraphics[width=0.19\linewidth]{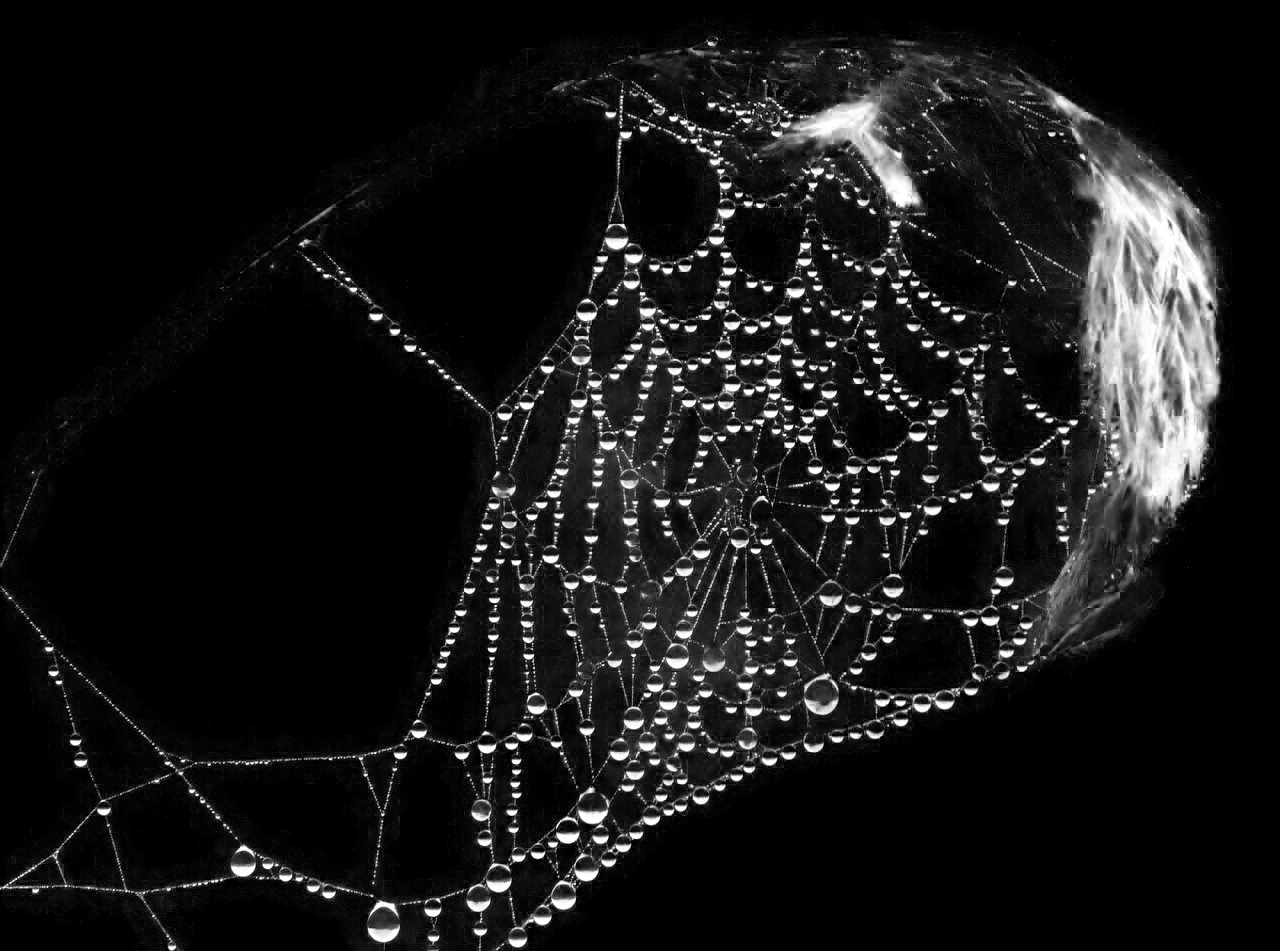} &
			\includegraphics[width=0.19\linewidth]{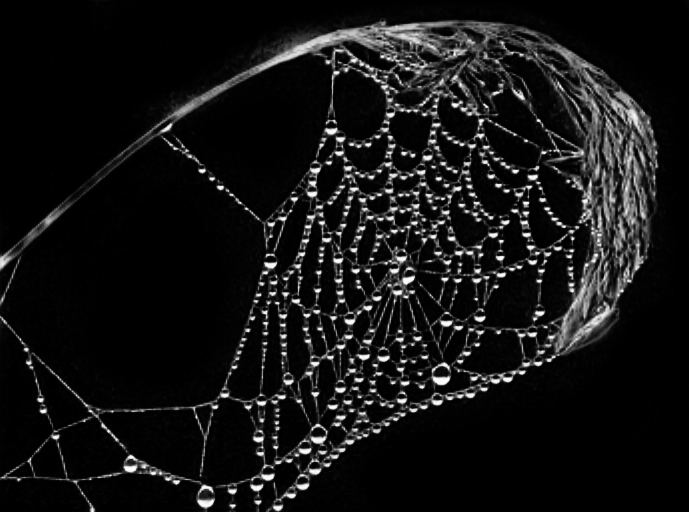} \\
			
			\includegraphics[width=0.19\linewidth]{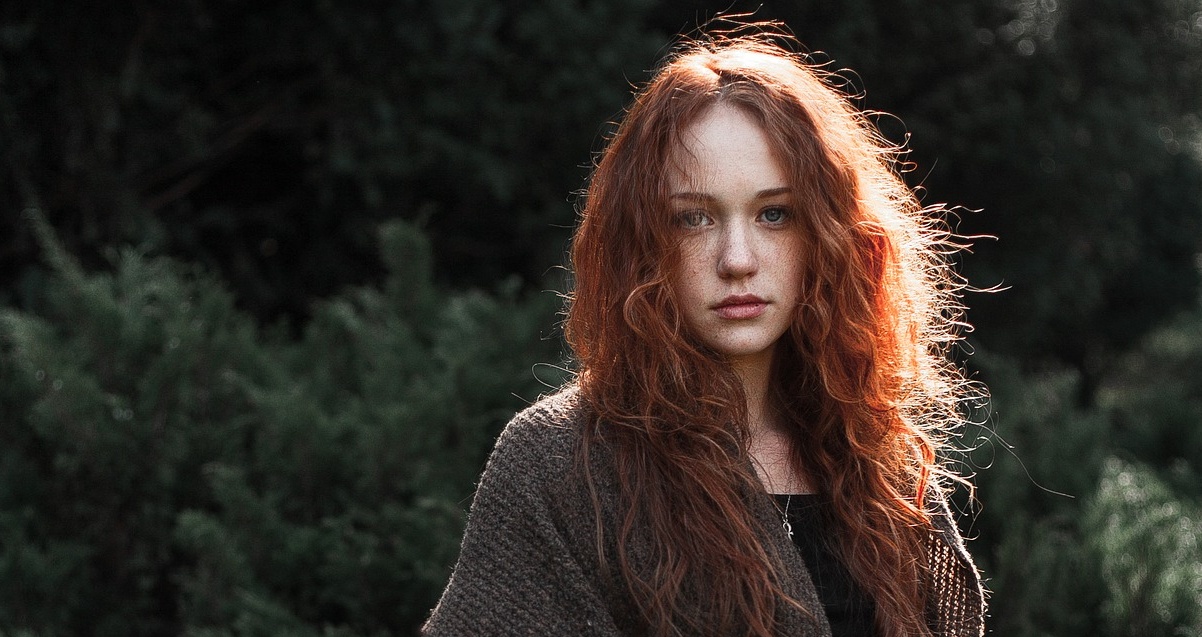} &
			\includegraphics[width=0.19\linewidth]{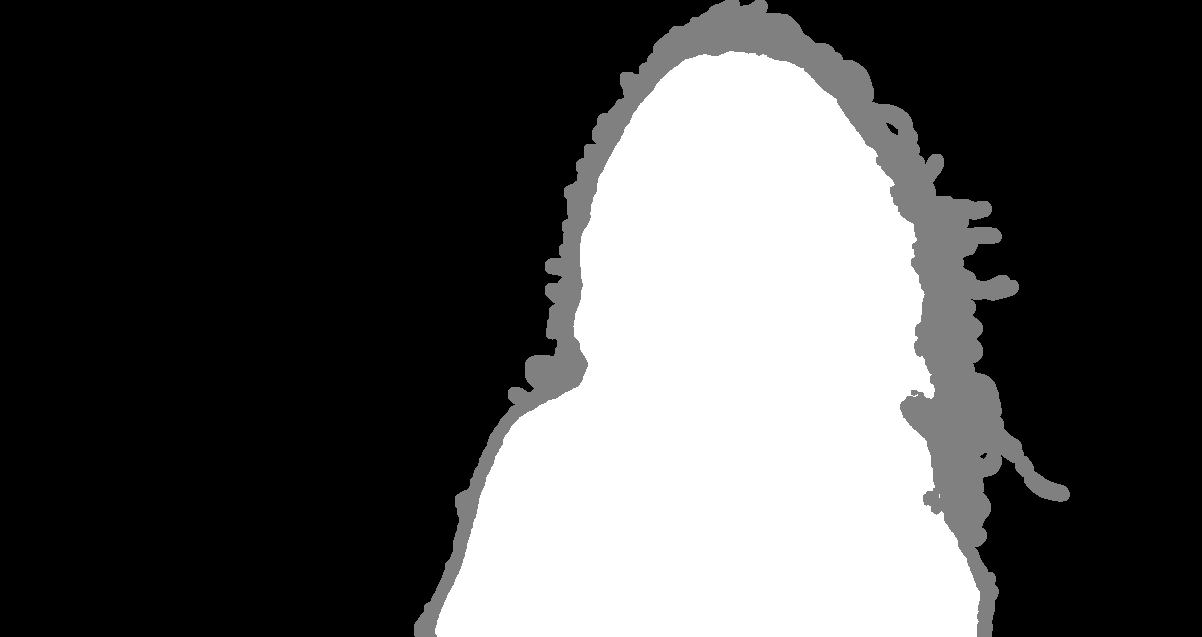} &
			\includegraphics[width=0.19\linewidth]{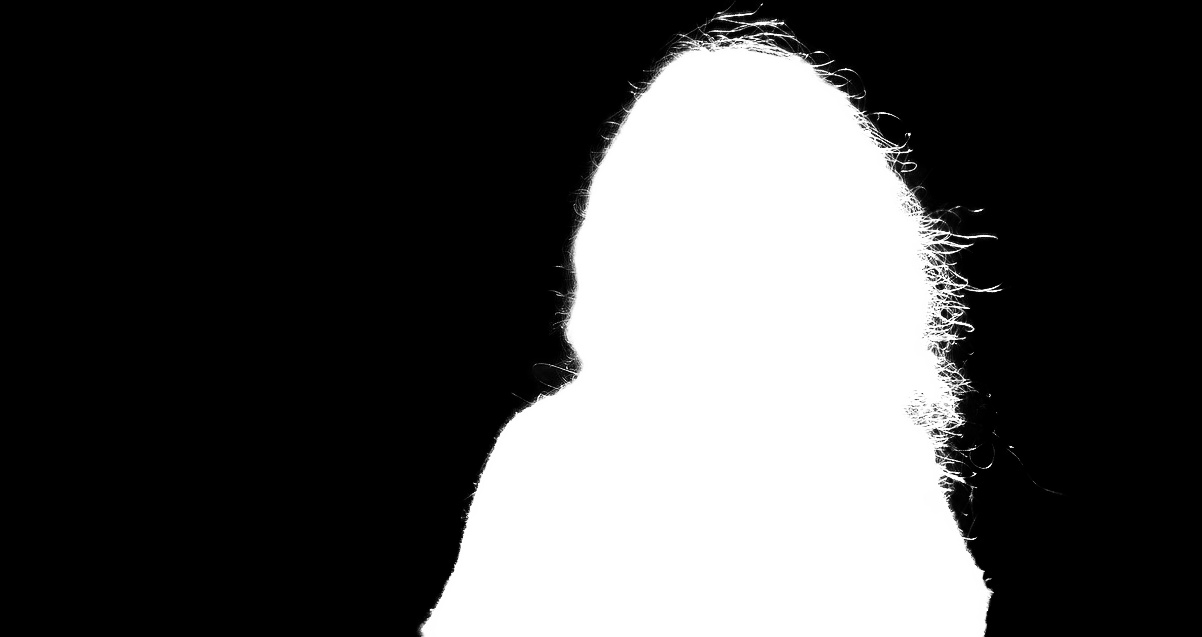} &
			\includegraphics[width=0.19\linewidth]{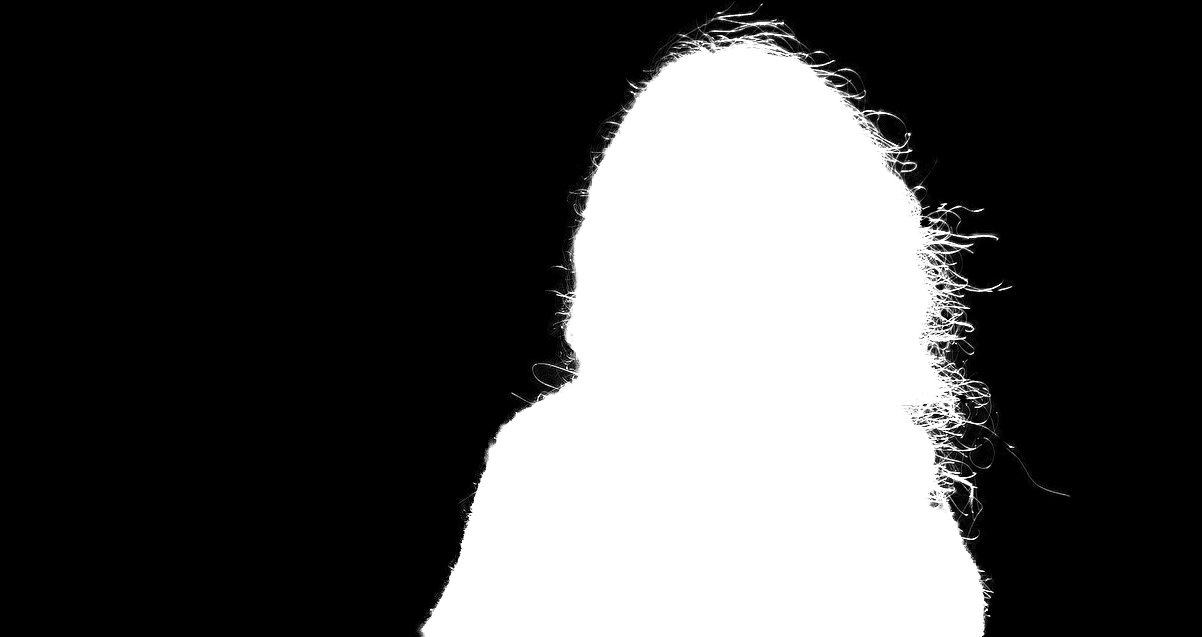} &
			\includegraphics[width=0.19\linewidth]{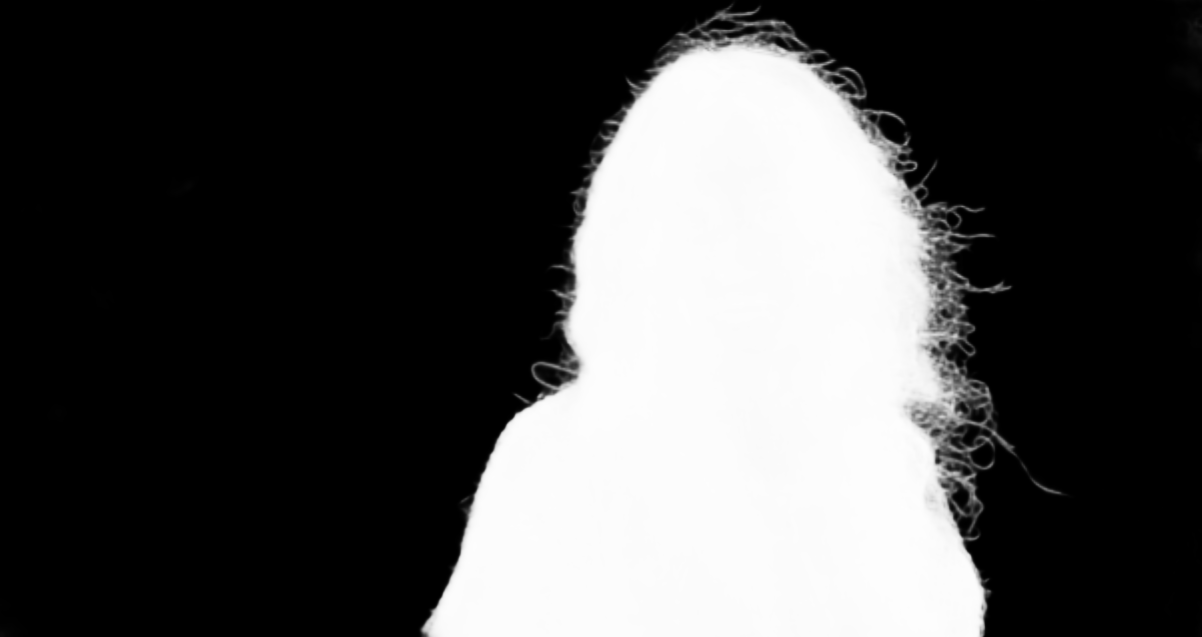} \\
			
			\includegraphics[width=0.19\linewidth]{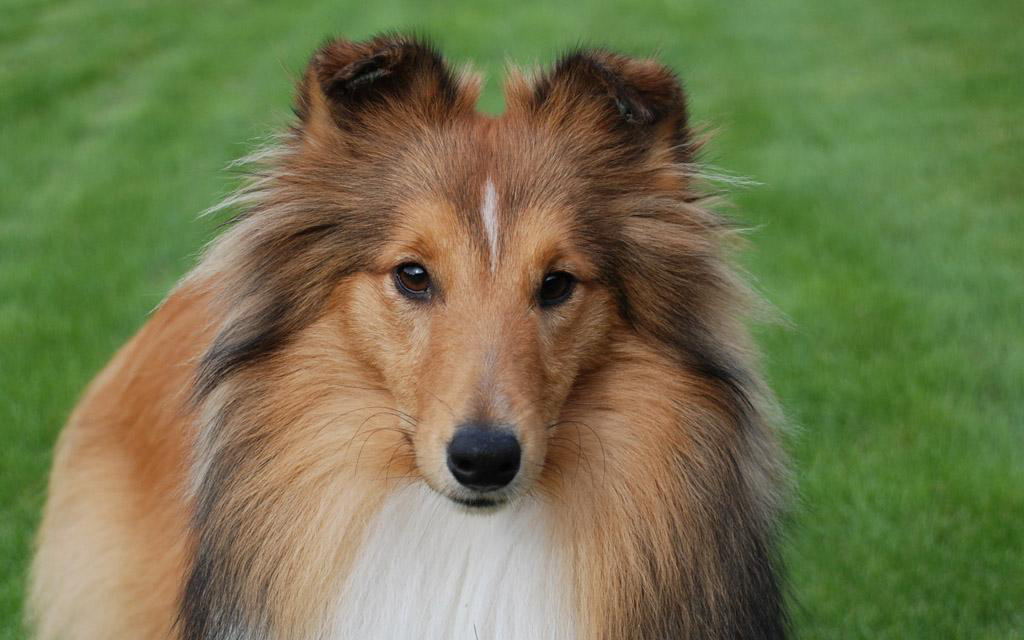} &
			\includegraphics[width=0.19\linewidth]{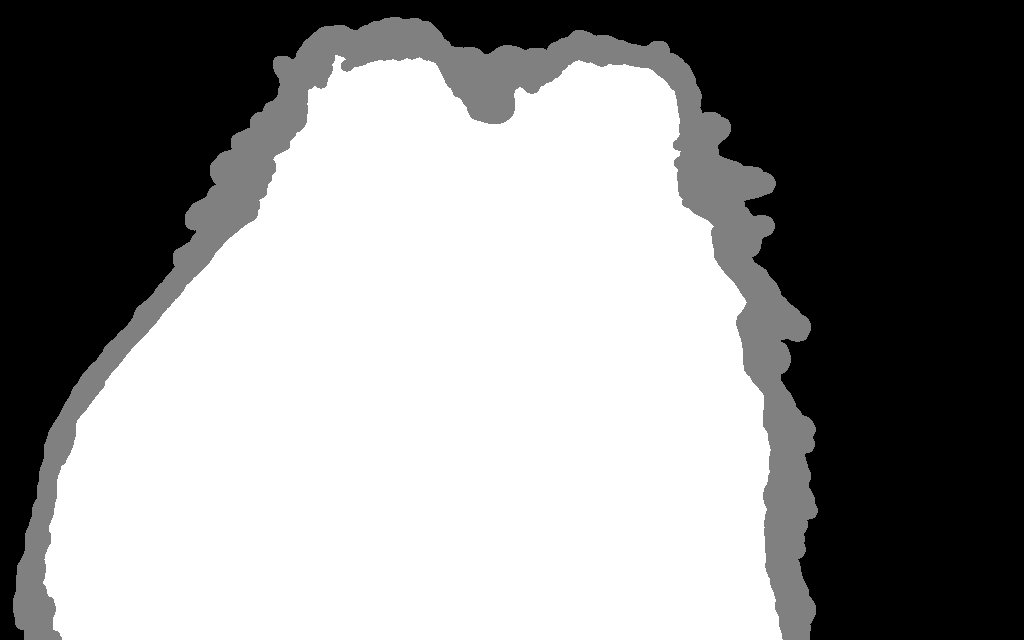} &
			\includegraphics[width=0.19\linewidth]{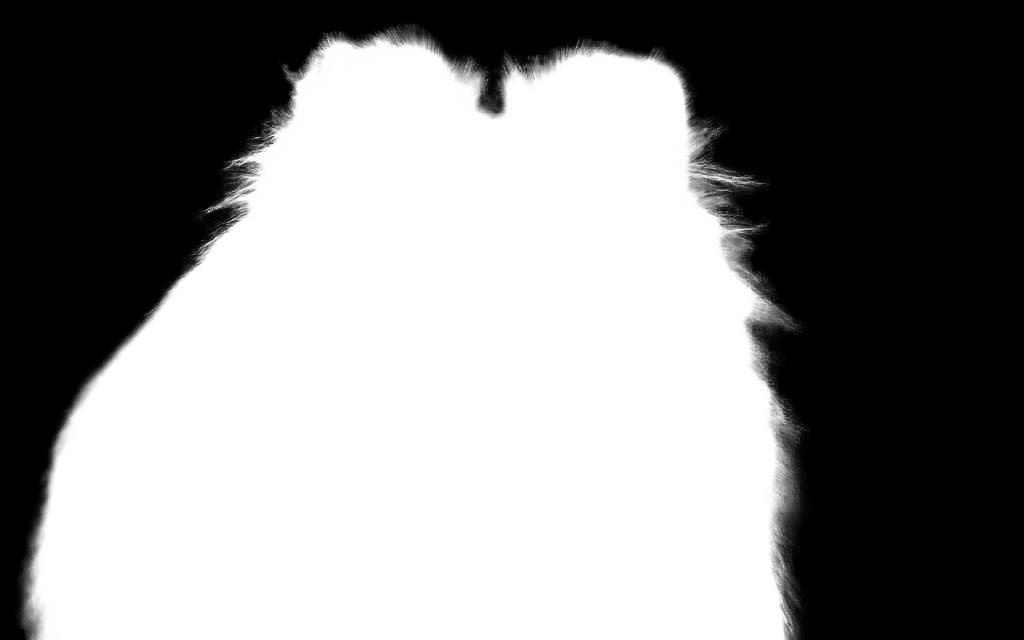} &
			\includegraphics[width=0.19\linewidth]{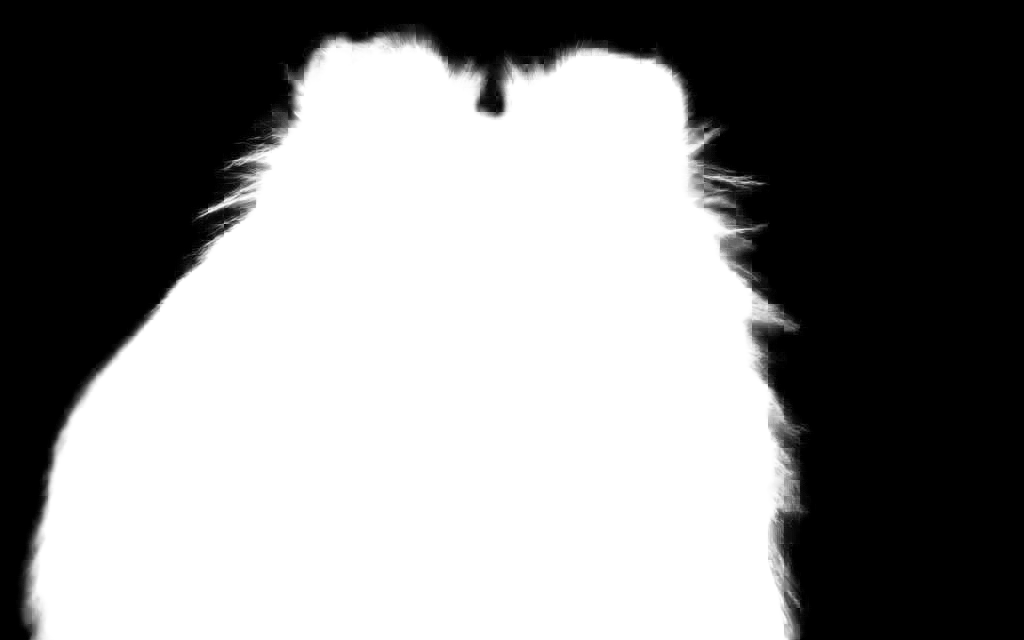} &
			\includegraphics[width=0.19\linewidth]{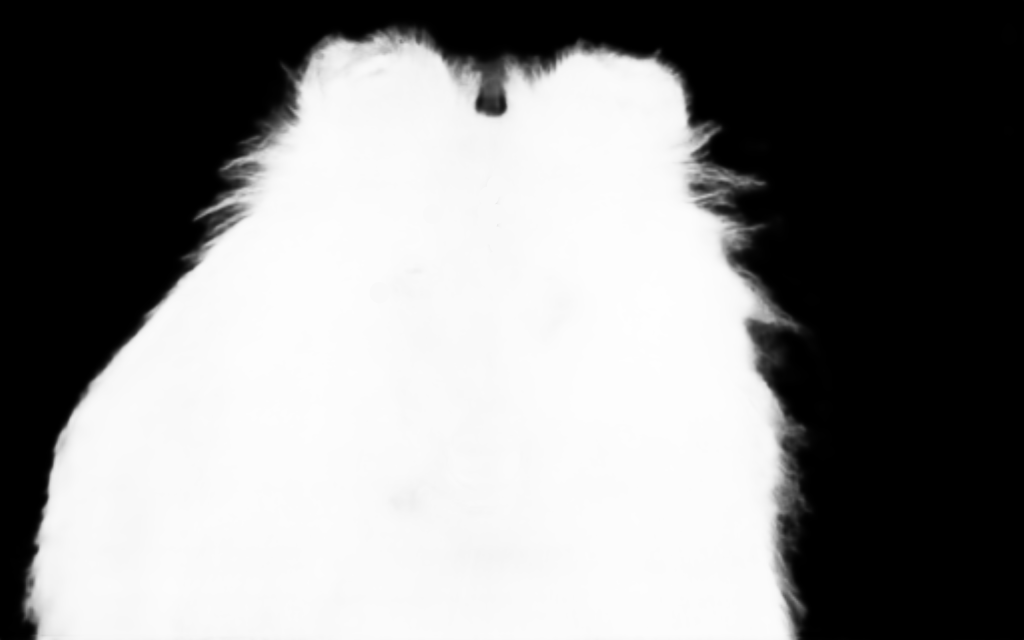} \\
			
			Input Image & Trimap & GCA~\cite{li2020natural} & A${^2}$U~\cite{dai2021learning} & Ours \\
	\end{tabular}}
	\caption{The results on real world images. }
	\label{fig:visual_real}
\end{figure*}

\textbf{HAttMatting + sentry supervision:} We downsample the ground truth alpha mattes by a factor of 4 and embed it at the back of pyramidal features distillation module as our sentry supervision. As shown in Table~\ref{tab:quantitative_adobe}, there has been a slight decline in all four metrics after we add sentry supervision. We attribute this improvement to the more pure extraction of advanced semantics. First, the sentry supervision can guide the channel-wise attention to further capture the matting-adapted semantics, and prevent the misclassification of pixels inside the \emph{FG}. This modification can promote the constraint of \emph{FG} and \emph{BG} regions, especially for our trimap-free motivation. Second, the resolution of the input images in matting are usually very high, and such middle-size sentry supervision can shield some unrelated small objects and enhance the main semantics of \emph{FG} subject. Third, according to~\ref{sssec:SA}, it can produce better guidance to assist the spatial attention to restrict unnecessary \emph{BG} noises. Additionally, the back propagation of the intermediate loss can accelerate the convergence of the whole model, while preventing the gradient from disappearing.

Consequently, the proposed model contains all components (i.e. channel-wise attention (CA), spatial  attention (SA) ,structural similarity loss (SSIM), sentry supervision loss and GAN loss) can achieve the best performance, which demonstrates that all the components listed above are necessary for the proposed method to generate better alpha mattes.
\begin{figure}[t]
\centering
\setlength{\tabcolsep}{1pt}\small{
	\begin{tabular}{cccc}
		\includegraphics[scale=0.0318]{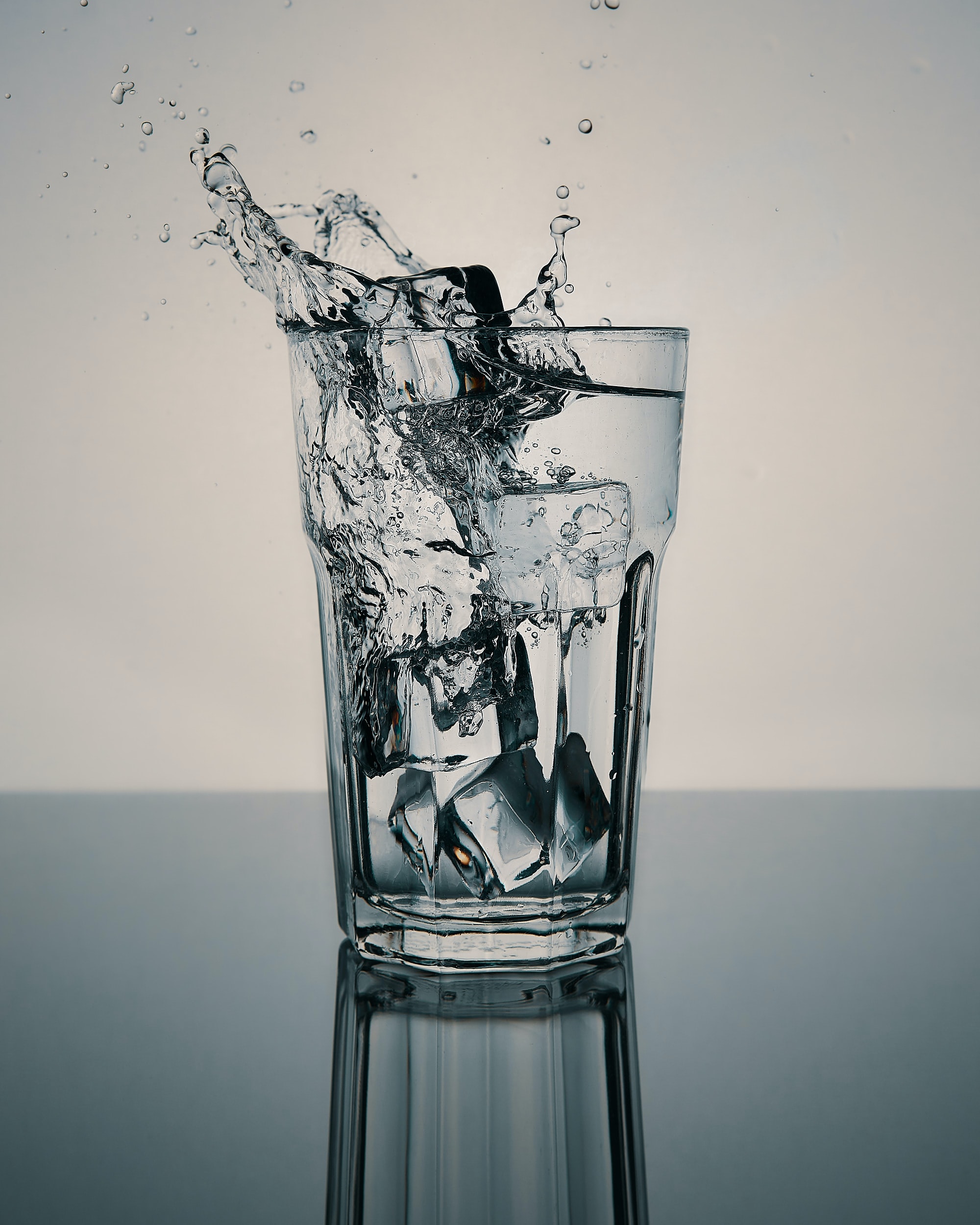} &
		\includegraphics[scale=0.124]{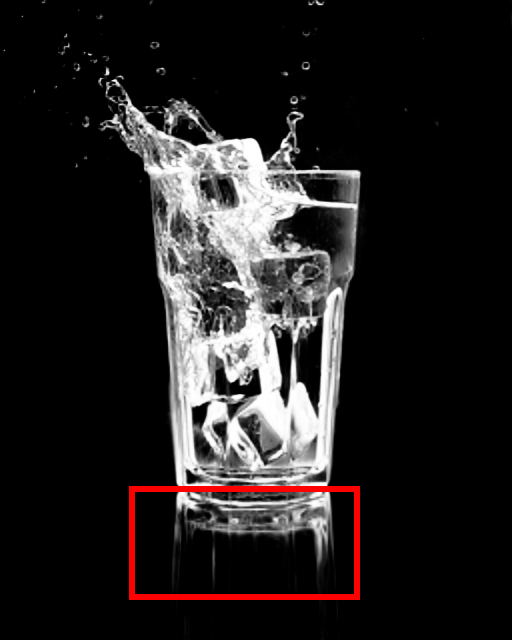} &
		\includegraphics[scale=0.388]{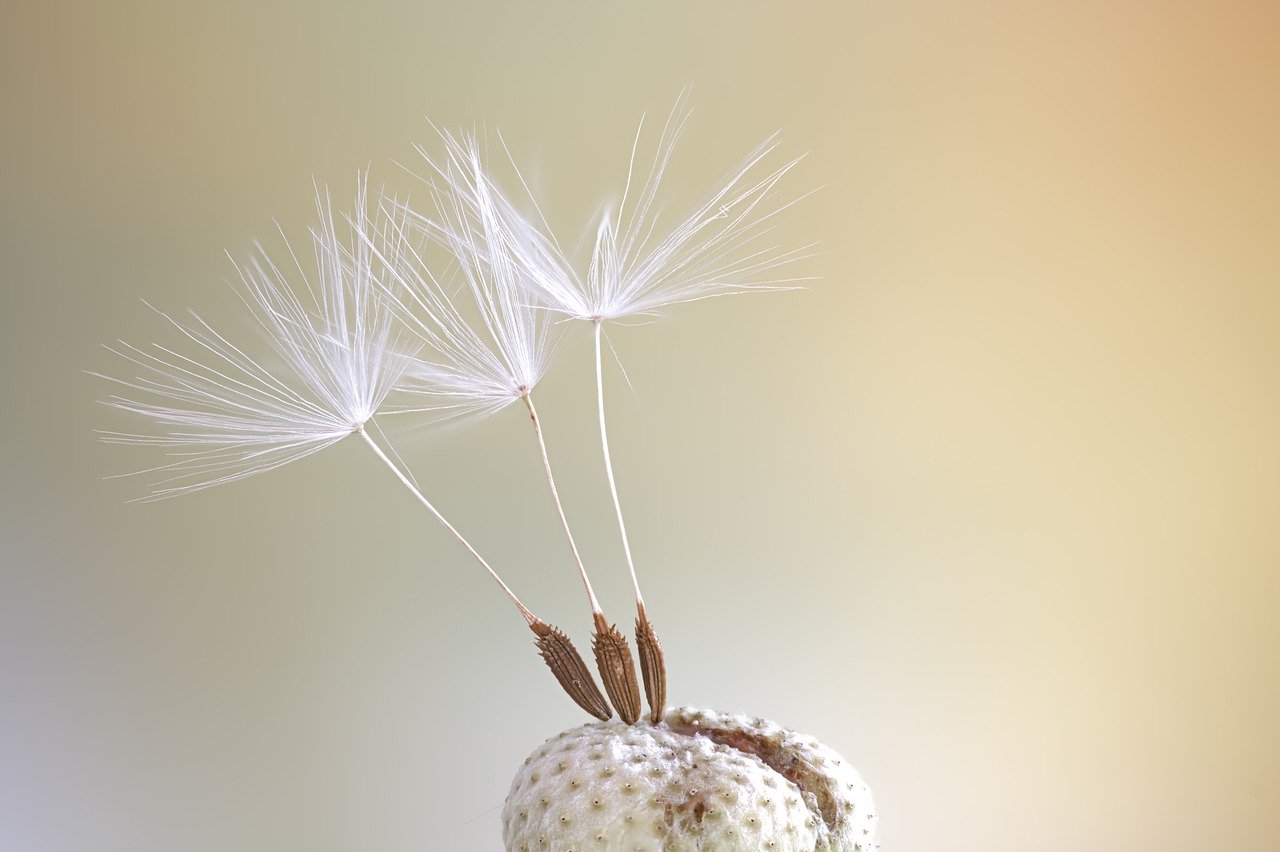} &
		\includegraphics[scale=0.155]{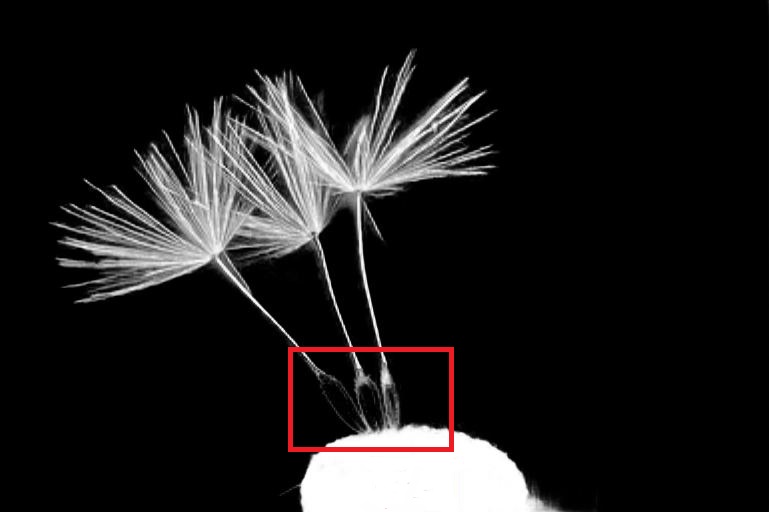} \\
		
		Input Image & Alpha Matte & Input Image & Alpha Matte \\
\end{tabular}}
\caption{Some failure cases on real-world examples. }
\label{fig:visual_fail}
\end{figure}

\subsection{Results on Real-world Images}
\label{ssec:real_images}
Fig.~\ref{fig:visual_real} shows our matting results on real-world images. All evaluation models are trained on the Composition-1k dataset. We dilate and erode our alpha mattes to generate high-quality trimaps (the second column), which are intractable for labeling from scratch. Based on these trimaps, we present the results of GCA~\cite{li2020natural} and A${^2}$U~\cite{dai2021learning} for visual comparison. \emph{HAttMatting++} can achieve high-quality alpha mattes without any external input or user interaction. The dense branches and the fine hairs can be predicted with high precision, which is benefited from our progressive appearance cues filtration and aggregation. By contrast, GCA and A${^2}$U ignore the stalk (the fourth row) even with high-quality trimaps.

\subsection{Failure Cases and Limitations}
There are also some limitations to our method, and two failure cases are illustrated in Fig.~\ref{fig:visual_fail}. The reflection, emptiness, blur, etc. of natural images can greatly influence our methods (the bottom of the cup is a struggle and the branches of the dandelion are poorly detected). The reflection and emptiness can provide pseudo information for our hierarchical and progressive features extraction, and result in misleading guidance for top-down aggregation. Besides, our \emph{HAttMatting++} also has other limitations: (1): the data generation and augmentation rules we refer to on the composite dataset can bring more artifacts and inconsistencies, which will expand the disparity between the natural images and the composite ones (other matting networks (\cite{Xu2017Deep}, \cite{Tang_2019_CVPR} and \cite{li2020natural} etc. may be also more or less affected in the same way); (2) our model can only handle simple scenes (similar to most existing matting methods~\cite{hou2019context,Zhang2019CVPR,cai2019disentangled}, and it will be better with trimaps as assistance when predicting multiple foreground objects simultaneously.

\section{Conclusions and Future work}
\label{sec:conclusion}
In this paper, we propose a Hierarchical and Progressive Attention Matting Network (\emph{HAttMatting++}), which can predict high-quality alpha mattes from single \emph{RGB} images. The \emph{HAttMatting++} employs channel-wise attention to extract matting-adapted semantics and performs spatial attention at different levels to filtrate progressive appearance cues. Extensive experiments demonstrate that our hierarchical and progressive structure aggregation can effectively distill high-level and low-level features from the input images, and achieve high-quality alpha mattes without external trimaps. 

In the future, we will explore how to improve the adaptability of the network to further increase the robustness of the model on natural images. The domain adaption~\cite{Sankaranarayanan_2018_CVPR} and GAN~\cite{xiangli2020real} may solve the problem of reducing the gap between synthetic data and natural images. We will also explore how to extract multi-target alpha mattes in a fashion of trimap-free method.

\begin{acks}
This work was supported in part by the National Natural Science Foundation of China under Grant $61972067/U21A20491$, the Innovation Technology Funding of Dalian (2020JJ26GX036), and the Open Research Fund of Beijing Key Laboratory of Big Data Technology for Food Safety (Project No. BTBD-2018KF).
\end{acks}

\bibliographystyle{ACM-Reference-Format}
\bibliography{sample-base}


\end{document}